\documentclass[final]{oist_thesis} 

\usepackage[dvipsnames]{xcolor}
\usepackage[english]{babel} 
\usepackage[utf8x]{inputenc} 

\usepackage[T1]{fontenc} 
\usepackage{lmodern}
\usepackage{graphicx} 
\graphicspath{{./Images/}} 

\usepackage{eso-pic} 
\usepackage{setspace} 
\usepackage{longtable} 
\usepackage{pdfpages} 
\usepackage{fancyhdr} 
\usepackage[colorlinks=true,allcolors=blue]{hyperref} 
\usepackage{epigraph} 

\usepackage{amsthm,amsmath,amssymb,amsfonts,bbm}
\usepackage{subcaption}
\usepackage{bm}
\usepackage{bbm}
\usepackage{empheq}
\usepackage{tikz}
\usepackage[export]{adjustbox}
\usepackage[makeroom]{cancel}
\usepackage{mathtools,nccmath}
\usepackage[ruled,vlined]{algorithm2e}
\usepackage{booktabs}
\usepackage{multirow}
\usepackage[normalem]{ulem}

\usepackage{epstopdf} 
\DeclareGraphicsExtensions{.pdf,.png,.jpg,.eps}

\newcommand{\stkout}[1]{\ifmmode\text{\sout{\ensuremath{#1}}}\else\sout{#1}\fi}
\usepackage{tikz}
\usetikzlibrary{decorations.pathreplacing}

\newcommand{\bea}{\begin{eqnarray}} 
\newcommand{\eea}{\end{eqnarray}}

\newtheorem{critere}{Criterion}

\usepackage[square, numbers, sort&compress]{natbib} 
\begin{document}
\pagestyle{empty} 
\frontmatter 

\puttitle{
	title=Robust Filtering – Novel Statistical Learning and Inference Algorithms with Applications, 
	name=Aamir~Hussain~Chughtai, 
	supervisor=Dr. Muhammad~Tahir, 
	cosupervisor=Dr. Momin~Uppal, 
	submissiondate={May, 2024}
}


\startpreamble

\unnumberedchapter{Declaration of Original and Sole Authorship} 
\chapter*{Declaration of Authorship} 
I, \name, declare that this thesis entitled \emph{\thesistitle} and the data presented in it are original and my own work. I confirm that:
\begin{itemize}
\item No part of this work has previously been submitted for a degree at this or any other university.
\item References to the work of others have been clearly acknowledged. Quotations from the work of others have been clearly indicated, and attributed to them.
\item In cases where others have contributed to part of this work, such contribution has been clearly acknowledged and distinguished from my own work.
\item None of this work has been previously published elsewhere, with the exception of the following: 

\subitem - Chughtai, A. H., Akram, U., Tahir, M., \& Uppal, M. (2020). Dynamic state estimation in the presence
of sensor outliers using MAP-based EKF. \textit{IEEE Sensors Letters}, 4(4), 1–4. \url{https://ieeexplore.ieee.org/document/9050910}
\subitem - Chughtai, A. H., Tahir, M., \& Uppal, M. (2020). A robust Bayesian approach for online filtering
in the presence of contaminated observations. \textit{IEEE Transactions on Instrumentation and
Measurement}, 70, 1–15. \url{https://ieeexplore.ieee.org/document/9239326}
\subitem - Chughtai, A. H., Tahir, M., \& Uppal, M. (2022). Outlier-robust filtering for nonlinear systems with
selective observations rejection. \textit{IEEE Sensors Journal}, 22(7), 6887-6897. \url{https://ieeexplore.ieee.org/document/9716131}
\subitem - Chughtai, A. H., Majal, A., Tahir, M., \& Uppal, M. (2022). Variational-based nonlinear Bayesian
filtering with biased observations. \textit{IEEE Transactions on Signal Processing}. \url{https://ieeexplore.ieee.org/document/9931968}
\subitem - Chughtai, A. H., Tahir, M., \& Uppal, M. (2024). Bayesian heuristics for robust spatial perception \textit{IEEE Transactions on Instrumentation and
	Measurement}. 
	\url{https://ieeexplore.ieee.org/document/10430181}
\subitem - Chughtai, A. H., Tahir, M., \& Uppal, M. (2024). EMORF/S: EM-Based Outlier-Robust Filtering and Smoothing With Correlated Measurement Noise (\textit{Under Review} ). \url{https://arxiv.org/abs/2307.02163} 

\end{itemize}

Date:  \submissiondate

Signature: \textbf{Aamir Hussain Chughtai}

\unnumberedchapter{Acknowledgment} 
\chapter*{Acknowledgment} 


\hspace{.5cm} I am blessed in sense to have access to resources to afford and pursue what I wanted academically and intellectually, eventually pursing a PhD in the field of my interest.  
\\

First of all I am indebted to my family who have been instrumental in this journey. I am thankful to my parents who have provided every resource in my educational endeavors and have supported me financially and emotionally. The roots of my reasoning aptitude can certainly be attributed to my father. I can vividly recall the childhood memory of the game of 5 Rs per math question we used to play. I thank my father who ensured every resource for my education. Then of course my mother who has been and remains a source of support, solace, warmth and a confidant. I also thank my sister who completes my family puzzle. Having grown together we have shared cherished memories. She has witnessed me closely in times of fail and success. My professional and academic journey would not have been possible had I not got the enabling family environment I was lucky enough to have. 
\\

I would also like to acknowledge my late high school maths teachers: Sir Ishtiaq and Taji. I am indebted to the training they provided me. They in a sense passed the baton to me and a part of them now lives through me. I also acknowledge the University of Engineering and Technology (UET), Lahore which has been essential in my professional struggle and gave me a reality check during the Bachelors program.
\\

I am very grateful to Lahore University of Management Sciences (LUMS) which has been formative in my personal and professional development. It is the place where I got exposed to the graduate level. The culture, environment, vibes and faculty is absolutely amazing. I am specially thankful to my PhD advisors Dr. Muhammad Tahir and Dr. Momin Uppal who have been wonderful mentors helping me navigating the process. Dr. Tahir has been really encouraging and supportive in every possible way. He gave me total freedom to express myself at the same time keeping me on track. Dr. Momin has been instrumental in setting the quality benchmark. His critical evaluation, objective feedback and insightful recommendations are worth acknowledging.  
\\

I would like to acknowledge my fellows in the AdCom Lab LUMS. First there were Nabeel and Azeem with whom I have spent quality time. Later Arslan, Zawar, Fasih, Talha, Mazhar and others joined the party. I have really enjoyed my work in their presence which otherwise had been too mechanical. I thank Arslan for his help in the simulations and experimental campaign in this work. Lastly, I would also like to acknowledge my friends Rehan, Saqib and Ahmad who have been appreciative and encouraging for my journey.
\\

I feel myself lucky for having opportunity to pursue the doctorate degree. It has been quite a ride with several ups and downs; a test of patience in a sense. It has been an opportunity to test myself at a high level in academics and to express myself. I am pleased with the journey.   

\cleardoublepage
\thispagestyle{empty} 

\vspace*{8cm} 

\hfill
\begin{parbox}{0.6\textwidth}{
\begin{flushright}

\textit{Dedicated to my mother who remains a source of warmth and support.
}
\end{flushright}}
\end{parbox}

\cleardoublepage
\thispagestyle{empty} 

\vspace*{10cm} 

\begin{parbox}{1\textwidth}{	
\begin{center}
\textit{``Only after disaster can we be resurrected.''}
\newline
\textit{--Chuck Palahniuk, Fight Club}
\end{center}}
\end{parbox}


\unnumberedchapter{Abstract} 
\chapter*{Abstract} 
\subsection*{\thesistitle}
\begin{figure*}[h!]
	\centering
	\includegraphics[width=0.7\linewidth,trim=4 4 4 4,clip,frame]{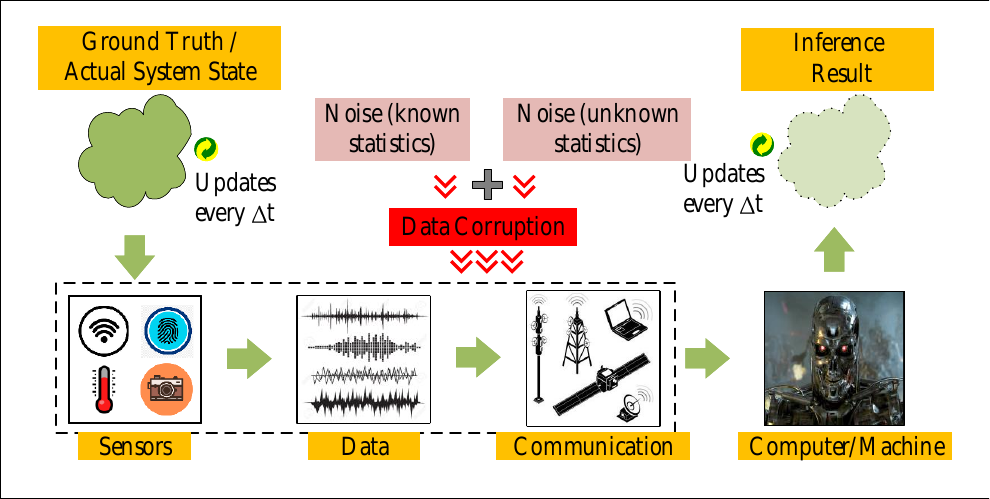}
	\label{fig:robustfiltering}
\end{figure*}


Statistical inference is a key task in many sophisticated modern solutions. Recent advances in computing, sensing, and communications technologies mean that we are dealing with unprecedented engineering complexities that call for developing efficient inference approaches. Essential goals of inference include estimating the model, structure, or latent system states using the obtained data. Subsequently, important actions based on the inference results can be performed. 

Filtering or state estimation of dynamical systems is a crucial inference task for providing new decision-making and system automation information in different applications. Examples include energy management systems, smart grids, robotics, healthcare, intelligent transportation systems, etc. The standard filtering setup assumes that the system dynamics are adequately captured by state space models, where the latent states evolve with first-order dynamics. Though not directly observable, the states manifest as a set of noisy external outputs measured by sensors. Considering the requirement of an online recursive solution in addition to the possibilities of underlying functional nonlinearities and dependency on noisy observations for inference the filtering problem is not trivial.  

The complexity of filtering is compounded when the measurements get corrupted with different abnormalities or {gross errors} having unknown or partially known statistics. This violates the assumption of standard filtering approaches where the nominal noise statistics are assumed to be known apriori. Gross errors encompass disturbances apart from the nominal noise, including outliers and biases/drifts. Generally modeled with zero-mean statistics, outliers are sudden sporadic surges in the measurements. Their appearance can be attributed to factors like sensor degradation, communication failures, environmental influences, front-end processing limitations, etc. On the other hand, biases disturb the data with non-zero mean noise statistics. Multiple factors in various applications lead to the appearance of biased observations. Examples include miscalibrations or malfunctioning of sensors and other configuration aberrations like sensor location, alignment, clock errors, etc.

Mitigating the effect of measurement abnormalities on state estimation calls for robust filtering methods, which form the basis of this dissertation. Our primary emphasis remains on devising and investigating robust nonlinear filters considering the occurrence of outliers and biases in the data. In some cases, peripheral instances of missing observations and drifts are also evaluated. We also extend one of our proposed robust filters to a general estimation setup in the presence of outliers. Also a smoothing extension for a designed method is also presented. To make our devised methods implementable on digital computers, we restrict our focus to discrete-time state space models. By proposing suitable models, we develop and study different state estimation approaches. We resort to tools from statistical (mostly Bayesian) inference theory, including the deterministic and stochastic methods for distributional approximations. To gauge the theoretical estimation limitations, we evaluate the Bayesian Cramer Rao Bounds for some of our models. For practical performance evaluation, we choose a variety of appropriate simulations and experimental setups to gauge the gains compared to the state-of-the-art approaches mostly in terms of estimation accuracy and computational strains.

\unnumberedchapter{Notations} 
\chapter*{Notations} 
\vspace{-1cm}
\begin{tabular}{ll}
&\hspace{-2.2cm} Here we present a general notation for the dissertation. Particular notation in each  \\
&\hspace{-2.2cm} chapter is defined in the local context.\\
&\\	
$a$& \hspace{.8cm} Scalar\\
$\mathbf{a}$& \hspace{.8cm} Vector\\
$\mathbf{A}$& \hspace{.8cm} Matrix\\
$\mathbbm{A}$& \hspace{.8cm} Space\\
${a}_k$\ $\mathbf{a}_k$\ $\mathbf{A}_k$& \hspace{.8cm} General scalar, vector and matrix respectively at time step $k$\\
${a}^i$\ $\mathbf{a}^i$\ $\mathbf{A}^i$& \hspace{.8cm} General $i$th scalar, vector and matrix respectively used in a similar  \\	
& \hspace{.8cm} context or to indicate exponent\\
${a}^{(i)}$\ $\mathbf{a}^{(i)}$\ $\mathbf{A}^{(i)}$& \hspace{.8cm} $i$th particle/sample or iteration value of a general scalar, vector and \\	
& \hspace{.8cm} matrix respectively\\
${a}(i)$& \hspace{.8cm} $i$th element of any vector $\mathbf{a}$\\
${A}(i,j)$& \hspace{.8cm}  Element at the $i$th row and $j$th column of any matrix $\mathbf{A}$\\
$\sqrt{\mathbf{A}}$ & \hspace{.8cm} Matrix such that $\mathbf{A}=\sqrt{\mathbf{A}} \sqrt{\mathbf{A}}^{\top} $\\
$[\mathbf{A}]_i$& \hspace{.8cm} $i$th column of matrix $\mathbf{A}$\\
$\mathbf{A}^{\top}$& \hspace{.8cm} Transpose of $\mathbf{A}$\\
$\mathbf{A}^{-1}$& \hspace{.8cm} Inverse of $\mathbf{A}$\\
$\mathbf{a}_{n:k}$& \hspace{.8cm} Set containing the vectors \{$\mathbf{a}_{n},...,\mathbf{a}_{k}$\}\\
$\mathbf{f}(\mathbf{a})$& \hspace{.8cm} Functional transformation of vector $\mathbf{a}$\\
${f}(\mathbf{a})(i)$& \hspace{.8cm} $i$th element of functional transformation $\mathbf{f}$ of vector $\mathbf{a}$\\
$\mathbf{f}^i(\mathbf{a})$& \hspace{.8cm} General $i$th functional transformation $\mathbf{f}$ of vector $\mathbf{a}$\\
$\frac{\partial}{\partial {a}}$& \hspace{.8cm} Partial derivative with respect to $a$\\
$\approx  $  & \hspace{.8cm} Approximately equal to \\
$\triangleq$  & \hspace{.8cm} Is defined equal to \\
%
$\delta(.)$& \hspace{.8cm} Delta function\\
%
%
%
%
%
%
%
%
%

$\odot  $& \hspace{.8cm} Hadamard product\\
{atan2}$(.)$& \hspace{.8cm} 2-argument arctangent\\
$\ll$& \hspace{.8cm} Much less than\\
$\gg$& \hspace{.8cm} Much greater than\\
$\mathrm{diag}$$(.)$& \hspace{.8cm}  Operator for vector to diagonal matrix conversion and vice versa\\
$\mathrm{mod}$& \hspace{.8cm}  Modulo\\
$|.|$& \hspace{.8cm} Determinant\\
$\|.\|$& \hspace{.8cm} L2 norm\\
$\|.\|_{F}$& \hspace{.8cm} Frobenius norm\\
${\text{tr}}(.)$& \hspace{.8cm} Trace\\
$\mathbf{x}\sim p(\mathbf{x})$& \hspace{.8cm} Random variable $\mathbf{x}$ follows the probability
distribution $p(\mathbf{x})$\\

$p(\mathbf{x})$& \hspace{.8cm} Probability distribution of $\mathbf{x}$\\

$p(\mathbf{x}|\mathbf{y})$ & \hspace{.8cm} Conditional distribution of $\mathbf{x}$ given $\mathbf{y}$ \\

$p(\mathbf{x}|\mathbf{y},\stkout{\mathbf{z}})$ & \hspace{.8cm} Conditional distribution of $\mathbf{x}$ given $\mathbf{y}$ being independent of $\mathbf{z}$\\
$\langle \mathbf{f}(\mathbf{x})\rangle_{p(\mathbf{x})} $& \hspace{.8cm} Expected value of $\mathbf{f}(\mathbf{x})$ with $\mathbf{x}\sim p(\mathbf{x})$


\end{tabular}

%
%
%
%
%
\newpage

\unnumberedchapter{Abbreviations} 
\chapter*{Abbreviations}

\begin{table}[thp!]
	\begin{tabular}{ll}	
		AB& \hspace{1cm} Academic Block\\
		AC& \hspace{1cm} Alternative Current\\
		ACM& \hspace{1cm} Approximate Conditional Mean\\
		AOA& \hspace{1cm} Angle of Arrival\\	
		ASOR & \hspace{1cm} Adaptive Selective Observations-Rejecting \\ 
		AWGN& \hspace{1cm} Additive White Gaussian Noise \\
		BDM& \hspace{1cm}  Bias Detecting and Mitigating\\
		BCRB& \hspace{1cm}  Bayesian Cramer-Rao Bound\\
		CKF & \hspace{1cm} Cubature Kalman Filter \\   
		CPS & \hspace{1cm} Cyber-Physical System \\   
		CUKF & \hspace{1cm} Constrained Unscented Kalman Filter  \\   
		EKF & \hspace{1cm} Extended Kalman Filter \\
		EM & \hspace{1cm} Expectation-Maximization \\
		EMORF & \hspace{1cm} EM-based outlier-robust filter\\
		EMORS & \hspace{1cm} EM-based outlier-robust smoother\\
		ESOR & \hspace{1cm} Extended Selective Observations-Rejecting \\ 
		FDI & \hspace{1cm} Fault Detection and Identification \\
		GAN & \hspace{1cm} Generative Adversarial Networks\\
		GP & \hspace{1cm} Gaussian Process \\ 	 
		GPS & \hspace{1cm} Global Positioning System \\
		HMM & \hspace{1cm} Hidden Markov Model \\    
		HV & \hspace{1cm} High-Voltage		 \\    
		IED	&  \hspace{1cm} Intelligent Electronic Devices \\ 
		i.i.d.	&  \hspace{1cm} Independent and Identically Distributed  \\ 
		IoT	&  \hspace{1cm} Internet of Things \\ 
		KF & \hspace{1cm} Kalman Filter \\
		KL & \hspace{1cm} Kullback-Leibler  \\ 
		kNN & \hspace{1cm} k-Nearest Neighbors \\
		LCKF& \hspace{1cm} Linearly Constrained Kalman Filter\\
		LOS& \hspace{1cm} Line-of-Sight \\ 
		LRT& \hspace{1cm} Likelihood Ratio Test \\ 
		LTI& \hspace{1cm} Linear Time-Invariant\\ 
		MAP& \hspace{1cm} Maximum a Posteriori\\
		MC& \hspace{1cm} Monte Carlo\\
		MEMS & \hspace{1cm} Microelectromechanical Systems \\
	\end{tabular}
\end{table}
\newpage
\begin{table}[thp!]
	\begin{tabular}{ll}
		MMSE& \hspace{1cm} Minimum Mean Squared Error\\
		MQTT& \hspace{1cm} 	Message Queuing Telemetry Transport\\		
		NLOS&\hspace{1cm} Non-Line-of-Sight \\
		OD&\hspace{1cm} Outlier-Detecting  \\ 
		ORGKF&\hspace{1cm} Outlier Robust Gaussian Kalman Filter\\
		PCA & \hspace{1cm} 	Principle Component Analysis \\ 
		PCB & \hspace{1cm} 	Printed Circuit Board \\ 
		PCRB & \hspace{1cm} Posterior Cramer Rao Bound \\ 
		PF& \hspace{1cm} Particle Filter\\
		PGM& \hspace{1cm} Probabilistic Graphical Model\\
		PGO& \hspace{1cm} Pose Graph Optimization\\
	    PMU	&  \hspace{1cm} Phasor Measurement Unit   \\ 
		P-SPKF	&  \hspace{1cm} Parallel Sigma Point Kalman Filter  \\
		QCQP & \hspace{1cm} Quadratically Constrained Quadratic Program \\
		RAM & \hspace{1cm} Random Access Memory \\
		RMSE & \hspace{1cm} Root Mean Squared Error \\
		ROR & \hspace{1cm} 	Recursive outlier-robust \\
		SDP & \hspace{1cm} Semidefinite Programming  \\ 
		SEM & \hspace{1cm} Switching Error Model \\ 
		SfM & \hspace{1cm} Structure from Motion  \\ 
		SKF & \hspace{1cm} Schmidt-Kalman Filter \\ 
		SLAM & \hspace{1cm} Simultaneous Localization and Mapping \\ 
		SMC & \hspace{1cm} Sequential Monte Carlo\\ 
		SOR & \hspace{1cm} 	Selective Observations-Rejecting \\ 
		SRD-SPKF & \hspace{1cm} Serial Re-draw Sigma Point Kalman Filter \\ 
		SSE & \hspace{1cm} School of Science and Engineering \\
		SSM & \hspace{1cm} State Space Model \\
		S-SPKF & \hspace{1cm} Serial Sigma Point Kalman Filter \\
		SVM & \hspace{1cm} Support-Vector Machines\\  
		TRMSE & \hspace{1cm} Time-Averaged Root Mean Squared Error \\ 
		USKF & \hspace{1cm} Unscented Schmidt Kalman Filter \\    		
		UKF & \hspace{1cm} Unscented Kalman Filter \\    
		UT & \hspace{1cm} Unscented Transform\\ 
		UWB & \hspace{1cm} Ultra-Wideband\\ 
		VAE	&  \hspace{1cm} Variational Autoencoders \\ 	
		VB	&  \hspace{1cm} Variational Bayes \\ 	
	\end{tabular}
\end{table}


\unnumberedchapter{Contents}
\tableofcontents 
\unnumberedchapter{List of Figures}
\listoffigures 
\unnumberedchapter{List of Tables}
\listoftables 


\addtocontents{toc}{\vspace{2em}} 
\mainmatter 


\numberedchapter 

\chapter{Introduction} \label{ch-1}

\section{Background}

\subsection{Inference Ability in Intelligent Life}

\begin{figure}[h!]
	\begin{subfigure}{.5\textwidth}
		\centering
		\includegraphics[width=7cm,height=4cm,frame]{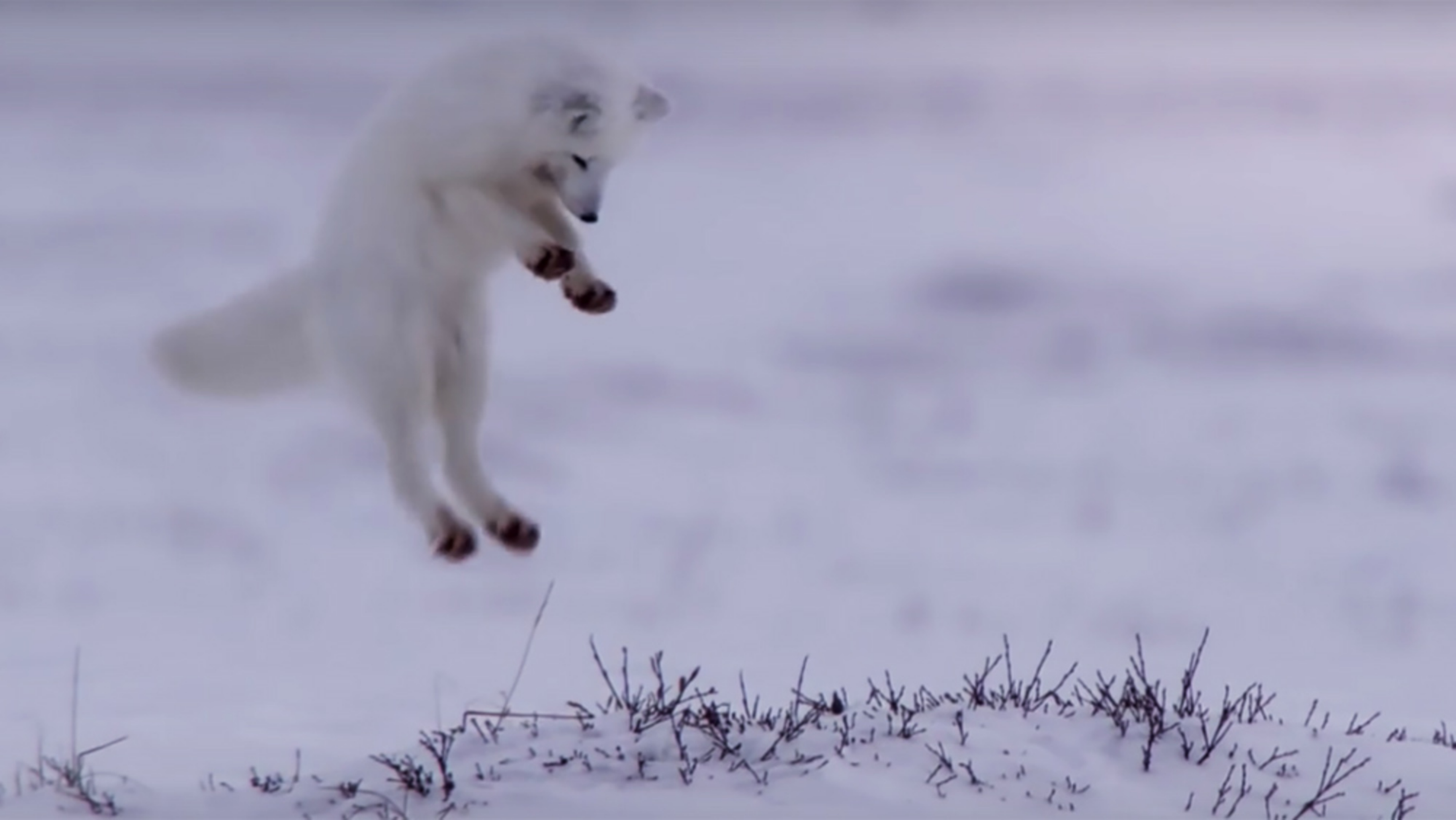}
		\caption{An Arctic fox on the hunt}
		\label{fig:sfig1}
	\end{subfigure}%
	\begin{subfigure}{.5\textwidth}
		\centering
		\includegraphics[width=7cm,height=4cm,frame]{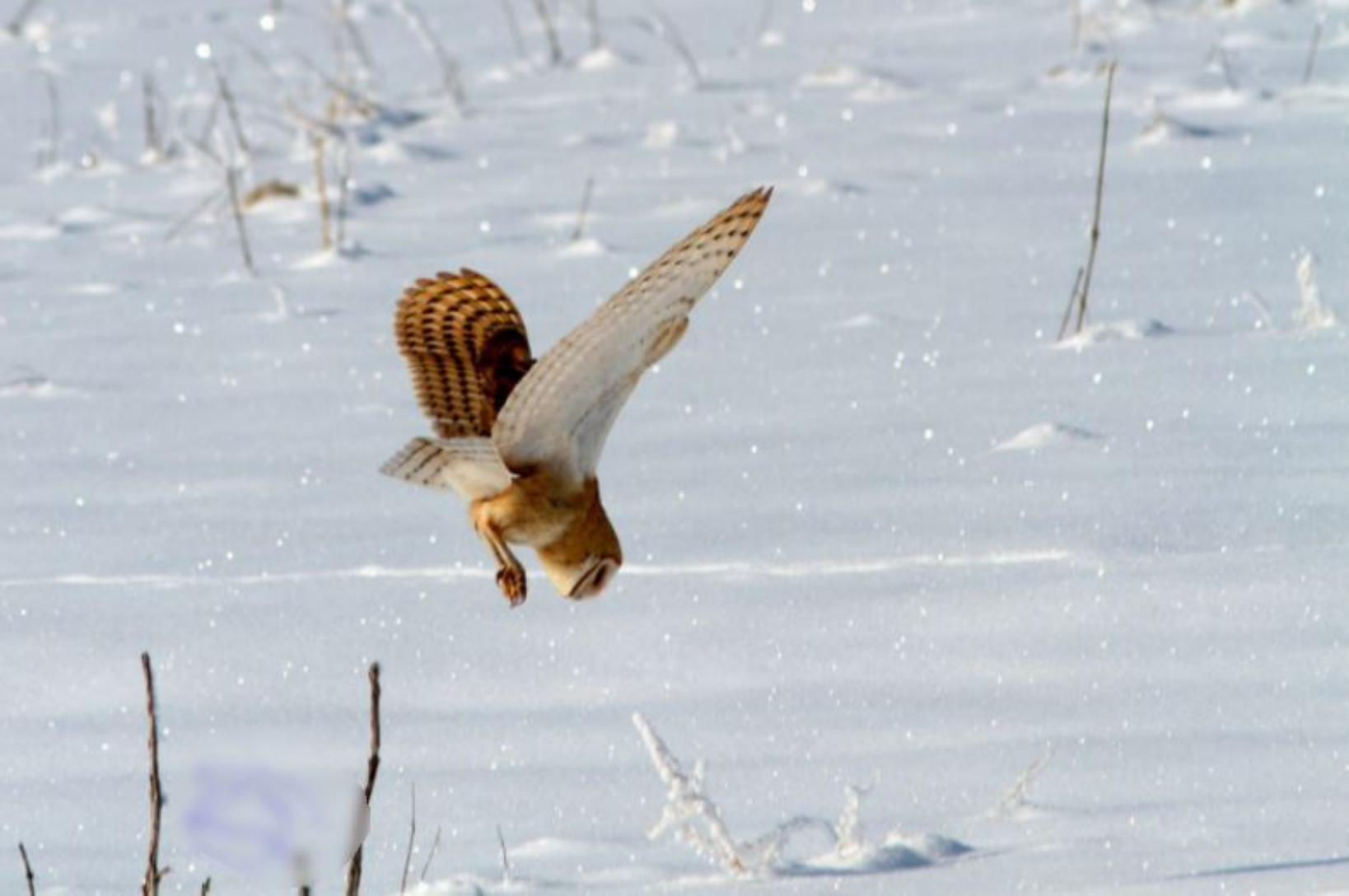}
		\caption{An owl on the hunt}
		\label{fig:sfig2}
	\end{subfigure}
	\caption{Animals using inference to devise their attacks to capture preys underneath snow}
	\label{fig:animal_attack}
\end{figure}

The ability to make inference is a prevalent characteristic in intelligent species and is an important component of intelligence. When faced with any situation where the ground truth is not directly observable or is obscure, exercising inference helps them to reach conclusions based on reasoning and evidence. This can provide key advantages in various tasks and can even be crucial for survival in some scenarios. It is a spectacular sight to see how different predators use inference during hunting to catch their preys and similarly how the targets use inference to outsmart the predators. For example, Arctic foxes and owls use their superior hearing abilities to infer the locations of their prey and use that to carry out attacks underneath snow e.g. as shown in Figure~\ref{fig:animal_attack}. Though the actual preys are not directly observable, the hunters use the sound signals along with their prior experiences to track the target position and subsequently attempt an attack. Such scenarios are commonly observed in the animal kingdom.

\begin{figure*}[t!]\centering
	\begin{subfigure}{.8\textwidth}
		\centering
		\includegraphics[width=10cm,frame]{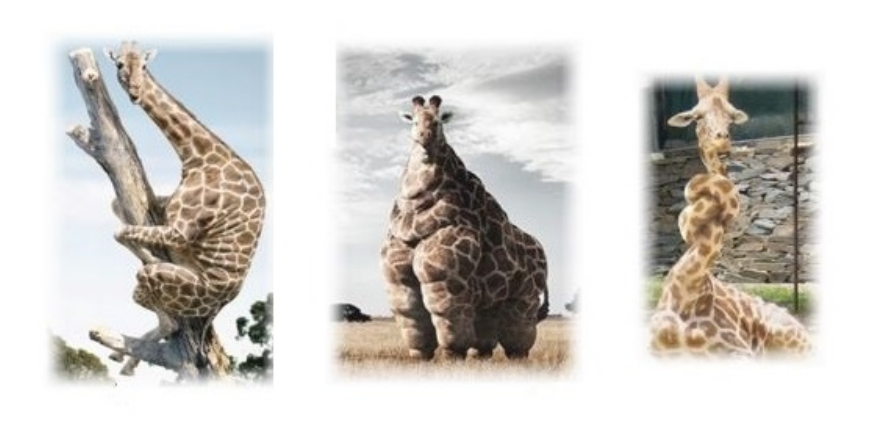}
		\caption{Guess the animal [Courtesy Thomas Serre]}
	\label{fig:giraffe}
	\end{subfigure}%
\\
	\begin{subfigure}{.8\textwidth}
		\hspace{-.1cm}
		\centering
		\includegraphics[width=10cm,height=4cm,trim=-10 0 10 0,clip,frame]{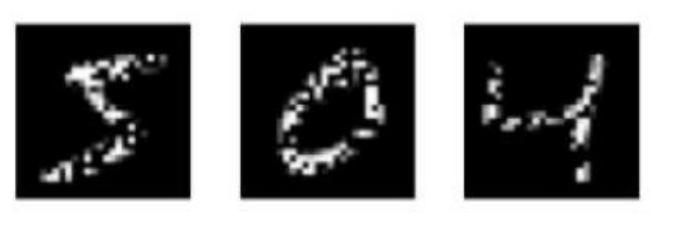}
		\caption{Guess the digits [Courtesy Nikhil Buduma]}
		\label{fig:hand}
	\end{subfigure}
	\caption{Examples of how we humans use inference for object recognition}
	\label{fig:guess}
\end{figure*}

As compared to animals, humans generally have superior inference abilities thanks to a complex combination of intelligence, memory and linguistic power, etc. We humans use inference subconsciously during our daily tasks. For example, while driving we use our visual abilities and mental faculties to infer the proximity of other vehicles and what different signboards mean. Similarly, we can recognize any specific bird outside our homes based on what sound we hear inside. Likewise, we can judge the similarity of faces and voices of different people. It is common in various competitions to infer what the next move of an opponent can be and plan accordingly. In ball sports, we infer the pace of the ball and predict its location and act accordingly. At any crime scene, detectives and police use different pieces of evidence to get to the perpetrator. If we analyze closely, inference ability is a very prevalent and important part of our social routine. An example of how we humans use inference during visual recognition is shown in Figure \ref{fig:guess}. Though the images are corrupted, by noise and distortions, we can still make an educated guess of what the actual object can be.

\subsection{Rise of the Machines}
\begin{figure*}[h!]
	\centering
	\includegraphics[width=0.5\linewidth,frame]{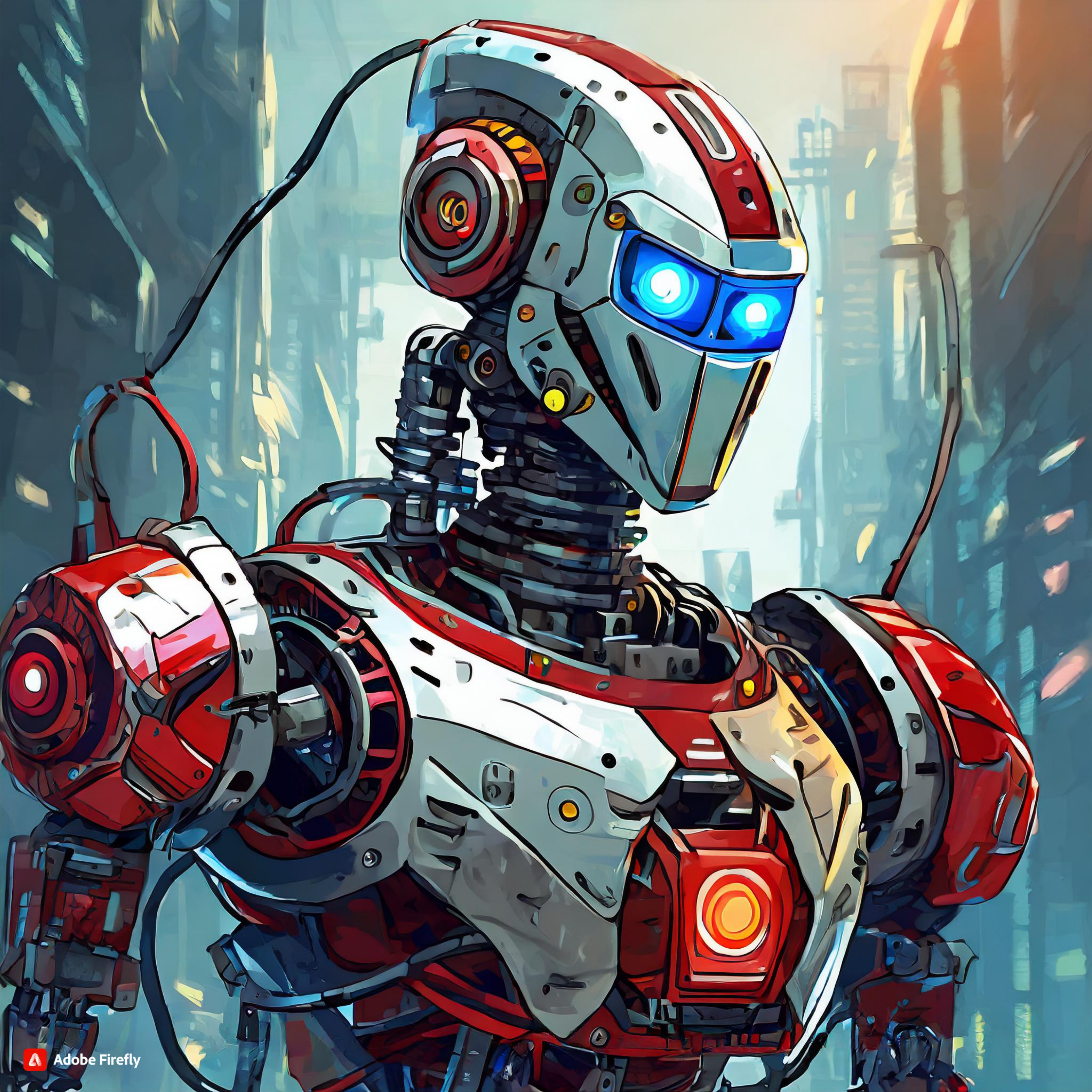}
	\caption{Intelligent machines are able to perform inference}
	\label{fig:terminator1}
\end{figure*}
\hspace{.5cm} Humans are unique in a way that they can build different tools and machines to perform certain tasks efficiently. This ability has been used for ages e.g. to develop weapons, vehicles, buildings, etc. More recently, we have been also been able to build machines - called computers - to perform computations and store data for us. Besides, we have also successfully developed certain devices to enhance our biological communication and sensory capabilities. Having an enhanced \textit{computation}, \textit{storage}, \textit{communication }and \textit{sensing} power at our disposal, we have been able to realize machines for inference tasks. We can now design \textit{artificial} systems to aid us in performing certain tasks or even completely delegate them to machines. The visuals which were portrayed only in sci-fi movies e.g. Figure \ref{fig:terminator1} are coming to reality in different forms now.

\subsection{Inference Process}

\begin{figure*}[h!]
	\centering
	\includegraphics[width=0.7\linewidth,trim=4 4 4 4,clip,frame]{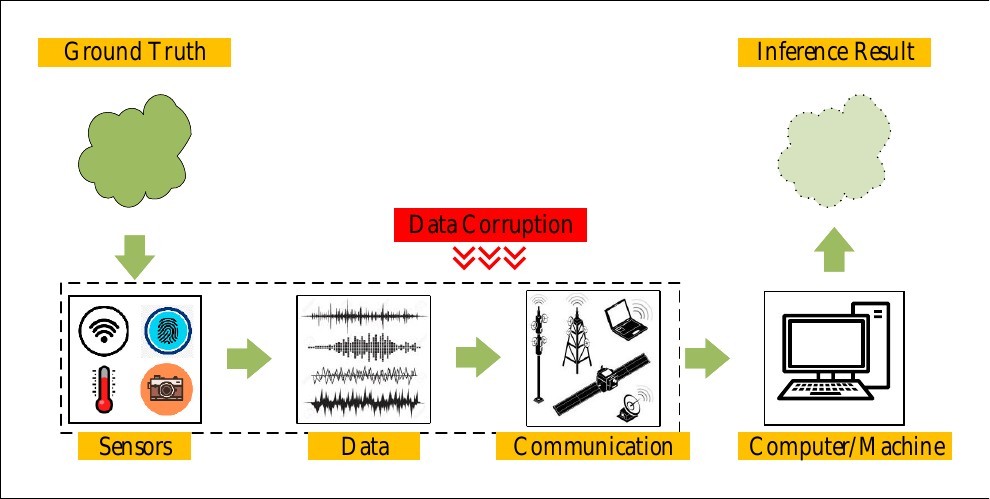}
	\caption{General inferential process using computers}
	\label{fig:inferencesystem}
\end{figure*}

For any inferential scheme to work effectively we should be able to map what we can sense and transmit (usually called the \textit{data}) to the target, subject or parameter, etc. of interest. In the process, we can use how the data was actually generated and any inherent limitations of the target, subject or parameter, etc. to our advantage. Mathematics provides us a systematic way for such mapping (modeling) and devise algorithms for inference. Subsequently, using the processing and storage power of computers we can execute  the algorithms. The general structure of performing inference \textit{artificially} using computers is shown in Figure \ref{fig:inferencesystem}. Some common terms that are commonly used in connection to the inference process are discussed here \cite{koller2009probabilistic,jordan2015machine,murphy2012machine}.

\subsubsection{Learning vs Inference}
\hspace{.5cm}The terminologies of learning and inference are used in connection in various contexts. These are even used interchangeably in some contexts. The traditional machine learning community generally makes a distinction between the two terms. They usually attribute the term \textit{learning} to the process of determining the model structure, topology, or parameters that explain the data. The term inference is used to describe the process of \textit{predicting} the sought variable or parameter of interest when any new data arrives. On the other hand, researchers from other communities like statistics use the terms interchangeably. For them learning the model parameters or structure is also a sort of inference. In this work, we resort to the latter interpretation.

Practically, the results of inference can be erroneous. The error depends on how much details the model can potentially capture. Moreover, it also depends on how tractable is the model for inference or how well inference can be performed on any chosen model using the techniques at hand.

\subsubsection{Supervised vs Unsupervised Learning}
\hspace{.5cm}Learning problems can be also be classified into supervised and unsupervised types. In supervised learning ground truth labels are available against the data. Using the labels and the data the model parameters or the structure are learned. Two main areas where supervised learning is employed include classification and regression problems. Unsupervised learning, on the other hand, does not resort to the availability of labels and rather learns the structure of the model for different tasks. Unsupervised methods are usually used for tasks like clustering, association, anomaly detection, etc. There also exist methods - known as semi-supervised methods - where some labels are available against the data.  

\subsubsection{Online vs Offline Problems}
\hspace{.5cm}There exists another class of distinction in learning and inference problems depending on whether the tasks performed have real-time requirements or not. The tasks which have to be executed on the go are online problems. Usually, in computer algorithms, these are performed on a per-sample basis i.e. inference has to be done within the sampling period before any new data arrives. Examples include object tracking, fault detection, heart rate monitoring, etc. Sometimes in online problems sequential batches of data are also processed. Offline problems, on the other hand, have no real-time requirements. This type of inference is performed on the complete batch of offline stored data or can be done sequentially on the stored data. Examples include image enhancements, classification, and segmentation, system identification, etc. 

\subsubsection{Deterministic vs Probabilistic Modeling}
\hspace{.5cm}Learning and inference models can be also categorized into deterministic and probabilistic types. As the name implies in deterministic modeling the structure is entirely based on deterministic variables and parameters. The learned variables are then used subsequently for inference. Mathematical tools used for learning and inference in these models are derived using methods from calculus, optimization, linear algebra, functional analysis, etc. Deterministic model examples include neural networks, principal component analysis (PCA) etc. Probabilistic or statistical models as opposed to deterministic models use probabilistic and statistical approaches in the construction of the models. The motivation primarily stems from the fact that some phenomenon in the real world can be effectively explained only scholastically. For example, sensors, communication channels, modeling assumptions, and approximations do transformations to their input data which cannot be solely represented deterministically and need statistical approaches for better representation and incorporate more information from data. Learning and inference in these models are mathematically done using tools of probability, statistics, information theory, optimization, linear algebra, etc. Probabilistic models are widely applied in practice e.g. hidden Markov models (HMMs), state space models (SSMs), variational autoencoders (VAEs) etc.

\subsubsection{Generative vs Discriminative Modeling}
\hspace{.5cm}Learning and inference tasks require mapping or transformation from data to some class or other parameter of interest. This can be achieved by generative or discriminative approaches. In discriminative methods, the (inverse) transformation is explicitly considered. For deterministic models, it means assuming the functional form and for probabilistic approaches, it means considering the posterior probability i.e. the probability of the target variable given the data or observation. Examples of these methods include neural networks, logistic regression, etc. In generative models, the data generation process is modeled for subsequent inference. For example, the underlying prior and likelihood distributions (employing the Bayes rule) are used to determine the posterior distributions of the latent parameters for inference. Examples of generative models include HMMs, Naive Bayes, Bayesian networks, etc. Generative models have the ability to generate synthetic data points.

\subsubsection{Parametric vs Non-parametric Modeling}
\hspace{.5cm}Another common distinction for learning and inference models is in terms of whether the structure of the learning paradigm is fixed \textit{apriori} or it is dynamic and determined using data. In parametric methods, usually, the learning and inference function is defined before the learning process. Subsequently, the parameters that define the functions are estimated. Examples include neural networks, HMMs, Logistic regression, etc. On the contrary, in non-parametric models, the function is not well-defined and can adapt in terms of the number of parameters. Sometimes these refer to \textit{distribution-free} methods where the underlying form of distribution are not specified but rather learned. Examples include k-nearest neighbors (kNN), support-vector machine (SVM), Gaussian process (GP) etc.

\subsubsection{Exact vs Approximate Inference}
\hspace{.5cm}Having a defined model the following arising question is the suitability and tractability of the model for learning and inference. Commonly, it is not practicable to minimize the actual loss function for learning or the solutions are too difficult to find. Consequently, approximate methods are employed to that end. For example, for deterministic loss functions techniques like convex relaxation, heuristic methods, functional approximations, etc. can be employed. In the case of stochastic loss functions functional approximations, variational inference, Monte Carlo methods, etc. are commonly used.  

\subsubsection{Cloud vs Edge Computing}
\hspace{.5cm}Devising inference algorithms depend on the application of the method and how many computational resources are available at hand. Usually, computing resources are available at the edge of the networks or in a centralized location known as the cloud. Edge computing resources are commonly allocated for processing real-time sensitive data and also where network connectivity is difficult. However, since the computational power in these devices is limited inference algorithms have to be devised keeping into consideration the computational resources. Cloud computing, in contrast, has a much larger processing power and is suitable for managing a large amount of data. For internet of things (IoT)  applications both edge and cloud resources are used to process data and perform tasks.

\subsection{Applications of Algorithmic Inference}
\hspace{.5cm}Inference finds applications is various fields. It is used in Physics-based applications, several areas of engineering including power and energy, robotics, vehicles, process industry, aerospace, defense and military applications, data science, neuroscience, healthcare, seismology, and many more. In the following, we discuss few examples of inference in different fields.

\subsubsection{Physics-based Applications}
\hspace{.5cm}Inference is widely used in various Physics-based applications for estimation and detection purposes. Examples include position estimation for alignment of high-energy laser beams, sequential detection of broadband ocean acoustic sources, detection of critical materials or chemicals, radioactive sources, etc \cite{candy2016bayesian}. Usually, complex Physics-based models are used to describe the underlying phenomenon of interest and the data it generates.  

\begin{figure}[h!]
	\centering
	\includegraphics[width=0.7\linewidth,frame]{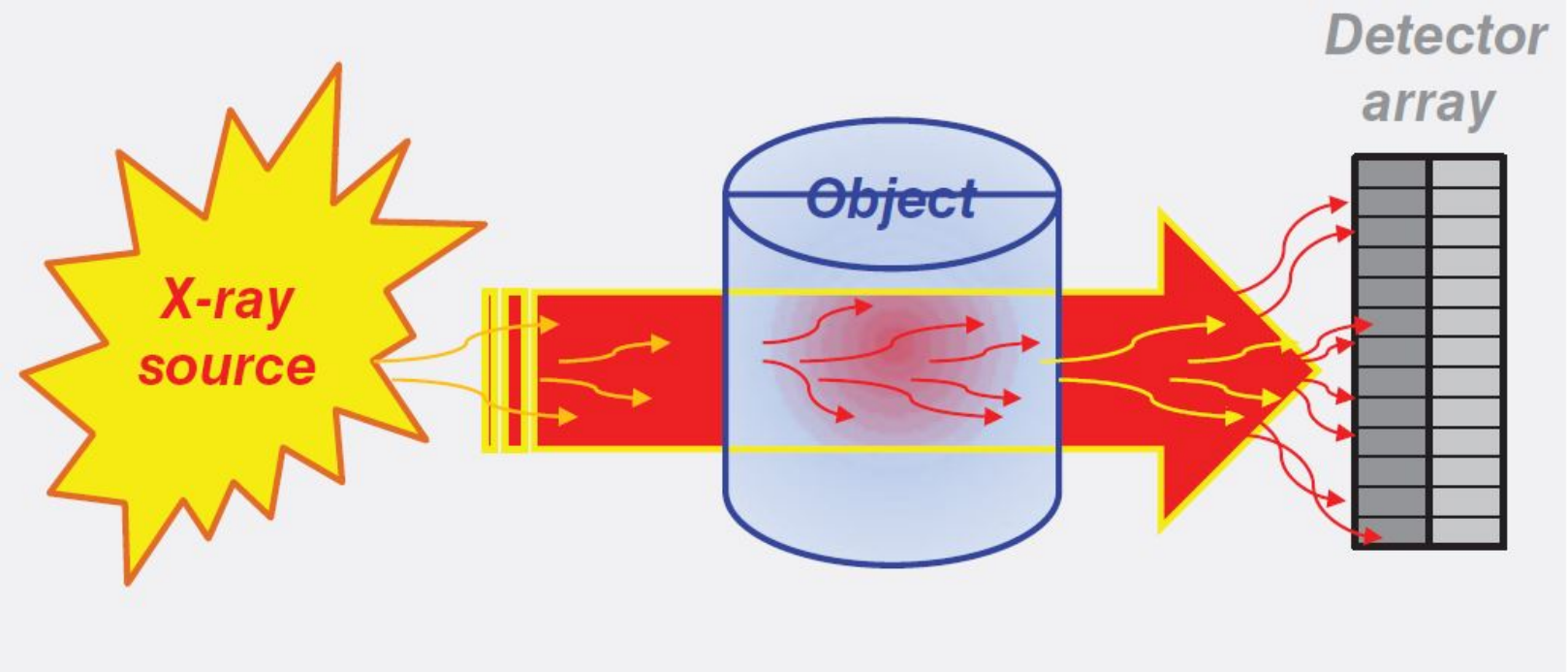}
	\caption{X-ray transmission physics: polyenergetic source, object, detector array
		measurements, and sequential inference for threat detection \cite{candy2016bayesian}}
	\label{fig:xray}
\end{figure}
\hspace{.5cm}Figure \ref{fig:xray} shows an example of application of inference to detect the presence of X-ray source as it passes through any container that carries it \cite{candy2016bayesian}. Based on the measurements the detector has to infer whether the radioactive material is present inside. Such devices are crucial for threat detection e.g. at entry points of security borders.

\subsubsection{Power and Energy}

\begin{figure}[h!]
	\centering
	\includegraphics[width=0.7\linewidth,frame]{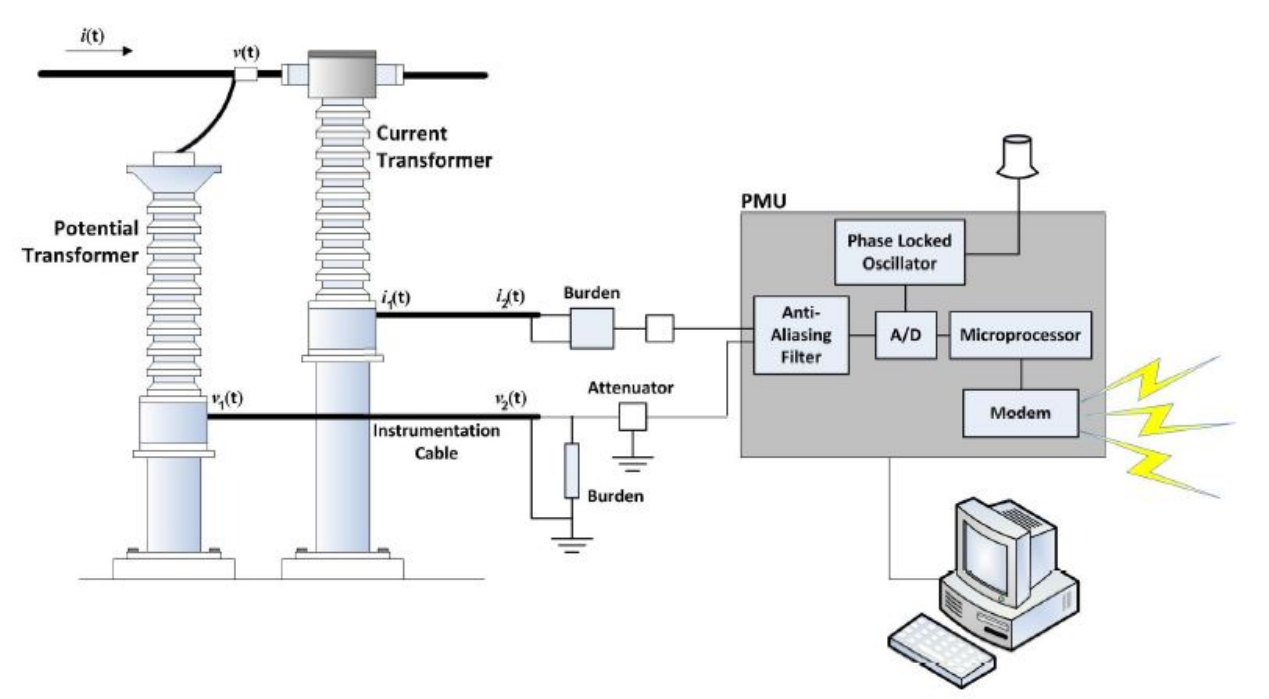}
	\caption{Inference is employed in power systems using dedicated hardware devices like PMUs \cite{sexauer2013phasor} }
	\label{fig:pmu}
\end{figure}
\hspace{.5cm}In power and energy engineering, dynamic state estimators are commonly employed for tasks in monitoring, protection, maintenance, etc. For example, dedicated hardware devices like phasor measurement units (PMUs), intelligent electronic devices (IEDs), and protective relays are programmed with specially designed inference algorithms for certain tasks. For example, PMUs provide real-time estimates of phasors for monitoring purposes as shown in Figure~\ref{fig:pmu}. Frequency relays have dedicated tasks to determine (based on inference) under and over frequency scenarios and issue tripping commands to circuit breakers.

\subsubsection{Robotics and Automation}
\begin{figure}[h!]
	\centering
	\includegraphics[width=0.7\linewidth,frame]{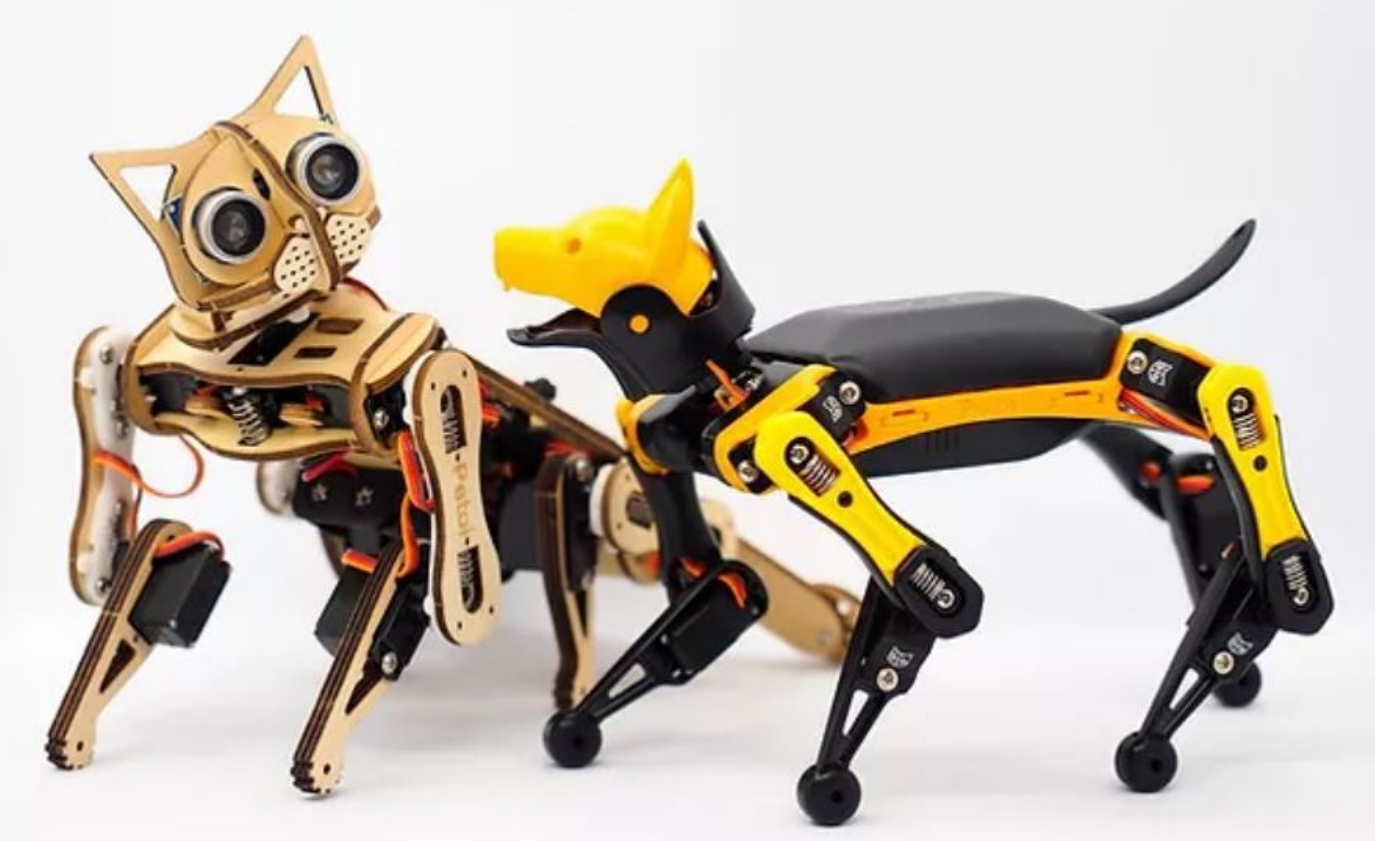}
	\caption{Inference algorithms allow different robots to process data and take actions}
	\label{fig:dog}
\end{figure}
\hspace{.5cm}Learning and inference algorithms are commonly deployed in robotics and automation applications. These methods allow different robots to process data and take appropriate actions in different scenarios. For example, a popular inference algorithm, known as simultaneous localization and mapping (SLAM) \cite{938381}, is deployed in robots for determining its location and at the same time generating maps of its surroundings. In addition, statistical inference is also used for automation purposes e.g. in the process industry to look for faults, abnormalities and also for predictive maintenance purposes.

\subsubsection{Defense}
 \hspace{.5cm}Statistical inference has key applications in modern defense areas. It lies at the core of several air and naval defense devices. For example, radar-based and sonar-based systems are commonly deployed to infer the position of any approaching adversary and threat.

\begin{figure}[h!]
	\centering
	\includegraphics[width=0.7\linewidth,height=.3\linewidth,frame]{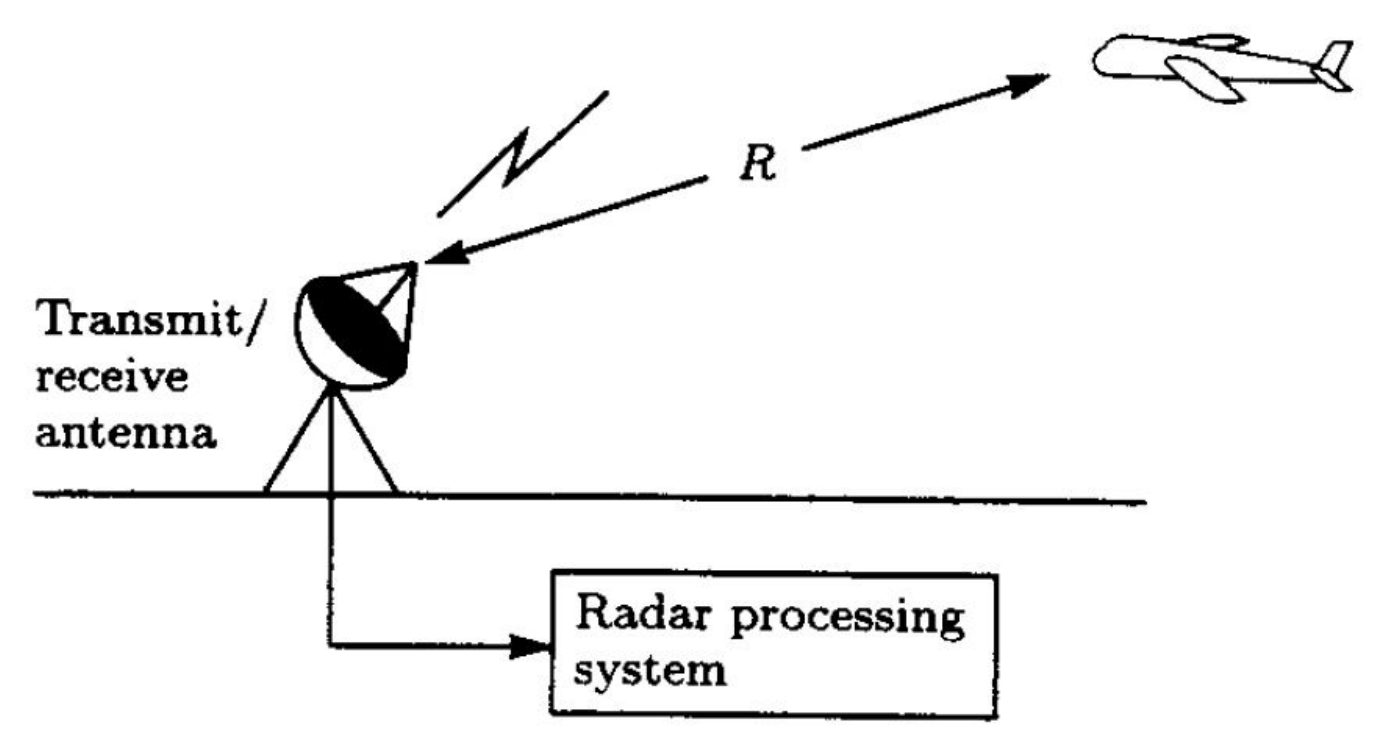}
	\caption{Radar systems allow inference of aircraft presence and position \cite{kay1993fundamentals}}
	\label{fig:radar}
\end{figure}

\begin{figure}[h!]
	\centering
	\includegraphics[width=0.7\linewidth,frame]{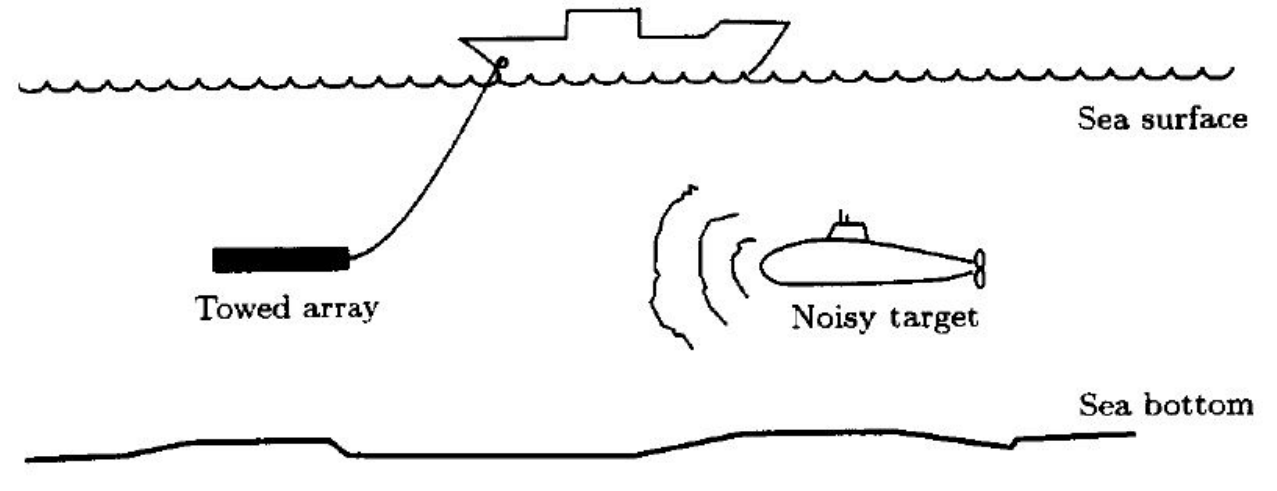}
	\caption{Sonar systems allow inference of underwater targets presence and position \cite{kay1993fundamentals}}
	\label{fig:sonar}
\end{figure}

Figure \ref{fig:radar} shows the basic scheme of how radar-based systems are used to estimate the presence and position of an aircraft. An electromagnetic pulse is sequentially transmitted and the signal at the receiver end is processed for inference. If an aircraft is presented a distorted form of the sent signal would be echoed back to the receiver. The signal can be processed for inference of the presence and range of the target. 

Similarly, Figure \ref{fig:sonar} depicts the basic scheme of how sonar-based systems are used to estimate the presence and position of an underwater target. The target radiates noise due to onboard machinery and propeller action. Sensor arrays can be used to obtain the underwater signals. The signals can be further be used to infer the relevant quantity of interest e.g. bearing, range, etc. 

There remains a constant struggle in military sectors of rivals to design attack systems that can fool adversary defense systems and also devise better protection apparatus to deter advanced forms of attacks.

\subsubsection{Smart Devices and Social Media}

\begin{figure}[h!]
	\centering
	\includegraphics[width=0.7\linewidth,frame]{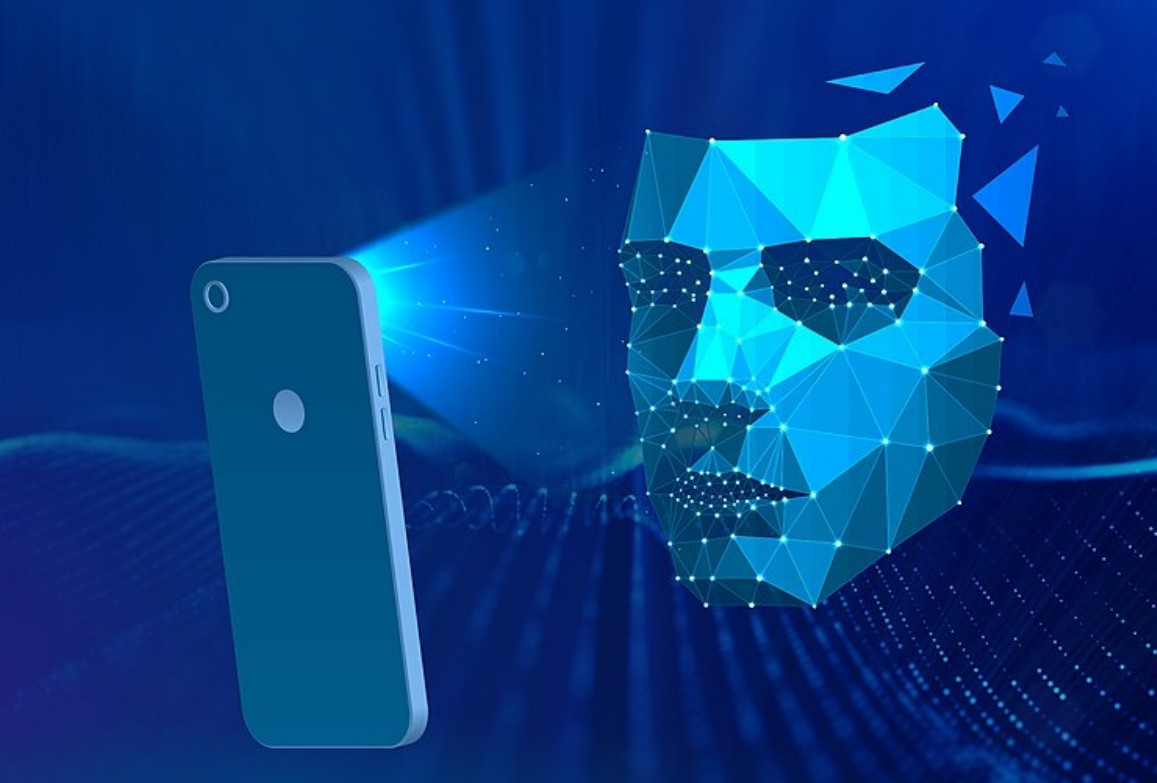}
	\caption{Facial recognition is a common feature in smartphones}
	\label{fig:facialrecognition}
\end{figure}

The world of today teems with smart devices and gadgets like smartphones, tablets, televisions, watches, wearable accessories, etc. Inference is an integral part of smartness in these devices. For example, pattern recognition is an important component in these devices. This can be in form of visual, audio, or haptic technologies. For example, facial (depicted in Figure \ref{fig:facialrecognition}), speech, or thumb impression recognition are now common features of smart devices.

Most of the smart devices can be connected to the internet and have access to social media platforms. Sophisticated inference algorithms are an important part of social media platforms' functionality. For example, these are used for targeted digital advertisement and marketing, search engine optimization, image tagging, data analysis, etc.

There are several other innumerable areas where learning and inference algorithms are now commonly used.

\subsection{Opportunities and Challenges} 

\subsubsection{Opportunities} 
\hspace{.5cm}With technological advances in various fields, there are innumerable opportunities and prospects associated with learning and inference methods. Below we discuss some major factors creating new opportunities.

{{\textbf{Computational power:}}}
There is more computational power available to us than ever before. More sophisticated supercomputers are being built and quantum computing is expected to take the processing power to unprecedented levels. Moreover, affordable computing power is now available for mass usage. This creates to more opportunities for edge computing, easier monitoring of events and phenomena, better automation of mundane and dedicated tasks, and more understanding of the universe, etc.   

\textbf{{Battery technology and energy harvesting:}} 
There has been a significant improvement in battery and energy storage technology in recent times. Even energy harvesting devices have come to life. These advances have resulted in better prospects in terms of the development of more hassle-free devices, reducing installation challenges, and deployment of apparatus in remote and inaccessible locations, etc.

\textbf{{Sensor technology}} 
Sensor technology has also seen remarkable advances in recent times in terms of accuracy, size, and cost. For example, microelectromechanical systems (MEMS) are now commonly available that can be used to sense different quantities of interest like pressure, acceleration, orientation, etc. The improvement in sensor technology has created opportunities like designing more precise and accurate systems, improved monitoring of events, creating a better user experience, and reducing the cost of automation, etc.  

\textbf{{Communication advances and IoT:}}
Like other domains, the field of communication has also witnessed massive advances. 6G, and IoT are common buzz words nowadays. These technologies are necessary to accommodate the widespread IoT devices and cyber-physical system (CPS), sources of a large amount of data, in the network. Newer standards and protocols are constantly emerging for higher data rates, more range, increased capacity, performance in environments with many connected devices, improved power efficiency e.g. WiFi 6, 6LoWPAN, LoRaWAN, Sigfox, etc. 

\textbf{{Open-source hardware and software:}} 
The development of solutions has not been easier than before especially with the advent of open-source hardware and software phenomenon. Moreover, several higher-level languages with different libraries are available that support rapid-prototyping and testing opportunities. In addition, services like one-stop PCB manufacturing are easily available that is making development easier. 

\textbf{{Theoretical advances in applied inference methods:} }
In addition to the enabling technology progress, several applied theoretical advances have also been made in recent history. There have been significant advances like the development of novel machine- and deep- learning architectures and algorithms e.g. generative adversarial networks (GANs) and their derivatives, application of Monte Carlo methods for learning and inference, development of heuristic methods for inference, advancement in statistical signal processing algorithms, etc. These allow us to work at increased effectiveness and devise more efficient learning and inference methods for the future.

\subsubsection{Challenges} 
\hspace{.5cm}Using these enabling technologies we can build an exciting future with smart cities, better healthcare, improved management and governance, more understanding of our biological selves and the universe, etc. However, more opportunities come with increased challenges. We discuss some of the associated challenges as follows. 

\textbf{{Cost:}}
Cost remains the most fundamental challenge in devising any new solution. In any viable project, the minimal cost is sought that meets the set requirement. Otherwise, requirements have to be relaxed in a given allocated budget. 

\textbf{{Algorithms complexity:}}
Another aspect associated with any given computational power is the complexity of the algorithms we are trying to execute. With any given processing power the least complex algorithms are sought. So developing simpler learning and inference algorithms remains a challenge in different fields. 

\textbf{{Real-time requirements:}} 
Some applications demand real-time or online inference of parameters of interest. Therefore, there is limited time within which the inference has to be performed before the next batch of data arrives. To meet this requirement can be challenging and specially designed algorithms are required to this end.

\textbf{{Data corruption and robustness:}}
A common issue with any data usually is the presence of noise and anomalies. This is because of the inherent nature of data, compounded with sensor quality and communication distortions. This poses a significant challenge in terms of developing \textit{robust} learning and inference algorithms.

\textbf{{Cybersecurity:}}
Internet and remote access to physical systems and IoT devices pose a major challenge in terms of cyber attacks and systems reliability. To tackle this challenge cybersecurity measures have to be devised on multiple fronts and at different levels for security.

\textbf{{Single-shot applications:}}
There are some applications which are single- or one-shot meaning there is not plenty of data available or abundant data cannot be extracted offline. Therefore, reliable online learning and inference is required in such applications that can accommodate new unseen scenarios.

\textbf{{Critical applications and contingency:}}
Some applications are highly critical and error tolerance is very low for these. Not only hardware redundancy but analytical redundancy is also required for these application. This means robust inference algorithms are required for these applications which can be challenging to devise.

\vfill
\section{Bayesian Inference}
\begin{figure}[h!]
	\centering
	\includegraphics[width=0.4\linewidth,frame]{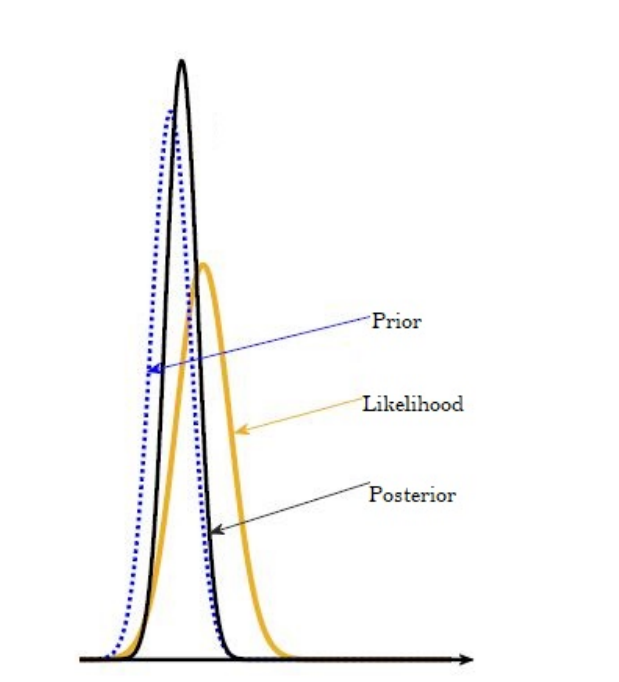}
	\caption[]{Bayesian inference leading to improvement in estimation quality}
	\label{fig:poster}
\end{figure}

\hspace{.5cm} Bayesian inference is a powerful statistical inference approach widely employed in many fields for making key decisions and executing actions accordingly. As the name suggests, the celebrated Bayes theorem is the basic workhorse for general Bayesian inference or estimation. It provides an inherent platform for fusing the \textit{prior} information of the desired parameter and the information from the observations (via the \textit{likelihood}) to obtain an estimate of the parameter of interest. The fusion of information from two sources generally leads to improvement in the estimation quality as depicted in Figure~\ref{fig:poster}.  

Suppose that an unknown parameter ${\theta}$ has a prior distribution $p({\theta})$. Measurements $\mathbf{y}$ (depending on ${\theta}$) are used for estimating ${\theta}$ with an estimator $\hat{{\theta}}=\hat{{\theta}}(\mathbf{y})$. For inference a loss function $C({\epsilon})$ is defined, where ${\epsilon}={\theta}-{\hat{{\theta}}}$ . Further, the Bayes risk $\mathcal{R}$ is defined as the expectation of the loss function 

\begin{flalign}
	\mathcal{R}&=\langle [C({\epsilon})] \rangle_{p({\theta},\mathbf{y})}\label{BE_eq1}\\
	\mathcal{R}&=\int \int [C({\theta}-{\hat{{\theta}}})]\ p({\theta},\mathbf{y})d{\theta}d\mathbf{y}\label{BE_eq2}\\
	\mathcal{R}&=\int \int [C({\theta}-{\hat{{\theta}}})]\ p({\theta}|\mathbf{y})\ d{\theta}\ p(\mathbf{y})d\mathbf{y}\label{BE_eq3}
\end{flalign}  

\begin{figure*}[h!]
	\centering
	\includegraphics[width=0.7\linewidth,frame]{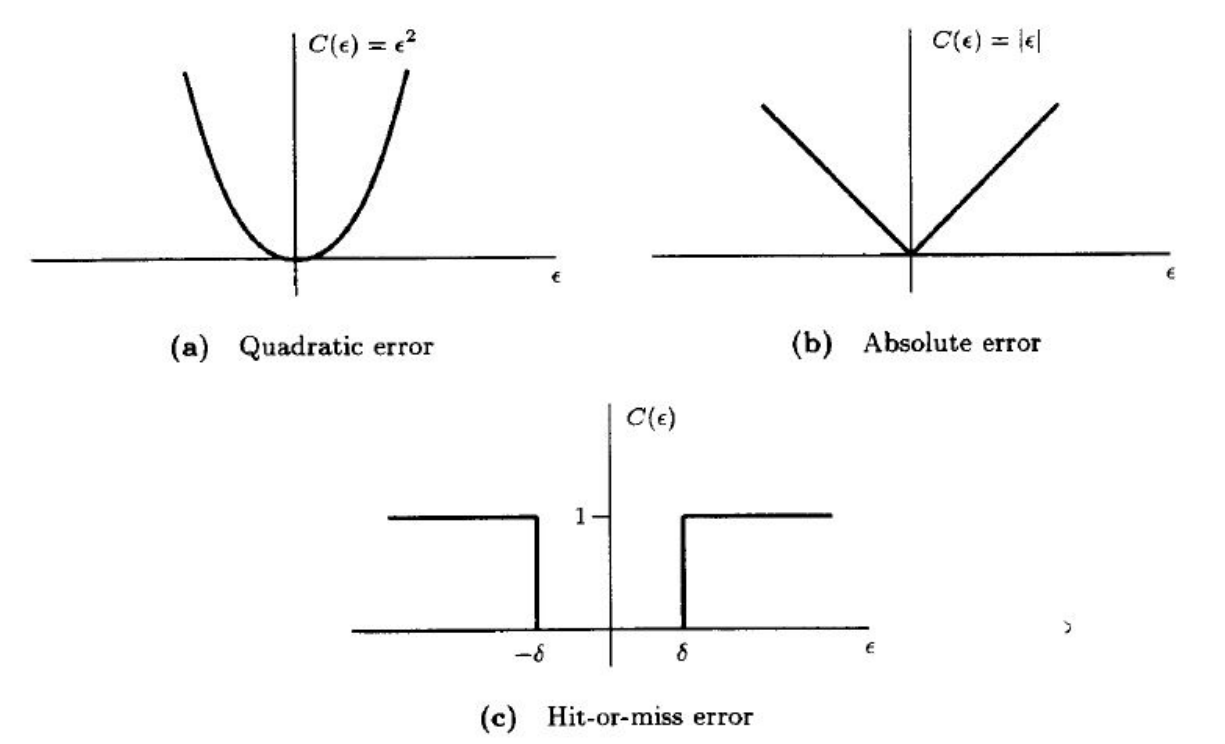}
	\caption{Different cost functions \cite{kay1993fundamentals}}
	\label{fig:lfunctions}
\end{figure*}

$\mathcal{R}$ has to be minimized for Bayesian estimation of $\theta$. Since p($\mathbf{y}$) is non-negative, the inner integral in \eqref{BE_eq3} is minimized, w.r.t $\hat{{\theta}}$, to devise Bayesian estimators. By the selection of different loss functions, various estimators can be derived. Commonly used loss functions include: quadratic error, absolute error and hit-or-miss error depicted in Figure~\ref*{fig:lfunctions}. The loss functions along with the corresponding resulting estimates are summarized in Table~\ref*{tab_BF1}.

\begin{table}[h!]
	\centering
	\caption{Loss functions and resulting estimates \cite{kay1993fundamentals}.\label{tab_BF1}}
	\begin{tabular}{|c|c|}
		\hline
		\textbf{Loss Function} & \textbf{Resulting Estimate }\\
		\hline
		Quadratic Error & Mean of posterior\\
		\hline
		Absolute Error & Median of posterior\\
		\hline
		Hit-or-miss Error & Mode of posterior\\
		\hline
	\end{tabular}
\end{table}

\begin{figure}[ht!]
	\centering
	\includegraphics[width=0.4\linewidth,,frame]{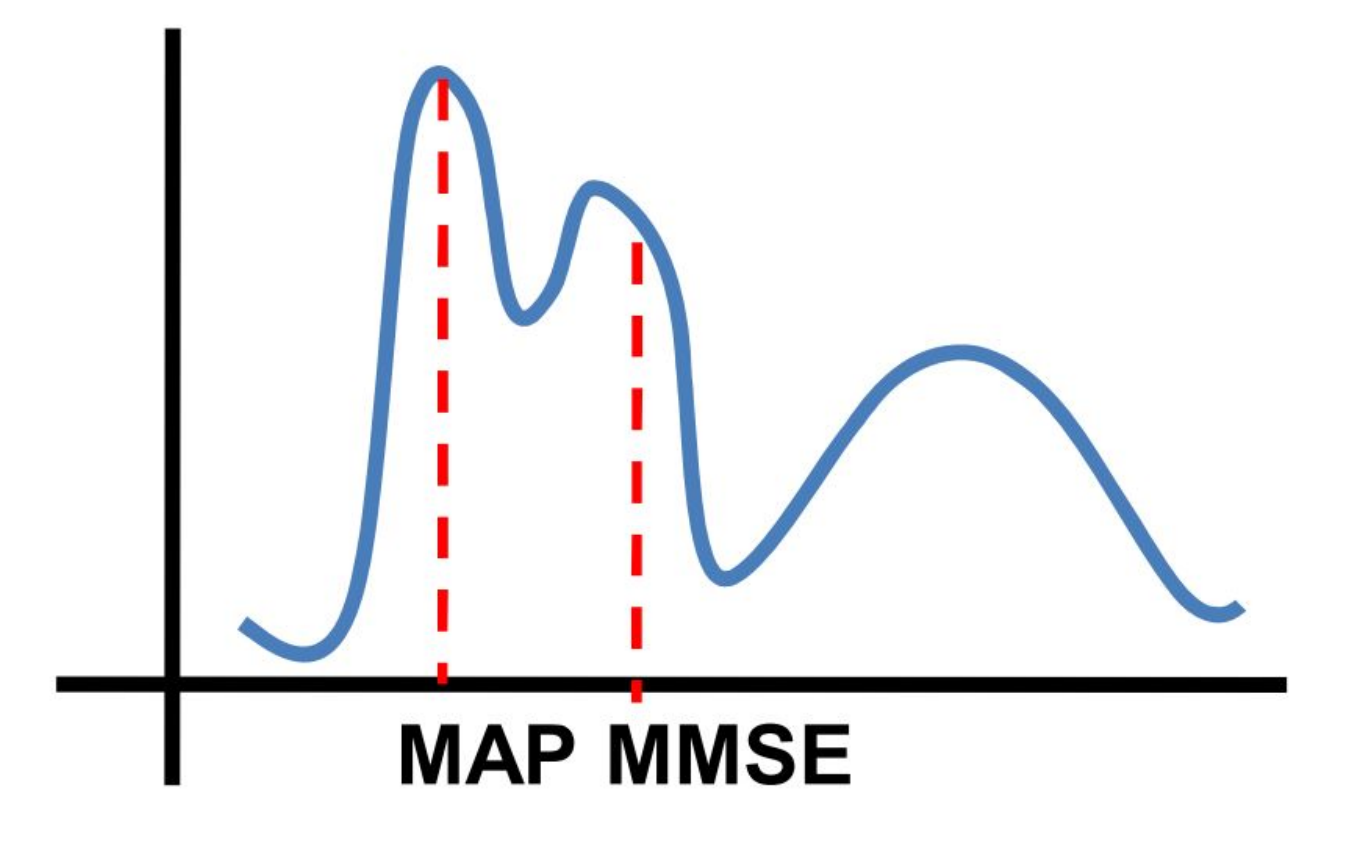}
	\caption{Common Bayesian estimators: MAP and MMSE \cite{rbe_doc}}
	\label{fig:mapmmse}
\end{figure}

The estimator derived using the quadratic loss function is commonly known as the minimum mean squared (MMSE) estimator. It also turns out to be the expected value of the posterior distribution. Furthermore, the estimator derived using the hit-or-miss loss function is referred to as the maximum a posteriori (MAP) estimator. These estimators are depicted in Figure~\ref{fig:mapmmse}

The process can be easily be extended to the case where an unknown parameter vector $\bm{\theta}$ needs to be estimated. In Bayesian inference, the primary objective is to determine the posterior distribution $p(\bm{\theta}|\mathbf{y})$ which is obtained using the Bayes rule as

\begin{equation}
	\overbrace{p(\bm{\theta}|\mathbf{y})}^{\text{Posterior}}= \overbrace{p(\mathbf{y}|\bm{\theta})}^{\text{Likelihood}} \overbrace{p(\bm{\theta})}^{\text{Prior}}/\overbrace{p(\mathbf{y})}^{\text{Normalizing\ factor}} 
	\label{BE_eq4}
\end{equation}

In this dissertation, we mostly resort to Bayesian tools for devising and studying inference approaches.

\section{Dynamical Systems Modeling and Inference}
\hspace{.5cm}The study of dynamical systems is of main interest in many fields owing to their ubiquity. In this dissertation, we keep our attention to state estimation in dynamical systems. We now discuss the state space models and common state estimation terminologies for these models. 
 
\subsection{State Space Models}
\hspace{.5cm}State space models(SSMs) are widely used to describe dynamical systems in diverse applications. These range from areas in economics \cite{ghysels2018applied}, statistics \cite{durbin2012time}, electrical engineering \cite{1100844,7435151}, and neuroscience \cite{doi:10.1162/089976603765202622}. They are popular due to their suitability for modeling diverse real-world scenarios. If we develop a state space model for a physical process, then we can devise and employ different methods to effectively control the underlying process and obtain useful information about the system. 

In an SSM, the system is described by latent states evolving with first-order (Markovian) dynamics related by first-order differential equations or difference equations. States are not directly observable. Rather, they manifest through a set of external outputs, measured by sensors. Most systems in the actual applications are described with continuous-time dynamics. However, presently state estimation and control algorithms are mostly implemented in digital electronics and computers. Therefore, discrete-time SSMs are commonly considered for analysis and design. Continuous-time systems can be satisfactorily transformed to discrete-time counterparts with suitable methods \cite{simon2006optimal}. Therefore, in this work, we restrict ourselves to discrete-time SSMs where the sampling rate of the system is constant.

SSMs are generally categorized into linear and nonlinear models. Although actual systems are mostly nonlinear, mathematical tools available for estimation and control are much more accessible and better understood for linear systems. Therefore, analysis of linear systems serves as the basis for analysis and design for nonlinear systems. Consequently, consideration of both models is important. SSMs are flexible in terms of modeling time-varying or time-invariant systems. However, we focus only on time-invariant systems as these can describe a large variety of actual systems. The results can easily be extended to time-varying systems as we.

\subsubsection{Linear SSMs}
\hspace{.5cm}The state-space model is defined by two equations. First is the state or process equation that captures the time evolution of the process. Second is the observation equation that describes the measurements generated. The actual dynamics of the real system cannot be perfectly accounted for, with \textit{deterministic} modeling, and some amount of uncertainty remains unmodeled. Consequently, the notion of \textit{process} and \textit{measurement} noises has to be introduces. These noise can manifest in the system in many ways, however, we only consider the common case where these noise add in the system dynamics.    

The general form of a discrete-time linear (time-invariant) SSM is given as:
\begin{align}
	\textbf{x}_k &=\textbf{F}\ \textbf{x}_{k-\text{1}}+ \textbf{q}_{k-\text{1}} \label{eqnSSM1}\\
	\textbf{y}_k &=\textbf{H}\ \textbf{x}_{k} + \textbf{r}_k   \label{eqnSSM2}
\end{align}
where  $\textbf{x}_k \in \mathbbm{R}^{n}$  is the state vector at time instant $k$, $\textbf{y}_k \in \mathbbm{R}^{m}$ is the measurement vector, $\textbf{q}_{k}$ describes the process noise vector, and $\textbf{r}_k$ is the measurement noise vector. $\textbf{F}$ and $\textbf{H}$ are known nonlinear functions that define the dynamics of process and observation equations, respectively.
\subsubsection{Nonlinear SSMs}
\hspace{.5cm}Similarly, the general form of a discrete-time nonlinear (time-invariant) SSM is given as:
\begin{align}
	\textbf{x}_k &=\textbf{f}\left(\textbf{x}_{k-\text{1}}\right)+ \textbf{q}_{k-\text{1}} \label{eqnSSM3}\\
	\textbf{y}_k &=\textbf{h}\left(\textbf{x}_{k} \right)+ \textbf{r}_k   \label{eqnSSM4}
\end{align}
where  $\textbf{x}_k \in \mathbbm{R}^{n}$  is the state vector at time instant $k$, $\textbf{y}_k \in \mathbbm{R}^{m}$ is the measurement vector, $\textbf{q}_{k}$ is the process noise vector, and $\textbf{r}_k$ is the measurement noise vector. Also, $\textbf{f}(.)$ and $\textbf{h}(.)$ are known nonlinear functions that define the dynamics of process and observation equations, respectively.

Generally, there can be an \textit{input} (or control) vector term in the process equation in the SSMs, that captures the effect of input excitation to the systems. The primary focus of this work is the \textit{inference} of latent variables and not \textit{control} of systems. Therefore, we generally consider SSMs without any input term in this study.

\begin{figure}[h!]
	\centering
	\includegraphics[width=0.7\linewidth,frame]{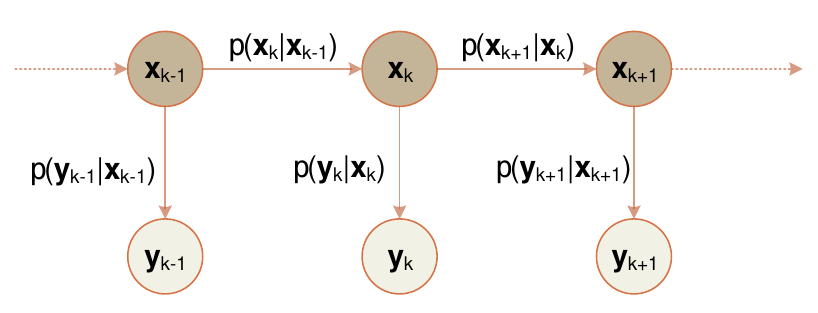}
	\caption{PGM of the structure of an SSM}
	\label{fig:pgmssm}
\end{figure}

SSMs are HMMs which can be viewed as a special probabilistic graphical models (PGMs). The PGM of the structure of an SSM is depicted in Figure~\ref{fig:pgmssm}.

\subsubsection{Noise Properties}
\hspace{.5cm}The noise affecting the system can be either White or colored. We define  these as follows

\textbf{White noise:}
If the noise is uncorrelated across time, it is known as White noise.

\textbf{Colored noise:}
If the noise is correlated across time, it is known as colored noise.

Generally, AWGN settings are assumed for common cases which leads to nice tractable properties in basic algorithms and serve many practical purposes.  

\subsubsection{Controllability and observability}
\hspace{.5cm}The two concepts of \textit{controllability} and \textit{observability} are commonly associated with SSMs defined as

\textbf{{Controllability:}}
The controllability condition means that it is possible to steer the states from an initial value to any desired final value, using valid inputs, within a finite period. 

\textbf{{Observability:}}
Observability is a measure of how well internal states of a system can be inferred by knowledge of its external outputs. 

Controllability and observability are mathematical dual concepts. Controllability and observability can be determined for linear and nonlinear systems using conditions specified with matrix algebra and calculus \cite{anderson2012optimal,hermann1977nonlinear}.

\subsection{Inference on SSMs}
\hspace{.5cm}Having defined an SSM to model a physical process, the primary question that follows is estimating the hidden state variables given the measurements. The presence of process noise and measurement noise makes the problem complicated. For successful state estimation the effect of noise on the estimates needs to be minimized.

\subsubsection{Filtering, Smoothing, and Prediction}
\hspace{.5cm}Formally, three major concepts of inference are relevant to SSMs. Along with filtering, smoothing and prediction are studied as well \cite{sarkka2023bayesian}. Here we explain what the terminologies specifically refer to.   

\textbf{{Filtering:}} Filtering generally refers to the inference of the desired quantities (state or parameters) at time index $k$ using measurements up to time $k$.

\textbf{{Smoothing:}} Smoothing generally refers to the inference of the desired quantity at time index $k$ where measurements derived later than time index $k$ can be used. Since more observations are used higher accuracy than filtering is expected. 

\textbf{{Prediction:}} Prediction as the name suggests refers to the inference of the desired quantity at time index $k$ where only measurements derived earlier than time index $k$ can be used. Prediction usually results in lower accuracy of inference as compared to filtering and smoothing.

In this dissertation, filtering remains our primary focus.

\section{Filtering}

\begin{figure*}[h!]
	\centering
	\includegraphics[width=0.7\linewidth,trim=4 4 4 4,clip,frame]{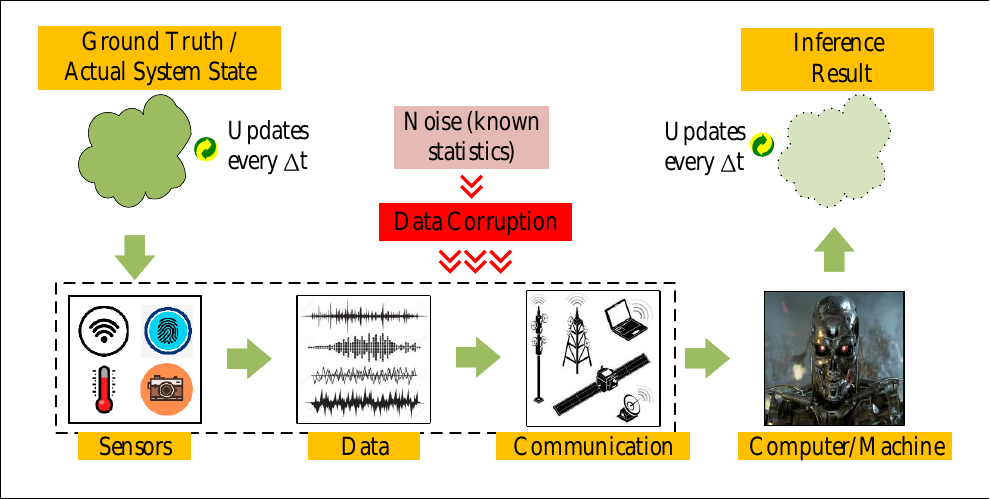}
	\caption{Standard filtering process using machines/computers}
	\label{fig:standardfiltering}
\end{figure*}

\hspace{.5cm}The notion of filtering in electrical engineering has been around for quite a time now. Analog filters have traditionally been employed in the context of filtering out unwanted signals from systems. With the advent of digital circuits and computers, the filtering required for metrological or control purposes has mostly shifted to the digital domain. Various advantages are gained in terms of cost, accuracy, and more adaptability of the digital filters.

It is important to mention that analog filters are not completely obsolete. In fact, these remain crucial where unwanted signals or noise need to be separated from the desired signals in the real (analog) world. For example, wave-trap in power systems is employed on high-voltage (HV) alternating current (AC) transmission power lines to prevent the transmission of high-frequency carrier signals of power line communication to unwanted destinations. 

Digital filters have been classically studied and designed in the frequency domain in general. Using the frequency domain distinction, the noise can be filtered out from the desired signals \cite{oppenheim1975digital}. More recently, the concept of \textit{statistical filtering} originated. The key idea is to exploit the statistical distinction between the desired signals and noise for filtering. 

The earliest ideas of Wiener and Kolmogorov \cite{wiener1950extrapolation,kolmogorov1962interpolation} in statistical filtering relate to stationary processes i.e. processes with unchanging statistical properties over time. Since for these processes, it is possible to relate the statistical properties of the
useful signal and unwanted noise with their frequency domain properties. Thus a conceptual connection between statistical filtering and classical filtering exists. 

The existence of non-stationary signals or noise called for other estimators. Thus filtering evolved explicitly in the time domain. In this regard, Bayesian filtering has been a popular choice \cite{candy2016bayesian,vsmidl2006variational,chen2003bayesian}. For linear dynamical systems with additive White Gaussian noise (AWGN), the Kalman filter (KF) is an optimal recursive estimator in the MMSE sense, resulting from Bayesian filtering \cite{kalman1960new, simon2006optimal}. 

Afterwards, derivatives of KFs followed for nonlinear systems e.g. extended Kalman filters (EKFs) \cite{1099709}. EKFs are based on derivatives of the nonlinear functions and subsequent approximations to transform the problem into a linear one. For nonlinear systems, derivative-free methods were introduced to avoid the problems of differentiation and for better approximation. For example, sigma-point methods exist that approximate the probability distributions using \textit{deterministic} sampling approximating the underlying distributions as Gaussian. Unscented Kalman filter (UKF) is one example of a sigma-point method where unscented transformation is used for filtering \cite{julier2004unscented}. The
unscented transform captures the higher-order moments caused by the nonlinear transform better than Taylor series based EKFs. Similarly, other sigma-point methods like Cubature Kalman filter (CKFs) have also been proposed in the literature. Another class of filtering methods for nonlinear systems, based on \textit{random} sampling of distributions, known as sequential Monte Carlo (SMC) methods or particle filters (PFs) also exist in literature. These are powerful methods demonstrated to work well for nonlinear systems with generic noise statistics (i.e. not limited to Gaussian). However, these methods are computationally intensive. Other concepts of \textit{approximate} inference e.g. VB inference methods - originated from statistical physics - have successfully been applied from filtering to devise lower computationally complicated methods. 

It is worth mentioning that time-domain filtering has been approached from other perspectives also. Major efforts using other tools have been exerted for this e.g. state observers have been extensively studied \cite{1657545}. Similarly, other techniques using optimization and more recently machine/deep learning have also been successfully used for filtering in different applications \cite{morimoto2007reinforcement,7365487}. The standard filtering process is shown in Figure \ref{fig:standardfiltering}. A common requirement for the correct working of standard filtering is the assumption that noise entering into the SSM has known statistics. In this work, we restrict our focus to the Bayesian perspective for devising novel filtering algorithms.  

\subsubsection{Bayesian Filtering}
\hspace{.5cm}In Bayesian filtering or \textit{online} state estimation for SSMs, at each time step $k$, we are interested in the posterior distribution $p(\mathbf{x}_k|\mathbf{y}_{1:k})$ of the state $\mathbf{x}_k$ conditioned on $\mathbf{y}_{1:k}$ which denotes the set of all previous measurements up to time $k$. For the standard SSM, given in \eqref{eqnSSM1}-\eqref{eqnSSM2} for the linear case or \eqref{eqnSSM3}-\eqref{eqnSSM4} for the nonlinear case, the posterior distribution, at each time step $k$,  using Bayes' rule can be expressed as
\begin{equation}
	\overbrace{p(\mathbf{x}_{k}|\mathbf{y}_{1:k})}^{\text{Posterior}}= \overbrace{p(\mathbf{y}_{k}|\mathbf{x}_{k},{\mathbf{y}_{1:{k-1}})}}^{\text{Conditional Likelihood}} \overbrace{p(\mathbf{x}_{k}|\mathbf{y}_{1:{k-1}})}^{\text{Prior}}/\overbrace{p(\mathbf{y}_{k}|\mathbf{y}_{1:{k-1}})}^{\text{Normalizing\ factor}} 
	\label{BF_eq1}
\end{equation} 
which is called the \textit{Update} step. The conditional likelihood $p(\mathbf{y}_{k}|\mathbf{x}_{k},\mathbf{y}_{1:{k-1}})$ can be obtained using the observation model and the previous measurements. $p(\mathbf{y}_{k}|\mathbf{y}_{1:{k-1}})$ denotes the normalizing factor. The predictive distribution $p(\mathbf{x}_{k}|\mathbf{y}_{1:{k-1}})$, which serves as the prior in \eqref{BF_eq1}, can be obtained using the \textit{Chapman-Kolmogorov} equation as
\begin{equation}
	{p(\mathbf{x}_{k}|\mathbf{y}_{1:{k-1}})}=\int p(\mathbf{x}_{k}|\mathbf{x}_{{k-1}},{\mathbf{y}_{1:{k-1}}})p(\mathbf{x}_{k-1}|\mathbf{y}_{{1:k-1}})d\mathbf{x_{k-1}}
	\label{BF_eq2}
\end{equation}
which is called the \textit{Prediction} step. In \eqref{BF_eq2},  the dynamic model and the previous measurements can be used to determine $p(\mathbf{x}_{k}|\mathbf{x}_{k-1},{\mathbf{y}_{1:{k-1}}})$ and $p(\mathbf{x}_{k-1}|\mathbf{y}_{1:k-1})$ is the posterior distribution determined at the previous time $k-1$. The normalizing factor $p(\mathbf{y}_{k}|\mathbf{y}_{1:{k-1}})$ in \eqref{BF_eq1} can be determined using 
\begin{equation}
	p(\mathbf{y}_{k}|\mathbf{y}_{1:{k-1}})=\int p(\mathbf{y}_{k}|\mathbf{x}_{{k}},\mathbf{y}_{1:{k-1}})p(\mathbf{x}_{k}|\mathbf{y}_{{1:k-1}})d\mathbf{x_{k}}
	\label{BF_eq3}
\end{equation}
\hspace{.5cm}Hence, at time step $k$, the posterior distribution $p(\mathbf{x}_{k}|\mathbf{y}_{1:k})$ can be completely determined from the posterior distribution at previous time step $p(\mathbf{x}_{k-1}|\mathbf{y}_{1:k-1})$ recursively, by using the dynamic and observation models and the historical measurements. However, the storage of previous measurements poses a significant challenge for tractable inference. Usually the noise properties are specified or assumed for tractability of the inference process. 
\begin{figure}[h!]
	\centering
	\includegraphics[width=0.9\linewidth,,frame]{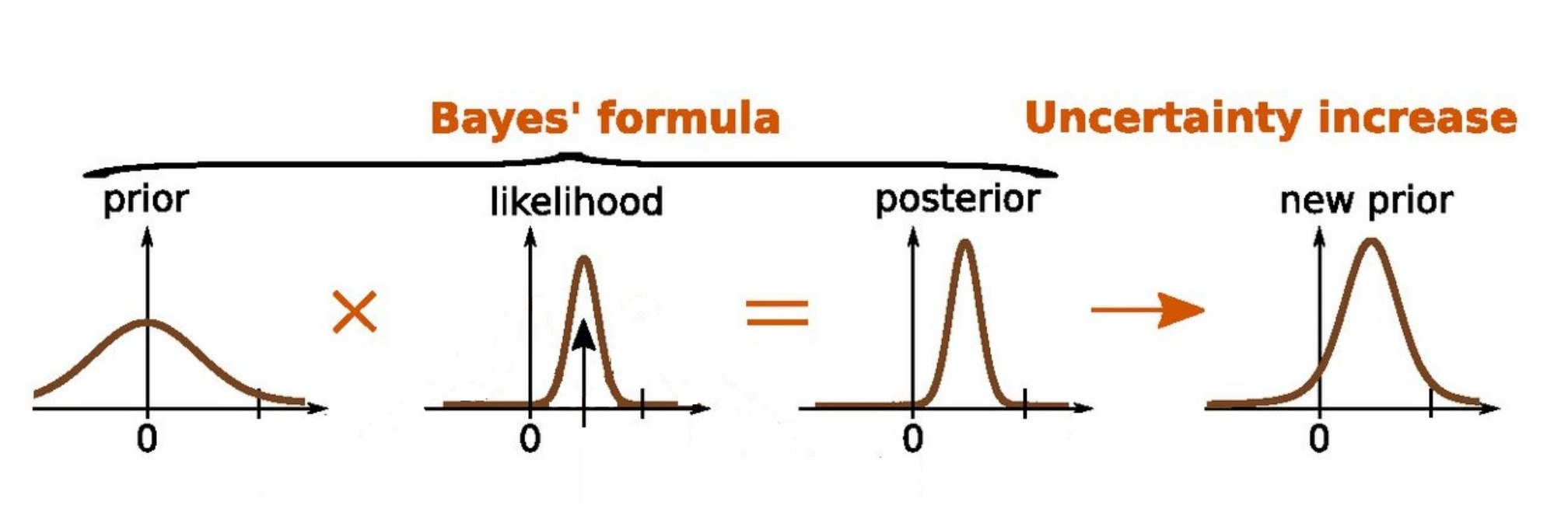}
	\caption{Bayesian Filtering Process \cite{zhou2018chance}}
	\label{fig:rbp}
\end{figure}

The recursive Bayesian filtering process is shown in Figure~\ref{fig:rbp}.


Generally the assumption of independence of noise across time is taken in the filtering algorithms. Also, the process noise and measurement noise are supposed to be statistically independent. 

This leads to the following simplifications in the recursive Bayesian process

\begin{align}
	{p(\mathbf{y}_{k}|\mathbf{x}_{k},{\mathbf{y}_{1:{k-1}})}}&={p(\mathbf{y}_{k}|\mathbf{x}_{k},\stkout{{\mathbf{y}_{1:{k-1}})}}}\label{NP_eq1}\\
	p(\mathbf{x}_{k}|\mathbf{x}_{{k-1}},{\mathbf{y}_{1:{k-1}}})&=p(\mathbf{x}_{k}|\mathbf{x}_{{k-1}},\stkout{\mathbf{y}_{1:{k-1}}})\label{NP_eq2}
\end{align}   

Consequently, the previous measurements do not need to be stored and inference becomes strictly online.

\subsection{Filtering for Linear SSMs}
\hspace{.5cm}For linear SSMs \eqref{eqnSSM1}-\eqref{eqnSSM2}, with certain assumptions on the noise properties \textit{optimal} filters can be devised. Suppose $\textbf{q}_{k}$ is White Gaussian with zero mean and covariance $\textbf{Q}_k$, and $\textbf{r}_k$ is White Gaussian with zero mean and covariance $\textbf{R}_k$. With these assumptions, the (exact) analytical expression of the posterior obtained in terms of the prior and likelihood as all the distributions involved become Gaussian. The moments of the posterior distribution at each time step are recursively updated. The MMSE criterion leads to the celebrated Kalman filter.

\subsubsection{Kalman Filter}
\hspace{.5cm}
Using state and observation models \eqref{eqnSSM1}-\eqref{eqnSSM2} we can write

\begin{align}
	p(\mathbf{x}_{k}|\mathbf{x}_{{k-1}})&=\mathcal{N}(\mathbf{x}_{{k}}|\mathbf{F}\  \mathbf{x}_{{k-1}},\textbf{Q}_{k-1})\label{KF_eq1}\\
	{p(\mathbf{y}_{k}|\mathbf{x}_{k})}&=\mathcal{N}(\mathbf{y}_{k}|\mathbf{H}\ \mathbf{x}_{k},\textbf{R}_k)\label{KF_eq2}
\end{align}  

Suppose $\mathbf{x_{0}}\sim\mathcal{N}(\hat{\mathbf{x}}^{+}_0,\mathbf{P}^{+}_0)$, we can analytically express

\begin{align}
	p(\mathbf{x}_{k}|\mathbf{y}_{{k-1}})&=\mathcal{N}(\mathbf{x}_{{k}}|\hat{\mathbf{x}}^{-}_k,\mathbf{P}^{-}_k)\label{KF_eq3}\\
	{p(\mathbf{x}_{k}|\mathbf{y}_{1:k})}&=\mathcal{N}(\mathbf{x}_{k}|\hat{\mathbf{x}}^{+}_k,\mathbf{P}^{+}_k)\label{KF_eq4}\\
	{p(\mathbf{y}_{k}|\mathbf{y}_{1:k})}&=\mathcal{N}(\mathbf{y}_{k}|\mathbf{H}\ \hat{\mathbf{x}}^{-}_k,\textbf{S}_k)\label{KF_eq5}
\end{align}  
where 
\begin{align}
	\mathcal{N}(\mathbf{x}|\mathbf{m},\mathbf{P})&=\frac{1}{(2 \pi)^{{n} / 2}|\mathbf{P}|^{1 / 2}}\exp \left(-\frac{1}{2}\left(\mathbf{x}-\mathbf{m}\right)^{\top} \mathbf{P}^{-1} \left(\mathbf{x}-\mathbf{m}\right)\right)\label{KF_eq6}
\end{align}
where $n$ denotes the number of elements in $\mathbf{x}$. 

The moments of the Gaussian distributions can be updated recursively using the prediction and update steps.

\textbf{Prediction step}

\begin{align}
	\hat{\mathbf{x}}_{k}^{-}&=\mathbf{F}~\hat{\mathbf{x}}^{+}_{k-1}\label{KF_eq7} \\
	\mathbf{P}_{k}^{-}&=\mathbf{F}\ \mathbf{P}^{+}_{k-1} \mathbf{~F}^{\top}+\mathbf{Q}_{k-1}\label{KF_eq8}
\end{align}

\textbf{Update step}

\begin{align}
	\mathbf{S}_{k} &=\mathbf{H}\ \mathbf{P}_{k}^{-} \mathbf{H}^{\top}+\mathbf{R}_{k}\label{KF_eq10} \\
	\mathbf{K}_{k} &=\mathbf{P}_{k}^{-} \mathbf{H}^{\top} \mathbf{S}_{k}^{-1}\label{KF_eq11} \\
	\hat{\mathbf{x}}^{+}_{k} &=\hat{\mathbf{x}}_{k}^{-}+\mathbf{K}_{k} \big(\mathbf{y}_{k}-\mathbf{H}\ \hat{\mathbf{x}}_{k}^{-}\big) \label{KF_eq12} \\
	\mathbf{P}^{+}_{k} &=\mathbf{P}_{k}^{-}-\mathbf{K}_{k} \mathbf{S}_{k} \mathbf{K}_{k}^{\top}\label{KF_eq13}
\end{align}

The recursion is started from the prior mean $\hat{\mathbf{x}}^{+}_{0}$ and covariance $\mathbf{P}^{+}_{0}$. The MMSE estimate for Kalman filtering, at time step $k$, is $\hat{\mathbf{x}}^{+}_{k}$ with confidence $\mathbf{P}^{+}_{k}$.

\subsection{Filtering for Nonlinear SSMs}\label{ch-2.4}
\hspace{.5cm}Linear SSMs with AWGN lead to nice analytical filtering solutions. However, in nonlinear SSMs it is not possible generally to find the exact analytical solutions thanks to the underlying functional nonlinearities. For these cases, different tractable methods have been reported in the literature.

Consider the nonlinear SSM in \eqref{eqnSSM3}-\eqref{eqnSSM4}. First, we present KF modifications for filtering in this case. For these algorithms, we suppose $\textbf{q}_{k}$ is White Gaussian with zero mean and covariance $\textbf{Q}_k$, and $\textbf{r}_k$ is White Gaussian with zero mean and covariance $\textbf{R}_k$. Then we present a more general method for filtering, based on the Monte Carlo approximation of distributions, i.e. the SMC method or the PF.

\subsubsection{Extended Kalman Filter (EKF)}
\hspace{.5cm}
First, the SSM given in \eqref{eqnSSM3}-\eqref{eqnSSM4} is linearized. Subsequently, the functions $\textbf{ f}(.)$ and $\textbf{h}(.)$ are approximated by the first two terms of their Taylor series expansion. The linearization of the functions  $\textbf{ f}(.)$ in \eqref{eqnSSM3} and $\textbf{h}(.)$  in \eqref{eqnSSM4} around $\bm{\mu}^\text{f}$ and $\bm{\mu}^\text{h}$, respectively, yields the following  model
\begin{align}
	\textbf{x}_k&=\textbf{f}(\bm{\mu}^\text{f})+ \textbf{F}(\bm{\mu}^\text{f}) \left(\textbf{x}_{k-\text{1}} - \bm{\mu}^{\text{f}} \right) + \textbf{q}_{k-\text{1}}  \label{EKF_eq1}\\
	\textbf{y}_k&=\textbf{h}(\bm{\mu}^\text{h})+ \textbf{H}(\bm{\mu}^\text{h}) \left(\textbf{x}_{k} - \bm{\mu}^\text{h} \right) + \textbf{r}_k  \label{EKF_eq2}
\end{align}
where $\textbf{F}(.)$ and $\textbf{H}(.)$ are the Jacobian matrices of $\textbf{ f}(.)$ and $\textbf{h}(.)$ respectively.

Once the SSM is linearized, the filtering equations can be derived using similar recursion steps as in KF. Following the standard Kalman methodology, we assume $\mathbf{x}_{k-\text{1}}$ to be normally distributed for tractable recursive inference. In particular, we assume that $p\left(\mathbf{x}_{k-{1}}|\textbf{y}_{\text{1:}{k-\text{1}}}\right) = \mathcal{N}\left(\textbf{x}_{k-\text{1}}{|}\ \hat{\textbf{x}}^\text{+}_{k-\text{1}} , \textbf{P}^\text{+}_{k-\text{1}} \right)$. Choosing $\bm{\mu}^\text{f} = \hat{\textbf{x}}^\text{+}_{k-\text{1}}$ in \eqref{EKF_eq1} it can be easily verified that $p\left(\textbf{x}_{k}|\textbf{y}_{\text{1:}{k-\text{1}}}\right)$ = $\mathcal{N}\left(\textbf{x}_{k}{|}
\hat{\textbf{x}}^-_{k} , \textbf{P}^-_{k} \right)$ where the parameters of the predictive density are given as

\textbf{Prediction Step}
\begin{align}
	\hat{\textbf{x}}^-_{k} &= \textbf{f}(\hat{\textbf{x}}^+_{k-\text{1}}) \label{EKF_eq3}\\ \textbf{P}^-_{k} &= \textbf{F}(\hat{\textbf{x}}^+_{k-\text{1}}) \textbf{P}^+_{k-\text{1}} \textbf{F}^{\top}(\hat{\textbf{x}}^+_{k-\text{1}}) + \textbf{Q}_{k-\text{1}}\label{EKF_eq4}
\end{align}

Assuming $\bm{\mu}^\text{h} = \hat{\textbf{x}}^{-}_{k}$ in \eqref{EKF_eq2}, we can obtain the parameters of the posterior distribution as follows

\textbf{{Update step}}

\begin{align} 
	\mathbf{S}_{k} &=\mathbf{H}\left(\hat{\mathbf{x}}_{k}^{-}\right) \mathbf{P}_{k}^{-} \mathbf{H}^{\top}\left(\hat{\mathbf{x}}_{k}^{-}\right)+\mathbf{R}_{k} \\ \mathbf{K}_{k} &=\mathbf{P}_{k}^{-} \mathbf{H}^{\top}\left(\hat{\mathbf{x}}_{k}^{-}\right) \mathbf{S}_{k}^{-1} \\ \hat{\mathbf{x}}^{+}_{k} &=\hat{\mathbf{x}}_{k}^{-}+\mathbf{K}_{k} \big( \mathbf{y}_{k}-\mathbf{h}(\hat{\mathbf{x}}_{k}^{-}) \big) \\ \mathbf{P}^{+}_{k} &=\mathbf{P}_{k}^{-}-\mathbf{K}_{k} \mathbf{S}_{k} \mathbf{K}_{k}^{\top} 
\end{align}

\subsubsection{Unscented Kalman Filter (UKF)}
\hspace{.5cm}Another popular nonlinear filtering algorithm is the UKF. The basic idea is that instead of approximating the functions in nonlinear SSMs the densities are approximated (as Gaussian) using deterministic sampling of the distributions. Subsequently, these samples are used in the prediction and update steps of the KF framework. 

Assuming $p\left(\textbf{x}_{k-\text{1}}|\textbf{y}_{\text{1:}{k-\text{1}}}\right) = \mathcal{N}\left(\textbf{x}_{k-\text{1}}|\ \hat{\textbf{x}}^\text{+}_{k-\text{1}} , \textbf{P}^\text{+}_{k-\text{1}} \right)$, the prediction and update steps can be summarized as

\textbf{{Prediction step}}
\begin{itemize}
	\item Form the sigma points as
	\begin{align} \mathcal{X}_{k-1}^{(0)} &=\hat{\mathbf{x}}^{+}_{k-1}\label{UKF_eq1} \\ \mathcal{X}_{k-1}^{(i)} &=\hat{\mathbf{x}}^{+}_{k-1}+\sqrt{{n}+\lambda}\left[\sqrt{\mathbf{P}^{+}_{k-1}}\right]_{i}\label{UKF_eq2} \\ \mathcal{X}_{k-1}^{(i+n)} &=\hat{\mathbf{x}}^{+}_{k-1}-\sqrt{n+\lambda}\left[\sqrt{\mathbf{P}^{+}_{k-1}}\right]_{i}, \quad i=1, \ldots, n \label{UKF_eq3}\end{align}
\end{itemize}
	where $[.]_i$ denotes the $i$th column of the matrix and $\lambda=\alpha^{2}(n+\kappa)-n$. $\lambda$ is a scaling parameter given in terms of the parameters $\alpha$ and $\kappa$ which determine the spread of the sigma points around
	the mean.
\begin{itemize}
	\item Propagate the sigma points through the dynamic model as
	\begin{equation}
		\hat{\mathcal{X}}_{k}^{(i)}=\mathbf{f}\left(\mathcal{X}_{k-1}^{(i)}\right), \quad i=0, \ldots, 2 n \label{UKF_eq4}
	\end{equation}
	\item Compute the predicted mean $\hat{\mathbf{x}}^{-}_{k}$ and the predicted covariance $\mathbf{P}^{-}_{k}$ as
	\begin{align} \hat{\mathbf{x}}_{k}^{-} &=\sum_{i=0}^{2 n} W_{i}^{(\mathrm{m})} \hat{\mathcal{X}}_{k}^{(i)} \label{UKF_eq5}\\ \mathbf{P}_{k}^{-} &=\sum_{i=0}^{2 n} W_{i}^{(\mathrm{c})}\left(\hat{\mathcal{X}}_{k}^{(i)}-\hat{\mathbf{x}}_{k}^{-}\right)\left(\hat{\mathcal{X}}_{k}^{(i)}-\hat{\mathbf{x}}_{k}^{-}\right)^{\top}+\mathbf{Q}_{k-1} \label{UKF_eq6}
	\end{align}
\end{itemize}where
	\begin{align} W_{0}^{(\mathrm{m})} &=\frac{\lambda}{n+\lambda}\label{UKF_eq7} \\ W_{0}^{(\mathrm{c})} &=\frac{\lambda}{n+\lambda}+\left(1-\alpha^{2}+\beta\right)\label{UKF_eq8} \\ W_{i}^{(\mathrm{m})} &=\frac{1}{2(n+\lambda)}, \quad i=1, \ldots, 2 n \label{UKF_eq9}\\ W_{i}^{(\mathrm{c})} &=\frac{1}{2(n+\lambda)}, \quad i=1, \ldots, 2 n \label{UKF_eq10}\end{align}
where $\beta$ is an additional parameter that can be used for incorporating more prior information.

\textbf{{Update step}}
\begin{itemize}
	\item Form the sigma points as
	\begin{align}
		\mathcal{X}_{k}^{-(0)} &=\hat{\mathbf{x}}_{k}^{-}\label{UKF_eq11} \\
		\mathcal{X}_{k}^{-(i)} &=\hat{\mathbf{x}}_{k}^{-}+\sqrt{n+\lambda}\left[\sqrt{\mathbf{P}_{k}^{-}}\right]_{i} \label{UKF_eq12}\\
		\mathcal{X}_{k}^{-(i+n)} &=\hat{\mathbf{x}}_{k}^{-}-\sqrt{n+\lambda}\left[\sqrt{\mathbf{P}_{k}^{-}}\right]_{i}, \quad i=1, \ldots, n \label{UKF_eq13}
	\end{align}
	\item Propagate the sigma points through the measurement model as
	\begin{equation}
		\hat{\mathcal{Y}}_{k}^{(i)}=\mathbf{h}\left(\mathcal{X}_{k}^{-(i)}\right), \quad i=0, \ldots, 2 n \label{UKF_eq14}
	\end{equation}
	\item Compute the predicted mean $\mathbf{\bar{y}}_{k}$, the predicted covariance of the measurement
	$\mathbf{S}_{k}$, and the cross-covariance of the state and the measurement as
	$\mathbf{C}_{k}$
	\begin{align}
		\mathbf{\bar{y}}_{k}&=\sum_{i=0}^{2 n} W_{i}^{(\mathrm{m})} \hat{\mathcal{Y}}_{k}^{(i)} \label{UKF_eq15}\\
		\mathbf{S}_{k}&=\sum_{i=0}^{2 n} W_{i}^{(\mathrm{c})}\left(\hat{\mathcal{Y}}_{k}^{(i)}-\mathbf{\bar{y}}_{k}\right)\left(\hat{\mathcal{Y}}_{k}^{(i)}-\mathbf{\bar{y}}_{k}\right)^{\top}+\mathbf{R}_{k}\label{UKF_eq16}\\
		\mathbf{C}_{k}&=\sum_{i=0}^{2 n} W_{i}^{(\mathrm{c})}\left(\mathcal{X}_{k}^{-(i)}-\hat{\mathbf{x}}_{k}^{-}\right)\left(\hat{\mathcal{Y}}_{k}^{(i)}-\mathbf{\bar{y}}_{k}\right)^{\top}\label{UKF_eq17}
	\end{align}
	\item Finally, compute the filter gain $\mathbf{K}_{k}$, the filtered state mean $\hat{\mathbf{x}}^{+}_{k}$ and the covariance
	$\mathbf{P}^{+}_{k}$ as 
	\begin{align}
		\mathbf{K}_{k} &=\mathbf{C}_{k} \mathbf{S}_{k}^{-1}\label{UKF_eq18} \\
		\hat{\mathbf{x}}^{+}_{k} &=\hat{\mathbf{x}}_{k}^{-}+\mathbf{K}_{k}\left[\mathbf{y}_{k}-\mathbf{\bar{y}}_{k}\right] \label{UKF_eq19}\\
		\mathbf{P}^{+}_{k} &=\mathbf{P}_{k}^{-}-\mathbf{K}_{k} \mathbf{S}_{k} \mathbf{K}_{k}^{\top}\label{UKF_eq20}
	\end{align}
\end{itemize}

\subsubsection{General Gaussian Filtering}
\hspace{.5cm}UKF can be viewed as a special case of Gaussian filter where the underlying densities are assumed to be Gaussian \cite{sarkka2023bayesian}. The distribution at each step are determined using moment matching. The prediction and update steps of the Gaussian filter are as follows

\textbf{{Prediction step}}
\begin{align}
	\hat{\textbf{x}}^{-}_{k}&=\int \mathbf{f}(\mathbf{x}_{k-1})\  {\mathcal{N}}\left(\mathbf{x}_{k-1}|\hat{\textbf{x}}^{+}_{k-1},\mathbf{P}^{+}_{k-1}\right)	d\mathbf{x}_{k-1}\label{GGF_p1}\\
	\mathbf{P}^{-}_{k}&=\int\Big((\mathbf{f}(\mathbf{x}_{k-1})-\hat{\textbf{x}}^{-}_{k})(\mathbf{f}(\mathbf{x}_{k-1})-\hat{\textbf{x}}^{-}_{k})^{\top} {\mathcal{N}}({\mathbf{x}_{k-1}}|\hat{\textbf{x}}^{+}_{k-1},\mathbf{P}^{+}_{k-1})\Big)d\mathbf{x}_{k-1}+\mathbf{Q}_{k-1}\label{GGF_p2}
\end{align}

\textbf{{Update step}}
\begin{align}
	\hat{\textbf{x}}^{+}_k&=\hat{\textbf{x}}^{-}_k+\mathbf{K}_k
	(\mathbf{y}_k-\bm{\mu}_k)	\label{GGF_u1}\\
	\mathbf{P}^{+}_{k}&=\mathbf{P}^{-}_{k}-\mathbf{C}_k\mathbf{K}^{\top}_k \label{GGF_u2}\\
	\mathbf{K}_k&=\mathbf{C}_k (\mathbf{U}_k+\mathbf{R}_k)^{-1}\label{GGF_u3}\\
	\bm{\mu}_k&=\int \mathbf{h}(\mathbf{x}_k)\  {\mathcal{N}}\left(\mathbf{x}_k|\hat{\textbf{x}}^{-}_{k},\mathbf{P}^{-}_{k}\right)	d\mathbf{x}_k\label{GGF_u4}\\
	\mathbf{U}_k&=\int(\mathbf{h}(\mathbf{x}_k)-\bm{\mu}_k)(\mathbf{h}(\mathbf{x}_k)-\bm{\mu}_k)^{\top}{\mathcal{N}}({\mathbf{x}_k}|\hat{\textbf{x}}^{-}_k,\mathbf{P}^{-}_{k})d\mathbf{x}_k\label{GGF_u5}\\
	\mathbf{C}_k&=\int(\mathbf{x}_k-\hat{\textbf{x}}^{-}_{k})(\mathbf{h}(\mathbf{x}_k)-\bm{\mu}_k)^{\top}{\mathcal{N}}({\mathbf{x}_k}|\hat{\textbf{x}}^{-}_{k},\mathbf{P}^{-}_{k})d\mathbf{x}_k\label{GGF_u6}
\end{align}

\subsubsection{Sequential Monte Carlo (SMC) based Particle Filter (PF)}
\begin{figure}[h!]
	\centering
	\includegraphics[width=0.7\linewidth,,frame]{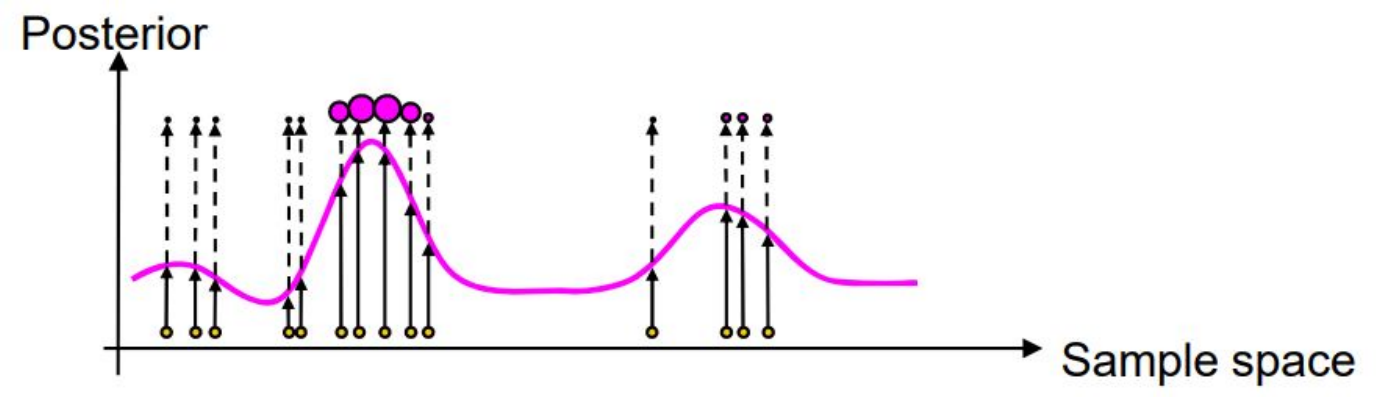}
	\caption{Particle generation in PF \cite{rbe_doc}}
	\label{fig:pf}
\end{figure}

\hspace{.5cm}
In the SMC method for filtering the idea is to approximate the posterior $p(\mathbf{x}_{k}|\mathbf{y}_{1:k})$ by an empirical distribution using Monte Carlo samples as depicted in Figure~\ref{fig:pf}.

However, drawing samples from an arbitrary distribution is
not trivial, hence an importance distribution is sought for
sampling. Conveniently, the importance distribution can be
chosen as $p(\mathbf{x}_{k}|\mathbf{y}_{1:{k-1}})$ which is not necessary in general \cite{tulsyan2016particle}. Consequently, we can empirically represent the required
distribution as

\begin{equation}
	p(\mathbf{x}_{k}|\mathbf{y}_{1:k}) \approx \sum_{l=1}^N w_{k}^{-(l)} \delta(\mathbf{x}_{k}-\hat{\mathbf{x}}_{k}^{-(l)}) \label{SMC_eq1}
\end{equation}
where total $N$ samples $\hat{\mathbf{x}}_{k}^{-(l)}$ are drawn from the importance distribution and the importance weights $w_{k}^{-(l)}$ cater for the mismatch between the target and sampling distribution. 

We can determine the importance weights as 
\begin{align}
	w_{k|{k-1}}^{(l)} \propto& \quad  p(\hat{\mathbf{x}}_{k}^{-(l)}|\mathbf{y}_{1:k})/p(\hat{\mathbf{x}}_{k}^{-(l)}|\mathbf{y}_{1:{k-1}}) \label{SMC_eq2}\\ &=p(\mathbf{y}_{k}|\hat{\mathbf{x}}_{k}^{-(l)})\label{SMC_eq3}
\end{align}
which are subsequently normalized to sum to unity. For recursion, we have
\begin{align}
	p(\mathbf{x}_{k}|\mathbf{y}_{1:{k-1}})=\frac{1}{N}\sum_{l=1}^{N}p(\mathbf{x}_{k}|\hat{\mathbf{x}}^{+(l)}_{{k-1}})\label{SMC_eq4}
\end{align}
where $\hat{\mathbf{x}}_{{k-1}}^{+(l)}$ denote samples drawn from $p(\mathbf{x}_{k-1}|\mathbf{y}_{1:{k-1}})$. Therefore, we  use the system model described by \eqref{eqnSSM3} to propagate $\hat{\mathbf{x}}_{{k-1}}^{+(l)}$ to obtain $\hat{\mathbf{x}}_{{k}}^{-(l)}$. The resulting bootstrap PF is summarized in Algorithm \ref{PF_algo1}. 

\begin{algorithm}[h!]
	\SetAlgoLined
	Generate \{$\hat{\mathbf{x}}^{+(l)}_{0}$\}$^N_{l=1}$ distributed according to the initial state density $p(\mathbf{x_0})$;
	
	\For{$k=1\  to\  K$}{
		Predict \{$\hat{\mathbf{x}}^{-(l)}_{k}$\}$^N_{l=1}$ according to $\hat{\mathbf{x}}^{-(l)}_{k}$\ $\sim p(\mathbf{x}_k$$|\hat{\mathbf{x}}^{+(l)}_{k-1})$, for $l=1,...,N$ using \eqref{eqnSSM3}\;
		Compute importance weights according to  $w^{-(l)}_{k}=\frac{p(\mathbf{y_k}|\hat{\mathbf{x}}^{-(l)}_{k})}{\sum_{l=1}^{N} p(\mathbf{y_k}|\hat{\mathbf{x}}^{-(l)}_{k})}$, for $l=1,...,N$ using \eqref{eqnSSM4}\;
		Compute the estimates as $\mathbf{\hat{x}}^+_{k}=\frac{1}{N}\sum_{l=1}^{N}w^{-(l)}_{k} \hat{\mathbf{x}}^{-(l)}_{k}$\;
		Resample particles as \{$\hat{\mathbf{x}}^{+(l)}_{k}$\}$^N_{l=1}$\; 
	}
	\caption{Bootstrap PF}
	\label{PF_algo1}
\end{algorithm}

\subsubsection{Other Methods for Nonlinear Filtering}
\hspace{.5cm}We have only summarized some of the common methods for nonlinear filtering. Numerous other algorithms have been devised and studied to this end. These include, for example, general Gaussian filtering methods, other (deterministic) sampling-based methods like CKF, several modifications of the PFs, etc \cite{sarkka2023bayesian}.

\section{Robust Filtering}

\begin{figure*}[h!]
	\centering
	\includegraphics[width=0.7\linewidth,trim=4 4 4 4,clip,frame]{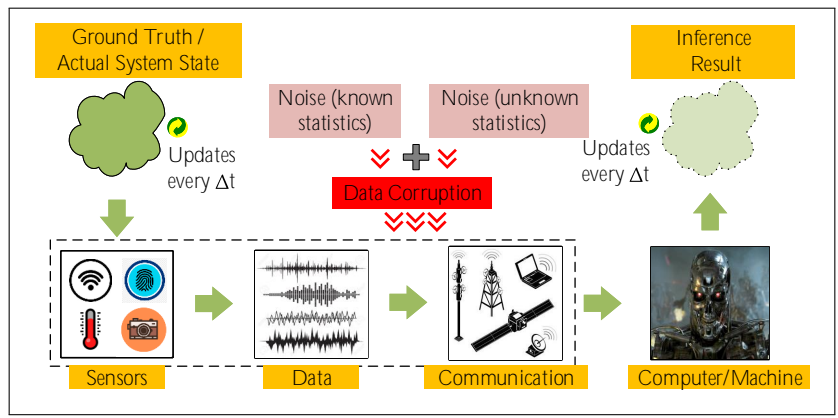}
	\caption{Robust filtering process using machines/computers}
	\label{fig:robustfiltering_Intro}
\end{figure*}

\subsection{Measurement Distortions}
\hspace{.5cm}
In addition to the nominal noise, the occurrence of measurement abnormalities is a common phenomenon in real-world applications that may arise due to reasons such as unexpected disturbances in the environmental conditions, temporary sensor failures, calibration errors, aging, and the inherent quality of sensors. For example, a momentary change in lighting conditions can cause anomalies in visual apparatus data, multi-path reflections can corrupt global positioning system (GPS) and ultra-wide band (UWB) measurements, and glint, speckle, and phase noise can contaminate radar data \cite{ting2007kalman,5979605,6629538,6266757, 8805384,du2015observation}. Similarly, the data from other sensors, like voltage and current sensors, barometer, and sonar, can also be infested with abnormalities \cite{8264764, gustafsson2000adaptive}. The measurement abnormalities are commonly referred to as \textit{gross errors} which majorly include outliers and biases \cite{narasimhan1999data}. 

Outliers are sudden sporadic surges in the measurements whose occurrence can be attributed to factors like sensor degradation, communication failures, environmental influences, front-end processing limitations, etc. \cite{9050910}. Outliers are generally modeled with zero-mean statistics. On the other hand, biases/drifts manifest in the data with non-zero mean noise statistics \cite{9050891}. Multiple factors in various applications lead to the appearance of biased observations. Examples include miscalibrations of sensors, other configuration aberrations like errors in sensor location or alignment, clock errors, or malfunctioning \cite{belfadel2019single}.

The occurrence of abnormalities disrupt the standard filtering process since the apriori knowledge of noise statistics gets compromised. This calls for robust filtering approaches which can sustain the appearance of data abnormalities. A general robust filtering scheme is shown in Figure \ref{fig:robustfiltering_Intro}.

\subsection{Existing Methods}
\hspace{.5cm}Robust filtering theory has been extensively investigated in the literature. Researchers have been interested in various dimensions of the problem including the design and evaluation of filters robust to different effects like anomalies (outliers and biases) in the process and measurement models, missing observations, system modeling errors, adversarial attacks on sensors, etc. Several robust filters have been proposed and evaluated for different applications resorting to techniques from diverse areas. Here, we briefly highlight some other robust filtering methods. 

The classical Wiener filter \cite{wiener1949extrapolation} has been successfully extended to its robust counterparts in the literature. The basic idea is to consider the least favorable power spectral densities (PSDs) for any specific uncertainty model assumed for the signal of interest and the noise. Subsequently, the optimal (using min-max MSE criterion) solution is sought to devise such filters \cite{1056875}. 

Similarly, the KF has numerous robust extensions apart from the ones discussed in the main introduction section. Some of its basic outlier-robust derivatives include $3\sigma$-rejection and score function type KFs \cite{zoubir2018robust}. These formulations use score functions applied to the residuals of the observed and predicted measurements to minimize the effect of outliers. Another robust KF derivative is the approximate conditional mean (ACM) filter which is based on the approximation of the conditional observation density prior to updating \cite{1100882}. Some other variations rely on a bank of KFs to gain robustness \cite{schick1994robust}. The cause of robust filtering has also been well served by the theory of M-estimation which has helped develop many of these methods \cite{zoubir2018robust}. In addition, regression-based KFs have also been proposed for achieving robustness \cite{5371933}. Similarly, guaranteed cost-based methods \cite{317138}, Krein space methods \cite{lee2004robust} and linearly constrained KFs (LCKFs) have also been proposed in this regard \cite{9638328}. Ambiguity sets for catering model distributional uncertainties have also been employed to this end \cite{shafieezadeh2018wasserstein}. Besides different information theoretic criteria have been used to devise robust KFs \cite{chen2017maximum}. KFs extensions to deal with system parametric uncertainties are also well-documented \cite{lewis2017optimal}.

Other types of robust filtering approaches have also been reported in the literature. $\text{H}_{\infty}$ is a popular approach that aims to minimize the worst-case estimation error by formulating a min-max problem using a smartly chosen objective function \cite{simon2006optimal}. Similarly, mixed Kalman/$\text{H}_{\infty}$ approaches have also been proposed leveraging the merits of both the methods \cite{7555348}. The use of finite impulse response (FIR) filters can also be found in this regard \cite{5428832,8355704,8744320}. Similarly, robust recursive estimators for SSMs resorting to sensitivity penalization-based methods have been devised \cite{zhou2010sensitivity}. The use of nonparametric techniques can also be found in the robust filtering literature as well \cite{zoubir2018robust}.

In this work, we focus our attention to mostly the Bayesian perspective of robust filtering approaches. In this regard, we highlight some of the filters for outliers and biases mitigation.

\subsubsection{Outlier Robust Filters}
\hspace{.5cm}Outlier-robust filtering methods can be classified as either compensation-based or rejection based methods.

Compensation-based methods, aim to utilize information from the outlier-ridden observations for inference updates. Some of these techniques assume prior statistics of the measurement noise or the residuals. These include methods based on robust statistics \cite{karlgaard2015nonlinear,7855662,chang2012multiple} and methods based on modeling the observation noise as Student-t or Laplacian distribution \cite{7812899,8009803}. Their performance is effective owing to the sporadic nature of outliers. However, since these methods are based on static loss functions based on design parameters, meticulous tuning of these parameters is required \cite{8869835}. Therefore, tuning-free learning-based compensation approaches have also been advocated in the literature \cite{6657710,du2015observation,6349794,8869835}. These methods assume a distribution to describe the measurement noise and subsequently aim to learn the parameters of the distribution. 

Rejection-based approaches, on the other hand, stem from the argument that generally outliers come from clutter and do not necessarily obey a well-defined distribution. Therefore, corrupted measurements should be completely discarded for state estimation. Traditionally, this is performed by comparing the normalized measurement residuals with some predefined thresholds \cite{6112697,6616007,8245799}. However, the selection of the threshold is mostly subjective. Some theoretical justifications for threshold selection are provided in an extended KF (EKF) based method \cite{mu2015novel}. However, the method is tested only for low outliers frequency and needs memory for past observations. Different learning-based rejection approaches have also been proposed in literature \cite{8398426,9239326}. These techniques aim to learn the parameters that determine whether to use (or discard) the observations for state estimation.

\begin{figure*}[h!]
	\centering
	\begin{subfigure}{.5\textwidth}
		\centering
		\includegraphics[width=.8\linewidth]{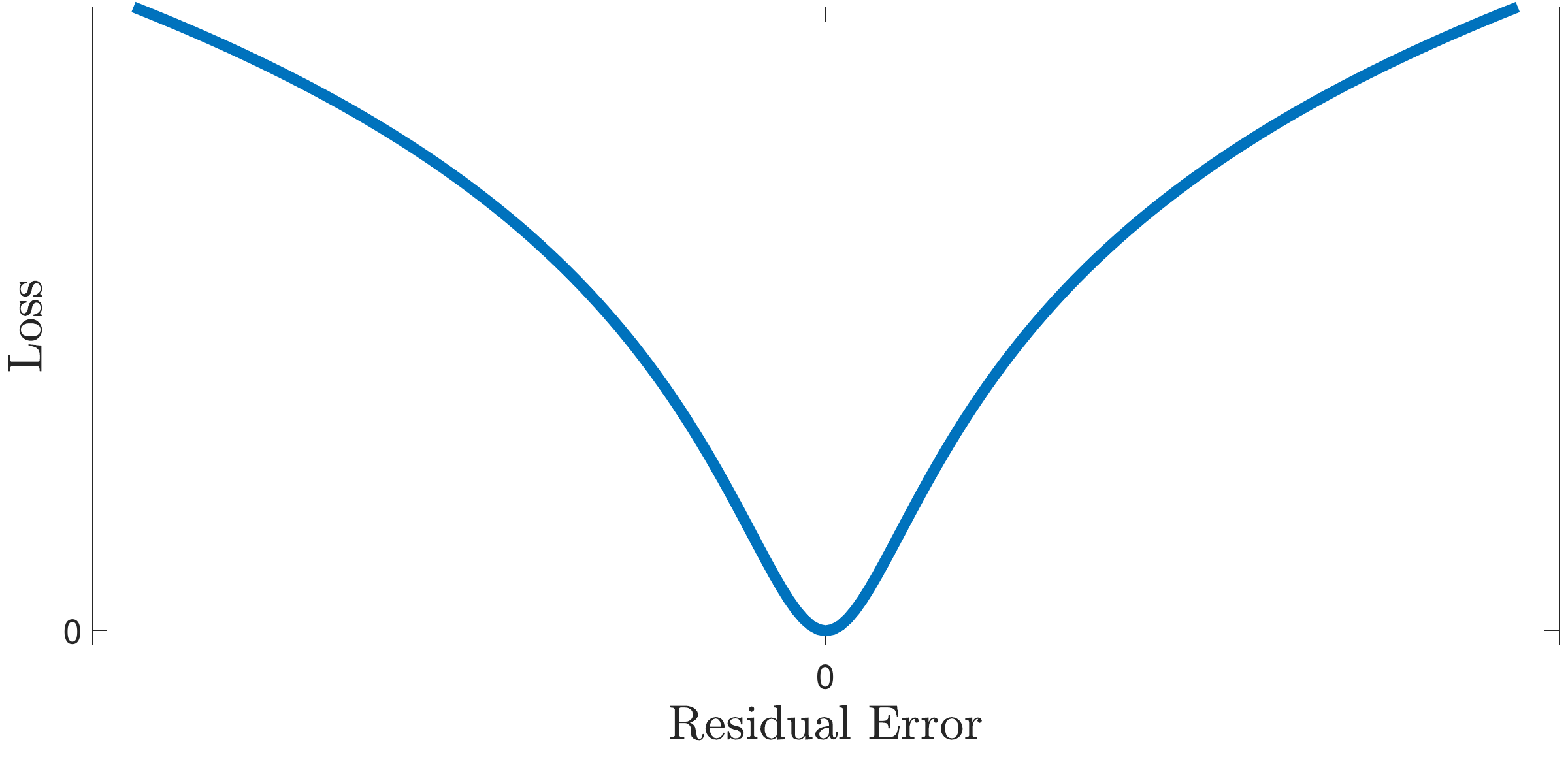}  
		\caption{Static loss functions in traditional methods}
		\label{fig:sub-first}
	\end{subfigure}
	\begin{subfigure}{.5\textwidth}
		\centering
		\includegraphics[width=.8\linewidth]{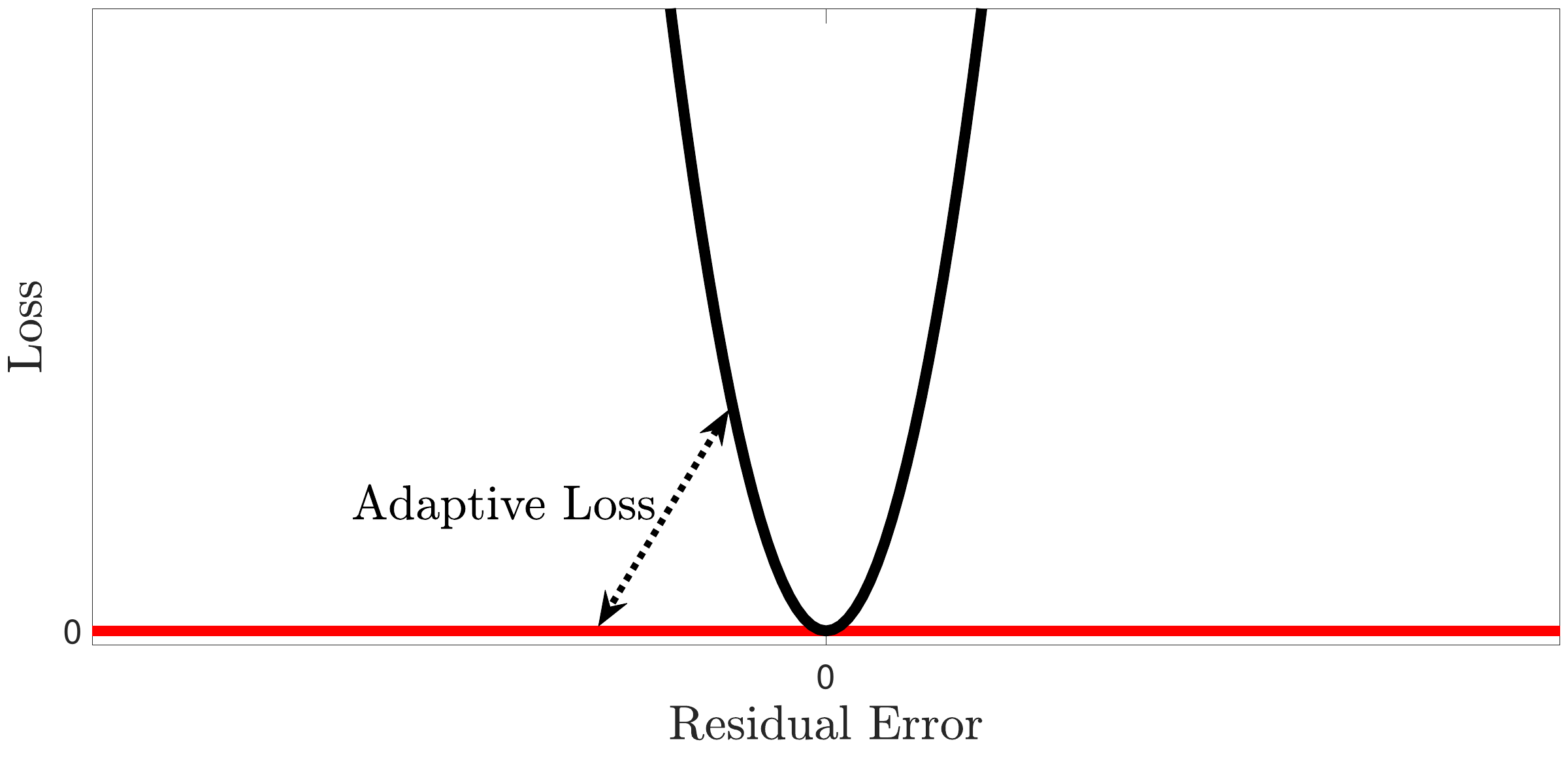}  
		\caption{Adaptive loss functions in learning-based methods}
		\label{fig:sub-second}
	\end{subfigure}
	\caption{Typical loss functions for outlier-robust state estimators. In traditional approaches, the loss function is static. In learning-based methods, the loss function adapts e.g. between a quadratic function and a constant to weight the data during inference.}
	\label{fig:loss}
\end{figure*}

Fig.~\ref{fig:loss} depicts the comparison of typical static loss functions in traditional approaches and dynamic loss functions in learning-based methods considering a uni-dimensional model for visualization \cite{8869835}. Resultingly, learning-based robust state estimators offer more advantages by reducing user input, being more general, and suiting better for one-shot scenarios.

\subsubsection{Bias Robust Filters}\hspace{.5cm}
Biases require separate treatment as compared to the outliers since they exhibit a different kind of behavior. Biases and drifts can sustain for prolonged periods as opposed to outliers that are known to be sporadic. Since the magnitude of biases in the data, the instances of their occurrence, and the particular measurement dimensions which get affected are unknown and only partial statistics describing such corruptions are available, the problem of filtering during their presence is nontrivial. The challenge is further compounded by the functional nonlinearities appearing in the SSMs. Given the significance of dealing with biases in the data during filtering, the topic has historically garnered the attention of various investigators. The approaches for devising bias-robust filters differ in the way such anomalies are neutralized in the filtering process.

Conventionally, biases are catered by assuming that the affected measurement dimensions are known in advance. Moreover, the bias evolution models are assumed to be simplistic or completely ignored during filtering. A straightforward approach is to jointly consider the state vector and bias vector for inference supposing the biases are described in a simple Markovian manner. With computational limitations at that time, earlier works attempted to reduce the processing overhead for such formulations \cite{1099223}. In a similar vein, Schmidt aimed to simplify the joint state and bias estimation in the SKF formulation resulting in the celebrated Schmidt Kalman Filter (SKF) \cite{schmidt1966application}. Interestingly, the bias is not estimated at each time step and only its correlations with the state are updated instead, making the SKF \textit{suboptimal} even if the bias transition can perfectly be modeled \cite{grewal2014kalman}. Ideas similar to the SKF have also been proposed to cater for biased measurements, in terms of exploiting partial information e.g. positivity of biases \cite{filtermobile}. These kinds of \textit{suboptimal} approaches are more useful in scenarios where it is safe to ignore the information regarding the evolution of \textcolor{black}{measurement biases}. With the advances in available processing power, the joint state and parameter estimation approach, in the KF framework, remains the standard go-to approach for catering biased observations \cite{zhang2021extended}.

Conventional methods are more relevant when the bias manifests in the observations consistently throughout the entire duration of system operation. However, these methods inherently assume prior knowledge regarding the affected measurement dimensions. For example, the authors do not consider bias estimation in the angle of arrival (AOA) measurements for filtering \cite{1326705}. Such information may be available beforehand for some applications. However, it must  generally be obtained from some \textit{detection} mechanism for the algorithms to work properly especially if the compromised dimensions vary over time. 

As a result, more sophisticated schemes have been proposed by integrating the detection process with the filtering framework. To this end, following two possibilities exist: 
\begin{enumerate}
	\item Use some external/separate detectors.
	\item Incorporate the detection/compensating process within a unified filtering framework.
\end{enumerate}
Both of these approaches have their merits and drawbacks. External detectors are particularly advantageous in terms of their off-the-shelf accessibility to several options. However, the performance of robust filtering is highly dependent on the functionality of these detectors and the way they integrate. On the other hand, internal detection methodologies are harder to design but obviate dependence on external algorithms.      

In the literature, several kinds of external bias detectors sometimes called fault detection and identification (FDI) algorithms, have been reported for several applications like tracking using UWB, GPS, and UMTS measurements. For example, in \cite{886790}, biased measurements are identified simply by comparison of the standard deviation of range measurements with a detection threshold. A similar approach is to use normalized residuals to detect the presence of any bias \cite{hu2020robust}. The method proposed in \cite{562692} uses a historical record of sequential observations and performs a hypothesis test for detection. For bias detection, the use of classical statistical hypothesis tests like likelihood ratio test (LRT) and other probability ratio tests, has also been documented \cite{686556,7080484,6549130,hu2020robust}. In addition, other methods resort to deep learning for the determination of affected measurements \cite{9108193}. There are several bias compensating filtering methods that rely on these kinds of separate detectors \cite{8805384,filtermobile,1583910}.

The other approach based on inherent bias detection and compensation for filtering is more challenging due to two underlying reasons. First, modeling bias inside the SSM, in a Markovian fashion, is tricky. As noted in \cite{jourdan2005monte} the bias transition cannot be simply modeled as a Gaussian centered at the current bias value. In \cite{jourdan2005monte}, the authors model the bias stochastically by assuming it remains clamped to the previous value with certain predefined probability and jumps with the remaining probability within a set range represented as a uniform distribution. In \cite{gonzalez2009mobile}, a similar model for describing bias is used. These kinds of models describe measurement biases effectively but make the use of KFs variants difficult since the underlying distributions for modeling do not remain Gaussian. Furthermore, the sole use of Gaussian approximations for inference no longer remains suitable. Therefore, the authors in \cite{jourdan2005monte,gonzalez2009mobile,9239326}, opt for the powerful PFs for the inference that can effectively handle arbitrary probability densities. In \cite{9050891} the authors consider the Student's-t-inverse-Wishart distribution to handle time-varying bias for linear systems.

\section{Organization and Contributions}
\begin{figure}[h!]
	\centering
	\includegraphics[width=1\linewidth,trim=2cm 0 5cm 0,clip]{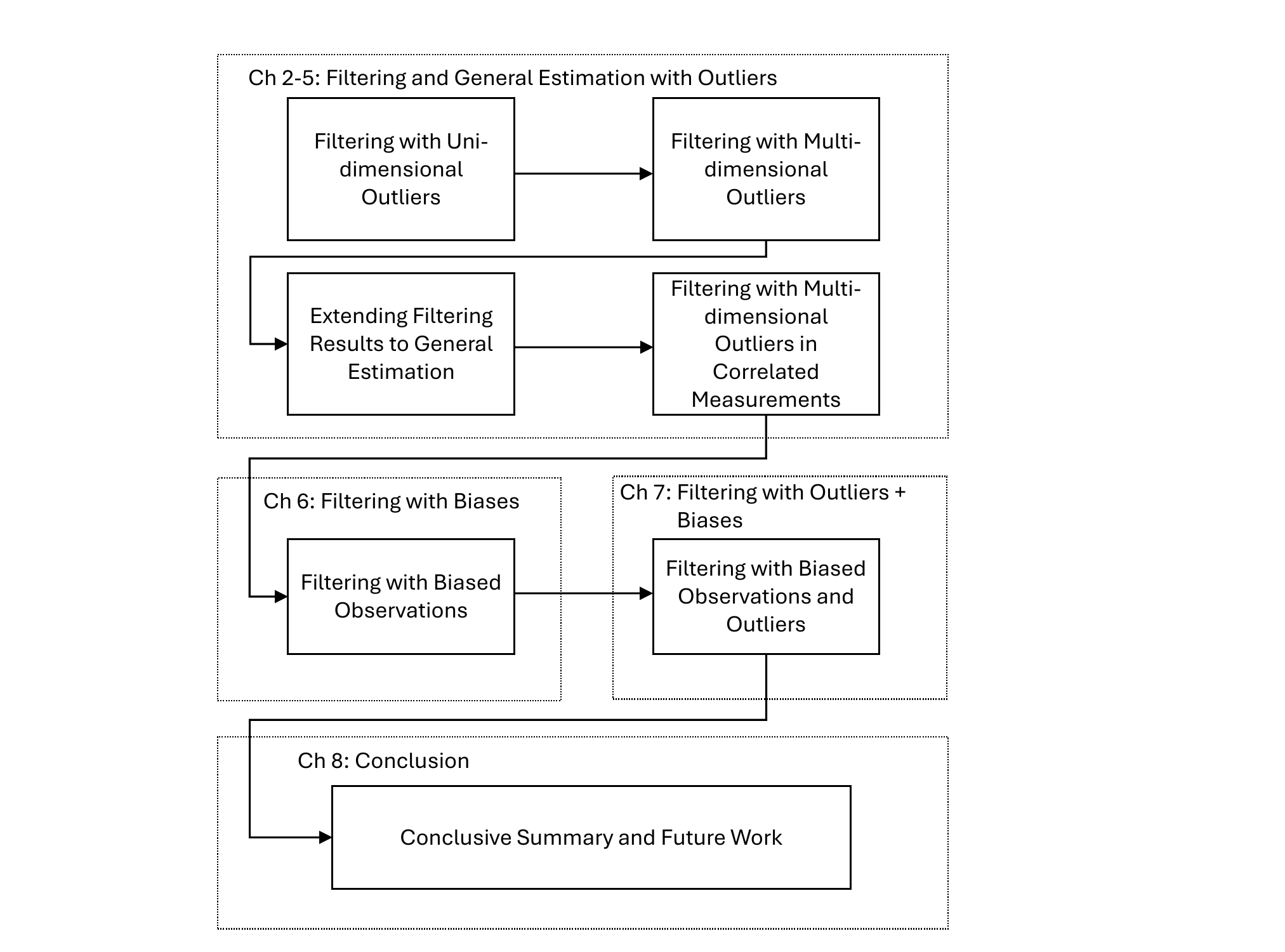}
	\caption{Dissertation Organization Overview}
	\label{fig:organisation}
\end{figure}

\hspace{.5cm}
Considering the limitations of the existing reported techniques in the literature, we devise and investigate robust Bayesian filters and derivative state estimators considering data abnormalities in this work. The organization of the remainder of the dissertation is summarized in Fig.~\ref{fig:organisation}. 

We first deal with outliers progressively in Chapters \ref{chap-3}-\ref{chap-6a}. Subsequently, we take the case of biases in Chapter \ref{chap-6b}. Subsequently, we consider simultaneous data corruption in Chapter \ref{chap-7}. Lastly, we conclude the dissertation and provide future directions in Chapter \ref{concl}.

Contents of the remaining dissertation are discussed as follows
\begin{itemize}
	\item Chapters \ref{chap-3}-\ref{chap-6a}: Filtering with data outliers
	\begin{itemize}
		\item In Chapter \ref{chap-3}, we consider the basic case of robust filtering considering uni-dimensional data corruption. Literature study shows that tractable learning-based outlier-robust filters that deal with outliers have a significant limitation i.e. these assume inherent dependency in measurement anomalies which leads to unnecessary loss of information during filtering. Aiming to address this issue, we propose a robust state estimator in the presence of sensor outliers using a MAP-based EKF. The proposal is tested using suitable target-tracking simulations.
		\item In Chapter \ref{chap-4}, we extend the robust filtering problem assuming outliers' appearance in multi-dimensions. The technique devised in Chapter \ref{chap-3} is limited in application and is not easily scalable to multi-dimensional corruption in large dimensions. Therefore, we resort to the Bayesian tools for approximate inference in particular the VB theory. We design the selective observation rejecting (SOR) filter considering independent measurements and present computational complexity analysis of the proposed and existing methods. Appropriate target-tracking simulations and indoor localization experimentation are used to verify our claims.
		\item In Chapter \ref{chap-5}, we expand the filtering results derived in Chapter \ref{chap-4} to general nonlinear estimation. Specifically, we consider the spatial perception problem and find that the filtering proposals in Chapter \ref{chap-4} are insufficient to tackle the problem. Therefore, we extend our techniques to extended SOR (ESOR) and adaptive SOR (ASOR) in Chapter \ref{chap-4}. Leveraging existing experimental datasets of 3D point cloud registration, mesh registration, and pose graph optimization, we conduct performance evaluations that reveal the merits of our proposals as compared to the state-of-the-art approaches.
		\item In Chapter \ref{chap-6a}, we extend the robust filtering problem assuming outliers arise in multi-dimensions with correlated measurements.  We use the EM theory for this general filtering setting catering to a wide variety of cases resulting in the EM-based outlier-rejecting filter (EMORF). We also present connections with the method in Chapter \ref{chap-4}. Moreover, we also extend the filtering technique to devise a smoothing method called EM-based outlier-rejecting smoother (EMORS), discussing the associated computational costs. In addition, Bayesian Cramer-Rao Bounds (BCRBs) for a filter and a smoother that can perfectly detect and reject outliers are presented which serve as useful theoretical benchmarks to gauge the error performance of different estimators. Lastly, various simulations, for an illustrative target tracking application, are carried out that indicate performance gains compared to the state-of-the-art outlier-rejecting state estimators.	
	\end{itemize}
	\item Chapter \ref{chap-6b}: Filtering with biased-observations
	\begin{itemize}
		\item In Chapter \ref{chap-6b}, we consider the problem of biased observations during filtering. Considering the limited availability of robust filters with biased observations that have a built-in bias detection mechanism, we propose the VB-based bias-detecting and mitigating (BDM) filter with this capability. Using a suitable model we derive the BCRBs to benchmark the performance of the proposed filter along with the comparative methods. Considering a target tracking problem in a wireless network where the range measurements are typically biased, we carry out simulations to back our proposals. 
	\end{itemize}
	
	\item Chapter \ref{chap-7}: Filtering with simultaneous outliers and biases appearance 
	\begin{itemize}
		\item In Chapter \ref{chap-7}, we consider a very general setting with the possibility of both outliers and biases appearing in the observations during filtering. Since the proposed model gets complicated, we resort to the sampling-based PFs for inference. Using simulations and experimental data of target tracking examples, we verify our claims.		
	\end{itemize}
	
	\item Chapter \ref{concl}: Conclusion and future work
	\begin{itemize}
		\item In Chapter \ref{concl}, we provide conclusive remarks for the dissertation. Moreover, we highlight other applications and possible future extensions of the work.   
	\end{itemize}

\end{itemize}
\chapter{Outlier-Robust Filtering Assuming Uni-dimensional Corruption} \label{chap-3}
\hspace{.5cm}
In this chapter, we consider the problem of dynamic state estimation  in scenarios where sensor measurements are corrupted with outliers. We propose a filter that utilizes MAP estimation in the EKF framework to identify and discard the outlier-ridden measurements from a faulty sensor at any given time instant. However, during this process, all those measurements that are not affected by  outliers, are still utilized for state estimation as opposed to several existing tractable methods. Using an illustrative example of dynamic target tracking, we demonstrate the effectiveness of the  proposed estimator.

\section{Problem Formulation}\hspace{.5cm}
Consider the general form of a discrete-time non-linear SSM:
\begin{align}
	\textbf{x}_k &=\textbf{f}\left(\textbf{x}_{k-\text{1}}\right)+ \textbf{q}_{k-\text{1}} \label{eqn1_}\\
	\textbf{y}_k &=\textbf{h}\left(\textbf{x}_{k} \right)+ \textbf{r}_k   \label{eqn2_}
\end{align}
where the subscript $k$ denotes the time index; $\mathbf{x}_k\in \mathbb{R}^n$ and $\textbf{y}_k \in \mathbb{R}^m$ are the state and measurement vectors respectively; the nonlinear functions  $\textbf{f}(.):\mathbb{R}^n\rightarrow\mathbb{R}^n$ and $\textbf{h}(.):\mathbb{R}^n\rightarrow\mathbb{R}^m$  define the process dynamics and observation equations respectively; $\textbf{q}_{k}\in \mathbb{R}^n$ and $\textbf{r}_k\in \mathbb{R}^m$ account for the process and measurement noise respectively. $\textbf{q}_{k}$ and $\textbf{r}_k$ are assumed to be statistically independent, White and normally distributed with zero mean and predetermined covariance matrices $\textbf{Q}_k$ and $\textbf{R}_k$ respectively. 

In this standard nonlinear form, several approaches exist for state estimation as given in the Chapter \ref{ch-1}. However, in the presence of observation outliers, the performance of these schemes degrades drastically. In such a scenario, we modify \eqref{eqn2_} as
\begin{align}
	\textbf{y}_k &=\textbf{h}\left(\textbf{x}_{k} \right)+ \textbf{r}_k +  \textbf{v}_k \label{MAP_EKF_eq1}
\end{align}
where $\textbf{v}_k $ represents the observation outlier noise vector assumed to be White. Now, it is reasonable to assume that $\textbf{v}_k $ is sparse as outliers occur with a very low probability in most practical systems. It is also rare that two or more sensors malfunction simultaneously. Generally there is no correlation in the reading of sensors, especially if these are of inherently different nature or are installed distantly from each other.  This enables us to make the assumption that the outlier can occur at most in one dimension of the measurement vector at any given time instant $k$. These kind of assumptions have been reported in earlier works, for instance authors in \cite{1406131} use similar assumption to deal with sensor biases. Further, we assume that the distribution of outliers in any single dimension is weighted sum of a delta function and a single mode of a normal distribution with zero mean and a very large predefined variance. With these assumptions and using the total law of probability, the distribution of $\textbf{v}_k $  can be written by conditioning it on the occurrence of observation
outliers. Let $\mathcal{I}_k(0)$ denote that the event that there is no outlier, and let $\mathcal{I}_k(i)$ denote the event that an outlier has occurred in the $i$th  dimension at time instant $k$. Given our assumptions, the events $\mathcal{I}_k(i)$ are mutually exclusive and exhaustive. Therefore, using the total law of probability, we arrive at
\begin{equation}
	p(\textbf{v}_k) = { \sum_{i=\text{0}}^{m} p(\textbf{v}_k | \mathcal{I}_k(i) )  p(\mathcal{I}_k(i))} =  \pi^\text{0} \ \delta(\textbf{v}_k) + \sum_{i=\text{1}}^{m} \pi(i) \ \mathcal{N}(\textbf{v}_k|\textbf{0},\bm{V}^i )  \label{MAP_EKF_eq2}
\end{equation}
where $\delta(\textbf{x})$ is a Dirac delta function, $\pi^\text{0}$ and $\pi(i)$ represent the probabilities of events $\mathcal{I}_k(0)$  and $\mathcal{I}_k(i)$  respectively and $\bm{V}^i$ is the covariance matrix of the form $\bm{V}^i=$ $\mathrm{diag}$$( \text{0}, ..,$\\ $\sigma_i^\text{2}, .., \text{0} )$ i.e. its only non-zero entry in the main diagonal is corresponding to $i$th dimension is $\sigma_i^\text{2}$. The notation $\mathcal{N}(\textbf{v}_k|\textbf{\text{0}},\bm{V}^i)$ represents a zero mean multivariate normal distribution with covariance $\bm{V}^i$, evaluated at $\textbf{v}_k$. The model in \eqref{MAP_EKF_eq2} uses the fact that in the absence of outliers,   $\textbf{v}_k $ is a zero vector, however when the outlier occurs in the $i$th dimension, the $i$th entry in the vector $\textbf{v}_k $ is non-zero and can be modeled as a zero mean Gaussian distribution with variance $\sigma_i^{\text{2}}$. Considering that no prior knowledge is available about the outlier statistics, we assume the values of $\sigma_i^{\text{2}}$ to be very large essentially resulting in uninformative priors for the outliers. 

\section{Proposed EKF-Based Estimator}
\subsection{Linearization}\hspace{.5cm}
To develop a Bayesian state-space estimator at each time step $k$, one needs to determine the posterior distribution of the state, conditioned on the set of all the previous observations $\textbf{y}_{\text{1:}k}$, denoted as $p\left(\textbf{x}_{k}| \textbf{y}_{\text{1:}k}\right)$. However, since the problem under consideration is nonlinear, we first linearize the SSM. Towards this objective, we follow the standard EKF approach of approximating the functions $\textbf{ f}(.)$ and $\textbf{h}(.)$ by the first two terms of their Taylor series expansion. The linearization of the functions  $\textbf{ f}(.)$ in \eqref{eqn1_} and $\textbf{h}(.)$  in \eqref{MAP_EKF_eq1} around the vectors $\bm{\zeta}^f$ and $\bm{\zeta}^h$, respectively, yields the following  model
\begin{align}
	\textbf{x}_k&=\textbf{f}(\bm{\zeta}^f)+ \textbf{F}(\bm{\zeta}^f) \left(\textbf{x}_{k-\text{1}} - \bm{\zeta}^f \right) + \textbf{q}_{k-\text{1}}  \label{MAP_EKF_eq3}\\
	\textbf{y}_k&=\textbf{h}(\bm{\zeta}^h)+ \textbf{H}(\bm{\zeta}^h) \left(\textbf{x}_{k} - \bm{\zeta}^h \right) + \textbf{r}_k +  \textbf{v}_k   \label{MAP_EKF_eq4}
\end{align}
where $\textbf{F}(.)$ and $\textbf{H}(.)$ are the Jacobians matrices of $\textbf{ f}(.)$ and $\textbf{h}(.)$ respectively.

\subsection{Prediction Step}\hspace{.5cm}
Given an approximation of the distribution of the state $\textbf{x}_{k-\text{1}}$ given all the previous measurements $\textbf{y}_{\text{1:}{k-\text{1}}}$, the prediction step entails inference of the state $\textbf{x}_{k}$ (without using the observation $\textbf{y}_k$). Following the standard Kalman methodology, we assume $\textbf{x}_{k-\text{1}}$ to be normally distributed for tractable recursive inference. In particular, we assume that $p\left(\textbf{x}_{k-\text{1}}|\textbf{y}_{\text{1:}{k-\text{1}}}\right) = \mathcal{N}\left(\textbf{x}_{k-\text{1}}|\ \hat{\textbf{x}}^\text{+}_{k-\text{1}} , \textbf{P}^\text{+}_{k-\text{1}} \right)$. Choosing $\bm{\zeta}^f = \hat{\textbf{x}}^\text{+}_{k-\text{1}}$ in \eqref{MAP_EKF_eq3} it can be easily verified that $p\left(\textbf{x}_{k}|\textbf{y}_{\text{1:}{k-\text{1}}}\right)$ = $\mathcal{N}\left(\textbf{x}_{k}|
\hat{\textbf{x}}^-_{k} , \textbf{P}^-_{k} \right)$ where $\hat{\textbf{x}}^-_{k} = \textbf{f}(\hat{\textbf{x}}^+_{k-\text{1}})$ and $\textbf{P}^-_{k} = \textbf{F}(\hat{\textbf{x}}^+_{k-\text{1}}) \textbf{P}^+_{k-\text{1}} \textbf{F}^{\top}(\hat{\textbf{x}}^+_{k-\text{1}}) + \textbf{Q}_{k-\text{1}}$.

\subsection{Update Step}\hspace{.5cm}
The final goal is to obtain the posterior density $p\left(\textbf{x}_{k} | \textbf{y}_{\text{1:}k}\right)$. By conditioning on the outlier events, the posterior density can be obtained as
\begin{align}
	p\left(\textbf{x}_k | \textbf{y}_{\text{1:}k} \right)  \propto&\ p\left(\textbf{x}_k , \textbf{y}_{k}|\textbf{y}_{\text{1:}{k-\text{1}}} \right)  =  \sum^{m}_{i=\text{0}} p\left(\textbf{x}_k, \textbf{y}_{k} |\mathcal{I}_k(i) ,\textbf{y}_{\text{1:}{k-\text{1}}} \right) \pi(i) \nonumber  \\
	=&  \underbrace{p\left(\textbf{y}_k|\textbf{x}_k, \mathcal{I}_k(0) \right) p\left(\textbf{x}_k|\textbf{y}_{\text{1:}k-\text{1}} \right)}_{p\left(\textbf{x}_k,\textbf{y}_{k} | \mathcal{I}_k(0) ,\textbf{y}_{\text{1:}{k-\text{1}}}\right)} \pi(0) +  \sum^{m}_{i=\text{1}} \underbrace{p\left(\textbf{y}_k|\textbf{x}_k,\mathcal{I}_k(i) \right) p\left(\textbf{x}_k|\textbf{y}_{\text{1:}k-\text{1}} \right)}_{p\left(\textbf{x}_k,\textbf{y}_{k} |\mathcal{I}_k(i) ,\textbf{y}_{\text{1:}{k-\text{1}}}\right)} \pi(i) \label{MAP_EKF_eq5}
\end{align}
where $ p\left(\textbf{y}_k|\textbf{x}_k, \mathcal{I}_k(0) \right) =  \mathcal{N}\left(\textbf{y}_k| \textbf{h}(\hat{\textbf{x}}^-_{k}) + \textbf{H}\left( \textbf{x}_k - \hat{\textbf{x}}^-_{k}  \right) , \textbf{R}_k \right)$,  $ p\left(\textbf{y}_k|\textbf{x}_k, \mathcal{I}_k(i) \right) $ $=$ \newline $ \mathcal{N}(\textbf{y}_k| \textbf{h}(\hat{\textbf{x}}^-_{k}) + \textbf{H}\left( \textbf{x}_k - \hat{\textbf{x}}^-_{k} \right), \textbf{R}_k + \bm{V}^i)$ assuming $\bm{\zeta}^h=\hat{\textbf{x}}^-_{k}$ in \eqref{MAP_EKF_eq4}.  Using these expressions, we can write $p\left(\textbf{x}_k,\textbf{y}_{k} | \mathcal{I}_k(0),\textbf{y}_{\text{1:}{k-\text{1}}}\right)$ as
\begin{equation}
	p\left(\textbf{x}_k,\textbf{y}_{k} |\mathcal{I}_k(0) ,\textbf{y}_{\text{1:}{k-\text{1}}}\right) = \mathcal{N}
	\left( \begin{pmatrix}\textbf{x}_k\\ \textbf{y}_k \end{pmatrix}| \begin{pmatrix}\hat{\textbf{x}}^{{-}}_k\\ \overline{\textbf{y}}^{\text{0}}_k \end{pmatrix}, \begin{pmatrix} \textbf{P}^-_k & \textbf{C}^{\text{0}}_k \\ {\textbf{C}^{\text{0}}_k}^{\text{T}} & \textbf{S}^{\text{0}}_k \end{pmatrix} 
	\right) \label{MAP_EKF_eq6}
\end{equation}
with $\overline{\textbf{y}}^\text{0}_k = \textbf{h}(\hat{\textbf{x}}_k^-)$, $\textbf{S}^\text{0}_k = \textbf{H}(\hat{\textbf{x}}_k^-) \textbf{P}_k^- \textbf{H}^{\top}(\hat{\textbf{x}}_k^-) + \textbf{R}_k$, and $\textbf{C}^\text{0}_k = \textbf{P}_k^- \textbf{H}^{\top}(\hat{\textbf{x}}_k^-)$. Alternatively, 
\begin{align}\hspace{.5cm}
	p\left(\textbf{x}_k,\textbf{y}_{k} | \mathcal{I}_k(0),\textbf{y}_{\text{1:}{k-\text{1}}}\right) &= p\left(\textbf{x}_k|\textbf{y}_{\text{1:}k} , \mathcal{I}_k(0) \right) p\left(\textbf{y}_k |\mathcal{I}_k(0) ,\textbf{y}_{\text{1:}{k-{1}}}\right)\nonumber\\ &= \mathcal{N}\left(\textbf{x}_k| \hat{\textbf{x}}_k^{{0}} , \textbf{P}_k^{\text{0}} \right) \mathcal{N}\left(\textbf{y}_k| \overline{\textbf{y}}^{\text{0}}_k , \textbf{S}^{{0}}_k \right) \label{MAP_EKF_eq7}
\end{align}
with $\hat{\textbf{x}}_k^\text{0} = \hat{\textbf{x}}_k^- + \textbf{K}_k^\text{0} \left( \textbf{y}_k - \overline{\textbf{y}}^\text{0}_k\right)$, $\textbf{K}_k^\text{0} = \textbf{C}^{0}_k \left( \textbf{S}^\text{0}_k\right)^{-\text{1}}$, and $\textbf{P}_k^\text{0}  = \textbf{P}_k^- + \textbf{K}_k^{0}  \textbf{S}^\text{0}_k {\textbf{K}_k^\text{0}}^{\top}$. Similarly, we can evaluate 
\begin{flalign}\hspace{.7cm}
	&p\left(\textbf{x}_k,\textbf{y}_k |\mathcal{I}_k(i) ,\textbf{y}_{\text{1:}{k-\text{1}}}\right) = \mathcal{N}\left(\textbf{x}_k| \hat{\textbf{x}}_k^i , \textbf{P}_k^i \right) \mathcal{N}\left(\textbf{y}_k| \overline{\textbf{y}}^i_k , \textbf{S}^i_k \right)& \label{MAP_EKF_eq8}
\end{flalign}
with $\hat{\textbf{x}}_k^i = \hat{\textbf{x}}_k^- + \textbf{K}_k^i \left( \textbf{y}_k - \overline{\textbf{y}}^i_k \right)$, $\overline{\textbf{y}}^i_k = \textbf{h}(\hat{\textbf{x}}_k^-)$, $\textbf{K}_k^i = \textbf{C}^i_k \left( \textbf{S}^i_k\right)^{-\text{1}}$, 
$\textbf{C}^i_k = \textbf{P}_k^- \textbf{H}^{\top}(\hat{\textbf{x}}_k^-) $, $\textbf{S}^i_k = \textbf{H}(\hat{\textbf{x}}_k^-)\textbf{P}_k^-   \textbf{H}^{\top}(\hat{\textbf{x}}_k^-) + \textbf{R}_k  + \bm{V}^i$, and $\textbf{P}_k^i  = \textbf{P}_k^- + \textbf{K}_k^i \textbf{S}^i_k {\textbf{K}_k^i}^{\top} $. Now, we can substitute \eqref{MAP_EKF_eq7} and \eqref{MAP_EKF_eq8} into the expression for $p\left(\textbf{x}_k | \textbf{y}_{\text{1:}k} \right)$ in \eqref{MAP_EKF_eq5}  to obtain the final expression for posterior density as
\begin{align}
	p\left(\textbf{x}_k | \textbf{y}_{\text{1:}k} \right)  \propto & \underbrace{\mathcal{N}\left(\textbf{x}_k| \hat{\textbf{x}}_k^\text{0} , \textbf{P}_k^\text{0} \right)  \mathcal{N}\left(\textbf{y}_k| \overline{\textbf{y}}^\text{0}_k , \textbf{S}^\text{0}_k \right)}_{p\left(\textbf{x}_k,\textbf{y}_k |\mathcal{I}_k(0) ,\textbf{y}_{\text{1:}{k-\text{1}}}\right)} \pi(0) + \sum^{m}_{i=\text{1}} \underbrace{\mathcal{N}\left(\textbf{x}_k| \hat{\textbf{x}}_k^i , \textbf{P}_k^i \right) \mathcal{N}\left(\textbf{y}_k| \overline{\textbf{y}}^i_k , \textbf{S}^i_k \right)}_{p\left(\textbf{x}_k,\textbf{y}_k | \mathcal{I}_k(i) ,\textbf{y}_{\text{1:}{k-\text{1}}}\right)} \pi(i)  \label{MAP_EKF_eq9} \\
	p\left(\textbf{x}_k | \textbf{y}_{\text{1:}k} \right)  \propto & \sum^{m}_{i=\text{0}} \mathcal{N}\left(\textbf{x}_k| \hat{\textbf{x}}_k^i , \textbf{P}_k^i \right) \mathcal{N}\left(\textbf{y}_k| \overline{\textbf{y}}^i_k , \textbf{S}^i_k \right) \pi(i) \label{MAP_EKF_eq10}
\end{align}
\subsubsection{State Estimates}\hspace{.5cm}
It is clear from \eqref{MAP_EKF_eq10} that the posterior distribution is multi-modal mixture Gaussian. It is evident that using this multi-modal distribution to compute the posterior at future time indices quickly becomes intractable as the number of modes grow exponentially with time. In order to maintain the tractability of the problem, we attempt to approximate the mixture Gaussian distribution of \eqref{MAP_EKF_eq10} with a uni-modal Gaussian distribution. We choose this distribution as the mode of the mixture distribution with the highest peak. We refer to this estimate as the MAP estimate under the assumption that the modes of the distributions do not have a significant overlap due to large values of outliers. It is obtained by finding the  $j$th mode of \eqref{MAP_EKF_eq10} which yields the maximum point of the distribution given as
\begin{equation}
	j = \underset{i}{\mathrm{argmax}} \ p\left(\textbf{x}_k = \hat{\textbf{x}}_k^i | \textbf{y}_{\text{1:}k} \right) \label{MAP_EKF_eq11}
\end{equation}
which leads to the state estimate as 
\begin{align}
	\hat{\textbf{x}}_k^+ = \hat{\textbf{x}}_k^j \ \ \text{ and} \ \ \textbf{P}_k^+ &= \textbf{P}_k^j \label{MAP_EKF_eq12}\\
	\text{Estimated state:} \ &\hat{\textbf{x}}_k^+ \label{MAP_EKF_eq13}
\end{align}

In the presence of outlier in the \(j\)th dimension of the measurement vector all the terms $\mathcal{N}\left(\textbf{y}_k| \overline{\textbf{y}}^i_k , \textbf{S}^i_k \right)$ in \eqref{MAP_EKF_eq10} become insignificant except for $i=j$. Hence, the dominant mode is $\mathcal{N}\left(\textbf{x}_k| \hat{\textbf{x}}_k^i , \textbf{P}_k^i \right)$ for $i=j$. Therefore, the estimator will incorporate the information only from the outlier-free dimensions of the measurement vector. The matrix $\bm{V}^i$ ensures that the \(i\)th entry in the main diagonal of $\textbf{S}^i_k$ is sufficiently large, so that the corresponding entries in $\left(\textbf{S}^i_k\right)^{-\text{1}}$ and $\textbf{K}_k^i$ are approximately 0. Hence, the outlier-ridden measurement dimension is ignored during the update step. In the absence of any outlier the term $\mathcal{N}\left(\textbf{y}_k| \overline{\textbf{y}}^i_k , \textbf{S}^i_k \right)$ in \eqref{MAP_EKF_eq10} becomes most significant for $i=\text{0}$. Hence, the dominant mode is $\mathcal{N}\left(\textbf{x}_k| \hat{\textbf{x}}_k^{\text{0}} , \textbf{P}_k^{\text{0}} \right)$ and the algorithm works as a standard EKF in the absence of any outlier. 
The pseudo-code of the proposed estimator is summarized in Algorithm \ref{algo_map}.
\vfill
\begin{algorithm}[h!]
	\SetAlgoLined
	Initialize $\hat{\textbf{x}}_{\text{0}}^+, \textbf{P}_{\text{0}}^+, \textbf{Q}_{k-\text{1}}, \textbf{R}_k, \pi(i),  \bm{V}^i \  \forall\  i$\;
	\For{{${k}=1\ \textnormal{to}\ K$, \textnormal{where \textit{K} = Total time samples,}} }{
		{Evaluate $\hat{\textbf{x}}^-_k$ and $\textbf{P}^-_\textbf{k}$ }\;
		{Determine all parameters of $p\left(\textbf{x}_k | \textbf{y}_{\text{1:}k} \right)$ in \eqref{MAP_EKF_eq10}}\;
		{Find $j$ using \eqref{MAP_EKF_eq11}}\;
		{Provide state estimates using \eqref{MAP_EKF_eq12} and \eqref{MAP_EKF_eq13}}	
	}
	\caption{Proposed MAP-based EKF state estimator}
	\label{algo_map}
\end{algorithm}

\section {Simulation Results}\hspace{.5cm}
We evaluate our proposed estimator on the problem of tracking a maneuvering target and compare its performance with that of the linearized version of outlier robust Gaussian Kalman filter (ORGKF) presented in \cite{8398426}. We point out that the linearization of the ORGKF refers to its implementation in the EKF framework rather than the cubature Kalman filter framework as proposed in the original work in order to ensure a fair comparison. The problem of the maneuvering target with an unknown turning rate can be modeled by the following dynamic system equation 
\begin{align}
	\mathbf{x}_k &= \begin{pmatrix} \text{1} & \frac{\text{sin}(\omega_{k}\Delta t)}{\omega_{k}} & \text{0} &  \frac{\text{cos}(\omega_{k}\Delta t)-\text{1}}{\omega_{k}} & \text{0} \\  \text{0} & \text{cos}(\omega_{k}\Delta t) &  \text{0} & -\text{sin}(\omega_{k}\Delta t) & \text{0}\\  \text{0} & \frac{\text{1}-\text{cos}(\omega_{k}\Delta t)}{\omega_{k}} & \text{1} &\frac{\text{sin}(\omega_{k}\Delta t)}{\omega_{k}} & \text{0} \\  \text{0} &  \text{\text{sin}}(\omega_{k}\Delta t) &  \text{0} &  \text{cos}(\omega_{k}\Delta t) & \text{0}\\  \text{0} & \text{0} & \text{0} & \text{0} &\text{1} \end{pmatrix} \mathbf{x}_{k-\text{1}} + \mathbf{q}_{k-\text{1}} \label{MAP_EKF_eq14} 
\end{align}

At any time instant $k$, the state vector $\textbf{x}_k = [a_k ,\dot{{a_k}},  b_k,  \dot{{b_k}},  \omega_{k}]^{\top}$ contains the 2D position, \(({a_k} , {b_k} )\), and velocities, 
\((\dot{{a_k}} , \dot{{b_k}} )\), along with the angular velocity $\omega_{k}$ of the target,  \( \Delta t \) is the sampling time, \(\textbf{q}_{k-\text{1}} \) represents the zero-mean Gaussian process noise with covariance matrix $\textbf{Q}_{k-\text{1}}$. Similarly, the measurement equation for this system gives us the polar coordinates of the target as
\begin{align}
	\textbf{y}_k = \begin{pmatrix} \sqrt{a_k^{\text{2}} + b_k^{\text{2}}} \\ \text{atan2}(b_k,a_k) \end{pmatrix} + \textbf{u}_k \label{MAP_EKF_eq15}
\end{align}
where  $\textbf{u}_k = [
	u_k({1}) , u_k({2})
]^{\top}$ is the simulated noise in the measurements. We evaluate the performance of both methods in terms of time-averaged root mean square error of position $\text{TRMSE}_\text{pos}$ defined as \cite{8398426} 
\begin{align}\hspace{1.5cm}
	\text{TRMSE}_\text{pos} = \frac{\text{1}}{K} \sum_{k=\text{1}}^{K} \sqrt{\frac{\text{1}}{L}  \sum_{l=\text{1}}^L  
		\left((a_k^l-\hat{a}_k^l)^\text{2}+(b_k^l-\hat{b}_k^l)^\text{2}\right)} \label{MAP_EKF_eq16}
\end{align}
where $\hat{a}^l_k$, and $\hat{b}_k^l$  are the estimated coordinates of the target at time instant $k$ for the $l$th MC run, and $L$  represents the total number of runs. We simulate the system using same values for the common parameters  as given in \cite{8398426}. Using $K=100$, and $L=1000$, we carry out the simulations for two different scenarios. Other parameters are set to the values as follows
\begin{align*}
	\pi(0) &= 0.6\\
	 \pi(1) &= 0.2\\
	  \pi(2) &= 0.2\\
	   \bm{V}^1 &= \text{diag}\left(10^{12}, 0\right) \\
	   \bm{V}^2 &= \text{diag}\left(0, 10\pi \right)
\end{align*}


\begin{figure}[t!]
	\centering
	\includegraphics[scale=0.35]{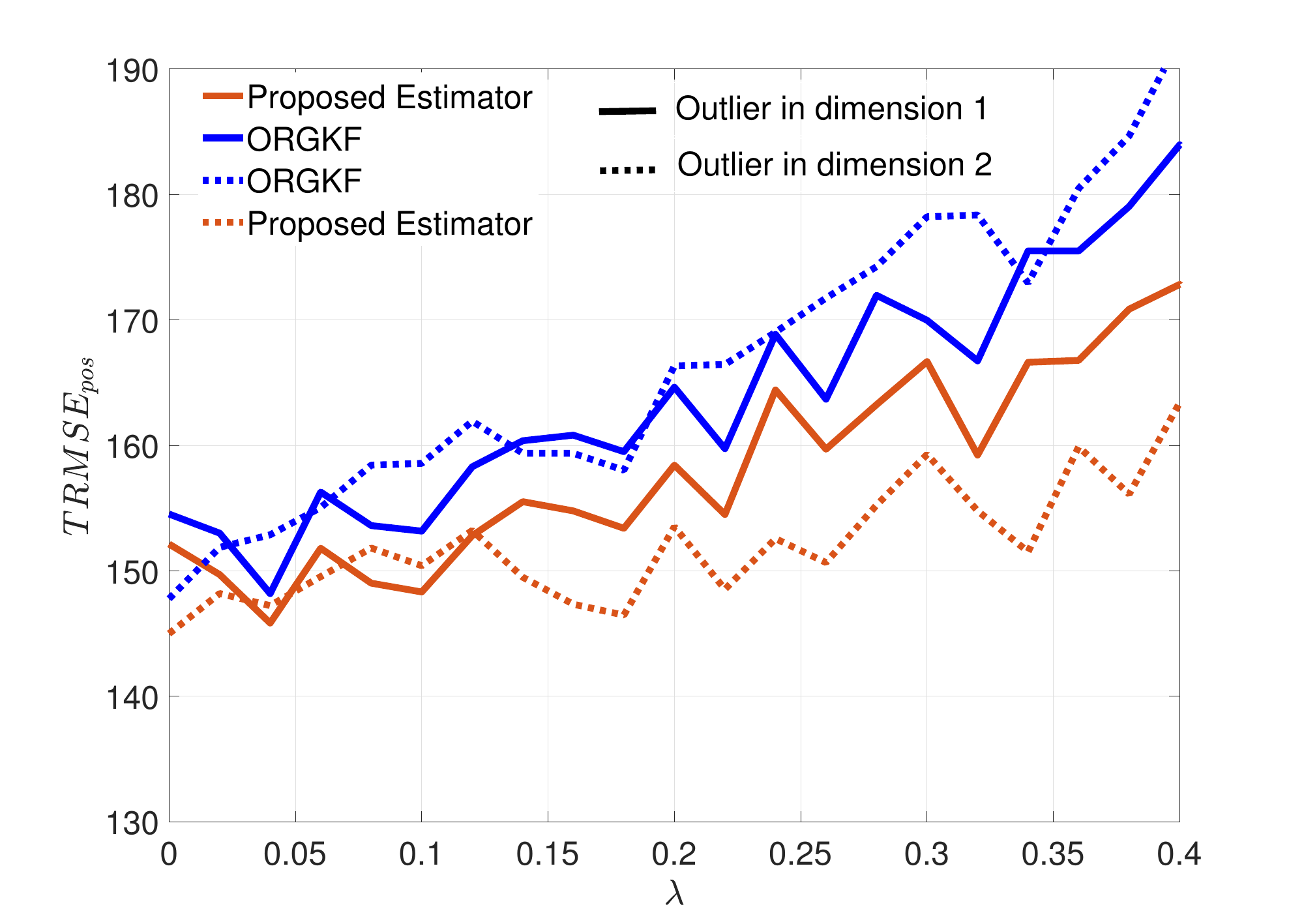}
	\caption{ $\text{TRMSE}_\text{pos}$ vs. $\lambda$ for heavy-tailed outliers depicting the effect of outlier contamination frequency on error of the estimators}
	\label{figI}
\end{figure}
\begin{figure}[t!]
	\centering
	\includegraphics[scale=0.35]{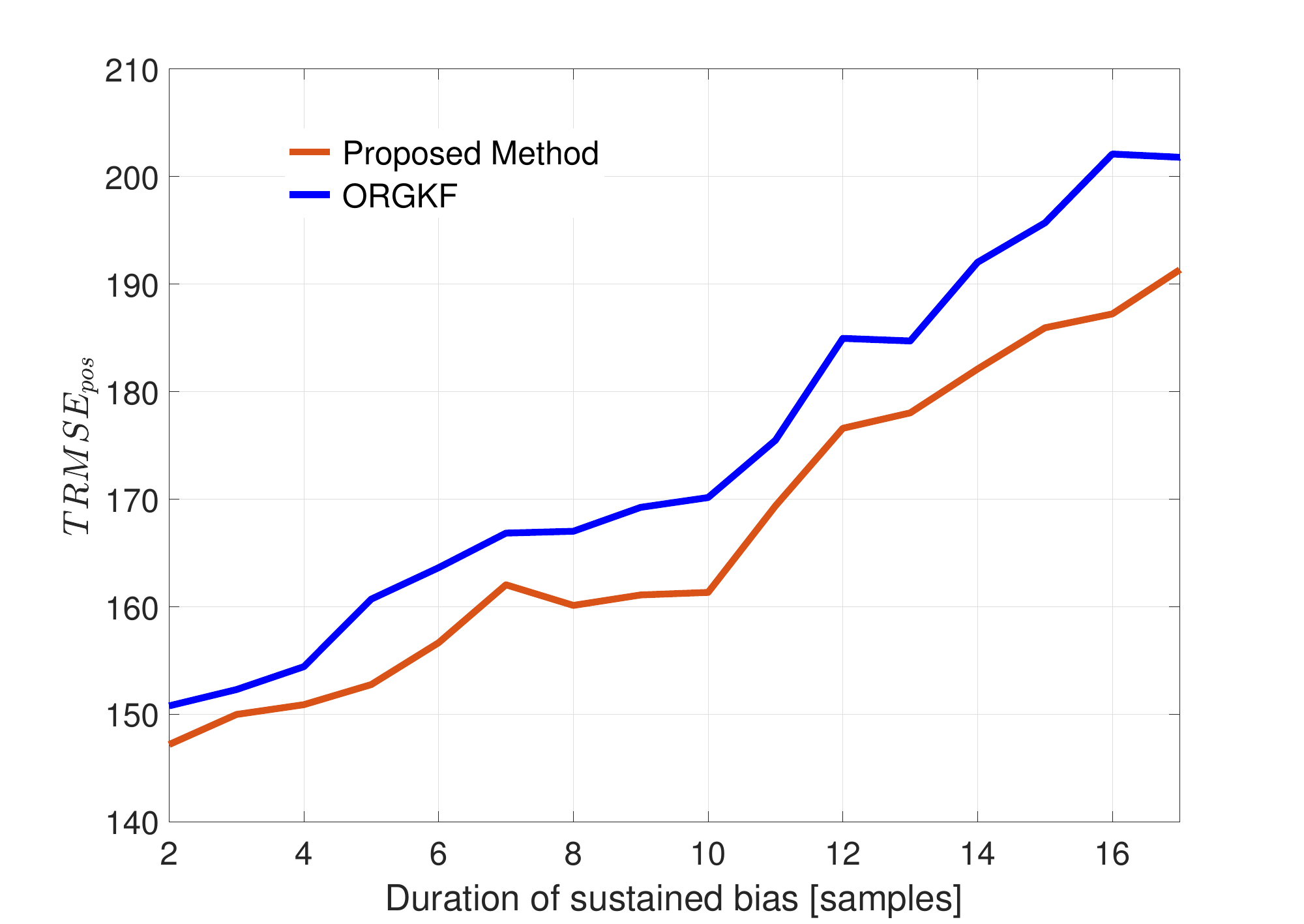}
	\caption{$\text{TRMSE}_\text{pos}$ vs. duration of sustained bias depicting the effect of persisting anomaly on error of the estimators}
	\label{figII}
\end{figure}

In the first scenario, we simulate the heavy-tailed noise by making only one dimension of  $\textbf{u}_k$ contaminated with outliers for the entire duration of one MC simulation run.  The total noise entering into $i$th dimension of  measurements follows the distribution as
\begin{equation*}
	u_k(i) \sim  \left(1 - \lambda\right) \mathcal{N}\left(0, \sigma^2(i) \right) + \lambda \mathcal{N}\left(0, \left(\gamma\sigma(i) \right)^2 \right)
\end{equation*}
where we assume the structure for $\textbf{R}_k = \text{diag}\left(\sigma^2(1) , \sigma^2(2)\right)$. We set  $\gamma = 100$, and vary the contamination ratio $\lambda$ to obtain a plot of $\text{TRMSE}_\text{pos}$. Since, our assumption is all the sensor measurements cannot be corrupted by outliers at any given time, we introduce heavy-tailed outliers into each dimension of the measurement vector separately. The results are shown in Figure \ref{figI} where it can clearly be observed that our estimator significantly outperforms the ORGKF. It can also be noted that our estimator is relatively stable as evident from the increasing gap between the two curves in both graphs at higher contamination ratios.

In the second scenario, the outliers sustain continuously for some duration. For each simulation run, we introduce these sustained biases in both dimensions but at different time instants such that there is no overlap between them at any given time. In these simulations, the bias has a magnitude of 10000 for first dimension  which  starts at time instant $k=26$. Similarly, the bias in the second dimension starts at $k=56$ with a magnitude of $5\pi$. Both of them last for certain duration which in the simulations is set to 16 samples.   Figure \ref{figII} shows the variation in $\text{TRMSE}_\text{pos}$ as we increase the duration of sustained bias. We observe that for all the outlier durations, our proposed estimator provides lower error as compared to ORGKF. This owes to the fact that the proposed MAP estimation is able to utilize the other disturbance-free measurement. Thus, our estimator sustains with more reliability until the pulse terminates, and both measurements become outlier-free again.

\section{Conclusion}\hspace{.5cm}
In this chapter, we have devised a novel estimator that utilizes MAP estimation to incorporate the error-free measurements at a given time instant inside an EKF framework. We demonstrate that the proposed estimator outperforms the linearized version of the state-of-the-art ORGKF, which discards all the measurements at a given time instant if one of them is contaminated by an outlier. It is the incorporation of the error-free measurements for inference that gives our proposed estimator an edge in cases where the outliers are unlikely to occur simultaneously in two or more measurements.  

\chapter{Outlier-Robust Filtering Considering Multi-dimensional Corruption in Independent Measurements} \label{chap-4}
\hspace{.5cm}
In Chapter \ref{chap-3}, we devised a state estimator that considered the possibility outliers occurrence in one dimension at a time. For some cases it is a useful estimator where we can assume the corruption of one measurement dimension at a point in time. The robust filter cannot be simply scaled to the possibility of occurrence of outliers in multiple dimensions owing to the combinatorial nature of the solution resulting in complexity issues. Specifically, to generally scale the proposed filter to $m$-dimensional measurement vector, we need to run $2^m$ Kalman filters resulting in computational complexity of $2^m m^3$ for the algorithm which is practical not feasible. Therefore, to get around the issue and devise learning-based filters robust to multi-dimensional data outliers, we utilize the tools from approximate inference. 

In this chapter, we present a novel outlier-robust filter for nonlinear dynamical systems considering a common case where measurements are obtained from independent sensors or measurement channels. The proposed method is devised by modifying the measurement model and subsequently using the theory of variational Bayes (VB) theory and general Gaussian filtering. We treat the measurement outliers independently for independent observations leading to selective rejection of the corrupted data during inference. By carrying out simulations for variable number of sensors we verify that an implementation of the proposed filter is computationally more efficient as compared to the proposed modifications of similar baseline methods still yielding similar estimation quality. In addition, experimentation results for various real-time indoor localization scenarios using UWB sensors demonstrate the practical utility of the proposed method.

\section{Modeling Details}\label{Sec_model}
\hspace{.5cm}
We consider a nonlinear discrete-time SSM of a dynamic physical process mathematically described by
\begin{align}
	\mathbf{x}_k&= \mathbf{f}(\mathbf{x}_{k-1})+\mathbf{q}_{k-1}	
	\label{eqn_model_1}\\
	\mathbf{y}_k&= \mathbf{h}(\mathbf{x}_{k})+\mathbf{r}_{k}
	\label{eqn_model_2}
\end{align}
where the subscript $k$ denotes the time index; $\mathbf{x}_k\in \mathbb{R}^n$ and $\textbf{y}_k \in \mathbb{R}^m$ are the state and measurement vectors respectively; the nonlinear functions  $\textbf{f}(.):\mathbb{R}^n\rightarrow\mathbb{R}^n$ and $\textbf{h}(.):\mathbb{R}^n\rightarrow\mathbb{R}^m$  define the process dynamics and observation equations respectively; $\textbf{q}_{k}\in \mathbb{R}^n$ and $\textbf{r}_k\in \mathbb{R}^m$ account for the process and measurement noise respectively. $\textbf{q}_{k}$ and $\textbf{r}_k$ are assumed to be statistically independent, White and normally distributed with zero mean and predetermined covariance matrices $\textbf{Q}_k$ and $\textbf{R}_k$ respectively. 


Practically, the observations from different sensors can be corrupted with outliers. This leads to the failure of conventional filtering based on the model \eqref{eqn_model_1}-\eqref{eqn_model_2}. In the following, we consider that observations are obtained from independent sensors, therefore, we model the outliers independently for each dimension. To mitigate the effect of outliers on the state estimation quality, we introduce an indicator vector  $\bm{\mathcal{I}}_k\in\mathbb{R}^m$ with Bernoulli elements. In particular, ${{\mathcal{I}}}_k(i)$ can assume two possible values $\epsilon$ (close to zero) and 1.  ${{\mathcal{I}}}_k(i)=\epsilon$ indicates the occurrence of an outlier in the corresponding dimension at time $k$. Since an outlier can occur independently at any instant, irrespective of the past and outliers in other dimensions, we assume that the elements of $\bm{{\mathcal{I}}}_k$ are statistically independent of each other and their history. Moreover, $\bm{\mathcal{I}}_k$ and $\mathbf{x}_k$ are also considered independent. Using $\theta_k(i)$ to denote the probability of no outlier in the $i$th observation, the distribution of $\bm{\mathcal{I}}_k$ is expressed as
\begin{equation}
	p(\bm{\mathcal{I}}_k)=\prod_{i=1}^{m}p({{\mathcal{I}}}_k(i))=\prod_{i=1}^{m} (1-{\theta_k(i)}) \delta({{{\mathcal{I}}}_k(i)}-\epsilon)+{\theta_k(i)}\delta( {{{\mathcal{I}}}_k(i)}-1)
	\label{eqn_model_3}
\end{equation}

Furthermore, the measurement likelihood conditioned on the current state $\mathbf{x}_k$ and the indicator $\bm{\mathcal{I}}_k$, independent of all the historical observations $\mathbf{y}_{1:{k-1}}$, is proposed to follow a Gaussian distribution
\begin{align}
	p(\mathbf{y}_k|\mathbf{x}_k,\bm{\mathcal{I}}_k)&={\mathcal{N}}\Big(\mathbf{y}_k|\mathbf{h}(\mathbf{x}_k),\bm{\Sigma}_k^{-1}\Big)\nonumber\\&=\frac{1}{\sqrt{(2 \pi)^{m}|\bm{\Sigma}_k^{-1}|}}\mathrm{exp}\big({-}\mfrac{1}{2}{(\mathbf{y}_k-\mathbf{h}(\mathbf{x}_k))}^{\top}\bm{\Sigma}_k(\mathbf{y}_k-\mathbf{h}(\mathbf{x}_k))\big)	
	\label{eqn_model_4}\\
	&=\prod_{i=1}^{m}\frac{1}{\sqrt{2 \pi {\mathrm{R}_k(i,i)}/{{\mathcal{I}}}_k(i)}}\ \mathrm{exp}\left( {-}\frac{{(\mathrm{y}_k(i)-\mathrm{h}(\mathbf{x}_k)(i))}^{2}}{2\mathrm{R}_k(i,i)}{{\mathcal{I}}}_k(i)\right) 
\end{align}
where $\bm{\Sigma}_k={\textbf{R}^{-1}_k}{diag}(\bm{\mathcal{I}}_k)$.  We assume statistical independence in the nominal noise adding in each of the measurement dimension as commonly observed especially in cases where the sensors are deployed independently. Therefore, $\textbf{R}_k$ is assumed to be diagonal and we are able to express the distribution as a product of univariate Gaussian distributions.

{Considering \eqref{eqn_model_3} and \eqref{eqn_model_4}, the modified measurement model incorporating the effect of outliers can be expressed as
	\begin{align}
		\mathbf{y}_k&= \mathbf{h}(\mathbf{x}_{k})+\bm{\nu}_{k}
		\label{eqn_model_2-m}
	\end{align}
	where the modified measurement noise assumes a Gaussian mixture model as
	\begin{align}
		\bm{\nu}_{k}\sim \underset{{\bm{\mathcal{I}}_k}}{\sum} \ {\mathcal{N}}(\bm{\nu}_k|\mathbf{0},\bm{\Sigma}_k^{-1}) p(\bm{\mathcal{I}}_k) \nonumber
	\end{align}

\begin{figure}[h!]
	\centering
	\includegraphics[trim={.5cm .5cm .5cm 0.7cm},clip,width=.6\linewidth]{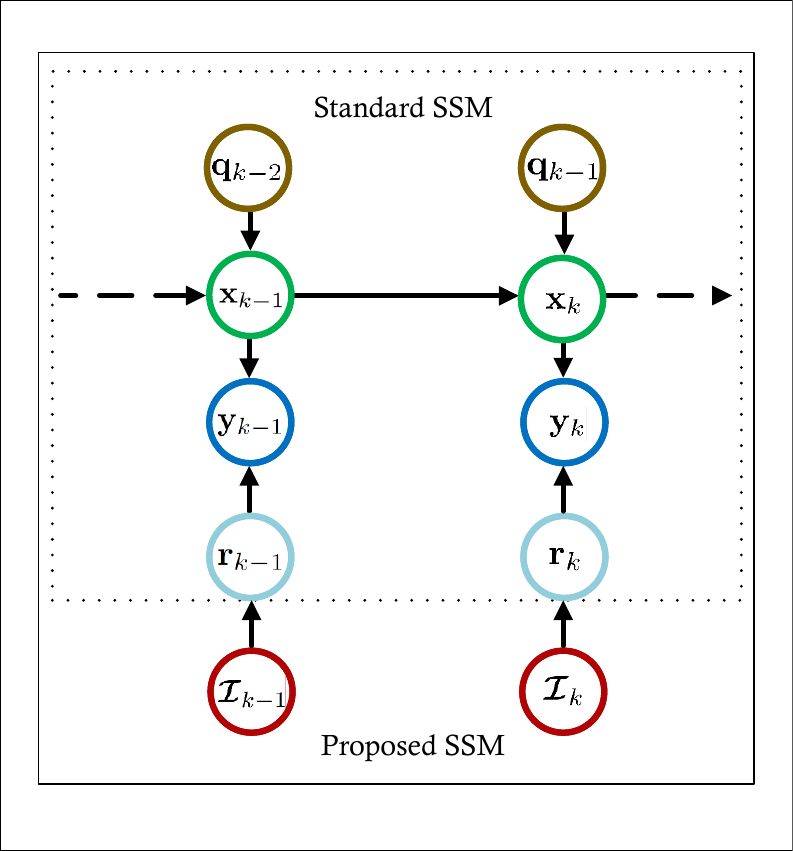}
	\caption{Probabilistic graphical model for the proposed method} 
	\label{fig:PGM_SOR}
\end{figure}

Fig.~\ref{fig:PGM_SOR} shows how the standard probabilistic graphical model (PGM) is modified into the proposed PGM for devising our outlier-robust state estimator. The suggested PGM meets the modeling aims of describing the nominal and corrupted data sufficiently while remaining docile for statistical inference.

\section{Variational Bayesian Inference}\label{Sec_VB_infer}
\hspace{.5cm}
With the proposed observation model, we can employ the Bayes rule recursively to obtain the analytical expression of the joint posterior distribution of $\mathbf{x}_k$ and $\bm{\mathcal{I}}_k$ conditioned on the set of all the observations $\mathbf{y}_{1:{k}}$
\begin{equation}
	p(\mathbf{x}_k,\bm{\mathcal{I}}_k|\mathbf{y}_{1:{k}})=\frac{p(\mathbf{y}_k|\bm{\mathcal{I}}_k,\mathbf{x}_{k})	p(\mathbf{x}_k|\mathbf{y}_{1:{k-1}})p(\bm{\mathcal{I}}_k)}{p(\mathbf{y}_k|\mathbf{y}_{1:{k-1}})}
	\label{eqn_vb_1_}
\end{equation} 

The joint posterior can further be marginalized to obtain $p(\mathbf{x}_k|\mathbf{y}_{1:{k}})$ for state inference. However, using this approach directly is computationally complex. Therefore, we resort to the standard VB method \cite{vsmidl2006variational}, {a technique for approximating intractable integrals arising in Bayesian inference}, where the joint posterior is approximated as a product of marginal distributions 
\begin{equation}
	p(\mathbf{x}_k,\bm{\mathcal{I}}_k|\mathbf{y}_{1:{k}})\approx q(\mathbf{x}_k)q(\bm{\mathcal{I}}_k)
	\label{eqn_vb_2}
\end{equation} 

The VB approximation aims to minimize the KL divergence between the product approximation and the true posterior. Accordingly, with $ \langle.\rangle_{q(\bm{\psi}_k)}$ denoting the expectation of the argument with respect to a distribution $q(\bm{\psi}_k)$, the variational distributions can be updated in an alternating manner as   
\begin{align}
	q(\mathbf{x}_k)&\propto \mathrm{exp}\big( \big\langle\mathrm{ln}(p(\mathbf{x}_k,\bm{\mathcal{I}}_k|\mathbf{y}_{1:{k}})\big\rangle_{q(\bm{\mathcal{I}}_k)}\big)\label{eqn_vb_3_}\\
	q(\bm{\mathcal{I}}_k)&\propto\mathrm{exp}\big( \big\langle\mathrm{ln}(p(\mathbf{x}_k,\bm{\mathcal{I}}_k|\mathbf{y}_{1:{k}})\big\rangle_{q(\mathbf{x}_k)}\big)\label{eqn_vb_4_}
\end{align} 

For tractability, we integrate general Gaussian filtering \cite{sarkka2023bayesian} results into the VB framework by assuming $p(\mathbf{x}_k|\mathbf{y}_{1:{k-1}})\approx {\mathcal{N}}\left(\mathbf{x}_k|\hat{\textbf{x}}^{-}_k,\mathbf{P}^{-}_{k}\right)$. Using the expressions of the prior distributions and measurement likelihood in \eqref{eqn_vb_1_}, the posterior distribution is approximated as a product of marginals derived as follows. In the derivation of $q(\mathbf{x}_k)$ and $q(\bm{\mathcal{I}}_k)$ we use the matrix algebra property: ${\text{tr}}(\mathbf{ABC})={\text{tr}}(\mathbf{CAB})={\text{tr}}(\mathbf{BCA})$, where the product $\mathbf{ABC}$ of matrices $\mathbf{A}, \mathbf{B}\ \text{and}\ \mathbf{C}$ is a scalar. 
\subsection{Derivation of $q(\mathbf{x}_k)$}
\hspace{.5cm}
Using \eqref{eqn_vb_3_} we can write
\begin{align}
	q(\mathbf{x}_k)\propto \mathrm{exp}&\left(-\mfrac{1}{2} (\mathbf{y}_k-\mathbf{h}(\mathbf{x}_k))^{\top}\mathbf{V}^{-1}_k(\mathbf{y}_k-\mathbf{h}(\mathbf{x}_k))-\mfrac{1}{2} (\mathbf{x}_k-\hat{\textbf{x}}^{-}_k)^{\top}({\mathbf{P}^{-}_{k}})^{-1}(\mathbf{x}_k-\hat{\textbf{x}}^{-}_k)\right)\label{eqn_vb_5_SOR}
\end{align} 
where
\begin{equation}
	\mathbf{V}^{-1}_k={\textbf{R}^{-1}_k}\left(\text{diag}\big(\big\langle\bm{\mathcal{I}}_k\big\rangle_{q(\bm{\mathcal{I}}_k)}\big)\right)
	\label{eqn_vb_6_}
\end{equation}

Using the results of Gaussian (Kalman) filter, $q(\mathbf{x}_k)$ can be approximated with a Gaussian distribution, $ {\mathcal{N}}\left(\mathbf{x}_k|\hat{\textbf{x}}^{+}_k,\mathbf{P}^{+}_{k}\right)$, with parameters given as
\begin{align}
	\hat{\textbf{x}}^{+}_k&=\hat{\textbf{x}}^{-}_k+\mathbf{K}_k
	(\mathbf{y}_k-\bm{\mu}_k)	\label{eqn_vb_7_}\\
	\mathbf{P}^{+}_{k}&=\mathbf{P}^{-}_{k}-\mathbf{C}_k\mathbf{K}^{\top}_k\label{eqn_vb_8_}
\end{align}
where {$\hat{\textbf{x}}^{-}_k$ and $\mathbf{P}^{-}_k$ denote the predicted mean and covariance matrix of a Kalman/Gaussian filter respectively and $\hat{\textbf{x}}^{+}_k$ and $\mathbf{P}^{+}_k$ denote the updated mean and covariance matrix of a Kalman/Gaussian filter respectively at the time step $k$ with}
\begin{align}
	\mathbf{K}_k&=\mathbf{C}_k (\mathbf{U}_k+\mathbf{V}_k)^{-1}=\mathbf{C}_k (\mathbf{V}^{-1}_k{-}\mathbf{V}^{-1}_k(\mathbf{I}+\mathbf{U}_k\mathbf{V}^{-1}_k)^{-1}\mathbf{U}_k\mathbf{V}^{-1}_k)\nonumber\\
	\bm{\mu}_k&=\int \mathbf{h}(\mathbf{x}_k)\  {\mathcal{N}}\left(\mathbf{x}_k|\hat{\textbf{x}}^{-}_{k},\mathbf{P}^{-}_{k}\right)	d\mathbf{x}_k\nonumber\\
	\mathbf{U}_k&=\int(\mathbf{h}(\mathbf{x}_k)-\bm{\mu}_k)(\mathbf{h}(\mathbf{x}_k)-\bm{\mu}_k)^{\top}{\mathcal{N}}({\mathbf{x}_k}|\hat{\textbf{x}}^{-}_k,\mathbf{P}^{-}_{k})d\mathbf{x}_k\nonumber
\end{align}
\begin{align}
	\mathbf{C}_k&=\int(\mathbf{x}_k-\hat{\textbf{x}}^{-}_{k})(\mathbf{h}(\mathbf{x}_k)-\bm{\mu}_k)^{\top}{\mathcal{N}}({\mathbf{x}_k}|\hat{\textbf{x}}^{-}_{k},\mathbf{P}^{-}_{k})d\mathbf{x}_k\nonumber
\end{align}

Note that the state estimates are updated with a modified measurement noise covariance $\mathbf{V}_k$ depending on the detection of outliers. When $\big\langle\bm{\mathcal{I}}_k\big\rangle_{q(\bm{\mathcal{I}}_k)}=\bm{1}$, corresponding to the ideal detection of no outlier, \eqref{eqn_vb_7_}-\eqref{eqn_vb_8_} become the standard Gaussian filtering equations. Likewise, when any $i$th entry of $\big\langle\bm{\mathcal{I}}_k\big\rangle_{q(\bm{\mathcal{I}}_k)}$ gets close to zero, the $i$th column of $\mathbf{K}_k$ approaches zero leading to the rejection of the corresponding $i$th measurement during inference. 
\subsection{\texorpdfstring{Derivation of $q(\boldsymbol{\mathcal{I}}_k)$}{}}
\hspace{.5cm}
Using \eqref{eqn_vb_4_} we can write
\begin{align}
	q(\bm{\mathcal{I}}_k) & \propto \prod_{i=1}^{m} \frac{\sqrt{{\mathcal{I}}_k(i)}}{\sqrt{2 \pi {\mathrm{R}_k(i,i)}}} \mathrm{exp}\left( {-}\frac{\mathrm{W}_k(i) {{\mathcal{I}}}_k(i) }{2\mathrm{R}_k(i,i)}\right)  \big((1-{\theta_k(i)}) \delta({{{\mathcal{I}}}_k(i)}-\epsilon)+{\theta_k(i)}\delta( {{{\mathcal{I}}}_k(i)}-1)\big)\label{eqn_vb_9}
\end{align} 
where $\mathrm{W}_k(i)$ is given as
\begin{equation}
	\mathrm{W}_k(i)=\big\langle(\mathrm{y}_k(i)-\mathrm{h}(\mathbf{x}_k)(i))^2\big\rangle_{q({\mathbf{x}}_k)} 
	\label{eqn_vb_9b}
\end{equation} 

We can further write 
\begin{align}
	q(\bm{\mathcal{I}}_k)&=\prod_{i=1}^{m}q({{\mathcal{I}}}_k(i))=\prod_{i=1}^{m} (1-{\Omega_k(i)}) \delta({{{\mathcal{I}}}_k(i)}-\epsilon)+{\Omega_k(i)}\delta( {{{\mathcal{I}}}_k(i)}-1)
	\label{eqn_vb_11_}
\end{align} 

It immediately follows that $q({{\mathcal{I}}}_k(i))$ is a Bernoulli distribution with the following probabilities. 
\begin{align}
	\Omega_k(i) &= c(i)  \ \theta_k(i)   \mathrm{exp}(-\frac{\mathrm{W}_k(i)}{2\mathrm{R}_k(i,i)}) \label{t1} \\
	1-\Omega_k(i) &= c(i) \ (1-\theta_k(i))\sqrt{\epsilon} \mathrm{exp}(-\frac{\mathrm{W}_k(i) \epsilon}{2\mathrm{R}_k(i,i)})   \label{t2}
\end{align} 
where $c(i)$ is a proportionality constant. Using \eqref{t1} and \eqref{t2}, we obtain 

\begin{align}
	\Omega_k(i) &=\frac{1}{1+{\sqrt{\epsilon}}(\frac{1}{\theta_k(i)}-1){\mathrm{exp}(\frac{\mathrm{W}_k(i)}{2\mathrm{R}_k(i,i)} (1-\epsilon))}}\label{eqn_vb_12_}  
\end{align}

\subsection{Choice of the Parameters $\theta_k(i)$ and $\epsilon$}
\hspace{.5cm}
For successful VB inference the choice of the parameters $\theta_k(i)$ and $\epsilon$ is important. Note that in \eqref{eqn_vb_12_} we cannot set $\theta_k(i)$ equal to $0$ or $1$ and $\epsilon$ as exactly 0. Otherwise, the parameter $\Omega_k(i)$ becomes independent of $\mathbf{W}_k$, i.e. the moment of $q(\mathbf{x}_k)$, making the VB updates impossible. We propose to set a neutral value of 0.5 or an uninformative prior for $\theta_k(i)$. Moreover, a value close to zero is proposed for $\epsilon$ since exact value of 0 denies the VB updates. Note that the use of uninformative prior has also been proposed in the literature for designing outlier-robust filters assuming no prior information about the existence of outliers is available \cite{8869835}.

\subsection{Predictive Distribution $p(\mathbf{x}_{k}|\mathbf{y}_{1:{k-1}})$}
\hspace{.5cm}
Lastly, to complete the recursive inference process, we need to obtain the predictive distribution $p(\mathbf{x}_{k}|\mathbf{y}_{1:{k-1}})$ from the posterior distribution at the previous instant $p(\mathbf{x}_{k-1}|\mathbf{y}_{1:{k-1}})$. With the posterior distribution $p(\mathbf{x}_{k-1}|\mathbf{y}_{1:{k-1}})$ approximated as Gaussian $q(\mathbf{x}_{k-1})\approx {\mathcal{N}}\left(\mathbf{x}_{k-1}|\hat{\textbf{x}}^{+}_{k-1},\mathbf{P}^{+}_{k-1}\right)$, we use the Gaussian filtering results to write $p(\mathbf{x}_k|\mathbf{y}_{1:{k-1}})\approx {\mathcal{N}}\left(\mathbf{x}_k|\hat{\textbf{x}}^{-}_k,\mathbf{P}^{-}_{k}\right)$ where
\begin{align}
	\hat{\textbf{x}}^{-}_{k}&=\int \mathbf{f}(\mathbf{x}_{k-1})\  {\mathcal{N}}\left(\mathbf{x}_{k-1}|\hat{\textbf{x}}^{+}_{k-1},\mathbf{P}^{+}_{k-1}\right)	d\mathbf{x}_{k-1}\label{eqn_vb_13_}\\
	\mathbf{P}^{-}_{k}&=\int (\mathbf{f}(\mathbf{x}_{k-1})-\hat{\textbf{x}}^{-}_{k})(\mathbf{f}(\mathbf{x}_{k-1})-\hat{\textbf{x}}^{-}_{k})^{\top} {\mathcal{N}}({\mathbf{x}_{k-1}}|\hat{\textbf{x}}^{+}_{k-1},\mathbf{P}^{+}_{k-1})  d\mathbf{x}_{k-1}+\mathbf{Q}_{k-1}\label{eqn_vb_14_}
\end{align}

For the convergence criterion, we suggest using the ratio of the L2 norm of the difference of the state estimates from the current and previous VB iterations and the L2 norm of the estimate from the previous iteration. This criterion has been commonly chosen in similar robust filters \cite{8398426}. The resulting SOR filter is presented in Algorithm \ref{Algo1_}.

%
 \begin{algorithm}[ht!]
	\SetAlgoLined
	Initialize\ $\hat{\textbf{x}}^{+}_0,\mathbf{P}^{+}_0$;
	
	\For{$k=1,2...K$}{
		Initialize $\theta_k(i)$, $\mathbf{Q}_k$, $\mathbf{R}_k$, $ {\mathbf{V}^{-1}_k}=\mathbf{R}^{-1}_k $ \;
		\textbf{Prediction:}\\
		Evaluate $\hat{\textbf{x}}^{-}_k,\mathbf{P}^{-}_k$ with \eqref{eqn_vb_13_} and \eqref{eqn_vb_14_}\;
		\textbf{Update:}\\
		\While{not converged}{
			Update ${\hat{\textbf{x}}_k^{+}}$ and  ${\mathbf{P}_k}^{+}$ with \eqref{eqn_vb_7_}-\eqref{eqn_vb_8_}\;
			Update ${\mathrm{W}_k(i)}$ with \eqref{eqn_vb_9b} and ${\Omega_k(i)}$ with \eqref{eqn_vb_12_} for each $i$\;
			Update ${\mathbf{V}^{-1}_k}$ with  \eqref{eqn_vb_6_}
		}
	}
	\caption{The proposed SOR filter}
	\label{Algo1_}
\end{algorithm}

\section{Performance Evaluation}\label{Eval}
\subsection{VB-based Outlier-robust Filters under Consideration}
\subsubsection*{Standard Methods}
\hspace{.5cm}
We resort to the UKF as the basic inferential engine, approximating the Gaussian integrals using the unscented transform \cite{wan2001unscented}, for all the considered methods. Therefore, we name the proposed method as selective observations-rejecting UKF (SOR-UKF). Similarly, the other VB-based outlier-robust baseline filters are called as recursive outlier-robust UKF (ROR-UKF) \cite{6349794}, switching error model UKF (SEM-UKF) \cite{8869835} and outlier-detecting  UKF (OD-UKF) \cite{8398426}. The choice of using UKF allows us to propose a computationally efficient modification of our devised filter based on an available efficient implementation of UKF for handling high-dimensional uncorrelated measurements reported in literature \cite{mcmanus2012serial}.

\subsubsection*{Proposed Modifications}
\hspace{.5cm}
Since the baseline methods, in their original form, do not treat the outliers selectively for each dimension (as these are under-parameterized in terms of outlier detection for each dimension), we modify each of these for such behavior. In particular, the measurements from each sensor at each time step are proposed to be used in turn i.e. the final estimate using any sensor's observation serves as the prediction for the update using the next sensor's measurement. Accordingly, the modified baseline algorithms are referred to as mROR-UKF, mSEM-UKF, and mOD-UKF.

Moreover, we propose using an alternate implementation of the proposed SOR-UKF, referred to as mSOR-UKF with lower complexity. Since $\mathbf{V}^{-1}_k$ is diagonal, the basic inferential UKF used within the proposed SOR-UKF can be modified from the standard parallel sigma point Kalman filter (P-SPKF) to the serial sigma point Kalman filter (S-SPKF) \cite{mcmanus2012serial} for use in mSOR-UKF. Note that S-SPKF can be also used in conjunction with ROR-UKF, SEM-UKF and OD-UKF, as originally reported, but these do not treat outliers selectively for each dimension hence compromising the estimation quality. mROR-UKF, mSEM-UKF and mOD-UKF, on the other hand, have analogous structures to the serial re-draw sigma point Kalman filter (SRD-SPKF) in \cite{mcmanus2012serial}.
\subsection{Theoretical Computational Complexity}
\begin{table}[h!]
	\caption{Theoretical complexity of different learning-based methods treating outliers selectively}
	\centering
	\begin{tabular}{@{}lll@{}}
		\toprule
		\ \ \textbf{{Method}}  & \textbf{{Complexity}} & \textbf{Computational Analogue}\\ \midrule
		\ \ {SOR-UKF}           & {$\mathcal{O}{(}(m+n)^3{)}$} & \quad \quad \quad P-SPKF \\
		\ \ {mOD/mSEM/mROR-UKF} & {$\mathcal{O}(m n^3)$}  & \quad \quad \quad SRD-SPKF  \\
		\ \ {mSOR-UKF}          &  {$\mathcal{O}(n^2(m+n))$} & \quad \quad \quad S-SPKF \\ \bottomrule
	\end{tabular}
	\label{Tab1}
\end{table}

The proposed SOR-UKF in the original form involves matrix inversions for Kalman gain evaluation and sigma points are drawn after processing the entire vector of measurements, resultingly it has a theoretical complexity of $\mathcal{O}{(}(m+n)^3{)}$. In mROR-UKF, mSEM-UKF, and mOD-UKF after updating the state estimate using one sensor's measurement, sigma points from the updated state distribution are regenerated, in VB iterations until convergence, for processing the next sensor's measurement during the same sampling interval. During updates, only scalar inversions are required for Kalman gain computations. Consequently, these methods have a theoretical complexity of $\mathcal{O}(m n^3)$. In mSOR-UKF, S-SPKF is used which has complexity of $\mathcal{O}(n^2(m+n))$. Moreover, all the $\mathrm{W}^{ii}_k$ terms can be computed with complexity not more than $m n$, we can implement mSOR-UKF with an overall theoretical complexity of $\mathcal{O}(n^2(m+n))$. Table~\ref{Tab1} summarizes the theoretical complexity of different learning-based methods treating multiple outliers selectively along with their standard analogous counterparts. Note that since the theoretical complexity depends on the exact functional forms of $\textbf{f}(.)$ and $\textbf{h}(.)$, we assume in the analysis that each entry of the functional mapping requires constant time independent of $n$. The assumption is true for several practical cases e.g. range data of a moving target from sensors depends on the position of target only.

\subsection{Simulation Results}
\subsubsection*{Simulation Setup}
\hspace{.5cm}
In all the simulations and experimental scenarios, we use MATLAB on an Intel i7-8550U processor powered computer and consider SI units. First, we choose to simulate a target tracking application for performance evaluation. 

\begin{figure}[!h]
	\centering
	\includegraphics[width=0.7\columnwidth]{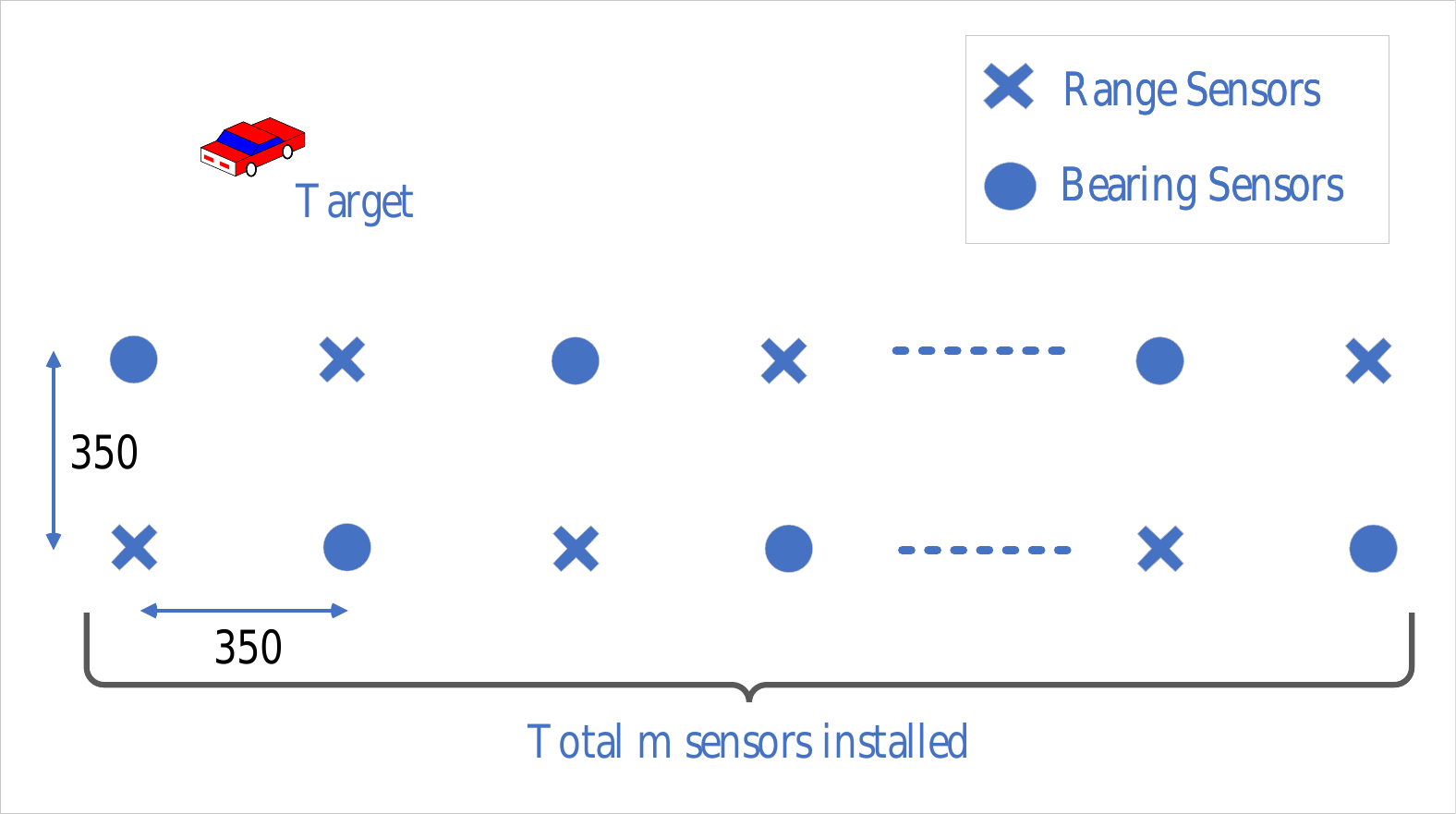}
	\caption{Setup of the target tracking test example}
	\label{results_setup}
\end{figure}

Fig.~\ref{results_setup} shows the testing setup for a target tracking example we consider in the simulations for performance evaluation. Maneuvering targets with unknown turn rates are commonly modeled as 
\begin{align}
	\mathbf{x}_k &= \begin{pmatrix} \text{1} & \frac{\text{sin}(\omega_{k}\Delta t)}{\omega_{k}} & \text{0} &  \frac{\text{cos}(\omega_{k}\Delta t)-\text{1}}{\omega_{k}} & \text{0} \\  \text{0} & \text{cos}(\omega_{k}\Delta t) &  \text{0} & -\text{sin}(\omega_{k}\Delta t) & \text{0}\\  \text{0} & \frac{\text{1}-\text{cos}(\omega_{k}\Delta t)}{\omega_{k}} & \text{1} &\frac{\text{sin}(\omega_{k}\Delta t)}{\omega_{k}} & \text{0} \\  \text{0} &  \text{\text{sin}}(\omega_{k}\Delta t) &  \text{0} &  \text{cos}(\omega_{k}\Delta t) & \text{0}\\  \text{0} & \text{0} & \text{0} & \text{0} &\text{1} \end{pmatrix} \mathbf{x}_{k-\text{1}} + \mathbf{q}_{k-\text{1}} \label{eqn_res1} 
\end{align}
where the state vector $\mathbf{x}_k= [a_k,\dot{{a_k}},b_k,\dot{{b_k}},\omega_{k}]^{\top}$
$ = \begin{bmatrix} a_k  & \dot{{a_k}} & b_k & \dot{{b_k}} & \omega_{k} \end{bmatrix}^{\top}$ 
contains the 2D position coordinates \(({a_k} , {b_k} )\), the corresponding velocities \((\dot{{a_k}} , \dot{{b_k}} )\), the angular velocity $\omega_{k}$ of the target at time instant $k$, \( \zeta \) is the sampling period, and $\mathbf{q}_{k-\text{1}} \sim N\left(0,\mathbf{Q}_{k-\text{1}}.\right)$ with $\mathbf{Q}_{k-\text{1}}$ given, in terms of scaling parameters $\eta(1)$ and $\eta(2)$, as \cite{8398426}
\begin{equation}
	\mathbf{Q}_{k-\text{1}}=\begin{pmatrix} \eta(1) \mathbf{M} & 0 & 0\\0 &\eta(1) \mathbf{M}&0\\0&0&\eta(2)
	\end{pmatrix}, \mathbf{M}=\begin{pmatrix} {\zeta}^3/3 & {\zeta}^2/2\\{\zeta}^2/2 &{\zeta}
	\end{pmatrix}\nonumber
\end{equation}

We consider that angle and range readings are obtained from sensors, installed around a rectangular area, at $m$ different locations. A total of $m/2$ independent sensors are used to provide angle readings where its $j$th sensor is present at the 2D coordinate $(a_{\theta}(j)=350(j-1),b_{\theta}(j)=350(j\mod 2))$. Similarly, range measurements are obtained from the other $m/2$ independent sensors where its $j$th sensor is located at $(a_{\rho}(j)=350(j-1),b_{\rho}(j)=350\ ((j-1)\mod2))$.

\begin{figure}[h!]
	\centering
	\hspace*{.2cm}\includegraphics[width=.7\linewidth]{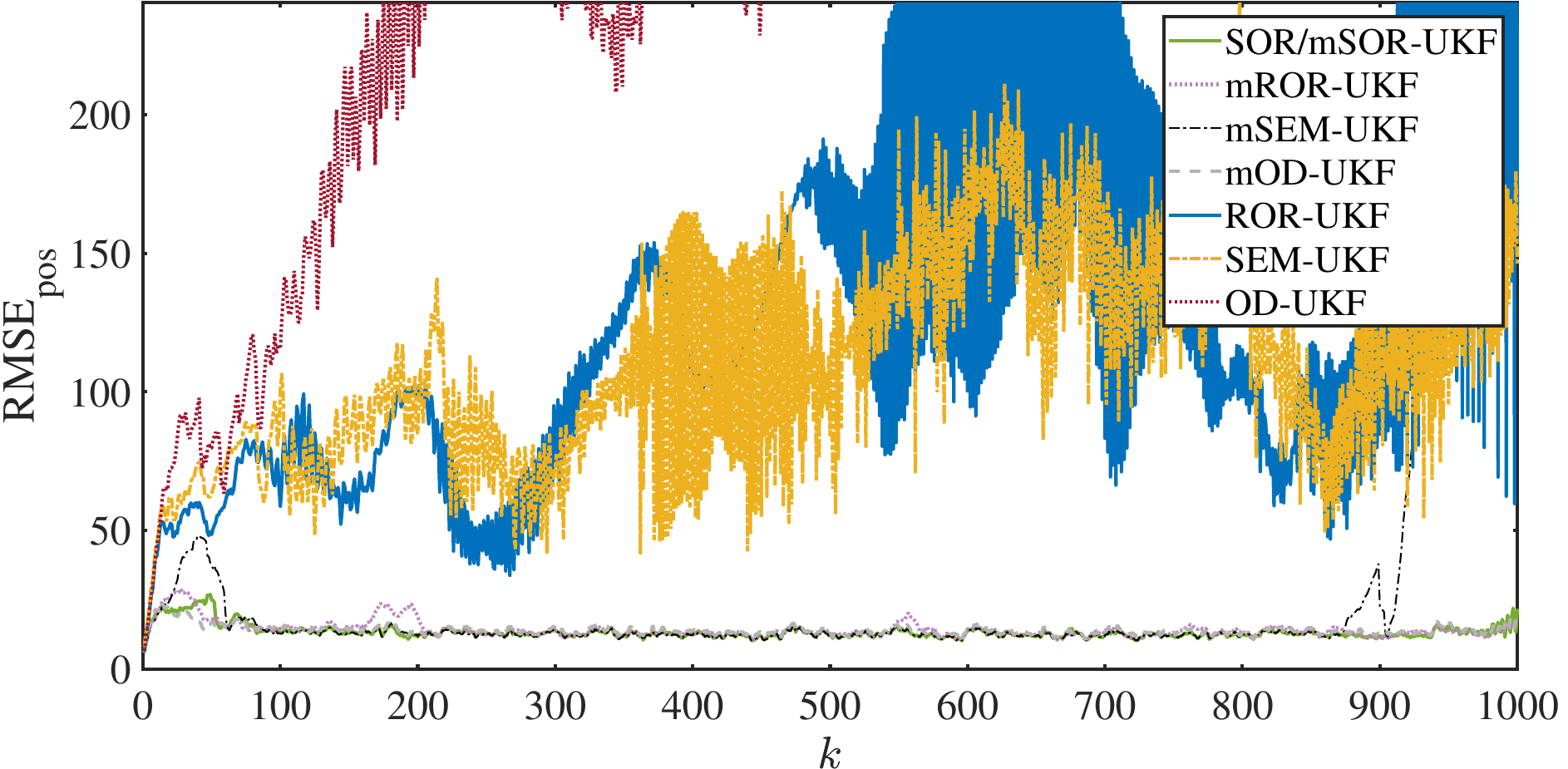}
	\caption{$\text{RMSE}_{\text{pos}}$ vs. $k$  ($\lambda=0.3, \gamma\sim \mathcal{U}(100,1000)$) showing how the error of the estimators vary over time with changes in variance of outliers}
	\label{fig:lambda6mixgauss_RMSE1000}
\end{figure}

\begin{figure}[!h]
	\centering
	\includegraphics[width=.8\columnwidth]{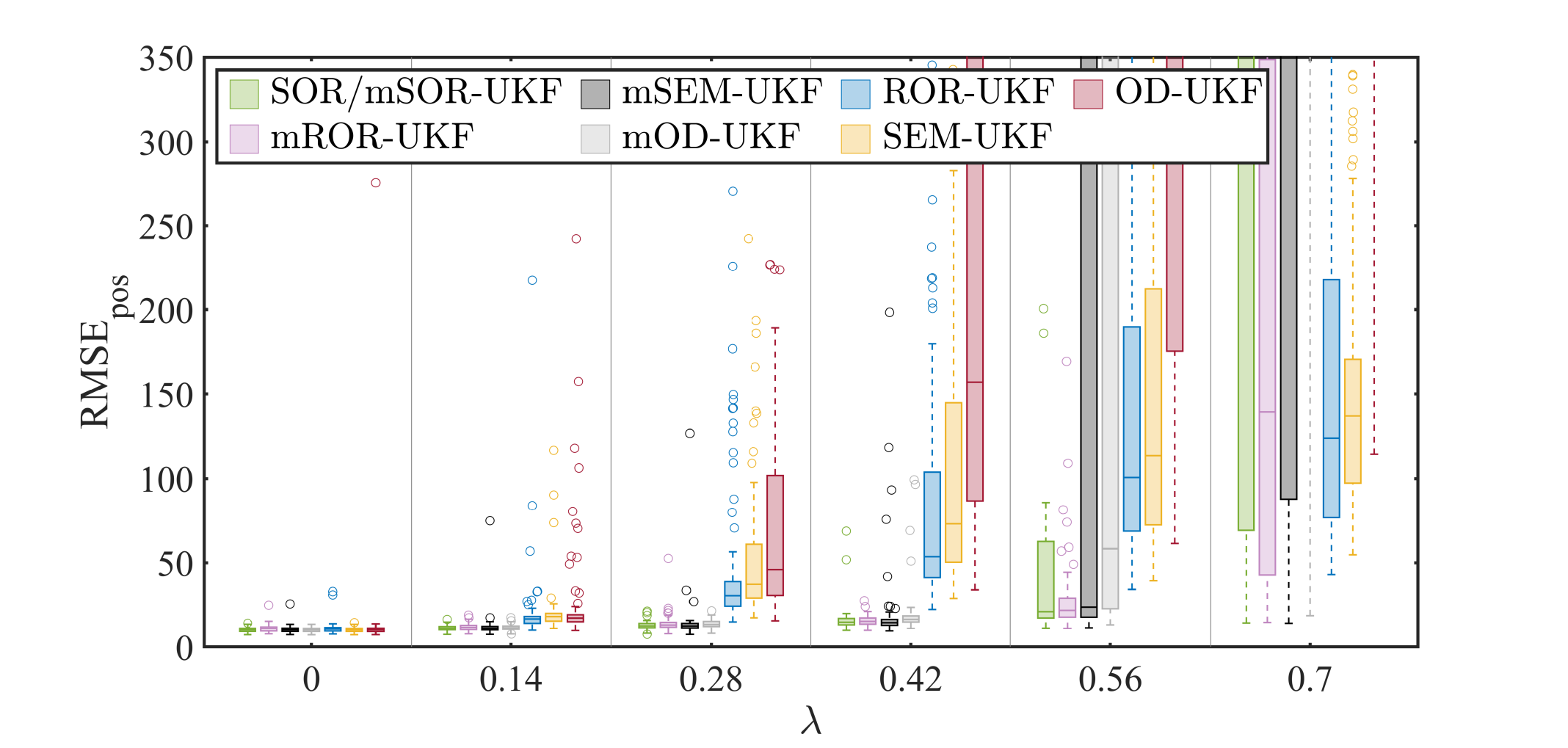}
	\caption{$\text{RMSE}_{\text{pos}}$ vs. $\lambda$  ($\gamma\sim \mathcal{U}(100,1000)$) showing the effect of outlier contamination frequency on error of the estimators}
	\label{results_2}
\end{figure}

\begin{figure}[!h]
	\centering
	\includegraphics[width=.8\columnwidth]{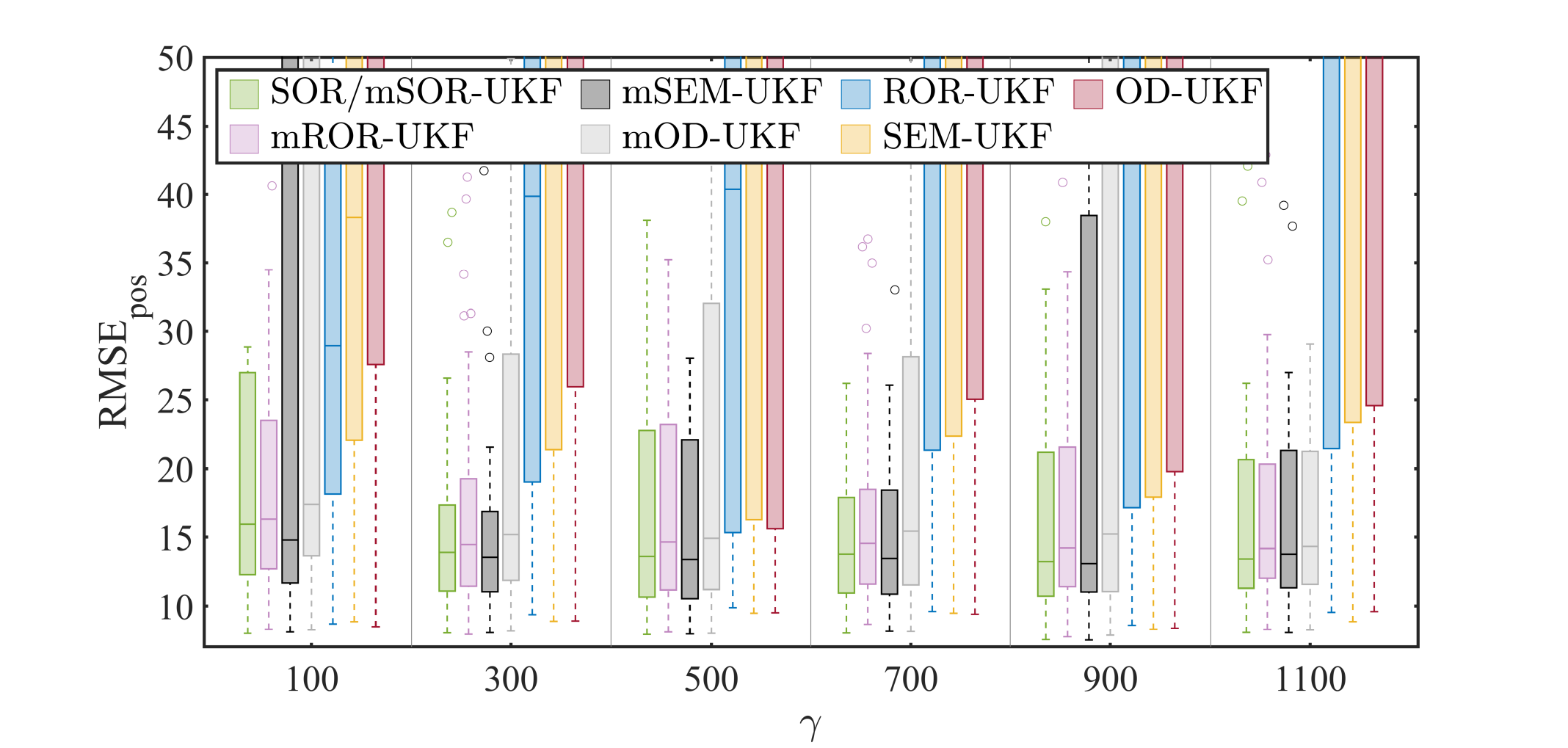}
	\caption{$\text{RMSE}_{\text{pos}}$ vs. $\gamma$  ($\lambda\sim \mathcal{U}(0,0.7)$) showing the effect of outlier contamination variance on error of the estimators} 
	\label{results_1}	
\end{figure}

\begin{figure}[!h]
	\centering
	\includegraphics[width=.8\linewidth]{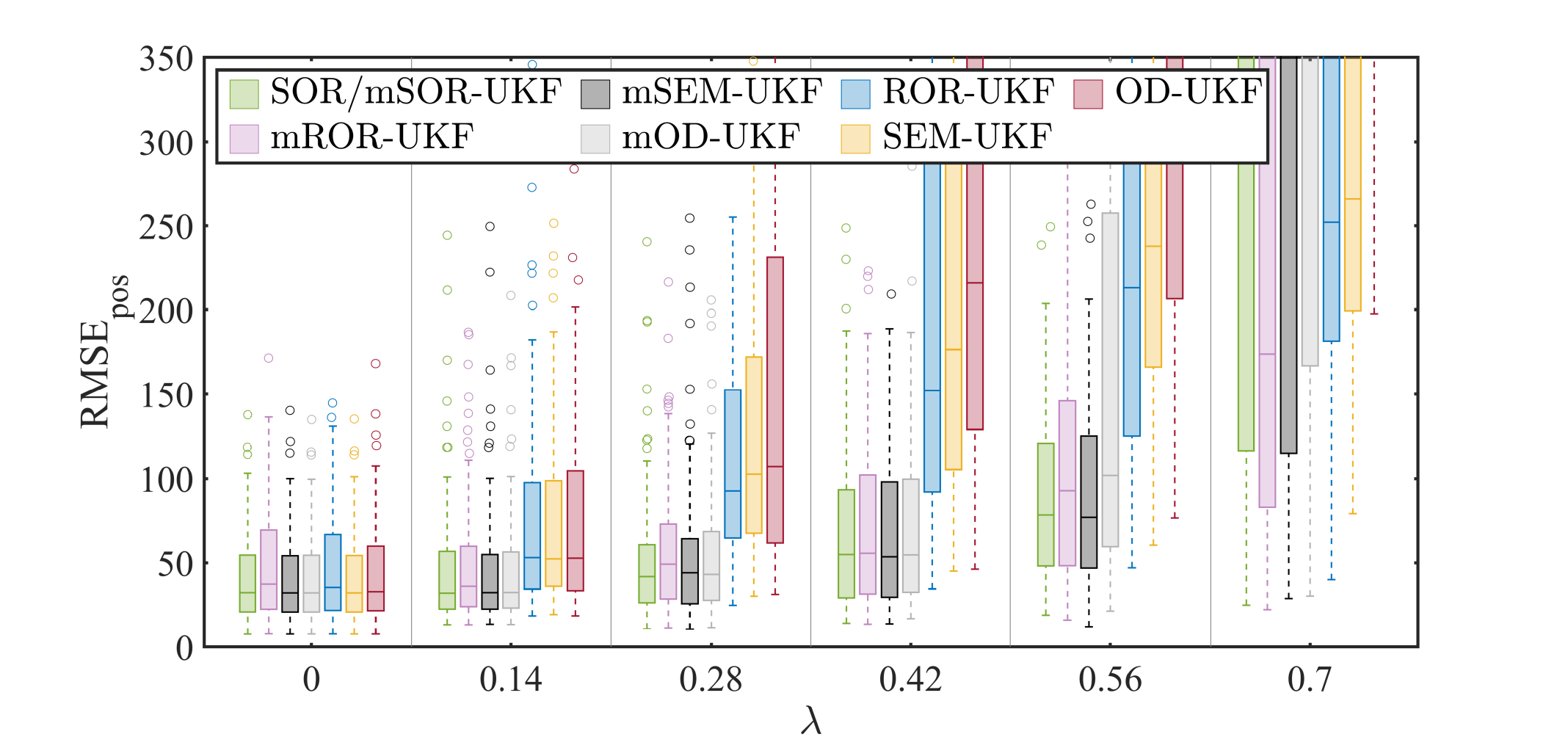}
	\caption{$\text{RMSE}_{\text{pos}}$ vs. $\lambda$ ($\gamma\sim \mathcal{U}(100,1000)$, ${\sigma_\theta}\sim \mathcal{U}(1.7\times10^{-3},3.5\times10^{-2})$, ${\sigma_\rho}\sim \mathcal{U}(5,50)$) depicting the effect of changes in outlier contamination variances and nominal noise statistics on error of the estimators}
	\label{fig:lambda6mixgauss_sigma}
\end{figure}

\begin{figure}[h!]
	\centering
	\includegraphics[width=.8\linewidth]{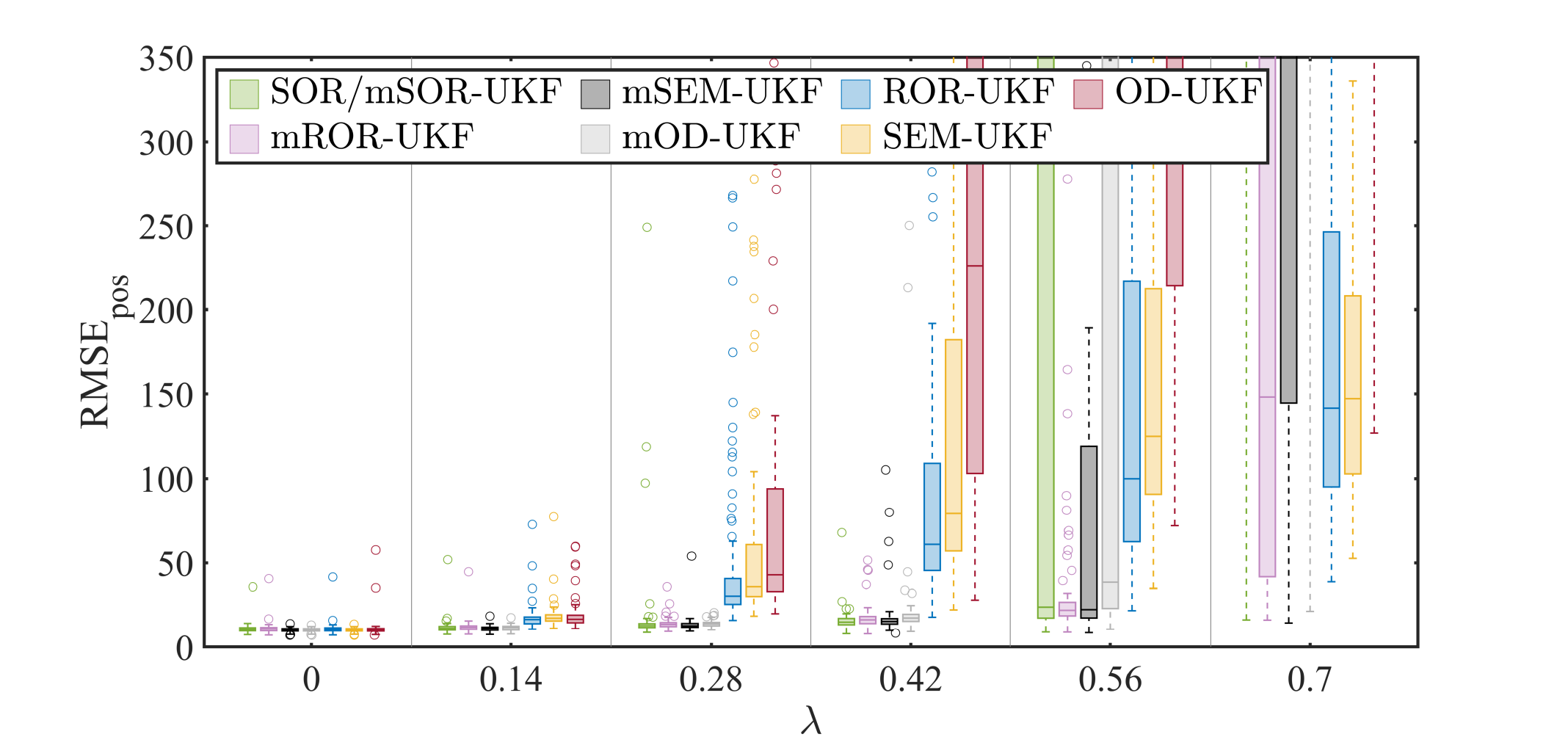}
	\caption{$\text{RMSE}_{\text{pos}}$ vs. $\lambda$ ($\gamma\sim \mathcal{U}(100,1000)$, $
		\epsilon \sim \mathcal{U}(10^{-7},10^{-3})$, $\theta_k(i)\sim \mathcal{U}(0.05,0.95)$) showing the effect of changes in SOR parameters on the error performance}
	\label{fig:lambda6mixgauss_randeps}
\end{figure}

\begin{figure}[h!]
	\centering
	\includegraphics[width=.8\linewidth]{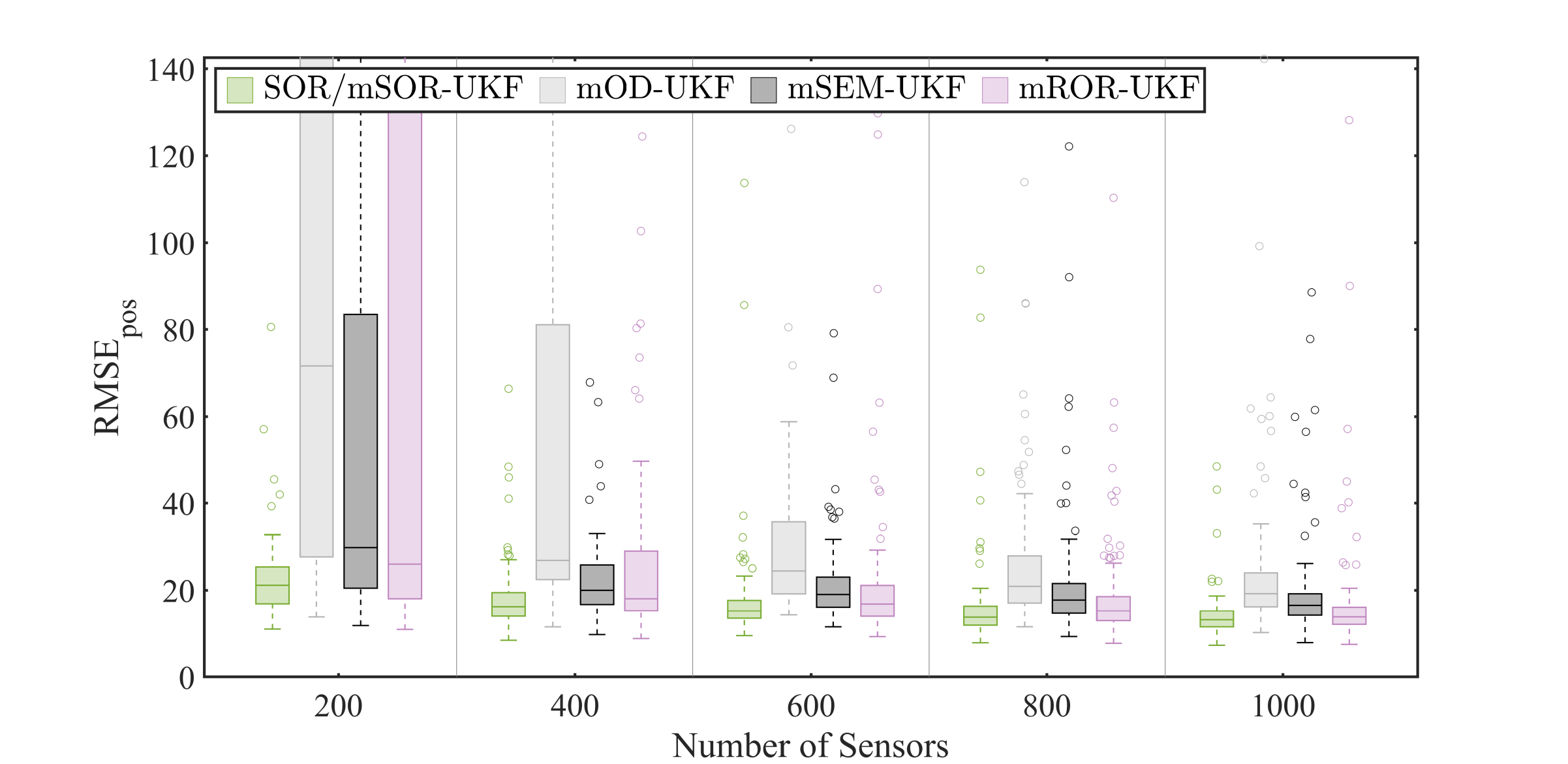}
	\caption{$\text{RMSE}_{\text{pos}}$ vs. $m$ ($\lambda=0.9$, $\gamma\sim \mathcal{U}(100,1000)$) depicting the effect of changes in the number of sensors on error of the estimators}
	\label{fig:lambda6mixgauss_lam_pntnine}
\end{figure}

\begin{figure}[h!]
	\centering
	\includegraphics[width=.8\linewidth]{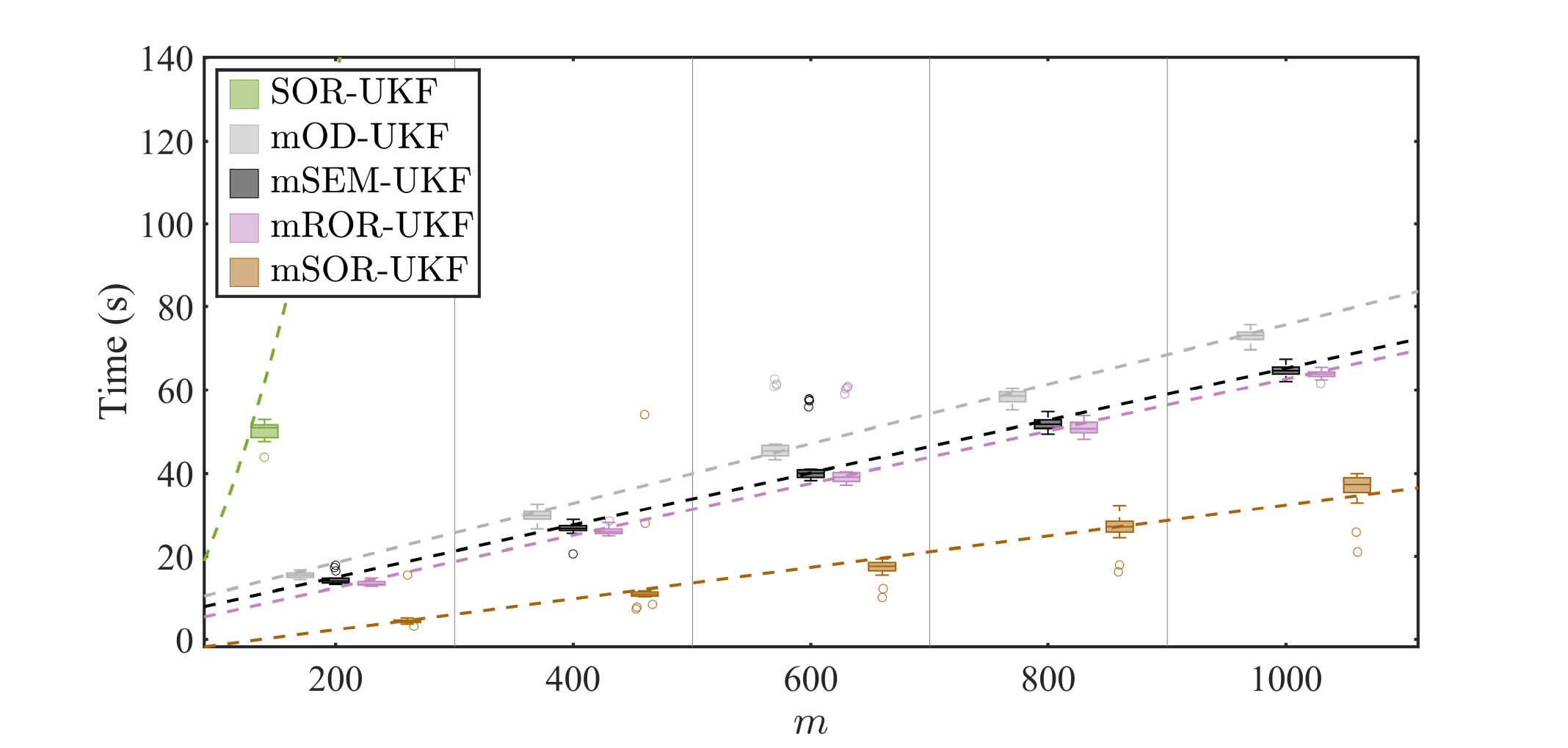}
	\caption{$\text{Computational overhead}$ vs. the number of sensors ($\lambda=0.9$,  $\gamma\sim \mathcal{U}(100,1000)$)}
	\label{fig:complexity}
\end{figure}

{For performance evaluation, two subcases of how measurements from different sensors get disturbed are considered.} {First, we suppose outliers explicitly add in the observations randomly. Subsequently, we evaluate the case of missing data. 
	
	\subsubsection*{{Outliers}}	
	\hspace{.5cm}
	We suppose a Gaussian mixture to model the additive effect of outliers commonly used for performance evaluation of robust filtering methods \cite{8869835,8398426}.} The measurement vector $\mathbf{y}_k$  considering the {additive} effect of independently occurring outliers in different dimensions {can be expressed} as
\begin{equation}
	\mathbf{y}_k=\mathbf{s}_k + \mathbf{o}_k \label{eqn_res2}
\end{equation}
where $\mathbf{s}_k=[s^{\theta}_k(1)...s^{\theta}_k(m/2),s^{\rho}_k(1)...s^{\rho}_k({m/2})] ^{\top}$ and $\mathbf{o}_k = [o^{\theta}_k(1)...o^{\theta}_k({m/2}),o^{\rho}_k(1)...o^{\rho}_k({m/2})]^{\top}$
denote the noise-free and noisy components of $\mathbf{y}_k$ respectively. The entries of $\mathbf{s}_k$, $s^{\theta}_k(j)$ and $s^{\rho}_k(j)$, which denote the noise-free values corresponding to $j$th bearing and range sensors respectively are given as

\begin{align}
	s^{\theta}_k(j)&=\text{atan2}(b_k-b_{\theta}(j),a_k-a_{\theta}(j))\label{eqn_res3}\\
	s^{\rho}_k(j)&=\left({(a_k-a_{\rho}(j))^{\text{2}} + (b_k-b_{\rho}(j))^{\text{2}}}\right)^{\tfrac{1}{2}}\label{eqn_res3b}
\end{align}

Similarly, the entries of $\mathbf{o}_k$, $o^{\theta}_k(j)$ and $o^{\rho}_k(j)$, control the noise content in the measurements from the $j$th bearing and range sensors. $\mathbf{o}_k$ is considered to obey the following distribution
\begin{align}
	p(\mathbf{o}_k)=\prod_{j=1}^{m/2}&\big(\lambda\   {\mathcal{N}}(o^{\theta}_k(j)|0,\gamma\ {\sigma_\theta}^2)+(1-\lambda)\  {\mathcal{N}}(o^{\theta}_k(j)|0,{\sigma_\theta}^2)\big)\nonumber\\  &. \left( \lambda\  {\mathcal{N}}(o^{\rho}_k(j)|0,\gamma\ {\sigma_\rho}^2)+(1-\lambda)\  {\mathcal{N}}(o^{\rho}_k(j)|0, {\sigma_\rho}^2) \right)\label{eqn_res4} 
\end{align}
where ${\sigma_\theta}^2$ and ${\sigma_\rho}^2$ are the variances of nominal noise in angle and range readings respectively. The parameters $\lambda$ and $\gamma$ control the frequency and variance of an outlier in each dimension respectively. 

\subsubsection*{{Base Parameters}}
\hspace{.5cm}
The following values of parameters are used ({unless stated otherwise}): the initial state $\mathbf{x}_0= [-10000,10,5000,-5,-0.0524] ^{\top}$, $\zeta=1$, $\eta(1)=0.1$, $\eta(2)=1.75\times10^{-4}$, ${\sigma_\theta}=3.5\times10^{-3}$, ${\sigma_\rho}=10$ {and $m=6$}. The initialization parameters of filters are: {${\hat{\textbf{x}}}^+_{0} \sim \mathcal{N}(\mathbf{x}_0,\mathbf{P}^+_{0}$), $\mathbf{P}^+_{0}=100\mathbf{Q}_{k}$}, $\epsilon=10^{-6}$ and $\theta_k(i)=0.5~\forall~i$. For each method, the UKF parameters \cite{wan2001unscented} are set as $\alpha=1$, $\beta=2$ and $\kappa=0$. Moreover, we use the same threshold of $10^{-4}$ for the convergence criterion for all the evaluated methods. Other parameters for the rival methods are assigned values as originally reported. All the simulations are repeated with a total time duration {$K=1000$} and $100$ independent MC runs.

{

	First, we assess the tracking performance over time of different filters. Fig.~\ref{fig:lambda6mixgauss_RMSE1000} shows the Root Mean Squared Error of the position estimates $\text{RMSE}_{\text{pos}}$ over time for this scenario assuming $\lambda=0.3, \gamma\sim \mathcal{U}(100,1000)$. The methods which treat outliers selectively for each dimension i.e. SOR-UKF, mSOR-UKF, mROR-UKF, mSEM-UKF, and mOD-UKF exhibit comparable tracking performance  whereas the other filters result in larger errors.

	We also assess the quality of estimates using different filters with a change in $\lambda$ with $\gamma\sim \mathcal{U}(100,1000)$. Fig.~\ref{results_2} shows the box plots of the position $\text{RMSE}$ ($\text{RMSE}_{\text{pos}}$) for this scenario. The methods dealing selectively with outliers exhibit comparable performance whereas the other filters result in a sharper rise of $\text{RMSE}_{\text{pos}}$ with an increase in $\lambda$.

	Subsequently, we vary $\gamma$ and observe $\text{RMSE}_{\text{pos}}$ with $\lambda\sim\mathcal{U}(0,0.7)$. Fig.~\ref{results_1} shows the distributions of the $\text{RMSE}_{\text{pos}}$ for this case using different filters. SOR-UKF, mSOR-UKF, mROR-UKF, mSEM-UKF, and mOD-UKF demonstrate comparable performance and outperform other methods.

	In addition, we evaluate how the change in nominal noise parameters ${\sigma_\theta}$ and ${\sigma_\rho}$ affects the performance of filters. Varying $\lambda$ the change in $\text{RMSE}_{\text{pos}}$ with $\gamma\sim \mathcal{U}(100,1000)$, ${\sigma_\theta}\sim \mathcal{U}(1.7\times10^{-3},3.5\times10^{-2})$, ${\sigma_\rho}\sim \mathcal{U}(5,50)$ is depicted in the box plots in Fig.~\ref{fig:lambda6mixgauss_sigma}. We observe a similar trend as in Fig.~\ref{results_2} except that the error magnitude levels increase.
	
	In addition, we evaluate the robustness of SOR-UKF and mSOR-UKF with change in filter parameters by assuming $\epsilon \sim \mathcal{U}(10^{-7},10^{-3})$ and $\theta_k(i)\sim \mathcal{U}(0.05,0.95)$ with $\gamma\sim\mathcal{U}(100,1000)$. Varying $\lambda$ the change in $\text{RMSE}_{\text{pos}}$ is depicted in the box plots in Fig.~\ref{fig:lambda6mixgauss_randeps}. We find the proposed filters quite robust to changes in filter parameters except at large values of $\lambda$ where the SOR-UKF and mSOR-UKF exhibit larger errors. 
	
	We also simulate a case where the effect of increase in the number of sensors on the estimation quality is observed. Fig.~\ref{fig:lambda6mixgauss_lam_pntnine} depicts the $\text{RMSE}_{\text{pos}}$ versus $m$, increased from 200 to 1000, for methods dealing each dimension selectively. We choose a high rate of outlier occurrence i.e. $\lambda=0.9$ with $\gamma\sim\mathcal{U}(100,1000)$. Owing to a large dimensionality of $m$, or more sources of information, we see that most of the algorithms result in low errors especially at higher values of $m$. We find SOR-UKF and mSOR-UKF comparatively more robust in this scenario. Note that non-selective methods exhibit higher errors so we skip them in the results.

	Lastly, we compare the computational cost for the selective methods exhibiting comparable lower $\text{RMSE}_{\text{pos}}$ values. We evaluate these methods in terms of the time taken to complete the MC simulations. Setting $\lambda=0.9$ with $\gamma\sim\mathcal{U}(100,1000)$, we vary $m$ and present the distributions of completion times of simulations using different filters in Fig.~\ref{fig:complexity}. With increasing $m$, the average completion times of the simulations increase for all the filters. We observe that the empirical computational overhead verifies the theoretical complexity. SOR-UKF exhibits cubic complexity whereas the other techniques have linear complexity in terms of $m$. The differences in the processing overhead among mROR-UKF, mSEM-UKF, and mOD-UKF depends on the their modeling parameters and inferential mechanism. Importantly, we note that with an increase in $m$, the computational costs of mROR-UKF, mSEM-UKF, and mOD-UKF rise more steeply than the mSOR-UKF as predicted by the theoretical complexity (analogous to how S-SPKF is computationally faster than the SRD-SPKF) in \cite{mcmanus2012serial}).

}

\subsubsection*{{Missing Data}}
\hspace{.5cm}
{We also evaluate the performance of different filters for the case of missing data. Missing observations can} {also be viewed as special case of outliers in a sense that the nominal model in \eqref{eqn_model_1}-\eqref{eqn_model_2} is unable to describe the observations and only the modified measurement equation \eqref{eqn_model_2-m} can with ${{\mathcal{I}}}_k(i)=\epsilon$ in any particular affected dimension $i$. Using the base parameters, as in the case of outliers, we simulate this case with $\lambda$ indicating the probability of missing observations in each dimension.

	\begin{figure}[h!]
		\centering
		\includegraphics[width=0.7\linewidth]{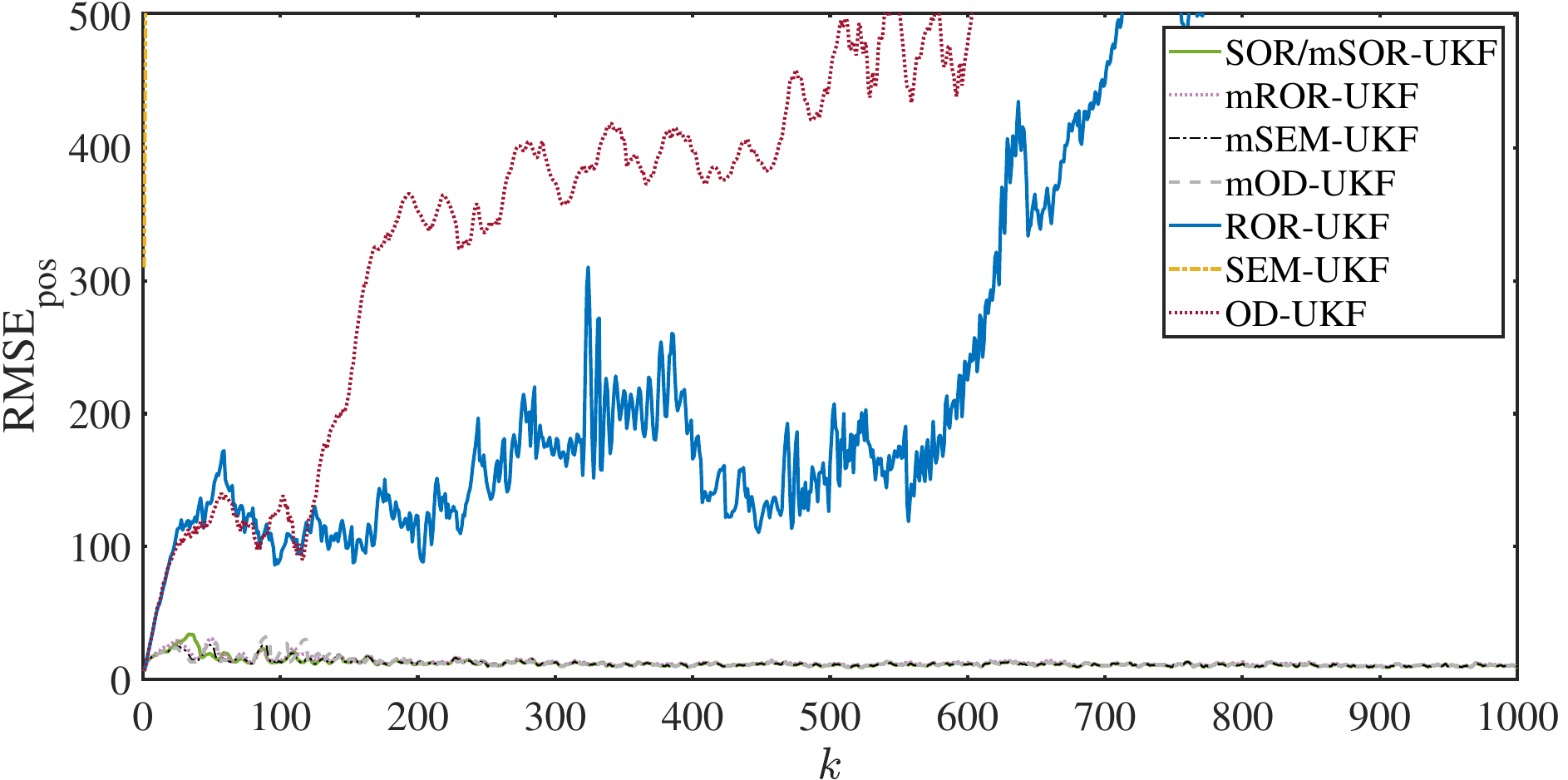}
		\caption{Error performance of different estimators over time with $\lambda=0.3$ }
		\label{fig:lambda6missing_RMSE1000}
	\end{figure} 
	
	\begin{figure}[h!]
		\centering
		\includegraphics[width=.8\linewidth]{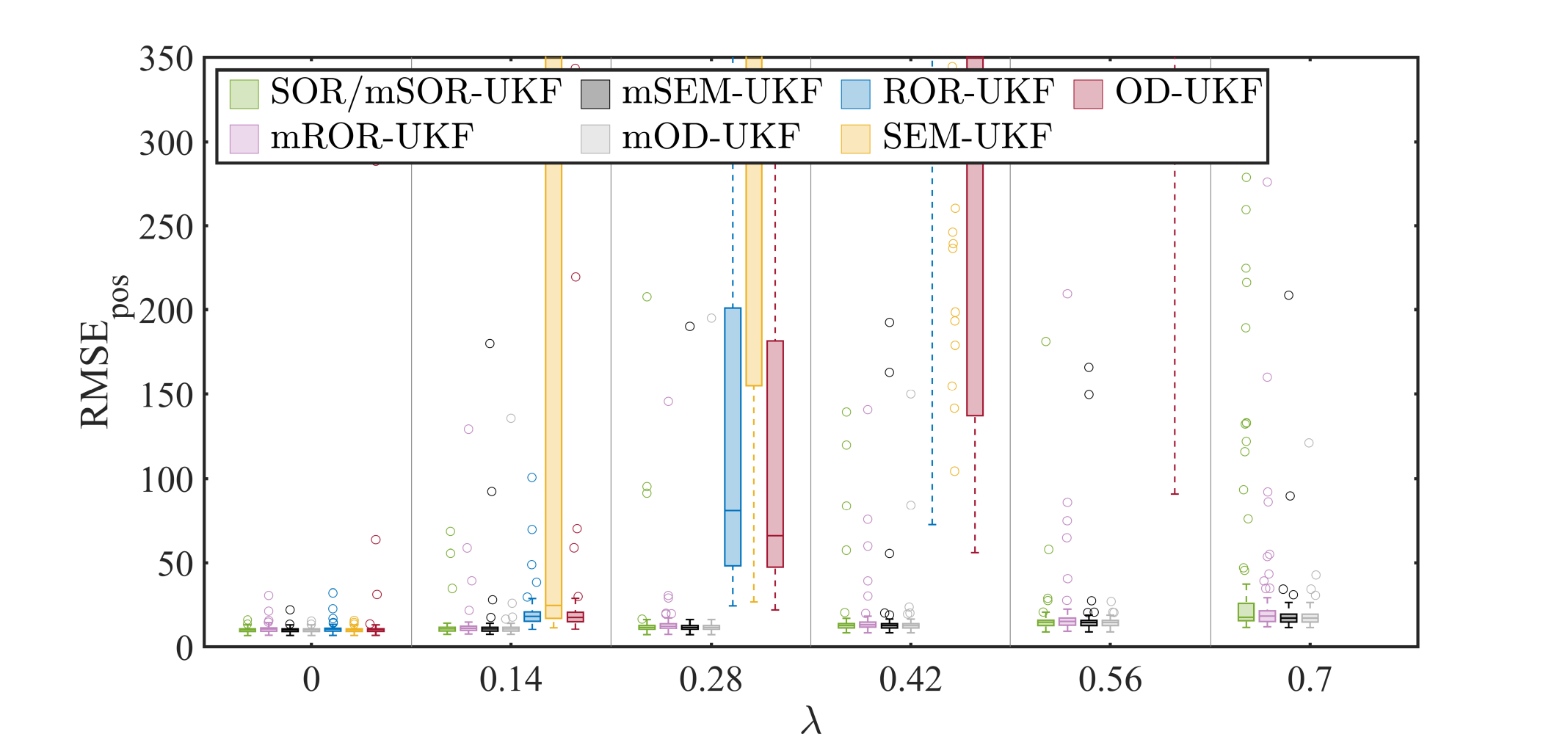}
		\caption{$\text{RMSE}_{\text{pos}}$ vs. $\lambda$ showing how changes in frequency of missing data affects the error performance of different estimators}
		\label{fig:lambda6missing}
	\end{figure}
	
	\begin{figure}[h!]
		\centering
		\includegraphics[width=.8\linewidth]{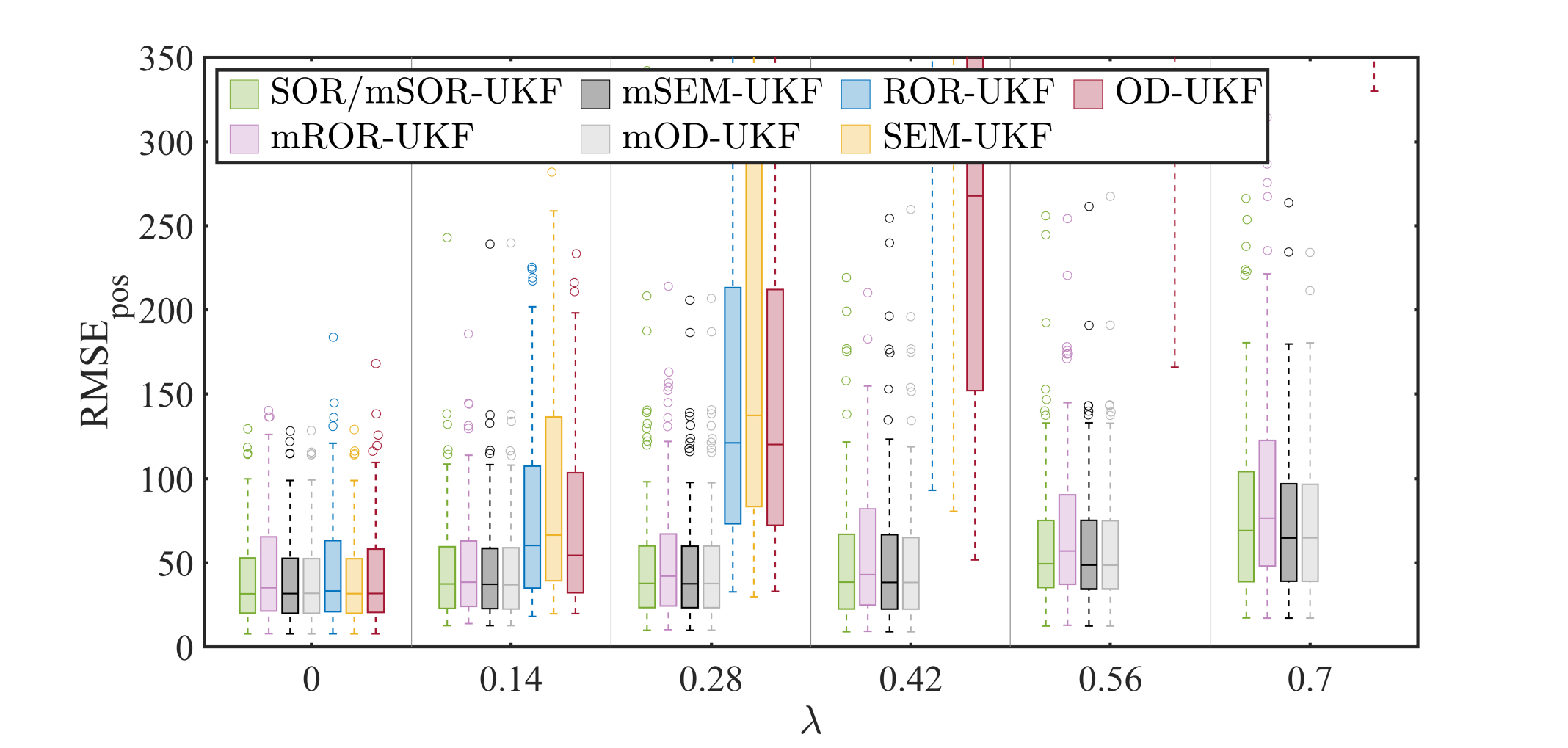}
		\caption{$\text{RMSE}_{\text{pos}}$ vs. $\lambda$, ${\sigma_\theta}\sim \mathcal{U}(1.7\times10^{-3},3.5\times10^{-2})$, ${\sigma_\rho}\sim \mathcal{U}(5,50)$) depicting how changes in nominal noise statistics affect the error performance of different estimators}
		\label{fig:lambda6missingsig}
	\end{figure}

	\begin{figure}[h!]
		\centering
		\includegraphics[width=.8\linewidth]{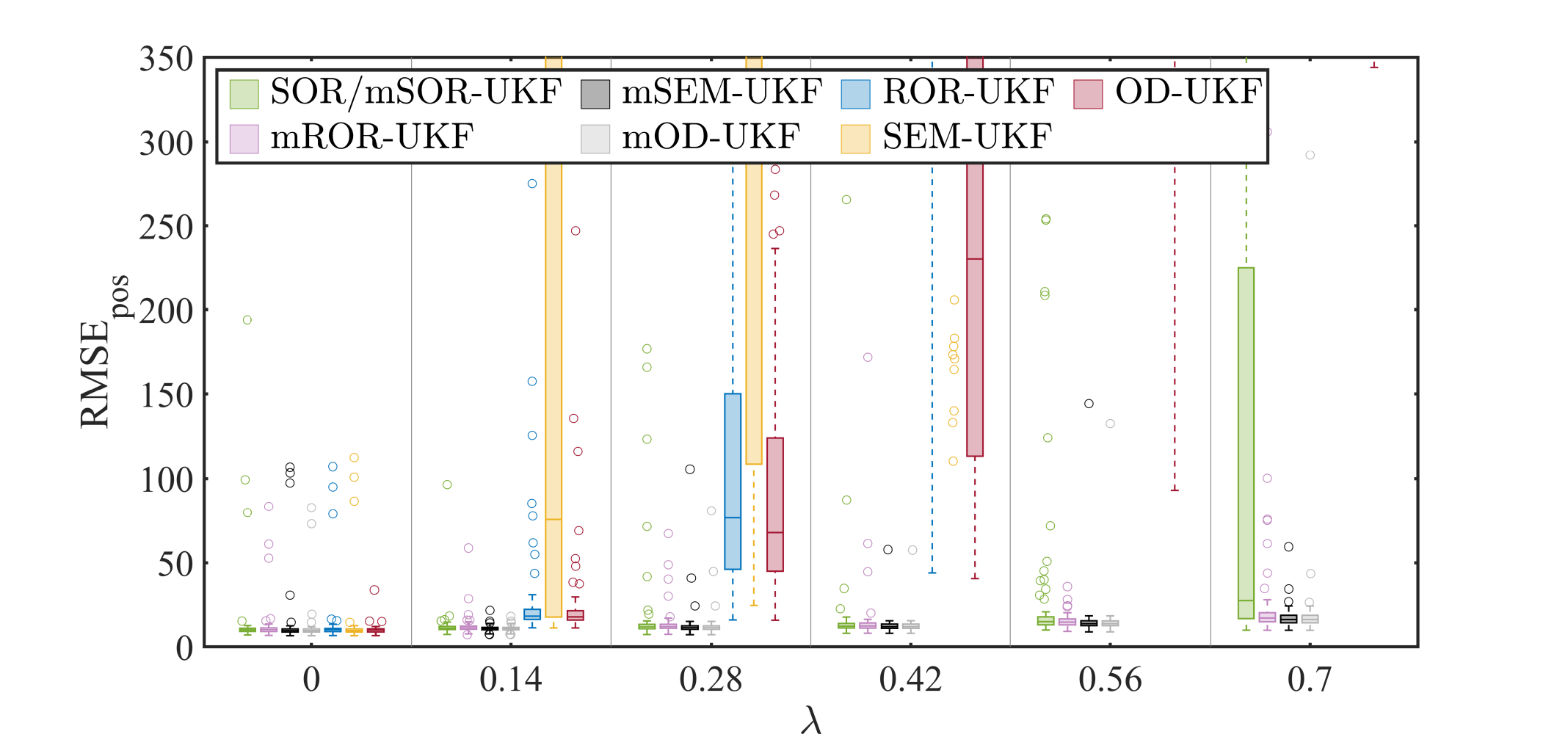}
		\caption{$\text{RMSE}_{\text{pos}}$ vs. $\lambda$, $
			\epsilon \sim \mathcal{U}(10^{-7},10^{-3})$, $\theta_k(i)\sim \mathcal{U}(0.05,0.95)$ depicting the robustness of SOR to changes in hyper-parameters}
		\label{fig:lambda6missingrandeps}
	\end{figure}

	\begin{figure}[h!]
		\centering
		\includegraphics[width=.8\linewidth]{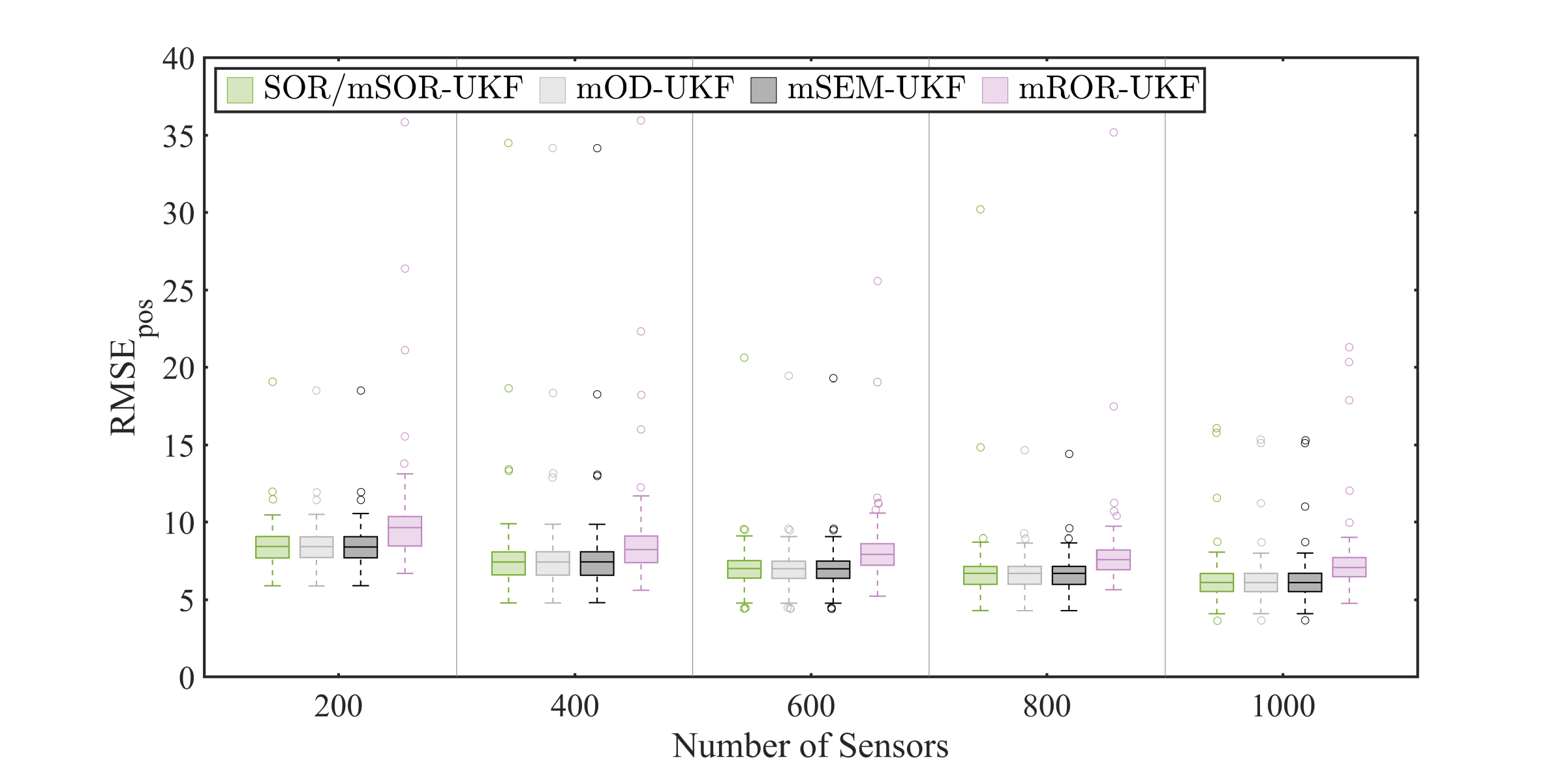}
		\caption{$\text{RMSE}_{\text{pos}}$ vs. the number of sensors ($\lambda=0.9$) }
		\label{fig:missing_lam_pntnine}
	\end{figure}
	
	We observe similar results as for the case of outliers. Fig.~\ref{fig:lambda6missing_RMSE1000} shows the tracking error of different algorithms over time with $\lambda=0.3$. We again find the selective methods exhibiting lower} {errors. 
	
	Fig.~\ref{fig:lambda6missing} shows the box plots of $\text{RMSE}_{\text{pos}}$ with change in $\lambda$ which reaffirms the superiority of methods dealing each dimension independently. However, we observe more} {degraded performances of non-selective methods compared to the case of outliers, drawn from a Gaussian mixture model, with increase in $\lambda$ especially the compensation-based methods i.e. SEM-UKF and ROR-UKF. These methods try to extract information from the measurement vector by learning the} {scaling parameters of their respective Gaussian covariance matrices given the abnormalities in data. This may be more suitable for certain applications and is not useful generally for unknown clutter or "nonsense outliers" in words of the} {authors of SEM-UKF. For more detailed discussion where the such non-selective compensating models are more useful the readers are referred to the Discussion Section of SEM-UKF pg. 8 \cite{8869835}. 
	
	Fig.~\ref{fig:lambda6missingsig} shows the box plots of $\text{RMSE}_{\text{pos}}$ with change in $\lambda$ by assuming ${\sigma_\theta}\sim \mathcal{U}(1.7\times10^{-3},3.5\times10^{-2})$ and ${\sigma_\rho}\sim \mathcal{U}(5,50)$ to depict how the change in nominal noise statistics affect the estimation performance. The results have a similar trend as in Fig.~\ref{fig:lambda6missing} with a change that the error magnitude levels increase.

	To test the robustness of the proposed filters with variations in filter parameters we assume $\epsilon \sim \mathcal{U}(10^{-7},10^{-3})$ and $\theta_k(i)\sim \mathcal{U}(0.05,0.95)$. With variation in $\lambda$, box plots of $\text{RMSE}_{\text{pos}}$ are given in Fig.~\ref{fig:lambda6missingrandeps}. Similar to the case of outliers, we observe the proposed algorithms to be quite robust to changes in filter parameters except at larger values of $\lambda$ where these produce more estimation errors.   
	
	As for the case of outliers in the previous case, we evaluate the comparative performance of methods dealing each dimension selectively with increase in $m$ and observe a similar trend for $\lambda=0.9$ as depicted in Fig.~\ref{fig:missing_lam_pntnine}. in this case the estimation quality remains similar for each method.
	
	For the computational complexity we have observed a similar trend for the case of missing observations as reported for the case of outliers as reported in Fig.~\ref{fig:complexity}.
	
	Lastly, note that a case might arise where any actual measurement is close to zero e.g. the target may be very close to any range sensor in the simulation example. For this case, the nominal model would be able to explain the data as a result the measurement would be used for inference instead of being discarded. We have also simulated this case by initializing the target very close to a range sensor's coordinates and repeating the missing observations simulations to test the robustness of the filters. For this case, we arrive at the similar results as reported above for the case of missing data simulations.}

\subsection{Experimental Results}

\begin{figure}[!h]
	\centering
	\includegraphics[width=.5\columnwidth]{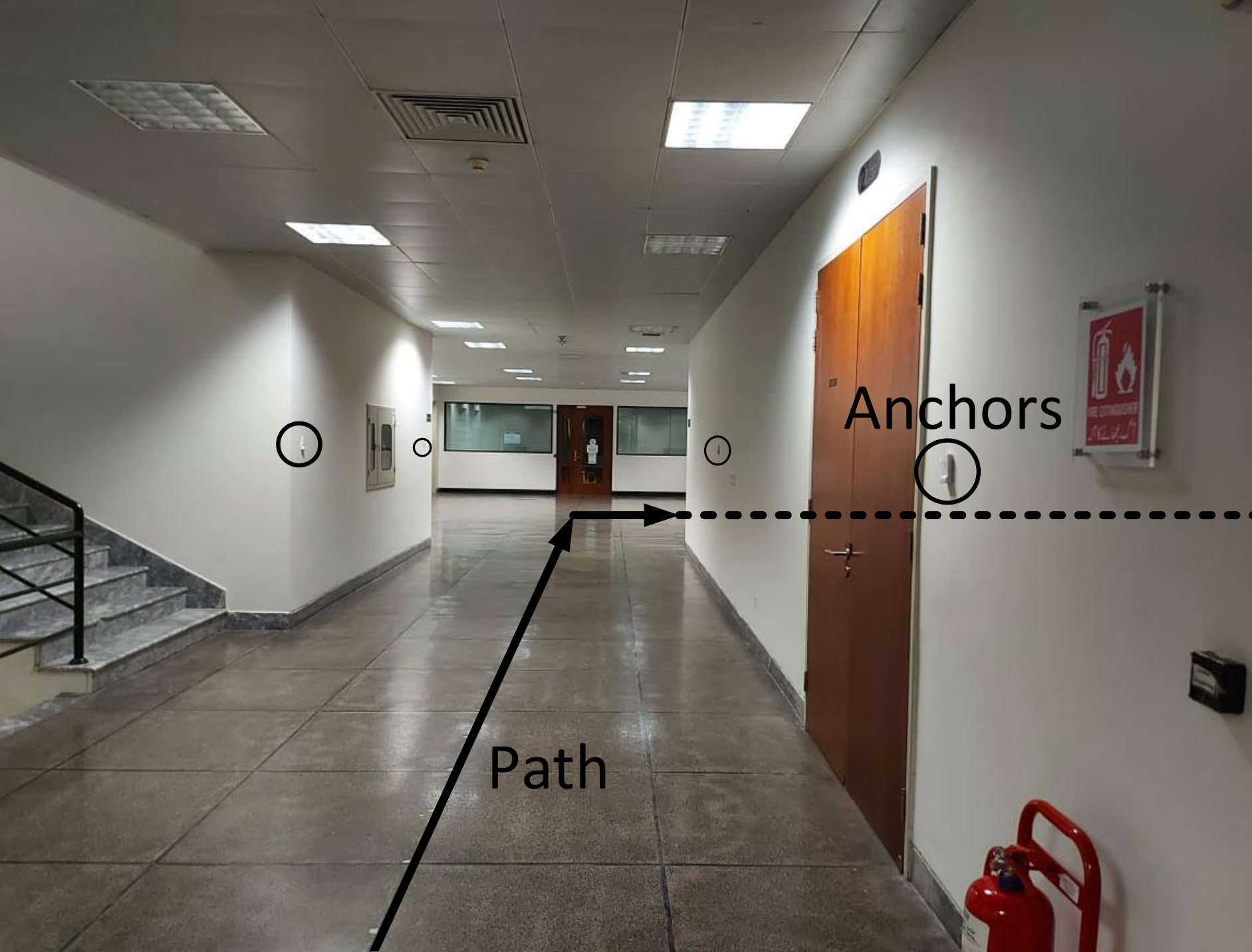} 
	\caption{Experimentation site 1: SSE corridor}
	\label{Exp_sc_results_1}	
\end{figure}

\begin{figure}[!h]
	\centering
	\includegraphics[width=.5\columnwidth]{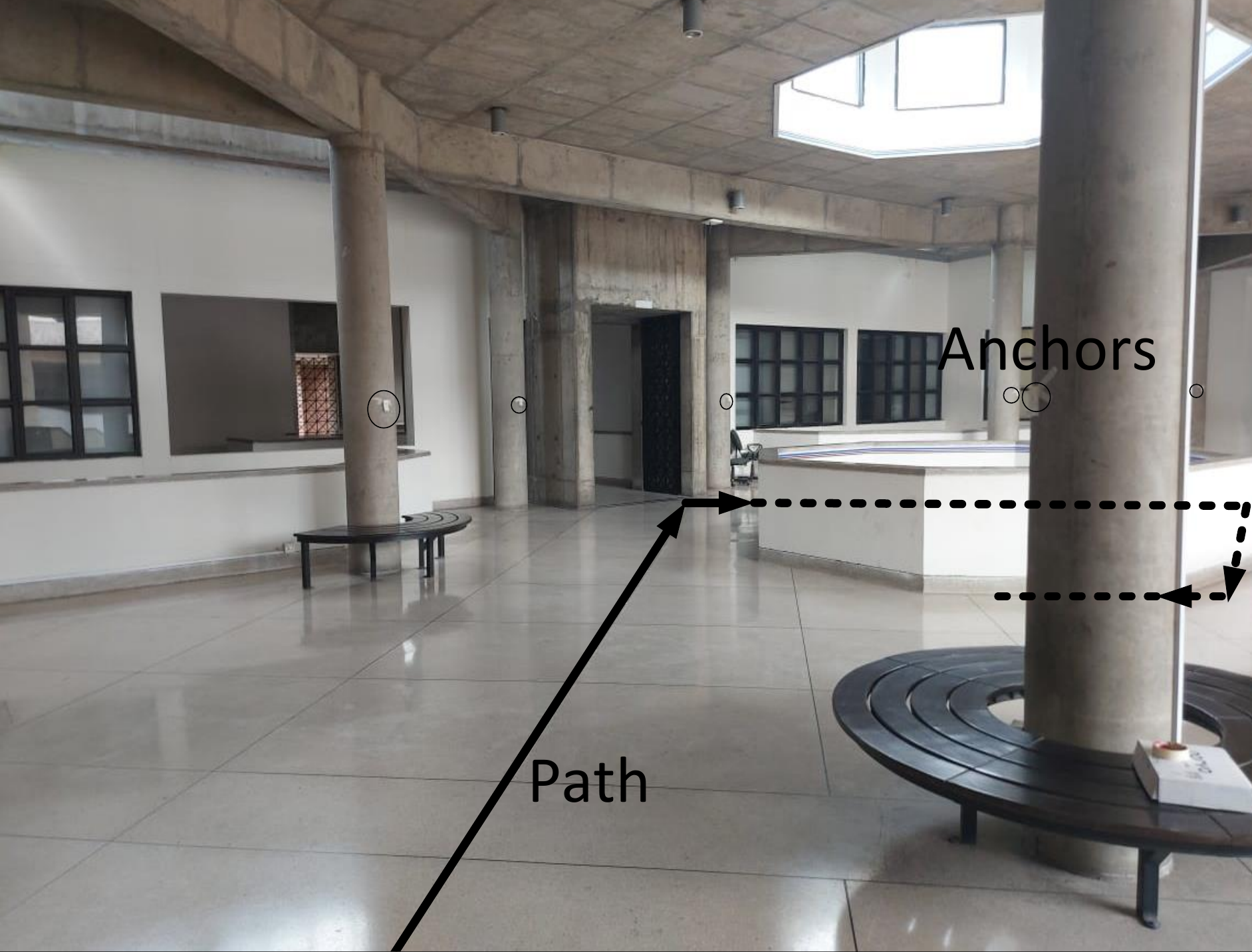}
	\caption{Experimentation site 2: AB corridor} 
	\label{Exp_sc_results_2}	
\end{figure}

\begin{figure}[!h]
	\centering
	\includegraphics[width=.5\columnwidth]{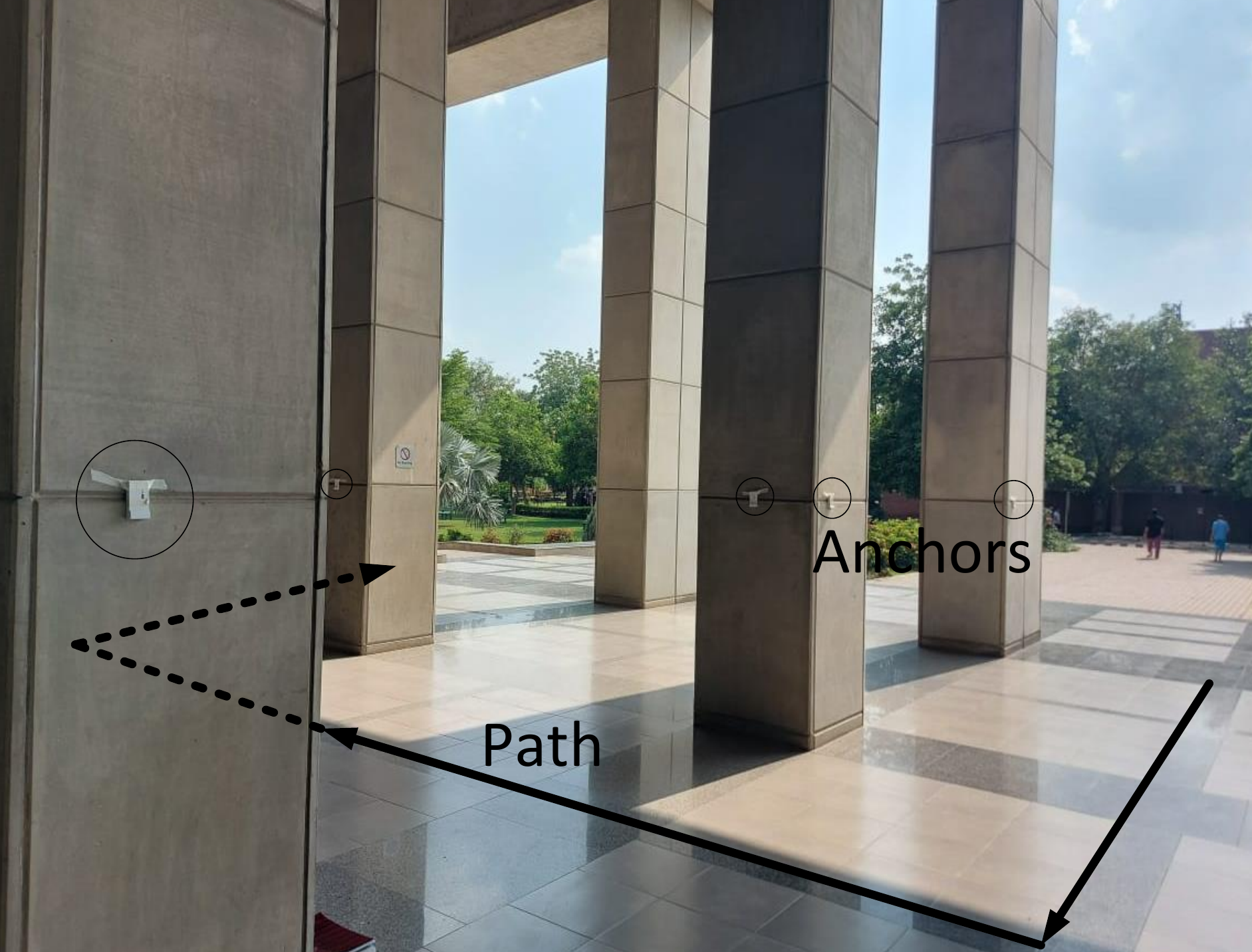}
	\caption{Experimentation site 3: SSE entrance}
	\label{Exp_sc_results_3}
\end{figure}

\subsubsection*{Experimental Campaigns}

\begin{table*}[h!]
	\centering
	\caption{Performance results of different VB-based outlier-robust filters for three experimental settings of indoor localization}
	\resizebox{\columnwidth}{!}{%
	\begin{tabular}{lcccccc} 
		\toprule
		\multirow{2}{*}{} & \multicolumn{2}{c}{~Scenario 1} & \multicolumn{2}{c}{~Scenario 2} & \multicolumn{2}{c}{~Scenario 3}  \\ 
		\cmidrule{2-7}
		& ~RMSE~ & Mean Run Time  & ~RMSE~ & Mean Run Time & ~RMSE~ & Mean Run Time   \\ 
		\midrule
		mROR-UKF          &      {0.16}      &     {0.24}              &    {0.16}      &   {0.21}                 &   {9.26}       &        {0.25}             \\
		mOD-UKF           &    {0.18}      &    {0.18}              &   {0.21}        &      {0.17}              &    {0.39}      &     {0.17}              \\
		mSEM-UKF          &     {0.17}       &      {0.28}              &   {0.11}         &     {0.24}         &     {0.40}     &       {0.23}               \\
		mSOR-UKF          &    {0.15}        &     {0.08}                &           {0.10} &      {0.07}             &   {0.36}       &       {0.06}            \\
		\bottomrule
	\end{tabular}}
	\label{Tab2}
\end{table*}

\hspace{.5cm}
To test the comparative performance of the proposed method in practical settings, we carry out experimental campaigns at three sites for real-time indoor positioning using UWB devices. We use MDEK1001 Development Kit, by Qorvo, which includes 12 UWB units based on the DWM1001 module. {The module’s on-board firmware drives the built-in UWB transceiver to form a network of anchor nodes and perform the two-way ranging exchanges with the tag nodes which enables each tag to compute its relative location to the anchors. For detailed information regarding the kit functionality readers can consult its freely available documentation}. Figs.~\ref{Exp_sc_results_1}~-~\ref{Exp_sc_results_3} show the experimental scenarios at three different locations including a corridor in the School of science and engineering (SSE) building, a corridor in the academic block (AB) and the entrance of the SSE building at the Lahore University of Management Sciences. For each of the experimental scenario, 1 unit is configured as a tag and rest of the 11 units are set in the anchor mode. In each scenario, all the 11 anchors are installed at predefined locations. Moreover, the tag traverses through a predetermined path whose step coordinates are known. The kit is used in its default configuration mode in which at a given time a maximum of 4 range readings are obtained from the closest anchors. The tag is attached to a laptop which logs the range data at each step traversed. The range data is obtained at 5Hz which is subsequently averaged to provide the final sensor readings at each step. The datasets generated from the experimental campaigns are available openly: \url{https://github.com/chughtaiah/UWB_Data}.
\subsubsection*{Sources of Outliers in Sensors' Data}
\hspace{.5cm}
There are two major sources of outliers in the range data. 
\begin{figure}[h!]
	\centering
	\includegraphics[width=.8\columnwidth]{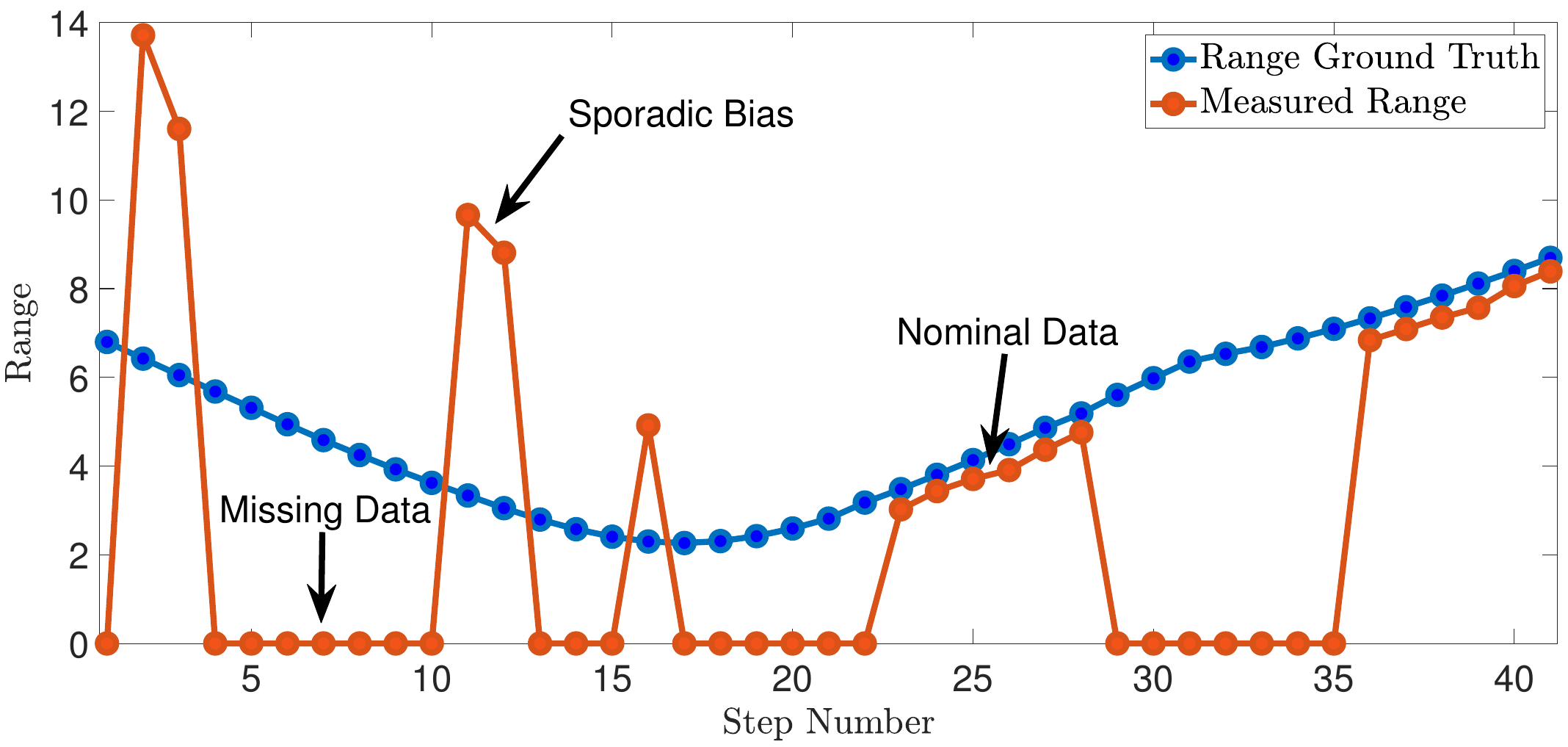}
	\caption{Example of range data corruption obtained from a UWB sensor during experimentation}
	\label{data_outliers}
\end{figure}
Firstly, there are missing observations in the data, since at most 4 out of 11 anchors provide data at a given time. The missing measurements are outliers in the data and contain no useful information. Secondly, UWB range data suffers from bias when the corresponding transceivers face physical obstruction during transmission known as the NLOS condition. Fig. \ref{data_outliers} shows example of how the range data obtained from a UWB sensor node during the experimental campaign is corrupted by the two types of outliers.

\subsubsection*{Performance Results}
\hspace{.5cm}
We consider random walk as the state mobility dynamic model which has historically been used for inference in different applications including mobile nodes in wireless sensor networks \cite{camp2002survey}. Keeping the 2D position of the target as our quantity of interest, the state vector $\mathbf{x}_k= [a_k,b_k]^{\top}$ evolves with $\textbf{f}(.)=\mathbf{I}$ in \eqref{eqn_model_1}. Moreover, the nominal measurement model, $\textbf{h}(.)$ in \eqref{eqn_model_2}, has a functional form of range data as in \eqref{eqn_res3b} including the term for the $z$-axis for the locations of the tag and the anchors. In addition, $\mathbf{Q}_{k-\text{1}}$ and $\mathbf{R}_{k}$ are diagonal matrices with entries as 0.1. For each case, we set $\mathbf{x}_0= [0,0]^{\top}$ {and initialize the filters randomly with ${\hat{\textbf{x}}}^+_{0} \sim \mathcal{N}(\mathbf{x}_0,\mathbf{P}^+_{0}$) where  $\mathbf{P}^+_{0}= 0.5 \mathbf{I}$. We} carry out 100 independent MC runs, keeping all other {applicable} parameters for the methods under consideration same as the {base parameters}.

\begin{figure}[!h]
	\centering
	\includegraphics[width=.8\columnwidth]{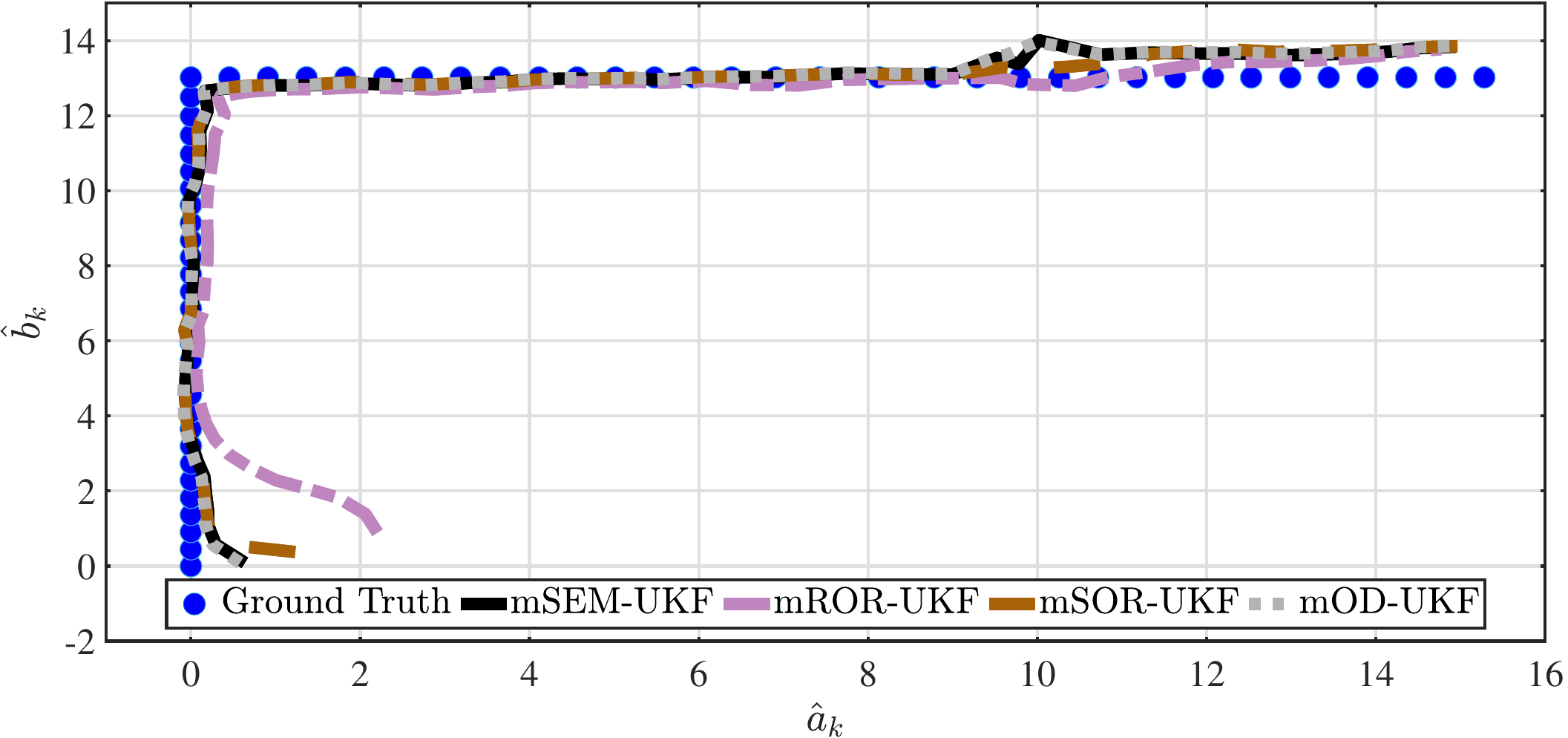}
	\caption{Tracking performance for Case 1}
	\label{Exp_results_1}
\end{figure}

\begin{figure}[!h]
	\centering
	\includegraphics[width=.8\columnwidth]{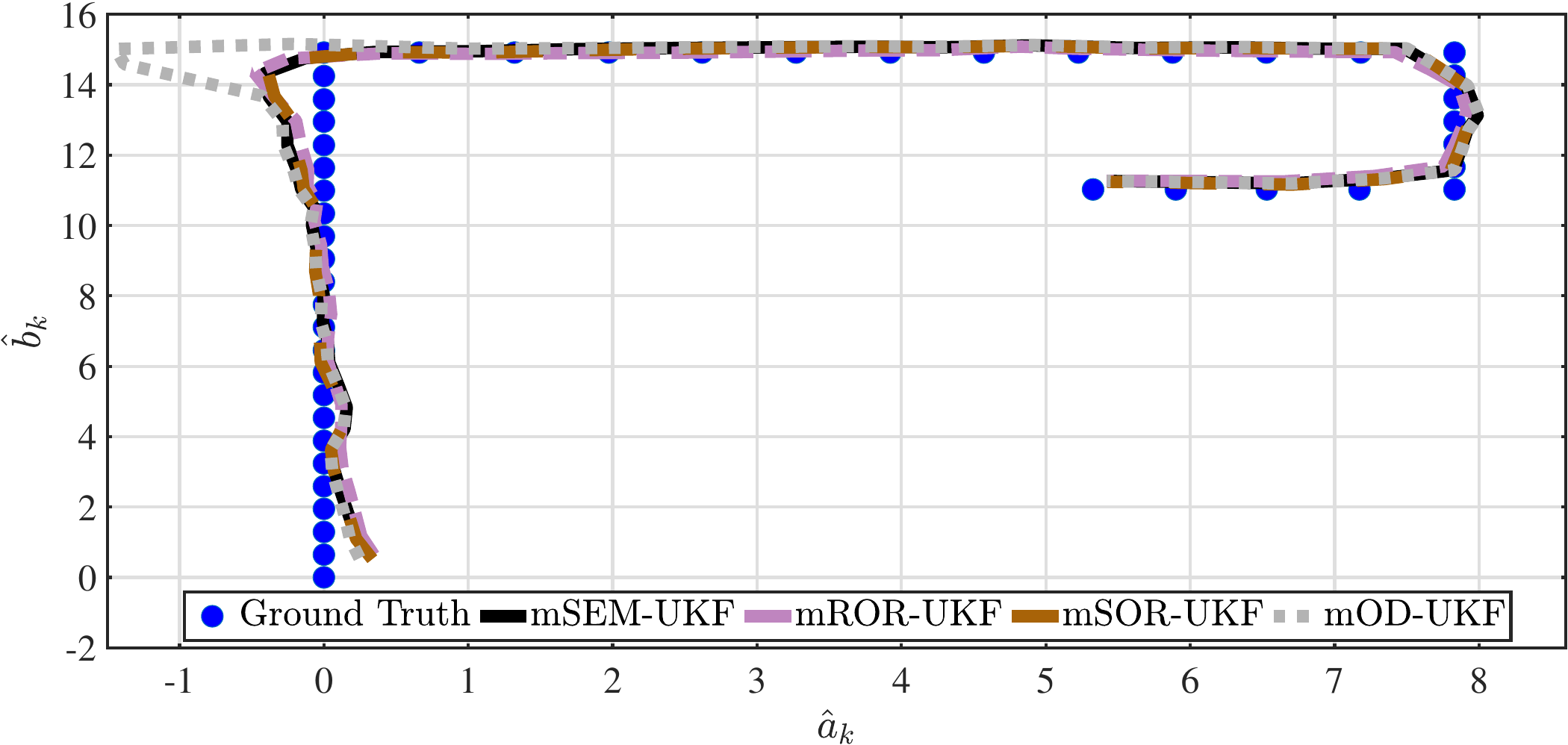}
	\caption{Tracking performance for Case 2}
	\label{Exp_results_2}
\end{figure}

\begin{figure}[!h]
	\centering
	\includegraphics[width=.8\columnwidth]{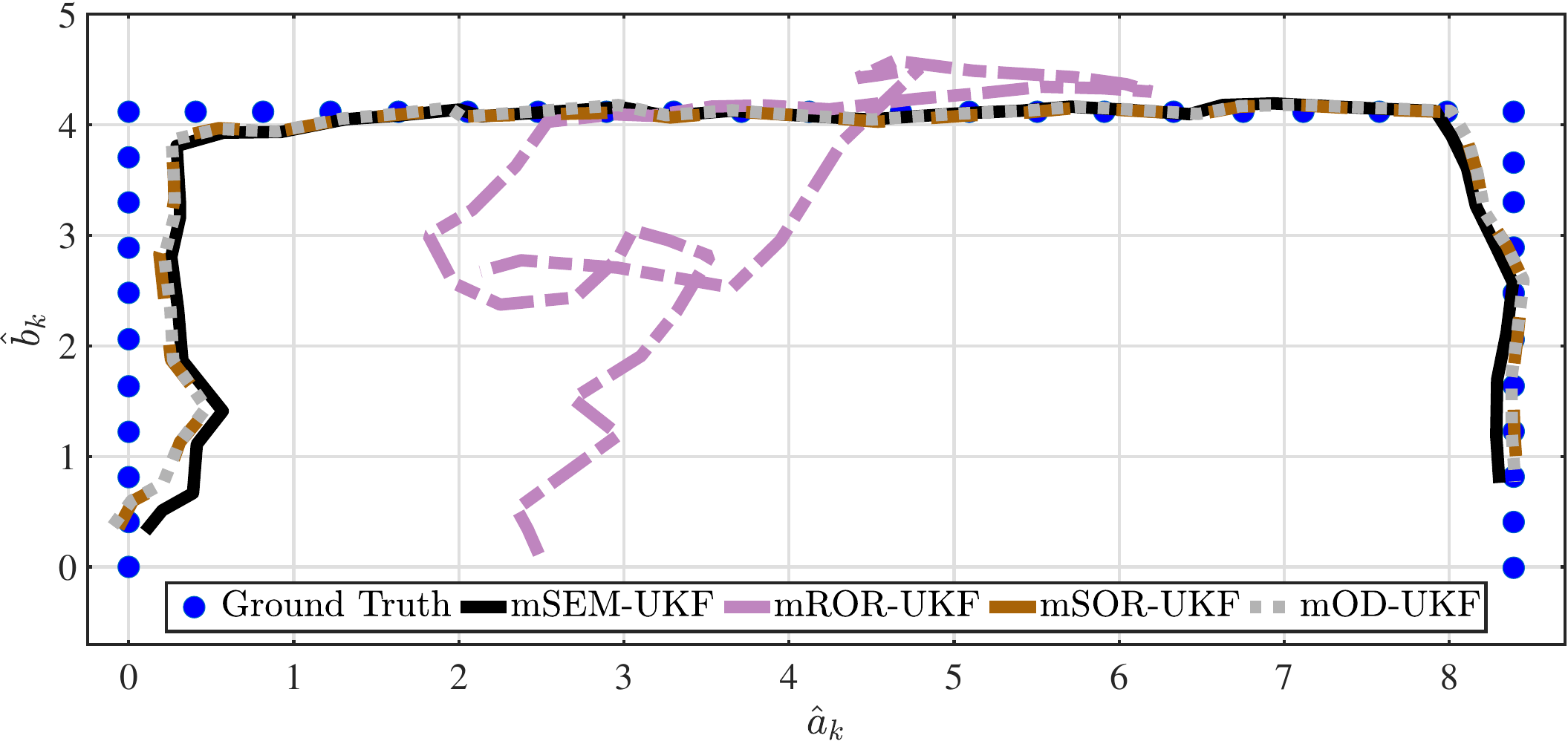}
	\caption{Tracking performance for Case 3}
	\label{Exp_results_3}
\end{figure}

We consider only the methods dealing outliers selectively with least computational complexity. Figs. \ref{Exp_results_1}~-~\ref{Exp_results_3} show the tracking performance of these filters. We observe that for cases 1 and 2 the methods exhibit similar tracking performances since the measurement mostly have missing observations as outliers for these cases. For case 3, the since data of some anchors is unusually contaminated with sporadic bias, in addition to the missing observations, the tracking error is larger in this case. In fact, tracking performance of mROR-UKF severely degrades for the third case due to unusual data corruption. 

Table \ref{Tab2} summarizes the performance evaluation results of the considered algorithms in the experimental settings in terms of RMSE and average computational run time for each case. We notice that the achieved RMSE in each case, normally remaining in the sub-meter range, is comparative for each algorithm. There is an exception for Scenario 3 where mROR-UKF exhibits larger errors due to more NLOS conditions resulting in occurrence of more bias in data. The proposed mSOR-UKF leads in terms of RMSE error and more importantly has the least processing overhead indicating its practical usefulness in real-world scenarios.

\section{Conclusion}\label{Concl}
\hspace{.5cm}
In this chapter, we have considered scenarios where a set of independent sensors provide observations for dynamical systems. We propose to model the outliers independently in each dimension for such cases and devise an outlier-robust filter, resulting in selective rejection of corrupted measurements during inference. In addition, we propose modifications to the existing tractable learning-based outlier robust filters to deal outliers selectively. Also a modification to the proposed method is presented which yield lower computational complexity. Simulations reveal that the techniques which treat outliers selectively exhibit comparative estimation quality which is better as compared to the other methods. Moreover, the theoretical computational overhead is verified during simulations. Lastly, experimentation in various indoor localization scenarios, using UWB modules, suggests the practical efficacy of the proposed method. The gains obtained in terms of computational overhead can be critical where 
\begin{enumerate}
	\item The data is obtained from a large number of sensors and data acquisition rates are very high. 
	\item The processing power is limited.  
	\item Energy savings are of prime concern for example in battery operated devices.
\end{enumerate}

\chapter{Extending Filtering Results to General Estimation Problems} \label{chap-5}
\hspace{.5cm}
In this chapter, we extend the filtering results from Chapter \ref{chap-4} to general nonlinear state estimation problem. Specifically we consider the problem of spatial perception which is a key task in several applications in several field e.g. robotics, intelligent systems, computer vision etc. In the presence of outliers the standard nonlinear least squared formulation results in poor estimates of the system state. Several methods have been considered in the literature to improve the reliability of the estimation process. Most methods are based on heuristics since guaranteed global robust estimation is not generally practical due to high computational costs. Recently general purpose robust estimation heuristics have been proposed that leverage existing non-minimal solvers available for the outlier-free formulations without the need for an initial guess. In this chapter, we propose two similar heuristics backed by Bayesian theory. We evaluate these heuristics in practical scenarios to demonstrate their merits in different applications including 3D point cloud registration, mesh registration and pose graph optimization. 

\section{Problem Setup}\hspace{.5cm}
Consider the following model where the information for inference is available in form of noisy observations which can generally be represented as transformations of the variable given as
\begin{equation}
	\mathbf{y}^i = \mathbf{h}^i(\mathbf{x})+\mathbf{\epsilon}^i \label{eq1}
\end{equation}
where the \textit{i}th measurement $\mathbf{y}^i$ ($i=1,\hdots,m$) is expressed as a known nonlinear function $\mathbf{h}^i(.)$ of the unknown variable of interest $\mathbf{x}$ corrupted by random noise $\mathbf{\epsilon}^i$. Under the assumption that $\mathbf{\epsilon}^i$, described with zero-mean Gaussian noise statistics with the precision matrix $\bm{\Omega}^i$ (inverse of the covariance matrix), is uncorrelated across each measurement channel, the MAP estimate has the following equivalent least square formulation \cite{7747236}
\begin{equation}
	\underset{\mathbf{x} \in \mathcal{X}}{\operatorname{argmin}} \sum_{i=0}^m\left\|\mathbf{y}^i-\mathbf{h}^i(\mathbf{x})\right\|_{\bm{\Omega}^i}^2=\underset{\mathbf{x} \in \mathcal{X}}{\operatorname{argmin}}\sum_{i=0}^m {\big(r(\mathbf{y}^i, \mathbf{x})(i)\big)}^2 \label{eq2}
\end{equation}
where $\mathcal{X}$ denotes the domain of $\mathbf{x}$, the notation $\|\mathbf{e}\|_{\bm{\Omega}}^2={\mathbf{e}}^{\top} \bm{\Omega} \mathbf{e}$ and the function $r(\mathbf{y}^i, \mathbf{x})(i)$ is the residual error for the \textit{i}th measurement. In \eqref{eq2} the residual corresponding to $i=0$ incorporates the regularizing term considering a Gaussian prior for $\mathbf{x}$. The cost function in \eqref{eq2} leads to brittle estimates in face of measurement outliers owing to unwanted over-fitting to the corrupted data. Owing to the underlying functional nonlinearities and nonconvexity of the domain, even solving \eqref{eq2} globally can be challenging which does not induce any robustness to outliers. However, several estimators have been devised in this regard for various applications including point cloud registration, mesh registration, pose graph optimization \cite{horn1987closed,8100078,rosen2019se} etc. These are commonly termed as \textit{non-minimal} solvers which utilize all the measurements for estimation. On the other hand, \textit{minimal} solvers use the smallest number of observations for estimation \cite{yang2020graduated}.

In this chapter, we extend two Bayesian methods for robust nonlinear estimation. These include the recursive outlier-robust (ROR) method \cite{6349794} and the selective observations rejecting (SOR) method (presented in Chapter \ref{chap-4}). We build on these two methods for the general nonlinear robust estimation problem enabling them to use existing non-minimal estimators in contrast to existing Bayesian approaches. To this end, we consider point estimates for $\mathbf{x}$ and use the EM framework. Since EM can be viewed as a special case of VB the adaption is possible. The resulting approaches have a similar structure as the state-of-the-art graduated non-convexity GNC methods \cite{yang2020graduated} with alternating variable and weight update steps.



\section{Relevant Bayesian methods and tools}\label{Sec_Bayes}\hspace{.5cm}
We now briefly discuss the background of related Bayesian methods focusing on the motivation of selection of the particular techniques. Moreover, we provide a short primer on the Bayesian tools that we leverage for devising the derivative methods in the upcoming section.   
\subsection{Choice of the Bayesian methods for robust estimation}\hspace{.5cm}
Recently, ROR and SOR methods have been successfully applied for devising robust nonlinear filtering techniques with satisfactory performance results. In these methods, estimation of $\mathbf{x}$ in \eqref{eq1} relies on modifying the measurement noise statistics. The choice is motivated by the inability of the nominal Gaussian noise to describe the data in face of outliers. Subsequently, the noise parameters are jointly estimated with $\mathbf{x}$. Bayesian theory offers attractive inferential tools for estimating the state and parameters jointly enabling iterative solutions \cite{cbst_book,sarkka2023bayesian}. We adapt these methods to the general nonlinear estimation context of \eqref{eq1} and \eqref{eq2} where non-minimal solvers for estimation are available for the outlier-free cases. The motivation for choosing these particular methods is twofold. First, the formulations of these filters lend their modification conveniently to the nonlinear problem at hand. Moreover, owing to the modeling simplicity, the choice of the hyperparameters for the noise statistics is intuitive for adaptation to our case.

\subsection{Expectation-Maximization as a special case of variational Bayes}\hspace{.5cm}
We aim to use non-minimal solvers, which have been developed and tested for different applications, for solving \eqref{eq2}. To that end, we need to cast the ROR and SOR algorithms in a way that existing nonlinear least squared solvers are invoked during inference. These Bayesian methods are devised using VB which leads to distributions for the state and parameters. However, the available solvers generally result in point estimates for the state. Therefore, to enable adoption of these Bayesian approaches for robust spatial perception applications, we adopt the EM method which as shown in the Bayesian literature can be viewed as a special case of the VB algorithm \cite{vsmidl2006variational}. We first present the VB method and then interpret EM method as its special case.

\subsubsection{Variational Bayes}\hspace{.5cm}
Suppose that we are interested in estimating multivariate parameter $\bm{\theta}$ from data $\mathbf{y}$. For tractability we can resort to the VB algorithm considering the mean-field approximation where the actual posterior is approximated with a factored distribution as \cite{murphy2013machine}
\begin{equation}
	p(\bm{\theta}|\mathbf{y})\approx \prod_{j=1}^{J} q(\bm{\theta}^j ) \label{mean_field}
\end{equation} 
where $J$ partitions of $\bm{\theta}$ are assumed with the $j$th partition given as $\bm{\theta}^j$. The VB marginals can be obtained by minimizing the Kullback-Leibler (KL) divergence between the product approximation and the
true posterior resulting in 
\begin{align}
	q(\bm{\theta}^j )&\propto e^{\big( \big\langle\mathrm{ln} ( p(\bm{\theta} |\mathbf{y}))\big\rangle_{ q(\bm{\theta}^{-j} ) } \big) } \ \forall \ j \label{VB_update}
\end{align} 
where ${ q(\bm{\theta}^{-j} ) }=\prod_{k\neq j} q(\bm{\theta}^k )$. The VB marginals can be obtained by iteratively invoking \eqref{VB_update} till convergence. 

\subsubsection{Expectation-Maximization}\hspace{.5cm}
From the Bayesian literature, we know that the EM method can be viewed as a special case of the VB algorithm considering point densities for some of the factored distributions in \eqref{mean_field}. In particular, the factored distributions which are assumed as point masses in \eqref{mean_field} can be written as delta functions
\begin{align}
	q(\bm{\theta}^n)=\delta(\bm{\theta}^n-\hat{\bm{\theta}}^n ) 
\end{align} 
with $n$ denoting the indices where such assumption is taken. Resultingly, we can update the parameter of $q(\bm{\theta}^n)$ using as  
\begin{equation}
	\hat{\bm{\theta}}^n=\underset{ \bm{\theta}^n} {\operatorname{argmax}} \big\langle\mathrm{ln} ( p(\bm{\theta} |\mathbf{y}))\big\rangle_{ q(\bm{\theta}^{-n} ) } \forall \ n	\label{M_step}
\end{equation}
\eqref{M_step} is formally known as the M-Step of the EM method. The remaining factored distributions not considered as point masses can be determined using \eqref{VB_update} where the expectation with respect to $q(\bm{\theta}^n)$ would simply result in sampling $\mathrm{ln} ( p(\bm{\theta} |\mathbf{y}) $ at $\hat{\bm{\theta}}^n$. This is formally called as the E-Step in the EM method.

Treating EM as particular case of VB allows us another advantage in addition to leveraging the existing non-minimal point estimators for the system state. It allows us the liberty to treat those parameters with point masses where the expectation evaluation with respect to that parameter would otherwise be unwieldy.

%

\section{Proposed Algorithms}\label{propose_algos}\hspace{.5cm}
Having chosen the two particular methods for application in robust perception tasks and having interpreted EM as a special case of VB we are in a position to present our proposals. In this section, we first present the standard ROR method and discuss its limitations. Investigating its drawbacks, we present its modified version. Then we shift our attention to the SOR method. We present the standard SOR technique and explain its shortcomings. Based on the drawn insights, we present two variants of the SOR method.
\subsection{ROR Methods}

\subsubsection{Standard ROR}
\hspace{.5cm}
We use the version of the ROR method as originally reported in Section 2.5 of \cite{6349794} where conditionally independent measurements are considered. The ROR method, as originally reported, assumes $\mathbf{y}^i$ (measurement from each channel) to be scalar but it can be a vector in general. Accordingly, the likelihood to robustify \eqref{eq1}  is the multivariate Student-t density. We denote the distribution as ${\mathrm{St(\mathbf{z}_s|\bm{\phi}_s,\mathbf{\Sigma}_s,\eta)}}$ where the random vector $\mathbf{z}_s$ obeys the Student-t density and the parameters include $\bm{\phi}_s$ (mean), $\mathbf{\Sigma}_s$ (scale matrix) and  $\eta$ (degrees of freedom) which controls the kurtosis or heavy-tailedness. Resultingly, we can write the likelihood density as \cite{6349794}
\begin{align}
	p(\mathbf{y}^i|\mathbf{x})&={\mathrm{St}}\big(\mathbf{y}^i|\mathbf{h}^i(\mathbf{x}),{\bm{\Omega}^i}^{-1},\nu(i) \big)=\int p(\mathbf{y}^i|\mathbf{x},\lambda(i))p(\lambda(i))  d\lambda(i) \label{eq_st1}
\end{align}
with the conditional likelihood following the multivariate Gaussian density given as
\begin{equation}
	p(\mathbf{y}^i|\mathbf{x},\lambda(i))=\mathcal{N}(\mathbf{y}^i|\mathbf{h}^i(\mathbf{x}),(\lambda(i) \bm{\Omega}^i)^{-1})
\end{equation}
where $\mathcal{N}(\mathbf{z}_n|\bm{\phi}_n,\mathbf{\Sigma}_n)$ symbolizes that the random vector $\mathbf{z}_n$ follows the Gaussian distribution parameterized by $\bm{\phi}_n$ (mean) and $\mathbf{\Sigma}_n$ (covariance matrix). $\lambda(i)$ in \eqref{eq_st1} obeys the univariate Gamma distribution given as \cite{6349794} 
\begin{equation}
	p(\lambda(i))=\mathcal{G}(\lambda(i)|\frac{\nu(i)}{2},\frac{\nu(i)}{2})
\end{equation}
where $\mathcal{G}({z}_g|a_g,b_g)$ denotes that the random variable $z_g$ follows the Gamma distribution with the shape parameter $a_g$ and the rate parameter $b_g$ \cite{murphy2007conjugate}.  The normalizing constant of the distribution is denoted as
\begin{equation}
	f(a,b)=\frac{b^a}{\Gamma(a)} \label{norm}
\end{equation} 
where ${\Gamma(a)}$ denotes the Gamma function.

Denoting $\bm{\lambda}$ as the vector with ${\lambda(i)}$ its $i$th element, we can write the following using the Bayes theorem
\begin{equation}
	p(\mathbf{x},\bm{\lambda}|\mathbf{y})\propto{p(\mathbf{y}|\bm{{\lambda}},\mathbf{x})	p(\mathbf{x})p(\bm{{\lambda}}) } \label{ROR_1}
\end{equation} 

Resultingly, the log-posterior, can be written as
\begin{align}
	&p(\mathbf{x},\bm{\lambda}|\mathbf{y})= \Big\{\sum_{i=1}^m \Big(-0.5 \lambda(i) {\big( r(\mathbf{y}^i, \mathbf{x})(i)\big)}^2 -0.5 \nu(i) \lambda(i)  \nonumber \\
	& + (0.5(\nu(i)+d)-1) \ln( {\lambda}(i))  \Big) -0.5 {\big(r(\mathbf{y}^{0}, \mathbf{x} )(0)\big)}^2 + constant \Big\} \label{log_ROR}
\end{align}

To proceed further, we seek the following VB factorization of the posterior distribution
\begin{equation}
	p(\mathbf{x},\bm{{\lambda}}|\mathbf{y})\approx q(\mathbf{x}) q(\bm{{{\lambda}}}) \label{ROR_2}
\end{equation}

Based on \eqref{M_step} and \eqref{log_ROR}, the state variable $\hat{\mathbf{x}}$ is estimated using the VB/EM theory as \cite{6349794}
\begin{equation}
	\hat{\mathbf{x}}=\underset{\mathbf{x} \in \mathcal{X}}{\operatorname{argmin}}\sum_{i=0}^m w(i) {\left(r(\mathbf{y}^i, \mathbf{x})(i)\right)}^2 \label{ROR_x}
\end{equation}
where $	w(i)=\langle{ {\lambda}}(i)\rangle_{q({{\lambda}}(i))} \forall\ i>0$. We have considered a point estimator for $\mathbf{x}$ i.e. $q(\mathbf{x})=\delta(\mathbf{x}-\hat{\mathbf{x}})$ to utilize existing solvers for spatial perception tasks mainly available in this form. 

Similarly, the weights can be updated as \cite{6349794}
\begin{equation}
	w(i)=\left({1+{\frac{{ { {\hat{r}^{2}(i)} } }-d}{\nu+d} }}\right)^{-1}\  \forall\ i>0 \label{ROR_w}
\end{equation}
where $d$ denotes the dimension of $\mathbf{y}^i$ and ${ {\hat{r}^{2}(i)} }={\big(r\left(\mathbf{y}^i, \hat{\mathbf{x}} \right)(i)\big)}^2$. We assume $\nu(i)=\nu\ \forall\ i$ and use $w(i)$ to weight the precision matrix instead of $\bar{\lambda}$ as in the original work to remain consistent with comparative works. Since we assume outliers occur only in the measurements, the weight for the regularizing term remains fixed as 1 i.e. $w(0)=1$ in \eqref{ROR_x}.


\subsection*{Limitations of standard ROR} \hspace{.5cm}
Starting with $w(i)=1\ \forall\ i$, the standard ROR invokes \eqref{ROR_x} and \eqref{ROR_w} iteratively till convergence. The technique has shown good performance in filtering context for different practical examples e.g. target tracking \cite{6349794} and indoor localization \cite{chughtai2022outlier}. This can be attributed to the appearance of well-regularized cost functions \eqref{eq2}. However, for advanced problems in robust spatial perception, the performance of standard ROR compromises at high outlier ratios since it fails to capture the outlier characteristics by fixing $\nu$ which governs the heavy-tailedness of noise or the characteristics of outliers. Empirical evidence confirms this observation. 
\begin{figure}[h!]
	\centering
	\includegraphics[width=.6\linewidth]{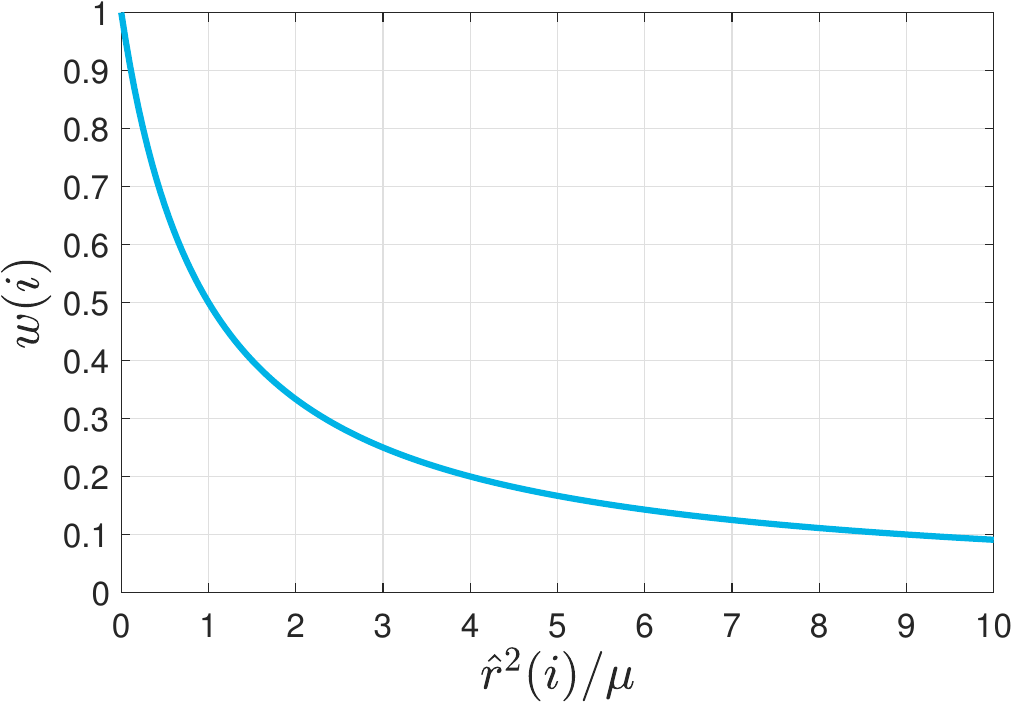}
	\caption{$w(i)$ vs ${\hat{r}^{2}(i)}/\mu$ in ROR.}
	\label{fig:rorweights}
\end{figure}

For further understanding of this limitation consider how $w(i)$ changes with parameters of Student-t distribution. The variation of $w(i)$ against ${\hat{r}^{2}(i)}/\mu$ is shown in Fig.~\ref{fig:rorweights} where  $\mu=\nu+d$. The plot of \eqref{ROR_w} would only be a shifted version of the plot in Fig.~\ref{fig:rorweights}. Since $\hat{r}^{2}(i) \gg d$ when outlier appears in the \textit{i}th dimension, we can simplify \eqref{ROR_w} as $w(i)=({1+{\frac{{\hat{r}^{2}(i)}}{\mu} }})^{-1} $ indicating the importance of the kurtosis $\nu$ and resultingly $\mu$. It can be observed that the residuals are gradually downweighted with increasing magnitude during estimation with $w(i)=0.5$ for ${\hat{r}^{2}(i)}=\mu$. Since the residuals are evaluated considering the state estimate using all the measurements initially, it is possible that the squared residuals even for the uncorrupted dimensions become greater than the prefixed $\mu$ downweighting them in the process. This can lead to performance issues. Therefore, this calls for adapting $\mu$ by considering the residuals evaluated with clean and corrupted measurements. 
Given the limitations, we propose a modification in the standard ROR method by adapting $\mu$ during iterations considering the residuals at each iteration. We call it Extended ROR or simply EROR.

\subsubsection{EROR}
\hspace{.5cm}
In EROR, we propose adaptation of $\mu$ during iterations considering the updated squared residuals based on the relationship of the $w(i)$ and $\mu$. The choice of $\mu$ is done such that weights assigned to residuals span the maximum portion of the zero to one range. In other words, the largest residuals need to be pruned with the smallest weights in the weighted least squared cost function and vice versa. In particular, for ${{\hat{r}^{2}(i)}}=\max(\hat{\mathbf{r}}^{2}) $ (s.t. $i>0$) with $w(i)\rightarrow0$ leads to $\mu\rightarrow0$ ($\mu=\frac{{\hat{r}^{2}(i)}}{1/{w(i)}-1}$). Practically, $\mu \ll \max(\hat{\mathbf{r}}^{2}) $. Similarly for the other extreme ${{\hat{r}^{2}(i)}}=\min(\hat{\mathbf{r}}^{2}) $ (s.t. $i>0$) with $w(i)\rightarrow1$ leads to $\mu\rightarrow\infty$. Practically, $\mu \gg \min(\hat{\mathbf{r}}^{2}) $. To cater for both extremes we propose $\mu= \mathrm{mean}(\max(\hat{\mathbf{r}}^{2}) ,\min(\hat{\mathbf{r}}^{2}) )$. Also, $\mu$ is lower bounded by $\chi$ to ensure residuals within a minimum threshold are not neglected during estimation. The notion of $\chi$ is similar to $\bar{c}^2$ as in \cite{yang2020graduated} which is set as the maximum error expected for the inliers. Note that adaptation of $\mu$ is intuitive which can be viewed as an additional step in the standard ROR devised using VB. EROR is presented as Algorithm \ref{Algo1}.


\begin{algorithm}[h!]
	\SetAlgoLined
	Initialize $w(i)=1\ \forall\ i  $\\
	
	\While{$\textnormal{not converged}$}{
		Variable update: $\hat{\mathbf{x}}=\underset{\mathbf{x} \in \mathcal{X}}{\operatorname{argmin}}\sum_{i} w(i) {(r\left(\mathbf{y}^i, \mathbf{x}\right) (i))}^2 $\\
		Residual update: ${\hat{r}^{2}(i)}={\left(r\left(\mathbf{y}^i, \hat{\mathbf{x}} \right)(i)\right)}^2$ $\forall\ i$\\
		Parameteric update: \\
		${\hat{r}_{\max}^{2}}= \max({\hat{\mathbf{r}}^{2} })\ \mathrm{s.t.}\ i>0 $\\
		${\hat{r}_{\min}^2 }= \min({\hat{\mathbf{r}}^{2} })\ \mathrm{s.t.}\ i>0 $ \\
		$\mu=\max(\mathrm{mean}({\hat{r}_{\max}^2},{\hat{r}_{\min}^2 }),\chi)$\\		
		Weight update: $w(i)=\frac{1}{1+({ {\hat{r}^{2}(i) } } /{\mu^{}}) }$ $\forall\ i>0$\\
	}
	\caption{The proposed estimator: EROR}
	\label{Algo1}
\end{algorithm}


\subsection{SOR Methods}
\subsubsection{Standard SOR}
\hspace{.5cm}
The SOR method in Chapter \ref{chap-4}, assumes $\mathbf{y}^i$ to be scalar but it can be a vector in general. In the original work, an indicator vector $\bm{\mathcal{I}}\in\mathbb{R}^m$ with Bernoulli elements is introduced to describe outliers in the measurements. In particular, ${{\mathcal{I}}}(i)=\epsilon$ indicates the occurrence of an outlier in the \textit{i}th dimension and ${{\mathcal{I}}}(i)=1$ is reserved for the no outlier case. Accordingly, the conditional likelihood to robustify \eqref{eq1} is a multivariate Gaussian density function
\begin{align}
	&p(\mathbf{y}^i|\mathbf{x},{{\mathcal{I}}}(i))= {\mathcal{N}}\Big(\mathbf{y}^i|\mathbf{h}^i(\mathbf{x}),({{\mathcal{I}}}(i) \bm{\Omega}^i)^{-1} \Big)=\frac{1}{\sqrt{ {(2 \pi)^{m}{|{\bm{\Omega}^i}^{-1}|}}  } }  e^{\left( {-}0.5 {{\mathcal{I}}}(i) {\left( r(\mathbf{y}^i, \mathbf{x})(i) \right)}^2 \right)} {{\mathcal{I}(i)}^{0.5}} \label{SOR_like}
\end{align}

${\mathcal{I}(i)}\ \forall \ i>0$ is assumed to have the following prior distribution
\begin{equation}
	p({{\mathcal{I}}}(i))=(1-{\theta(i)}) \delta({{{\mathcal{I}}}(i)}-\epsilon)+{\theta(i)}\delta( {{{\mathcal{I}}}(i)}-1)
\end{equation}
where $\theta(i)$ denotes the prior probability of having no outlier in the $i$th measurement channel. $\epsilon$ has the role of catering for describing the anomalous data in effect controlling the covariance of outliers. Using the Bayes theorem we can write 
\begin{equation}
	p(\mathbf{x},\bm{\mathcal{I}}|\mathbf{y})\propto{p(\mathbf{y}|\bm{\mathcal{I}},\mathbf{x})	p(\mathbf{x})p(\bm{\mathcal{I}})}
\end{equation} 
where $\bm{\mathcal{I}}$ denotes the vector with ${\mathcal{I}(i)}$ its $i$th element. 

As a result, the log-posterior is given as
\begin{align}
	&\ln (p(\mathbf{x},\bm{\mathcal{I}},b|\mathbf{y}))= \Big\{\sum_{i=1}^m \Big(-0.5 {{\mathcal{I}}}(i) {\big(r(\mathbf{y}^i, \mathbf{x})(i)\big)}^2 + 0.5 \ln( {{\mathcal{I}}}(i) ) \nonumber \\
	& + \ln\left( (1-{\theta(i)})\delta( {{{\mathcal{I}}}(i)}-\epsilon)  +{\theta(i)}\delta( {{{\mathcal{I}}}(i)}-1) \right) \Big) -0.5 {\big(r(\mathbf{y}^{0}, \mathbf{x})(0)\big)}^2 \nonumber\\
	&+ constant \Big\} \label{log_ESOR}
\end{align}

For tractable inference we resort to the following VB factorization of the posterior distribution
\begin{equation}
	p(\mathbf{x},\bm{\mathcal{I}}|\mathbf{y})\approx q(\mathbf{x}) q(\bm{{\mathcal{I}}}) \label{VB_SOR}
\end{equation}

Based on \eqref{M_step} and \eqref{log_ESOR}, the state estimate $\hat{\mathbf{x}}$ is updated using the VB/EM theory as
\begin{equation}
	\hat{\mathbf{x}}=\underset{\mathbf{x} \in \mathcal{X}}{\operatorname{argmin}}\sum_{i=0}^m w(i) {(r(\mathbf{y}^i, \mathbf{x})(i))}^2 \label{SOR_x}
\end{equation}
with $w(0)=1$ and 
\begin{align}
	w(i)&=\langle{{\mathcal{I}}}(i)\rangle_{q({{\mathcal{I}}}(i))}\  =\Omega(i)+(1-\Omega(i))\epsilon \approx \Omega(i)\ \forall\ i>0 
\end{align}
where $\Omega(i)$ parameterizes the VB posterior marginal ${q({{\mathcal{I}}}(i))}$ corresponding to $\theta(i)$ in $p({{\mathcal{I}}}(i))$. $\Omega(i)$ is updated using \eqref{VB_update} and \eqref{log_ESOR} as


\begin{align}
	\Omega(i) &=\frac{1}{1+{\sqrt{\epsilon}}(\frac{1}{\theta(i)}-1){ e^ { \left( 0.5 {\hat{r}^{2}(i) } (1-\epsilon) \right) } }} 
\end{align}

Since the hyperparameter $\epsilon$ is assumed to be a small positive number we denote it as $\epsilon=\exp({{-{\rho}}^2})$. Assuming a neutral prior for the occurrence of an outlier in the $i$th dimension i.e. ${\theta(i)}=0.5$ and noting that $\epsilon=\exp({{-{\rho}}^2}) \approx 0$ we can write 

\begin{equation}
	w(i)\approx\Omega(i)\approx\frac{1}{1+{ e^{\left(0.5({\hat{r}^{2}(i) }-\rho^{2} ) \right)} }} \forall\ i>0 \label{SOR_w}
\end{equation}

\subsection*{Limitations of standard SOR} 
\hspace{.5cm}
Starting with $w(i)=1\ \forall\ i$, the standard SOR invokes \eqref{SOR_x} and \eqref{SOR_w} iteratively till convergence. The technique has shown good performance in the filtering context but has limited ability in robust spatial perception tasks as the standard ROR method. This drawback compromises the performance especially at high outlier ratios since the standard SOR fails to capture the outlier characteristics by fixing $\rho^2$ (or $\epsilon$) which governs the covariance of outliers. Experimental evidence verifies this limitation. 

\begin{figure}[h!]
	\centering
	\includegraphics[width=.6\linewidth]{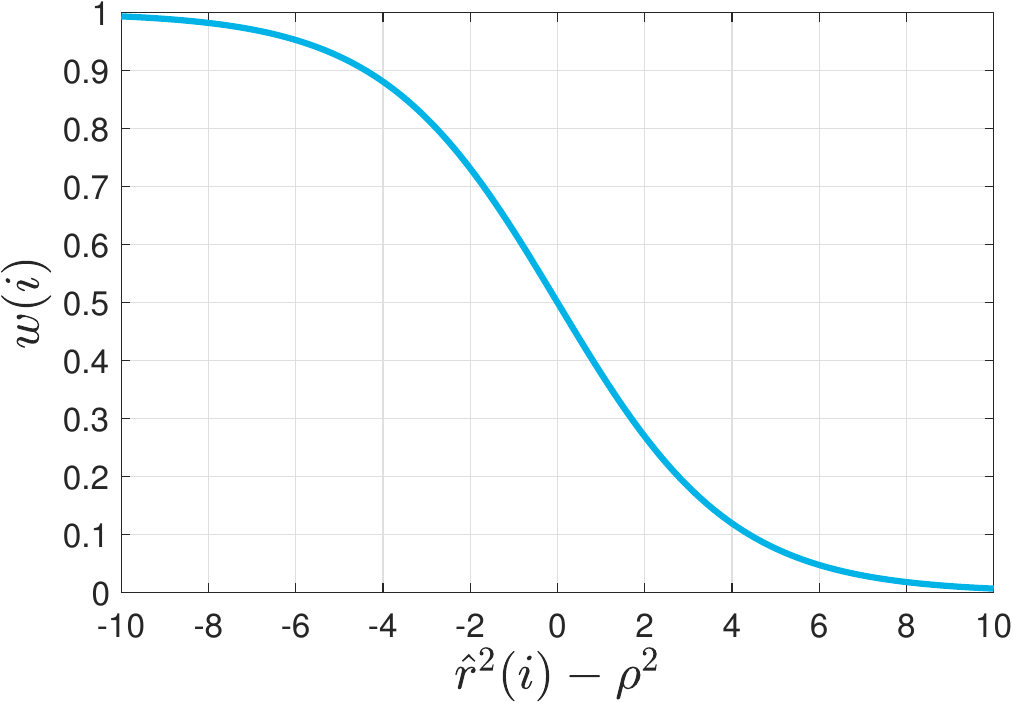}
	\caption{$w(i)$ vs ${\hat{r}^{2}(i)}-\rho^2$ in SOR.}
	\label{fig:sorweights}
\end{figure}

To further appreciate this limitation consider how $w(i)$ changes with ${\hat{r}^{2}(i)}-\rho^2$ as shown in Fig.~\ref{fig:sorweights}. It can be observed that the residuals are gradually downweighted with increasing magnitude during estimation with $w(i)=0.5$ for ${\hat{r}^{2}(i)}=\rho^2$. Since the residuals are evaluated considering the state estimate using all the measurements initially, it is possible that the squared residuals even for the uncorrupted dimensions become greater than the prefixed $\rho^2$ downweighting them during the inference process. This can lead to performance issues. Therefore, this calls for adapting $\rho^2$ by considering the residuals evaluated with clean and corrupted measurements. 

Given the limitations, we present two methods based on the standard SOR technique. Firstly, we propose a modification in the SOR method by adapting $\rho^2$ during iterations considering the residuals at each iteration. We call it Extended SOR  or simply ESOR. Secondly, by modifying the basic SOR model with a notion of adaptive unknown covariance of outliers, we propose Adaptive SOR or simply ASOR where the parameters controlling the covariance of outliers are \textit{learnt} within the inferential procedure.


\subsubsection{ESOR}
\hspace{.5cm}
In ESOR, we modify $\rho^2$ during iterations taking into account the updated squared residuals. In particular, we propose to select $\rho^2={\sum_{i=0}^m} w(i) {\hat{r}^{2}(i) }/{\sum_{i=0}^m} w(i)$. In other words, $\rho^2$ is selected as the effective centroid of points given as squared residuals considering the weights assigned. With such an intuitive choice, the probability of declaring an outlier in the $i$th dimension is 0.5 when ${\hat{r}^{2}(i)}$ equals the effective mean of squared residuals. The residuals greater than $\rho$ become the candidates for downweighing and vice versa during an iteration. Lastly, $\rho^2$ is lower bounded by $\gamma$ to ensure residuals within a minimum threshold are not neglected during estimation. ESOR is presented as Algorithm \ref{Algo_ESOR}.
\begin{algorithm}[h!]
	\SetAlgoLined
	Initialize $w(i)=1\ \forall\ i  $\\
	
	\While{$\textnormal{not converged}$}{
		Variable update: $\hat{\mathbf{x}}=\underset{\mathbf{x} \in \mathcal{X}}{\operatorname{argmin}}\sum_{i} w(i) {(r(\mathbf{y}^i, \mathbf{x})(i))}^2 $\\
		Residual update: ${\hat{r}^{2}(i)}={\big(r^i\left(\mathbf{y}^i, \hat{\mathbf{x}} \right)\big)}^2$ $\forall\ i$\\
		Parameteric update: $\rho^{2}=\max(\frac{{\sum_i} w(i){\hat{r}^{2}(i)}}{{\sum_i} w(i)},\gamma)$\\
		Weight update: $w(i)=\frac{1}{1+{e^{(0.5({\hat{r}^{2}(i)}-\rho^{2} ))} }}$ $\forall\ i>0$\
	}
	\caption{The proposed estimator: ESOR}
	\label{Algo_ESOR}
\end{algorithm}

\subsubsection{ASOR}
\hspace{.5cm}
For devising ESOR we adapt $\rho^2$ (or $\epsilon$) during iterations to capture the characteristics of outliers. Nevertheless, the choice of selection of $\rho^2$ which controls the covariance of outliers is entirely intuitive. This can be viewed as an additional step within the standard SOR method, not falling under the standard VB approach. However, the insights drawn from ESOR with sound experimental performance suggest the merits of considering or \textit{learning} the characteristics of outliers during inference. With these observations, we now present ASOR which is devised with the standard VB approach. In contrast to EROR and ESOR, we jointly \textit{estimate} the covariance controlling factor along with the state and the weights in ASOR.     

The conditional likelihood for designing ASOR remains same as for the standard SOR given in \eqref{SOR_like}. Building on the insights from EROR and ESOR, we need to adapt covariances to describe the outliers. In particular, we assume that for outlier occurrence in the $i$th observation channel i.e. ${{\mathcal{I}}}(i)\neq 1$ in \eqref{SOR_like}, ${{\mathcal{I}}}(i)$ obeys a Gamma probability density being supported on the set of positive real numbers. Resultingly, we write the hierarchical prior distribution of ${{\mathcal{I}}}(i)$ given as
\begin{align}
	p({{\mathcal{I}}}(i)|b)&=(1-{\theta(i)}) \underbrace{f(a,b) { {{\mathcal{I}}}(i) } ^{a-1}  e^{-b {{{\mathcal{I}}}(i)} }}_{ \mathcal{G}({{\mathcal{I}}}(i)|a,b) } +{\theta(i)}\delta( {{{\mathcal{I}}}(i)}-1) \label{pIb}
\end{align}
where we assume the two components of $p({{\mathcal{I}}}(i)|b)$ in \eqref{pIb} as disjoint with the Gamma density defined as zero for ${{{\mathcal{I}}}(i)}=1$ without losing anything. This assumption helps in subsequent derivation. The parameter $b$ is the factor that captures the common effect of outliers in each observation channel. From the Bayesian theory we know that the conjugate prior of $b$ is also a Gamma distribution given as \cite{fink1997compendium} 
\begin{align}
	p(b)&=f(A,B) {b}^{A-1}  e^{-B b } \label{pb_ASOR}
\end{align}
where $A$ and $B$ are the parameters of this Gamma distribution. We are now in a position to invoke the Bayes theorem given as
\begin{equation}
	p(\mathbf{x},\bm{\mathcal{I}},b|\mathbf{y})\propto {p(\mathbf{y}|\bm{\mathcal{I}},\mathbf{x})	p(\mathbf{x})p(\bm{\mathcal{I}}|b)p(b)}
\end{equation} 

Resultingly, the log-posterior, which is used in subsequently derivation, is given as
\begin{align}
	&\ln (p(\mathbf{x},\bm{\mathcal{I}},b|\mathbf{y}))= \Big\{\sum_{i=1}^m \Big(-0.5 {{\mathcal{I}}}(i) {\left(r(\mathbf{y}^i, \mathbf{x})(i)\right)}^2 + 0.5 \ln( {{\mathcal{I}}}(i) ) \nonumber \\
	& + \ln\left( (1-{\theta(i)}) f(a,b) { {{\mathcal{I}}}(i)} ^{a-1}  e^{-b {{{\mathcal{I}}}(i)} } +{\theta(i)}\delta( {{{\mathcal{I}}}(i)}-1) \right) \Big) \nonumber \\
	& -0.5 {\big(r\left(\mathbf{y}^{0}, \mathbf{x}\right)(0)\big)}^2 + (A-1)\ln(b)- B b + constant \Big\} \label{log_ASOR}
\end{align}

To proceed further we resort to the VB factorization given as
\begin{equation}
	p(\mathbf{x},\bm{\mathcal{I}},b|\mathbf{y})\approx q(\mathbf{x}) q(\bm{{\mathcal{I}}}) q({b})
\end{equation}

Using the VB/EM theory and with the assumption that $q(\mathbf{x})=\delta(\mathbf{x}-\hat{\mathbf{x}})$ we obtain the following resorting to \eqref{M_step} and \eqref{log_ASOR} 
\begin{equation}
	\hat{\mathbf{x}}=\underset{\mathbf{x} \in \mathcal{X}}{\operatorname{argmin}}\sum_{i=0}^m w(i) { \left( r^i(\mathbf{y}^i, \mathbf{x}) \right) }^2 
\end{equation}
where $w(0)=1$ and
\begin{align}
	w(i)&=\langle{{\mathcal{I}}}(i) \rangle_{q({{\mathcal{I}}}(i))} \ \forall\ i>0  \label{w(i)_sora}
\end{align}

Thanks to the notion of VB-conjugacy \cite{vsmidl2006variational}, it turns out that $q(\mathcal{I}(i))$ has a same functional form as of $p({{\mathcal{I}}}(i)|b)$ in \eqref{pIb}. $q(\mathcal{I}(i))$ is parameterized by $\alpha$, $\beta(i)$ and $\Omega(i)$ corresponding to $a$, $b$ and $\theta(i)$ in $p({{\mathcal{I}}}(i)|b)$ respectively. Resultingly, we can evaluate the expectation in \eqref{w(i)_sora} as
\begin{align}
	w(i)=\Omega(i)+(1-\Omega(i))\alpha/\beta(i) \ \forall\ i>0
\end{align}

The VB marginal $q(\bm{\mathcal{I}})$ can be obtained using \eqref{VB_update} and \eqref{log_ASOR} as
\begin{align}
	q(\bm{\mathcal{I}})	 \propto&  \prod_{i=1}^m  {{\mathcal{I}(i)}^{0.5}} e^{-0.5 {\hat{r}^{2}(i) } \mathcal{I}(i)} ( (1-{\theta(i)}) f(a,b) {{{\mathcal{I}}}(i)}^{a-1}  e^{-\hat{b} {{{\mathcal{I}}}(i)} }  + {\theta(i)}\delta( {{{\mathcal{I}}}(i)}-1) )  
\end{align}
where we have assume a point estimator for $b$ i.e. $q(b)=\delta(b-\hat{b})$ to simplify the arising expectations. 

We can further write  
\begin{align}
	q(\bm{\mathcal{I}})	 	 =&  \prod_{i=1}^m      k(i) (1-{\theta(i)} ) f(a,\hat{b})  {{{\mathcal{I}}}(i)}^{\alpha-1 } e^{  - \beta(i) \mathcal{I}(i) } + k(i) {\theta(i)} e^{-0.5 {\hat{r}^{2}(i) }}  \delta( {{{\mathcal{I}}}(i)}-1)  \label{k(i)} 
\end{align}
where $k(i)$ is the proportionality constant for the $i$th dimension, $\beta(i)={0.5 {\hat{r}^{2}(i)}  +\hat{b} } $ and $\alpha=a+0.5$. 

Proceeding ahead we can write
\begin{align}
	q(\bm{\mathcal{I}})	& =  \prod_{i=1}^m    \overbrace{(1-{\Omega(i)} )\underbrace{ f(\alpha,\beta(i)) {{{\mathcal{I}}}(i)}^{\alpha-1 } e^{  - \beta(i) \mathcal{I}(i) }}_{q^1({{\mathcal{I}}}(i))}  + {{\Omega(i)} \delta( {{{\mathcal{I}}}(i)}-1)} }^{q(\mathcal{I}(i))} \nonumber   
\end{align}
where 
\begin{equation}
	{\Omega(i)}=k(i) e^{-0.5 {\hat{r}^{2}(i) }} \theta(i) \label{omega_sora}
\end{equation}

To determine  $k(i)$, we note that the distribution in \eqref{k(i)} should integrate to $1$. Therefore, the following should hold
\begin{equation}
	k(i) {\theta(i)} e^{-0.5 {\hat{r}^{2}(i)}}   +   k(i) (1-{\theta(i)} ) \frac{ f(a,\hat{b})}  {f(\alpha,{\beta(i)})  } =1
\end{equation}

Leading to 
\begin{equation}
	k(i)=\frac{1}{{\theta(i)} e^{-0.5 {\hat{r}^{2}(i)}}   +   (1-{\theta(i)} ) \frac{ f(a,\hat{b})}  {f(\alpha,{\beta(i)})  }} \label{k_sora}
\end{equation}

Resultingly, using \eqref{norm} and \eqref{omega_sora}, \eqref{k_sora} we arrive at 
\begin{equation}
	\Omega(i)=\frac{1}{1+ \zeta \frac {{\hat{b}}^a} {{{\beta(i)}^\alpha }} e^{0.5 {\hat{r}^{2}(i)} }    } \ \forall\ i>0
\end{equation}
where $\zeta=(\frac{1}{\theta(i)}-1) \frac{\Gamma(\alpha)}{\Gamma(a)}$.

Lastly, in a similar manner using the VB/EM approach we can determine $q(b)=\delta(b-\hat{b})$ where using \eqref{M_step} and \eqref{log_ASOR}
\begin{equation}
	\hat{b}=\underset{b}{\operatorname{argmax}} \big\langle\mathrm{ln} ( p(\hat{\mathbf{x}},\bm{\mathcal{I}},b |\mathbf{y})) \big\rangle_{ q(\bm{\mathcal{I}}) } \label{b_ASOR}
\end{equation}

The expected log-posterior in \eqref{b_ASOR} can be written as 
\begin{align}
	\big\langle\mathrm{ln} ( p(\hat{\mathbf{x}},\bm{\mathcal{I}},b |\mathbf{y})) \big\rangle_{ q(\bm{\mathcal{I}}) }=& {\sum_{i=1}^m v(b)(i)}+{(A-1)}\ln (b){-Bb}+ constant \label{log_post_b}
\end{align}
where
\begin{align}
	v(b)(i)& ={   \langle \ln ((1-{\theta(i)} ) f(a,b) {{{\mathcal{I}}}(i)}^{a-1 } e^{  - b \mathcal{I}(i) }  + {\theta(i)} \delta( {{{\mathcal{I}}}(i)}-1))  \rangle_{q({{\mathcal{I}}}(i))} }
\end{align}
which can further be written as follows considering only the terms dependent on $b$ 
\begin{align}
	v(b)(i) =    (1-\Omega(i)) ( \ln ( f(a,b)) - b  \langle{{{\mathcal{I}}}(i)}\rangle_{q^1({{\mathcal{I}}}(i))}) + constant  \label{v_b}
\end{align}
where we assume $q^1(\mathcal{I}(i))$ is defined as zero for ${{{\mathcal{I}}}(i)}=1$ as the prior Gamma density in \eqref{pIb}.
Given $\langle{{{\mathcal{I}}}(i)}\rangle_{q^1({{\mathcal{I}}}(i))}={\alpha}/{\beta(i)}$ and using the expressions in \eqref{norm} and \eqref{v_b}, we can write \eqref{log_post_b} as

\begin{align}
	\big\langle  \mathrm{ln} ( p(\hat{\mathbf{x}},\bm{\mathcal{I}},b |\mathbf{y})) \big\rangle_{ q(\bm{\mathcal{I}}) } =
	&{(\bar{A}-1)}\ln (b) -\bar{B}b + constant \label{lnb_ASOR} 
\end{align}
where
\begin{align}
	\bar{A} &= A+\sum_{i=1}^m a (1-\Omega(i)) \label{A_bar}\\
	\bar{B}&=B+\sum_{i=1}^m (1-\Omega(i)) \frac{\alpha}{\beta(i)} \label{B_bar}
\end{align}

Maximizing \eqref{lnb_ASOR} using differentiation, we obtain $\hat{b}$ according to \eqref{b_ASOR} as
\begin{equation}
	\hat{b}=\frac{\bar{A}-1}{\bar{B}} \ \ \ \mathrm{s.t.} \ \ \ \bar{A}>1 \label{hb_ASOR}
\end{equation}
where $\bar{A}>1$ owing to the requirement of positivity of parameter $b$ of Gamma distribution in \eqref{pIb} being approximated in \eqref{hb_ASOR}. Also noting that the parameter $a>0$ for validity of the Gamma distribution in \eqref{pIb}, $A>1$ is a sufficient condition for \eqref{hb_ASOR} to hold considering \eqref{A_bar}. Lastly, note that since the rate parameter of distribution in \eqref{pb_ASOR} is positive i.e. $B>0$ any numerical errors in \eqref{hb_ASOR} are avoided. 
The resulting method namely ASOR is given as Algorithm \ref{Algo2_ASOR}.

\begin{algorithm}[h!]
	\SetAlgoLined
	Initialize $w(i)=1\ \forall\ i$\ and $A, B, a, \hat{b}$, $\theta(i)\ \forall\ i$ \\
	Evaluate $\alpha=a+0.5$ and $\zeta=(\frac{1}{\theta(i)}-1) \frac{\Gamma(\alpha)}{\Gamma(a)}$\\ 
	\While{$\textnormal{not converged}$}{
		Variable update: $\hat{\mathbf{x}}=\underset{\mathbf{x} \in \mathcal{X}}{\operatorname{argmin}}\sum_{i} w(i) {(r(\mathbf{y}^i, \mathbf{x})(i))}^2 $\\
		Residual update: ${\hat{r}^{2}(i)}={ \big( r\left(\mathbf{y}^i, \hat{\mathbf{x}} \right)(i)\big)}^2$ $\forall\ i$\\
		Parameteric updates: \\
		$\beta(i)={0.5 {\hat{r}^{2}(i)}  +\hat{b} } $ \\
		$	\Omega(i)=\frac{1}{1+ \zeta \frac {{\hat{b}}^a} { { {\beta(i)}^\alpha }} e^{0.5 {\hat{r}^{2}(i)} } } $ \\
		$\hat{b}=\mfrac{A-1+\sum_i a (1-\Omega(i))}{B+\sum_i(1-\Omega(i)) \frac{\alpha}{\beta(i)} }$ \\
		Weight update: $ w(i)=\Omega(i)+(1-\Omega(i))\alpha/\beta(i) $  $\forall\ i>0$\ \\
	}
	\caption{The proposed estimator: ASOR}
	\label{Algo2_ASOR}
\end{algorithm}

\subsection*{Remarks}\hspace{.5cm}
It is interesting to note that the variable update steps in EROR, ESOR and ASOR is the same as in GNC reflecting their ability to employ non-minimal solvers during inference. We propose using the change in ${\sum_i} w(i) {\hat{r}^{2}(i)}$ during consecutive iterations as the standard convergence criterion. We lastly remark that in EROR and ESOR an exception to break the iterations can be added for seamless operation when the sum of the weights gets close to zero which we experienced in certain applications with very high ratios of outliers.

\section{Experiments}\label{Sec_Exp}\hspace{.5cm}
In this section, we discuss the performance results of the proposed methods for different spatial perception applications including 3D point cloud registration (code: MATLAB), mesh registration (code: MATLAB), and pose graph optimization (PGO) (code: C++) on an Intel i7-8550U processor-based computer and consider SI units. We consider GNC-GM and GNC-TLS as the benchmark methods which have shown good results in these applications outperforming methods including the classical RANSAC and recently introduced ADAPT. For EROR and ESOR we consider $\chi=\gamma=\bar{c}^2$ where $\bar{c}^2$, dictating the maximum error expected for the inliers, is set as specified in the original work \cite{yang2020graduated}. In ASOR we resort to the choice of initialization which performs the best across different applications given as $a=0.5,A=10000,B=1000,\hat{b}=10000,\theta(i)=0.5\ \forall \ i$. We use the normalized incremental change of $10^{-5}$ in  $\underset{i}{\sum} w(i) {\hat{r}^{2}(i)}$ during consecutive iterations as the convergence criterion for GNC-TLS, EROR, ESOR and ASOR. For GNC-GM the convergence criterion remains the same as originally reported.

\subsection{3D Point Cloud Registration}
\begin{figure}[h!]
	\centering
	\includegraphics[width=.6\linewidth,trim=1cm 3cm 2.5cm 3cm,clip]{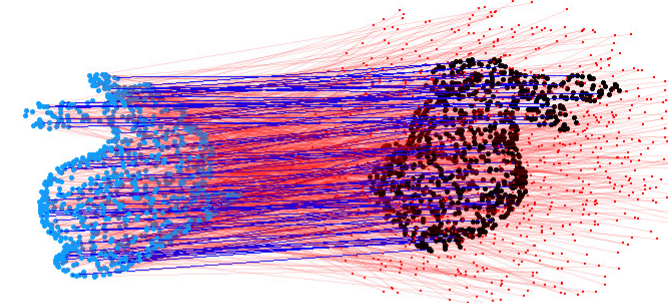}
	\caption{Point clouds with correspondences in 3D point cloud registration for the \textit{Bunny} dataset \cite{curless1996volumetric}.}
	\label{fig:bunny}
\end{figure}
\begin{figure*}[ht!]
	\centering
	\begin{subfigure}{0.5\textwidth}
		\includegraphics[width=\textwidth]{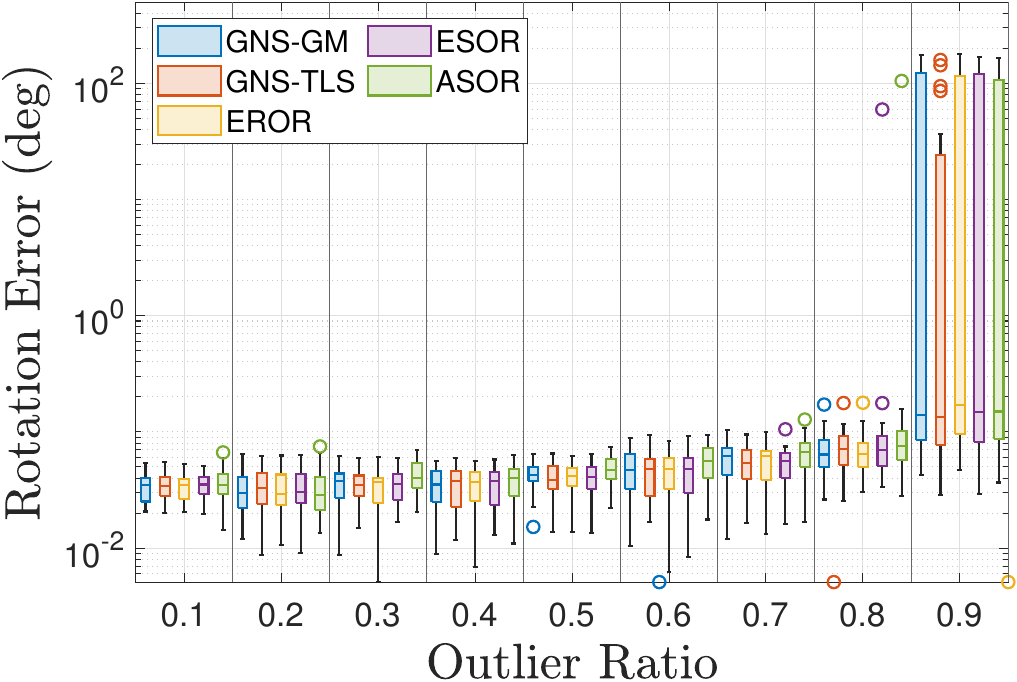}
		\caption{Rotation error.}
		\label{fig:first_horn}
	\end{subfigure}
	\hfill
	\begin{subfigure}{0.5\textwidth}
		\includegraphics[width=\textwidth]{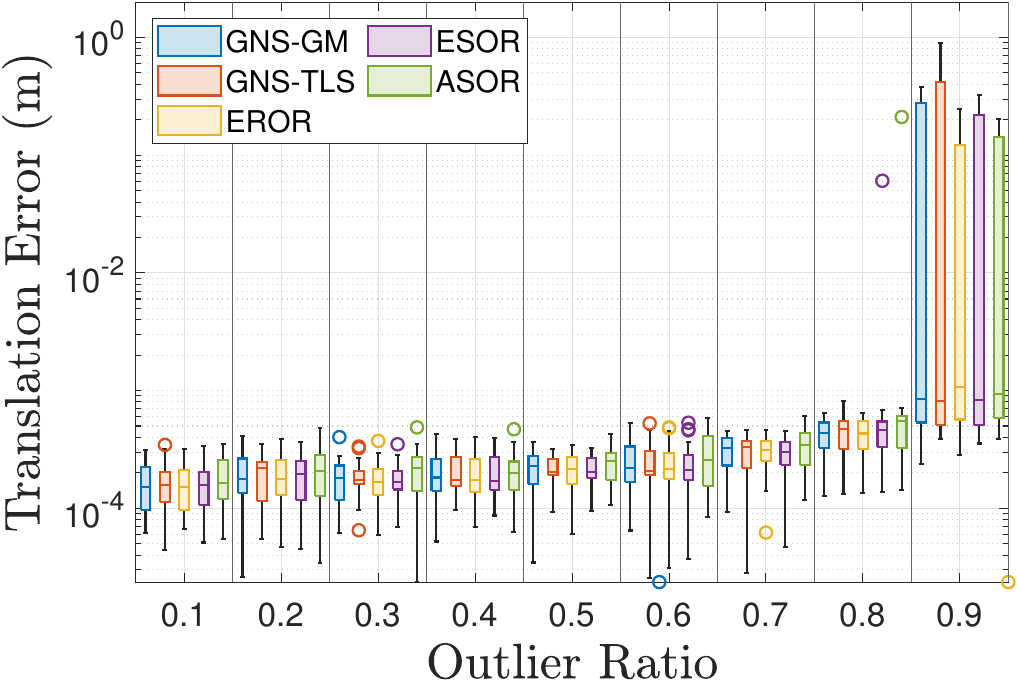}
		\caption{Translation error.}
		\label{fig:second_horn}
	\end{subfigure}
	\hfill
	\begin{subfigure}{0.5\textwidth}
		\includegraphics[width=\textwidth]{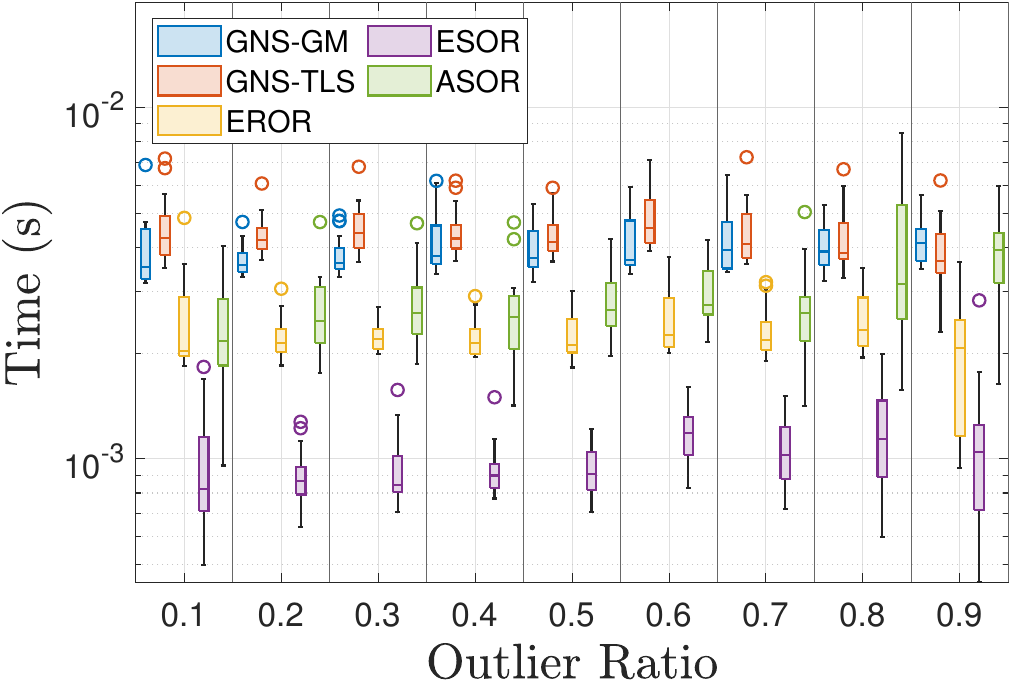}
		\caption{Computational time.}
		\label{fig:third_horn}
	\end{subfigure}
	\caption{Performance of robust estimators for 3D point cloud registration considering the \textit{Bunny} dataset \cite{curless1996volumetric}.}
	\label{fig:figures_horn}
\end{figure*}

In 3D point cloud registration, we assume that a set of 3D points $\mathbf{p}^i\in \mathbb{R}^3, i = 1, . . . , m$ undergo a transformation, with rotation $\mathbf{R} \in \mathrm{SO(3)}$  and translation $\mathbf{t} \in \mathbb{R}^3$, resulting in another set of 3D points $\mathbf{q}^i\in \mathbb{R}^3, i = 1, . . . , m$. The putative correspondences $(\mathbf{p}^i,\mathbf{q}^i)$ can be potentially infested with outliers. Fig.~\ref{fig:bunny} depicts how the \textit{Bunny} point cloud from the Stanford repository \cite{curless1996volumetric} undergoes a random transformation in a point cloud registration setup (blue lines: inliers, red lines: outliers). The objective is to estimate $\mathbf{R}$ and $\mathbf{t}$ that best aligns the two point clouds by minimizing the effect of outliers. The problem can be cast in form of \eqref{eq2} where the \textit{i}th residual is the Euclidean distance between $\mathbf{q}^i$ and $\mathbf{R}\mathbf{p}^i+\mathbf{t}$. We resort to the renowned Horn's method as the non-minimal solver for this case which provides closed form estimates in the outlier-free case \cite{horn1987closed}.

Using the proposed EROR, ESOR and ASOR we robustify the Horn's method and report the results for the \textit{Bunny} dataset. We downsample the point cloud to $m=100$ points and restrict it within a $[-0.5, 0.5]^3$ box before applying a random rotation $\mathbf{R} \in \mathrm{SO(3)}$ and a random translation $\mathbf{t}\ (\|\mathbf{t}\|_2\leq3)$. The inliers of the transformed points are corrupted with independent noise samples drawn from $\mathcal{N}(0,0.001^2)$ whereas the outliers are randomly generated being contained within a sphere of diameter $\sqrt{3}$. 20 Monte Carlo (MC) runs are used to capture the error statistics for up to $90\%$ outlier contamination ratio.

Figs.~\ref{fig:figures_horn} showcase the performance results of our proposed methods in comparison to the GNC methods. The rotation and translational error statistics as depicted in Figs.~\ref{fig:first_horn}~-~\ref{fig:second_horn} are very similar in all the cases with errors generally increasing for all the methods at large outlier ratios. In terms of computational complexity, the gains of the Bayesian heuristics can be observed for this case in Fig.~\ref{fig:third_horn} which is a key evaluation parameter for runtime performance. ESOR exhibits much faster comparative behavior followed by EROR and ASOR for the entire range of outlier ratios. We have observed similar error performance of the methods for other values of $m$ (upto maximum number of available points). Also we have noticed that with an increase in the inlier noise magnitude and number of points, ASOR slows down becoming computationally comparative to the GNC methods. Moreover, we have noted that increasing the parameter $a$ generally reduces the processing time at the cost of slightly increased errors at higher outlier ratios. 

Other specialized solvers like TEASER \cite{yang2019polynomial} also exist for point cloud registration problems which are certifiably robust and can sustain higher outlier contamination better. However, it has scalability issues thanks to the involvement of sluggish semidefinite programming (SDP) based solution for rotation estimation. An improved version of TEASER namely TEASER++ \cite{9286491} was recently introduced that leverages the heuristic, GNC-TLS, in its inferential pipeline for rotation estimation while still certifying the estimates. Owing to the well-devised estimation pipeline, GNC-TLS has to deal with a smaller percentage of outliers making TEASER++ faster and improving its practicality. Since the error performance of the Bayesian heuristics is similar to the GNC methods and these are generally found to be faster in various scenarios of the point cloud registration problem this indicates their utility as standalone estimators. Moreover, these can potentially be useful in other inferential pipelines like TEASER++ (while enjoying certifiable performance) but need a thorough evaluation.

\subsection{Mesh registration}
\begin{figure}[h!]
	\centering
	\includegraphics[width=.6\linewidth,trim=1.4cm 4cm 1.4cm 3cm,clip]{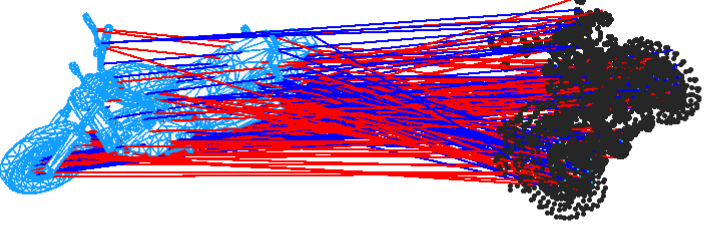}
	\caption{Mesh and point cloud with correspondences in mesh registration for the \textit{motorbike} mesh model \cite{6836101}}
	\label{fig:meshbike}
\end{figure}
In the mesh registration problem, the points $\mathbf{p}^i$ from the a point cloud are transformed to general 3D primitives $\mathbf{q}^i$ including points, lines, and/or planes. Fig.~\ref{fig:meshbike} shows the result of a random transformation of a point cloud to the \textit{motorbike} mesh model from the PASCAL dataset \cite{6836101} (blue lines: inliers, red lines: outliers). The aim is to estimate the $\mathbf{R}$ and $\mathbf{t}$ by minimizing the squared sum of residuals which represent the corresponding distances. We resort to \cite{8100078} as the basic non-minimal solver which has been proposed to find the globally optimal solution for the outlier-free case.
\begin{figure*}[h!]
	\centering
	\begin{subfigure}{0.5\textwidth}
		\includegraphics[width=\textwidth]{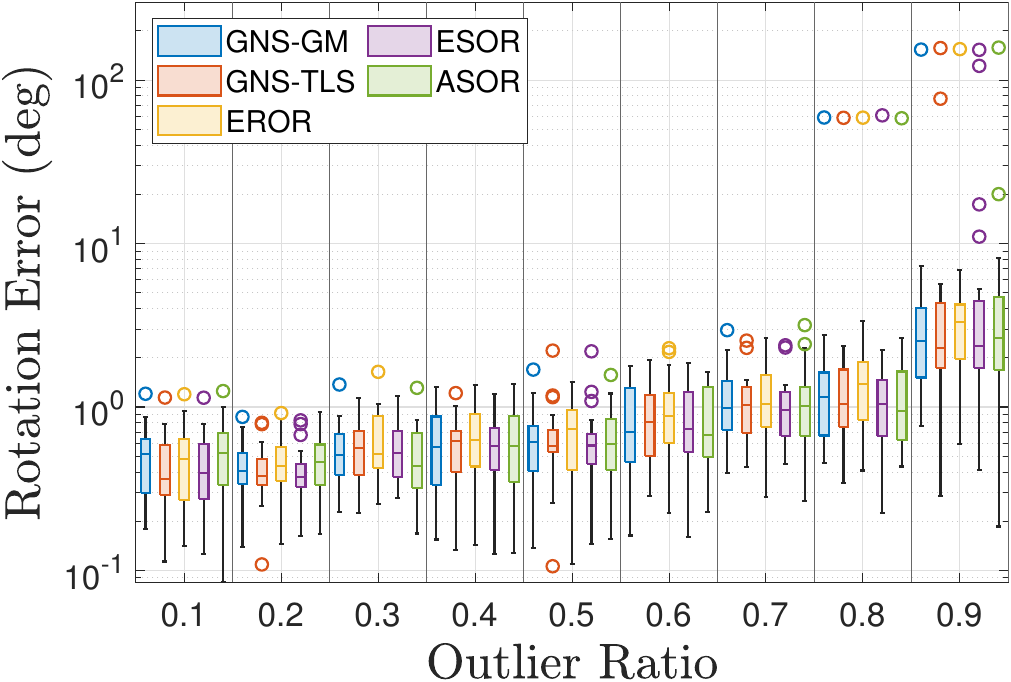}
		\caption{Rotation error.}
		\label{fig:first_mesh}
	\end{subfigure}
	\hfill
	\begin{subfigure}{0.5\textwidth}
		\includegraphics[width=\textwidth]{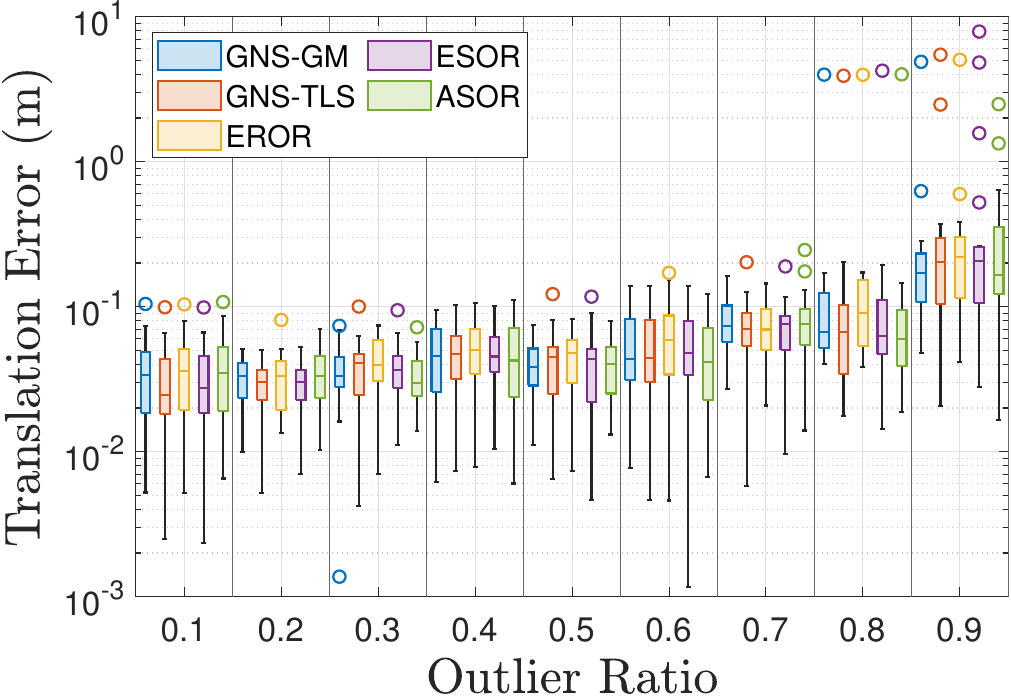}
		\caption{Translation error.}
		\label{fig:second_mesh}
	\end{subfigure}
	\hfill
	\begin{subfigure}{0.5\textwidth}
		\includegraphics[width=\textwidth]{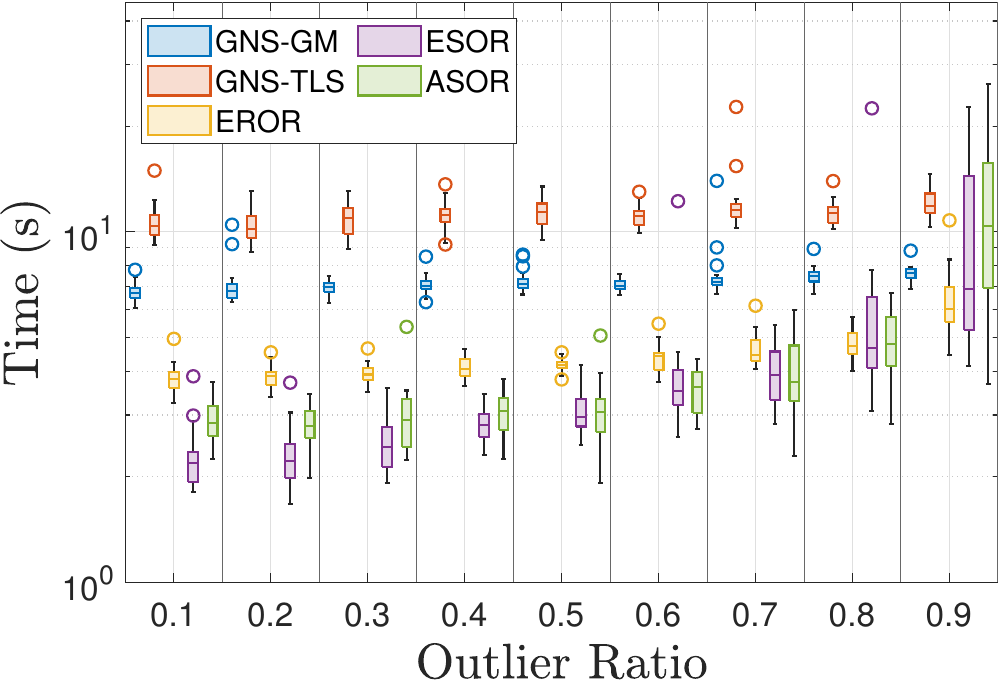}
		\caption{Computational time.}
		\label{fig:third_mesh}
	\end{subfigure}
	\caption{Performance of robust estimators for mesh registration considering the \textit{motorbike} model \cite{6836101}.}
	\label{fig:figures_mesh}
\end{figure*}  
Using the presented EROR, ESOR and ASOR heuristics we robustify the solver and report the results for the \textit{motorbike} mesh model. To create point cloud we randomly sample points on the vertices, edges and faces of the mesh model, and then apply a random transformation $(\mathbf{R} \in \mathrm{SO(3)}$, $\mathbf{t}\ (\|\mathbf{t}{\|}_2\leq5))$ and subsequently add independent noise samples from $\mathcal{N}(0,0.01^2)$. Considering $100$ point-to-point, $80$ point-to-line and $80$ point-to-plane correspondences, we create outliers by random erroneous correspondences. For this case also, 20 MC runs are carried out to generate the error statistics for up to $90\%$ outlier contamination ratio. We see a trend similar to the point cloud registration case as depicted in Figs.~\ref{fig:first_mesh}~-~\ref{fig:second_mesh}. In particular, the performance in terms of errors is similar for all the algorithms. However, the Bayesian heuristics exhibit a general computational advantage, except at very high outlier ratios where the performance becomes comparable. The Bayesian heuristics generally have similar runtimes with SOR modifications having a general advantage. We observed similar performance of the methods for other combinations of correspondences in this case.

During the experiments of point cloud and mesh registration, we have noticed that ROR and SOR modifications generally get computationally more advantageous, with estimation quality remaining similar, when the outliers become larger in comparison to the nominal noise samples.

\subsection{Pose graph optimization}
\begin{figure}[h!]
	\centering
	\hspace{0cm}\subfloat[\centering {INTEL}]{{\includegraphics[width=4.2cm,trim=1cm 1cm 1cm 1cm,clip]{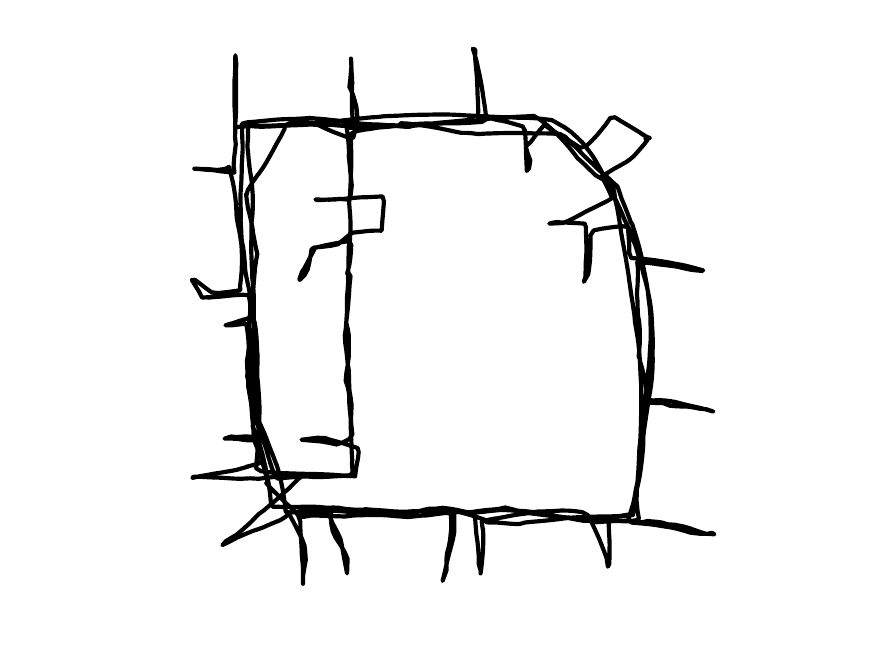} }}%
	\hspace{0cm}\subfloat[\centering {CSAIL}]{{\includegraphics[width=4.2cm,trim=2cm 1cm 2cm 1cm,clip]{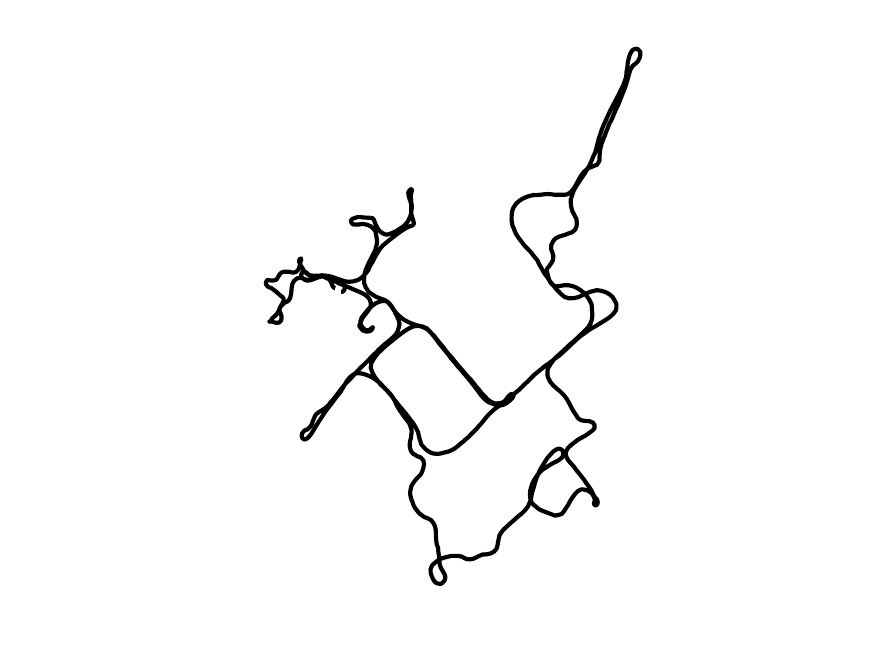} }}%
	\caption{Ground truth paths of datasets considered in pose graph optimization.}%
	\label{fig:PGO_path}%
\end{figure}
\begin{figure*}[h!]
	\centering
	\begin{subfigure}{0.5\textwidth}
		\includegraphics[width=\textwidth]{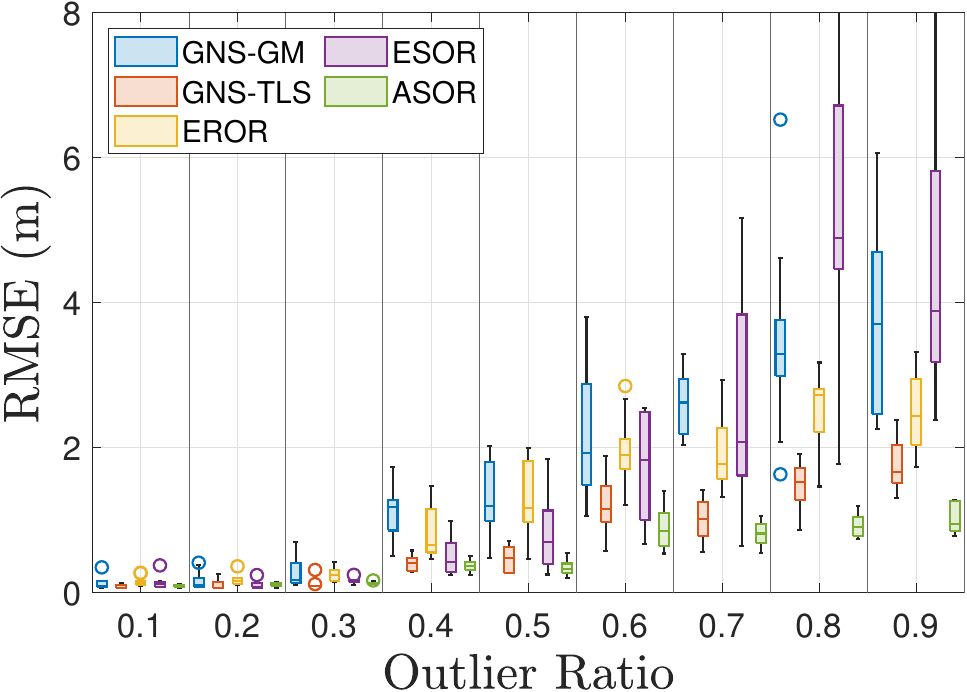}
		\caption{RMSE (INTEL).}
		\label{fig:first_PGO}
	\end{subfigure}
	\hfill
	\begin{subfigure}{0.5\textwidth}
		\includegraphics[width=\textwidth]{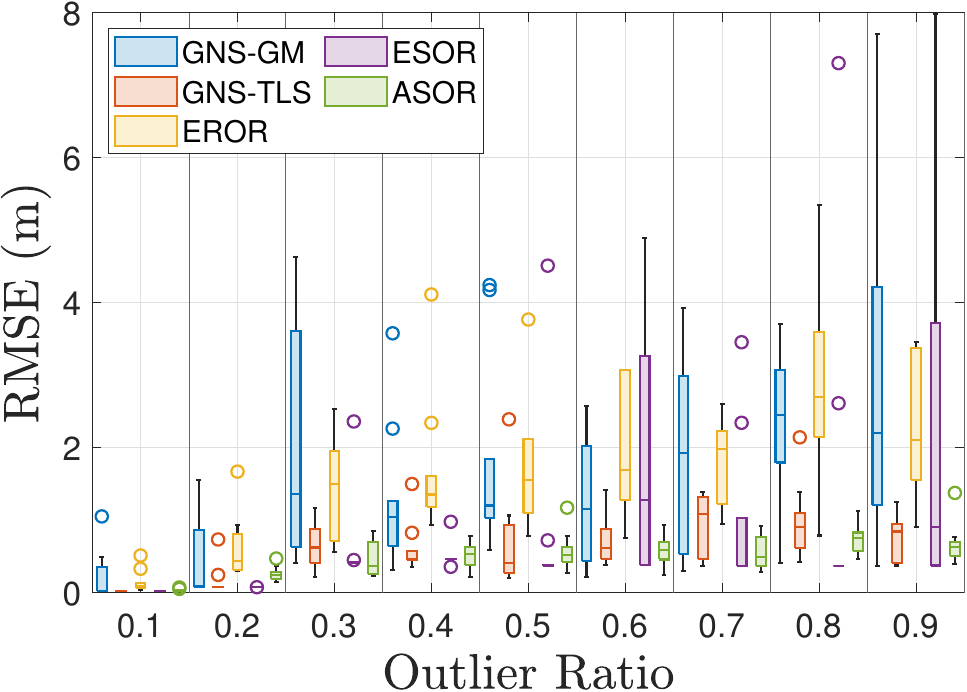}
		\caption{RMSE (CSAIL).}
		\label{fig:second_PGO}
	\end{subfigure}
	\hfill
	\begin{subfigure}{0.5\textwidth}
		\includegraphics[width=\textwidth]{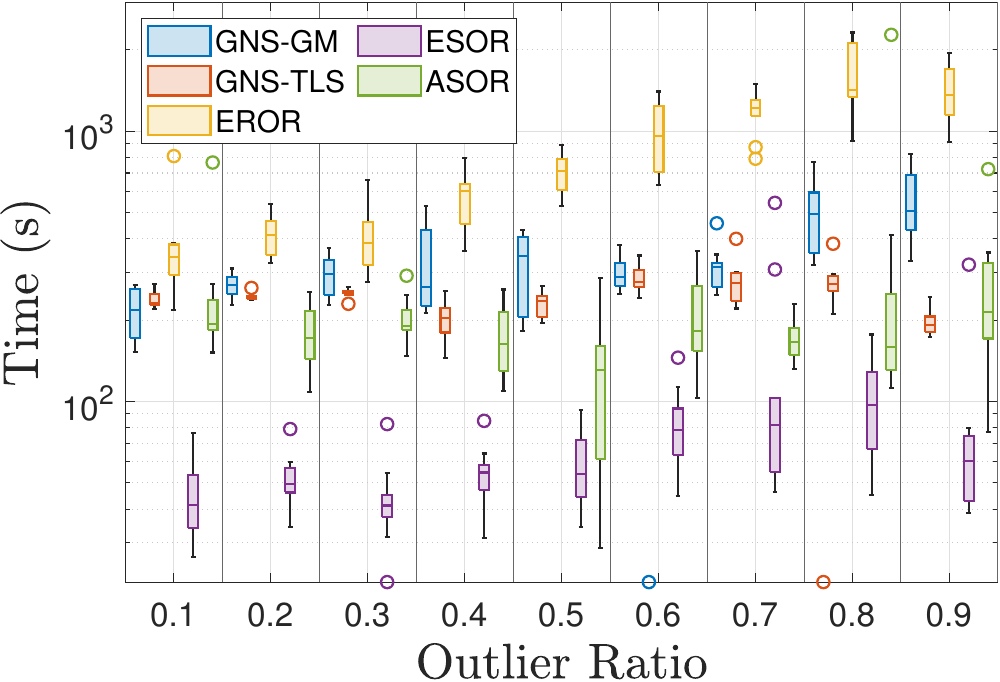}
		\caption{Computational time (INTEL).}
		\label{fig:third_PGO}
	\end{subfigure}
	\caption{Performance of robust estimators for pose graph optimization considering Intel and CSAIL datasets.}
	\label{fig:figures_PGO}
\end{figure*}  
PGO is typically employed for several problems arising in robotic and computer vision applications like SLAM and structure from motion (SfM) \cite{9346012}. The objective is to estimate a set of poses $(\mathbf{t}^i,\mathbf{R}^i), i = 1, . . . , m$ using pairwise relative measurements $(\tilde{\mathbf{t}}^{i j}, \tilde{\mathbf{R}}^{i j})$. Relative observations can result in consecutive pose constraints (e.g. from odometry measurements) or non-successive pose constraints (e.g. from scan matching) also known as loop closures. The residual error for this case is given as 
\begin{equation}
	r\left(\mathbf{R}^i, t^i\right)=\sqrt{\kappa^{i j}\|\mathbf{R}^j-\mathbf{R}^i \tilde{\mathbf{R}}^{i j}\|_F^2+\tau^{i j}\|t^j-t^i-\mathbf{R}^i \tilde{t}^{i j}\|_2^2}\nonumber
\end{equation}
where $\kappa^{i j}$ and $\tau^{i j}$ encode the measurement noise statistics and $\|.\|_{F}$ denotes the Frobenius norm. We resort to SE-Sync \cite{rosen2019se} as the non-minimal solver for this case and use the Python binding of C++ open-sourced by the authors. Adopting the same experimentation setup of GNC we randomly corrupt loop closures and retain odometry observations. For benchmarking we consider 2D datasets including INTEL and CSAIL which are available openly (path ground truth plotted in Fig.~\ref{fig:PGO_path}). Since simulations take much more time as compared to the previous case we carry out 10 MC runs to obtain the error statistics. Fig.~\ref{fig:first_PGO} showcases the root mean squared (RMSE) of the trajectory considering the INTEL dataset. The proposed Bayesian heuristics in this case also exhibit comparative performance to the GNC methods. At very higher outlier rates ESOR has slightly compromised performance. In Fig.~\ref{fig:second_PGO} the RMSE for the CSAIL dataset are depicted reflecting the same pattern. In the PGO examples, GNC-TLS and ASOR outperform other methods with ASOR generally having the least error for different outlier ratios. As far as the computational performance is concerned ESOR has the smallest runtime, followed by ASOR while ROR is the slowest for both cases. Fig.~\refeq{fig:third_PGO}  depicts the computational runtime statistics for the INTEL dataset. CSAIL has similar computational time statistics with ESOR being relatively much faster as compared to other methods.     

%
%
%
Lastly, we also evaluated the Bayesian heuristics using $\max^i({w(i) {\hat{r}^{2}(i)}})<\bar{c}^2$ as the stopping criteria which resulted in much faster performance but with slightly degraded error performance in the considered scenarios of the spatial perception applications.

\section{Conclusion}\label{Sec_conc}\hspace{.5cm}
In this chapter, we have proposed extensions of ROR and SOR heuristics for general purpose nonlinear estimation for spatial perception problems enabling them to invoke existing non-minimal solvers. Evaluations in several experiments demonstrate their merits as compared to similar general-purpose GNC heuristics. In particular, in the 3D point cloud and mesh registration problems EROR, ESOR and ASOR have similar estimation errors over a wide range of outlier ratios. However, Bayesian heuristics have a general advantage in computational terms. For the PGO setups, we generally find Bayesian methods to compete with GNC in estimation quality. The devised ROR and SOR modifications are found to be the least and most computationally efficient for this case. Using another suggested criteria can lead to improvement in computational terms at expense of estimation quality. In short, the proposed methods provide general purpose options, in addition to the GNC heuristics, to robustify the existing non-minimal solvers against outliers in different spatial perception applications indicating their usefulness. Empirical evidence suggests that the proposed approaches provide a general edge in computational terms while remaining competitive in terms of error. The actual possible gains depend on whether the solvers are used standalone or in an inferential pipeline for the particular application scenarios and should be evaluated for the case under consideration before deployment.

\chapter{Outlier-Robust Filtering and Smoothing With Correlated Measurement Noise} \label{chap-6a}
\hspace{.5cm}
In this chapter, we consider the problem of outlier-robust state estimation where the measurement noise can be correlated. This a much general setting as compared to previous chapters. We use insights drawn from Chapter \ref{chap-4} and also present useful connections to the work reported there. 

Noise correlation emerges in several real-world applications e.g. sensor networks, radar data, GPS-based systems, etc. We consider the data corruption and noise correlation effects in system modeling which is subsequently used for inference. We employ the Expectation-Maximization (EM) framework to derive both outlier-resilient filtering and smoothing methods, suitable for online and offline estimation respectively. The standard Gaussian filtering and the Gaussian Rauch–Tung–Striebel (RTS) smoothing results are leveraged to devise the estimators. In addition, Bayesian Cramer-Rao Bounds (BCRBs) for a filter and a smoother which can perfectly detect and reject outliers are presented. These serve as useful theoretical benchmarks to gauge the error performance of different estimators. Lastly, different numerical experiments, for an illustrative target tracking application, are carried out that indicate performance gains compared to similarly engineered state-of-the-art outlier-rejecting state estimators. The advantages are in terms of simpler implementation, enhanced estimation quality, and competitive computational performance.

\section{State-space modeling}\label{EM_model}
\subsection{Standard modeling}\hspace{.5cm}
Consider a standard nonlinear discrete-time state-space model (SSM) to represent the dynamics of a physical system given as
\begin{align}
	\mathbf{x}_k&= \mathbf{f}(\mathbf{x}_{k-1})+\mathbf{q}_{k-1}	
	\label{eqn_model_1_EM}\\
	\mathbf{y}_k&= \mathbf{h}(\mathbf{x}_{k})+\mathbf{r}_{k}
	\label{eqn_model_2_EM}
\end{align}
where $\mathbf{x}_k\in \mathbb{R}^n$ and $\textbf{y}_k \in \mathbb{R}^m$ denote the state and measurement vectors respectively; the nonlinear functions  $\textbf{f}(.):\mathbb{R}^n\rightarrow\mathbb{R}^n$ and $\textbf{h}(.):\mathbb{R}^n\rightarrow\mathbb{R}^m$  represent the process dynamics and observation transformations respectively; $\textbf{q}_{k}\in \mathbb{R}^n$ and $\textbf{r}_k\in \mathbb{R}^m$ account for the additive nominal process and measurement noise respectively. $\textbf{q}_{k}$ and $\textbf{r}_k$ are assumed to be statistically independent, White, and normally distributed with zero mean and known covariance matrices $\textbf{Q}_k$ and $\textbf{R}_k$ respectively. We consider that $\textbf{R}_k$ can be a fully enumerated matrix capturing the correlations between the measurement noise entries. 

\subsection{Modeling outliers for inference}\hspace{.5cm}
The model in \eqref{eqn_model_1_EM}-\eqref{eqn_model_2_EM} assumes that the measurements are only affected by nominal measurement noise $\textbf{r}_k$. However, the observations in every dimension can be corrupted with outliers leading to the disruption of standard state estimators as the measurement data cannot be described by the regular model. Therefore, data outliers need to be appropriately modeled within the generative SSM with two basic objectives. Firstly, the model should sufficiently capture the effect of outlier contamination in the data. Secondly, the model should remain amenable to inference. 

To model the outliers in SSM, we consider an indicator vector $\bm{\mathcal{I}}_k\in\mathbb{R}^m$ having Bernoulli elements where its $i$th element ${{\mathcal{I}}}_k(i)$ can assume two possible values: $\epsilon$ (close to zero) and 1. ${{\mathcal{I}}}_k(i)=\epsilon$ denotes the presence of an outlier, whereas ${{\mathcal{I}}}_k(i)=1$ indicates no outlier in the $i$th dimension at time $k$. Since outliers can occur independently at any instant, we assume that the elements of $\bm{{\mathcal{I}}}_k$ are statistically independent of their past. Additionally, we assume that the entries of $\bm{{\mathcal{I}}}_k$ to be independent of each other since generally no knowledge of correlations between outliers is available which are not easy to model anyway. Moreover, this choice is motivated by the goal of inferential tractability. We also consider $\bm{\mathcal{I}}_k$ and $\mathbf{x}_k$ to be statistically independent since the outlier occurrence does not depend on the state value. The assumed distribution of $\bm{\mathcal{I}}_k$ is given as
\begin{equation}
	p(\bm{\mathcal{I}}_k)=\prod_{i=1}^{m}p({{\mathcal{I}}}_k(i))=\prod_{i=1}^{m} (1-{\theta_k(i)}) \delta({{{\mathcal{I}}}_k(i)}-\epsilon)+{\theta_k(i)}\delta( {{{\mathcal{I}}}_k(i)}-1)
	\label{eqn_model_3_EM}
\end{equation}
where $\theta_k(i)$ denotes the prior probability of no outlier in the $i$th observation at time $k$. Further, the conditional measurement likelihood given the current state $\mathbf{x}_k$ and the indicator $\bm{\mathcal{I}}_k$, is proposed to be normally distributed as

\begin{align}
	p(\mathbf{y}_k|\mathbf{x}_k,\bm{\mathcal{I}}_k)&={\mathcal{N}}\big(\mathbf{y}_k|\mathbf{h}(\mathbf{x}_k),\mathbf{R}_k({ \bm{\mathcal{{I}}}_k}) \big)\nonumber\\
	&=\frac{1}{\sqrt{(2 \pi)^{m}|\mathbf{R}_k({ \bm{\mathcal{{I}}}_k})|}}\mathrm{exp} \left({-}\mfrac{1}{2}{(\mathbf{y}_k-\mathbf{h}(\mathbf{x}_k))}^{\top}\mathbf{R}_k^{-1}(\bm{\mathcal{I}}_k) (\mathbf{y}_k-\mathbf{h}(\mathbf{x}_k)) \right) 		\label{eqn_model_4_EM}
\end{align}
where $\mathbf{R}_k(\boldsymbol{\mathcal{I}}_{k})$ is given as
\begin{align}
\mathbf{R}_k(\boldsymbol{\mathcal{I}}_{k})=
	\begin{bmatrix}
		{R_k}({1,1}) / {{{\mathcal{I}}}_k(i)}  & \dots &  {R_k}({1,m}) \delta( {{{\mathcal{I}}}_k(i)} {-}1) \delta( {{{\mathcal{I}}}_k(m)}{-}1) \\
		\vdots &  \ddots& \vdots\\
		{R_k}({m,1}) \delta( {{{\mathcal{I}}}_k(m)}{-}1) \delta( {{{\mathcal{I}}}_k(i)}{-}1)  & \cdots & {R_k}({m,m}) / {{{\mathcal{I}}}_k(m)}
	\end{bmatrix} 
	\label{eqn_model_5_EM}
\end{align}
$\mathbf{R}_k({ \bm{\mathcal{{I}}}_k})$ is the modified covariance matrix of the measurements considering the effect of outliers. The effect of $\mathbf{R}_k({ \bm{\mathcal{{I}}}_k})$ on the data generation process can be understood by considering the possible values of ${\mathcal{I}}_k(i)$. In particular, ${\mathcal{I}}_k(i)=\epsilon$ leads to a very large $i$th diagonal entry of $\mathbf{R}_k({ \bm{\mathcal{{I}}}_k})$, while placing zeros at the remaining $i$th row and column of the matrix. Resultingly, when an outlier occurs in the $i$th dimension its effect on state estimation is minimized. Moreover, the $i$th dimension no longer has any correlation with any other entry, ceasing to have any effect on any other dimension during inference. This is in contrast to ${\mathcal{I}}_k(i)=1$ which ensures the diagonal element and the off-diagonal correlation entries with other non-affected dimensions are preserved. Lastly, note that the conditional likelihood is independent of the batch of all the historical observations $\mathbf{y}_{1:{k-1}}$.    

\section{Proposed Robust Filtering}\label{filter}\hspace{.5cm}
For filtering, we are interested in the posterior distribution of $\mathbf{x}_k$ conditioned on all the observations $\mathbf{y}_{1:{k}}$ that have been observed till time $k$. For this objective, we can employ the Bayes rule recursively. Given the proposed observation model, the analytical expression of the joint posterior distribution of $\mathbf{x}_k$ and $\bm{\mathcal{I}}_k$ conditioned on the set of all the observations $\mathbf{y}_{1:{k}}$ is given as
\begin{equation}
	p(\mathbf{x}_k,\bm{\mathcal{I}}_k|\mathbf{y}_{1:{k}})=\frac{p(\mathbf{y}_k|\bm{\mathcal{I}}_k,\mathbf{x}_{k})	p(\mathbf{x}_k|\mathbf{y}_{1:{k-1}})p(\bm{\mathcal{I}}_k)}{p(\mathbf{y}_k|\mathbf{y}_{1:{k-1}})}
	\label{eqn_fl_1}
\end{equation}

Theoretically, the joint posterior can further be marginalized to obtain the required posterior distribution $p(\mathbf{x}_k|\mathbf{y}_{1:{k}})$. Assuming $p(\mathbf{x}_k|\mathbf{y}_{1:{k-1}})$ as a  Gaussian distribution, we need to run $2^m$ Kalman filters corresponding to each combination of $\bm{\mathcal{I}}_k$ to obtain the posterior. This results in computational complexity of around $\mathcal{O}(2^{m} m^3)$ where $m^3$ appears due to matrix inversions (ignoring sparsity). Therefore, this approach clearly becomes impractical and unsuitable. 

To get around the problem, we can possibly employ the standard VB method where the product of VB marginals is conveniently used to approximate the joint posterior. We assume the following factorization of the posterior 
\begin{equation}
	p(\mathbf{x}_k,\bm{\mathcal{I}}_k|\mathbf{y}_{1:{k}})\approx q^f(\mathbf{x}_k)\prod_i q^f({\mathcal{I}}_k(i))
	\label{eqn_fl_2}
\end{equation} 

The VB approximation aims to minimize the Kullback-Leibler (KL) divergence between the product approximation and the true posterior and leads to the following marginals \cite{murphy2012machine} 
\begin{align}
	q^f(\mathbf{x}_k)&\propto \exp \big( \big\langle\mathrm{ln} ( p(\mathbf{x}_k,\bm{\mathcal{I}}_k|\mathbf{y}_{1:{k}}))\big\rangle_{ q^f({{\bm{\mathcal{I}}}_k})  } \big)\label{eqn_vb_1_EM}\\
	q^f({\mathcal{I}}_k(i))&\propto \exp \big( \big\langle \mathrm{ln} ( p(\mathbf{x}_k,\bm{\mathcal{I}}_k|\mathbf{y}_{1:k}))\big\rangle_{q^f(\mathbf{x}_k)  q^f(\bm{\mathcal{I}}^{i-}_k)  }\ \big) \ \forall\ i \label{eqn_vb_2_EM}
\end{align} 

Using \eqref{eqn_vb_1_EM}-\eqref{eqn_vb_2_EM} alternately the VB marginals can be updated iteratively until convergence. The procedure provides a useful way to approximate the true marginals of the joint posterior by approximating these as $p(\mathbf{x}_k|\mathbf{y}_{1:{k}}) \approx {q^f(\mathbf{x}_k)}$ and $	p(\bm{\mathcal{I}}_k|\mathbf{y}_{1:{k}})\approx \prod_i q^f({\mathcal{I}}_k(i))$.

For our model, \eqref{eqn_vb_1_EM} becomes computationally unfriendly if we use the standard VB approach. In fact, the same complexity order of $\mathcal{O}(m^3 2^m)$ appears as with the basic marginalization approach making this approach intractable too. We elaborate more on it in the upcoming subsection.
\subsection*{Expectation-Maximization as a particular case of variational Bayes}\label{em}\hspace{.5cm}
To deal with the complexity issue, instead of considering distributions we can resort to point estimates for ${{\mathcal{I}}}_k(i)$. In particular, consider $q^f({\mathcal{I}}_k(i))= \delta({{\mathcal{I}}}_k(i)-{\hat{{\mathcal{I}}}}_k(i))$ where ${\hat{{\mathcal{I}}}}_k(i)$ denotes the point approximation of ${{{\mathcal{I}}}}_k(i)$. Consequently, the variational distributions can be updated in an alternating manner in the Expectation (E) and Maximization (M) steps in the EM algorithm given as \cite{vsmidl2006variational}

\subsubsection*{\textnormal{E-Step}}
\begin{align}
	{q^f(\boldsymbol{\mathbf{x}_k})}&=p(\mathbf{x}_k|\mathbf{y}_{1:{k}},\hat{\bm{\mathcal{I}}}_k)	\propto	p(\mathbf{x}_k,\hat{\bm{\mathcal{I}}}_k|\mathbf{y}_{1:{k}})	\label{eqn_fl_E}
\end{align}
\subsubsection*{\textnormal{M-Step}}
\begin{equation}
	{\hat{\mathcal{I}}}_k(i)= \underset{{{\mathcal{I}}}_k(i)}{\mathrm{argmax}}\big\langle\mathrm{ln}(p(\mathbf{x}_k,{\mathcal{I}}_k(i),\hat{\bm{\mathcal{I}}}_k(i-)|\mathbf{y}_{1:{k}})\big\rangle_{q^f(\mathbf{x}_k)} \label{eqn_fl_M}
\end{equation}
where all ${\hat{\mathcal{I}}}_k(i)$ in the M-Step are successively updated using the latest estimates. 
\subsection{Prediction}\label{pre_fil}
\hspace{.5cm}
We first obtain the predictive distribution $p(\mathbf{x}_{k}|\mathbf{y}_{1:{k-1}})$ using the posterior distribution at the previous instant $p(\mathbf{x}_{k-1}|\mathbf{y}_{1:{k-1}})$ approximated as Gaussian $q^f(\mathbf{x}_{k-1})\approx {\mathcal{N}}\left(\mathbf{x}_{k-1}|\hat{\mathbf{x}}^{+}\hat{\mathbf{x}}^{+}_{k-1},\mathbf{P}^{+}_{k-1}\right)$. Using Gaussian (Kalman) filtering results we make the following approximation \cite{sarkka2023bayesian}
\begin{equation}
	p(\mathbf{x}_k|\mathbf{y}_{1:{k-1}})\approx {\mathcal{N}}\left(\mathbf{x}_k|\hat{\mathbf{x}}^{-}_k,\mathbf{P}^{-}_{k}\right)\label{eqn_fl_pr0}
\end{equation}
where
\begin{flalign}
	\hat{\mathbf{x}}^{-}_{k}=&\int \mathbf{f}(\mathbf{x}_{k-1})\  {\mathcal{N}}\left(\mathbf{x}_{k-1}|\hat{\mathbf{x}}^{+}_{k-1},\mathbf{P}^{+}_{k-1}\right)	d\mathbf{x}_{k-1} \label{eqn_fl_pr1}\\
	\mathbf{P}^{-}_{k}=&\int ( \mathbf{f}(\mathbf{x}_{k-1})-\hat{\mathbf{x}}^{-}_{k})(\mathbf{f}(\mathbf{x}_{k-1})-\hat{\mathbf{x}}^{-}_{k})^{\top}{\mathcal{N}}({\mathbf{x}_{k-1}}|\hat{\mathbf{x}}^{+}_{k-1},\mathbf{P}^{+}_{k-1})  d\mathbf{x}_{k-1}+\mathbf{Q}_{k-1} \label{eqn_fl_pr2}
\end{flalign}
\subsection{Update}\label{upd_fil} 
\hspace{.5cm}
Using the expressions of the prior distributions from \eqref{eqn_model_3_EM} and \eqref{eqn_fl_pr0} along with the conditional measurement likelihood in \eqref{eqn_model_4_EM}, we can express the joint posterior distribution as 
\begin{align}
	\ p\left(\mathbf{x}_{k}, \boldsymbol{\mathcal{I}}_{k}|\mathbf{y}_{1:k}\right) \propto& \mfrac{\mathcal{N}(\mathbf{x}_{k} | \hat{\mathbf{x}}_{k}^{-}, \mathbf{P}_{k}^{-})}{\sqrt{(2 \pi)^{m}\left|{\mathbf{R}}_{k}(\boldsymbol{\mathcal{I}}_{k})\right|}} \exp \Big\{ {-}\mfrac{1}{2}\left(\mathbf{y}_{k}-\mathbf{h}\left(\mathbf{x}_{k}\right)\right)^{\top}{\mathbf{R}}_{k}^{-1}(\boldsymbol{\mathcal{I}}_{k})\left(\mathbf{y}_{k}-\mathbf{h}\left(\mathbf{x}_{k}\right)\right)\Big\} \nonumber\\
	&\Big\{ \prod_{i} \left(1-\theta_{k}(i)\right) \delta\left(\mathcal{I}_{k}(i)-\epsilon\right)+
	\theta_{k}(i) \delta\left( {\mathcal{I}}_{k}(i) -1\right) \Big\} \label{eqn_fl_up1_x}
\end{align}
\subsubsection{Derivation of $q^f(\mathbf{x}_{k})$}
\hspace{.5cm}
With the E-Step in \eqref{eqn_fl_E} we can write 
\begin{align}
	\ q^f\left(\mathbf{x}_{k}\right) \propto& \exp \Big\{{-}\mfrac{1}{2}\left(\mathbf{y}_{k}-\mathbf{h}\left(\mathbf{x}_{k}\right)\right)^{\top} {\mathbf{R}}_{k}^{-1} (\boldsymbol{\hat{\mathcal{I}}}_{k}) \left(\mathbf{y}_{k}-\mathbf{h}\left(\mathbf{x}_{k}\right)\right) \nonumber\\
	&{-}\mfrac{1}{2}\left(\mathbf{x}_{k}-\hat{\mathbf{x}}_{k}^{-}\right)^{\top} ({\mathbf{P}_{k}^{-}})^{-1}\left(\mathbf{x}_{k}-\hat{\mathbf{x}}_{k}^{-}\right)\Big\}\label{eqn_fl_up2_x}
\end{align}
where ${\mathbf{R}}_{k}^{-1}({\boldsymbol{ \hat{\mathcal{I}} }_{k}})$ assumes a particular form resulting from the inversion of ${\mathbf{R}}_{k}({\boldsymbol{ \hat{\mathcal{I}} }_{k}})$ as described in Appendix \ref{FirstAppendix_EM}. 

Note that we avoid evaluating $\big\langle{\mathbf{R}}_{k}^{-1} (\boldsymbol{{\mathcal{I}}}_{k})\big\rangle_{ q^f({{\bm{\mathcal{I}}}_k}) }$ that would be required in the standard VB approach. This means that we are able to evade the complexity level of around $\mathcal{O}(m^3 2^m)$ since matrix inversion for each of the $2^m$ combinations are required to evaluate the expectation. However, thanks to EM, we are now working with  ${\mathbf{R}}_{k}^{-1}({\boldsymbol{ \hat{\mathcal{I}} }_{k}})$ which can be evaluated with the maximum complexity of $\mathcal{O}(m^3)$ (considering a fully populated matrix).


To proceed further, we use the results of the general Gaussian filter, to approximate $q^f(\mathbf{x}_k)$ with a Gaussian distribution, $ {\mathcal{N}}\left(\mathbf{x}_k|\hat{\mathbf{x}}^{+}_k,\mathbf{P}^{+}_{k}\right)$, with parameters updated as
\begin{align}
	\hat{\mathbf{x}}^{+}_k&=\hat{\mathbf{x}}^{-}_k+\mathbf{K}_k
	(\mathbf{y}_k-\bm{\mu}_k)	\label{eqn_fl_up4b_x}\\
	\mathbf{P}^{+}_{k}&=\mathbf{P}^{-}_{k}-\mathbf{C}_k\mathbf{K}^\top_k \label{eqn_fl_up5_x}
\end{align}
where
\begin{align}
	\mathbf{K}_k=&\ \mathbf{C}_k (\mathbf{U}_k+{\mathbf{R}}_{k}({\boldsymbol{ \hat{\mathcal{I}} }_{k}}) )^{-1}= \mathbf{C}_k ( {\mathbf{R}}_{k}^{-1}({\boldsymbol{ \hat{\mathcal{I}} }_{k}}) {-} {\mathbf{R}}_{k}^{-1}({\boldsymbol{ \hat{\mathcal{I}} }_{k}})(\mathbf{I}+\mathbf{U}_k {\mathbf{R}}_{k}^{-1}({\boldsymbol{ \hat{\mathcal{I}} }_{k}}) )^{-1}\mathbf{U}_k {\mathbf{R}}_{k}^{-1}({\boldsymbol{ \hat{\mathcal{I}} }_{k}}) ) \nonumber\\
	\bm{\mu}_k=&\int \mathbf{h}(\mathbf{x}_k)\  {\mathcal{N}}\left(\mathbf{x}_k|\hat{\mathbf{x}}^{-}_{k},\mathbf{P}^{-}_{k}\right)	d\mathbf{x}_k\nonumber\\
	\mathbf{U}_k=&\int(\mathbf{h}(\mathbf{x}_k)-\bm{\mu}_k)(\mathbf{h}(\mathbf{x}_k)-\bm{\mu}_k)^{\top}{\mathcal{N}}({\mathbf{x}_k}|\hat{\mathbf{x}}^{-}_k,\mathbf{P}^{-}_{k})d\mathbf{x}_k\nonumber\\
	\mathbf{C}_k=&\int(\mathbf{x}_k-\hat{\mathbf{x}}^{-}_{k})(\mathbf{h}(\mathbf{x}_k)-\bm{\mu}_k)^{\top}{\mathcal{N}}({\mathbf{x}_k}|\hat{\mathbf{x}}^{-}_{k},\mathbf{P}^{-}_{k})d\mathbf{x}_k \nonumber
\end{align}
\subsubsection{Derivation of $\hat{\mathcal{I}}_k(i)$}\hspace{.5cm}
With the M-Step in \eqref{eqn_fl_M} we can write 
\begin{align}
	{\hat{\mathcal{I}}}_k(i)=&\ \underset{{{\mathcal{I}}}_k(i)}{\mathrm{argmax}}\label{eqn_fl_up1_i} \big\langle\mathrm{ln}(p(\mathbf{x}_k,{\mathcal{I}}_k(i),\hat{\bm{\mathcal{I}}}_k(i-)|\mathbf{y}_{1:{k}} )\big\rangle_{q^f(\mathbf{x}_k)}&
\end{align}

Using the Bayes rule we can proceed as
\begin{flalign}
	{\hat{\mathcal{I}}}_k(i)=&\  \underset{{{\mathcal{I}}}_k(i)}{\mathrm{argmax}} \big(\big\langle\mathrm{ln}(p(\mathbf{y}_{{k}}|\mathbf{x}_{k},{\mathcal{I}}_k(i),\hat{\bm{\mathcal{I}}}_k(i-), \stkout{\mathbf{y}_{1:{k-1}}}))\big\rangle_{q^f(\mathbf{x}_{k})}\nonumber\\
	&+\mathrm{ln} (p({\mathcal{I}}_k(i)|\stkout{ {\mathbf{x}_{k}},\hat{\bm{\mathcal{I}}}_k(i-),\mathbf{y}_{1:{k-1}} } )) \big)+const. & \label{eqn_fl_up2_i}
\end{flalign}
where $const.$ is some constant and $p(\mathbf {x}|\mathbf {y},\stkout {\mathbf {z}})$ denotes the conditional independence of $\mathbf{x}$ and $\mathbf{z}$ given $\mathbf{y}$. We can further write
\begin{align}
	{\hat{\mathcal{I}}}_k(i)=&\  \underset{{{\mathcal{I}}}_k(i)}{\mathrm{argmax}} \Big\{ {-}\mfrac{1}{2} \mathrm{tr} \big( \mathbf{W}_k {\mathbf{R}}_{k}^{-1}( {\mathcal{I}_k(i)} , \hat{\bm{\mathcal{I}}}_k(i-) )     \big) {-}\mfrac{1}{2}	\ln|{\mathbf{R}}_{k}( {\mathcal{I}_k(i)} , \hat{\bm{\mathcal{I}}}_k(i-) ) |\nonumber \\
	&+\ln\big( (1-\theta_{k}(i))  \delta(\mathcal{I}_{k}(i)-\epsilon)+\theta_{k}(i) \delta ({\mathcal{I}}_{k}(i)-1) \big) \Big\}\label{eqn_fl_up3_i}
\end{align}
where ${\mathbf{R}}_{k}( {\mathcal{I}_k(i)} , \hat{\bm{\mathcal{I}}}_k(i-) )$ denotes ${\mathbf{R}}_{k}({\bm{\mathcal{I}}}_k)$ evaluated at ${\bm{\mathcal{I}}}_k$ with it $i$th element as ${\mathcal{I}_k(i)}$ and remaining entries $\hat{\bm{\mathcal{I}}}_k(i-) $ and 
\begin{align}
	\mathbf{W}_k=\int \left(\mathbf{y}_{k}-\mathbf{h}\left(\mathbf{x}_{k}\right)\right) \left(\mathbf{y}_{k}-\mathbf{h}\left(\mathbf{x}_{k}\right)\right)^{\top} {\mathcal{N}}({\mathbf{x}_k}|\hat{\mathbf{x}}^{+}_{k},\mathbf{P}^{+}_{k})d\mathbf{x}_k \nonumber
\end{align}

Resultingly, $\hat{\mathcal{I}}_k(i)$ can be determined as
\begin{align}
	{ \hat{\mathcal{I}}}_k(i) &= 
	\begin{cases}
		1 & \text{if } \hat{\tau}_k(i) \leq 0,\\
		\epsilon & \text{if } \hat{\tau}_k(i) >0
	\end{cases}\label{If}
\end{align}
with
\begin{flalign}
	\hat{\tau}_k(i) =&\  \mathrm{tr} \big( \mathbf{W}_k \triangle{\hat{\mathbf{R}}}_{k}^{-1}     \big) +\ln\Big(\frac{|{\mathbf{R}}_{k}( {\mathcal{I}_k(i)=1} , \hat{\bm{\mathcal{I}}}_k(i-) ) |}{|{\mathbf{R}}_{k}( { \mathcal{I}_k(i)=\epsilon} , \hat{\bm{\mathcal{I}}}_k(i-) )|}\Big)+2\ln\Big(\frac{1}{\theta_k(i)}-1  \Big)  & \label{eqn_fl_up5_i}
\end{flalign}
where
\begin{align}
	\triangle{\hat{\mathbf{R}}}_{k}^{-1}=  ({\mathbf{R}}_{k}^{-1}( {\mathcal{I}_k(i)=1} , \hat{\bm{\mathcal{I}}}_k(i-) )-{\mathbf{R}}_{k}^{-1}( {\mathcal{I}_k(i)=\epsilon} , \hat{\bm{\mathcal{I}}}_k(i-) ))\label{eqdelR}
\end{align}

Using the steps outlined in Appendix \ref{SecondAppendix_EM}, we can further simplify $\hat{\tau}_k(i)$ as
\begin{flalign}
	\hat{\tau}_k(i)= &\  \Big\{ \mathrm{tr} \big( \mathbf{W}_k \triangle{\hat{\mathbf{R}}}_{k}^{-1}     \big) 
	+\ln \Bigg|\textbf{I}-\frac{\mathbf{R}_k(-i,i) \mathbf{R}_k(i,-i) ( \hat{\mathbf{R}}_{k}(-i,-i) )^{-1}}{R_k(i,i)}\Bigg|  \nonumber \\ & +\ln(\epsilon)+2\ln\Big(\frac{1}{\theta_k(i)}-1  \Big) \Big\} &\label{eqn_fl_up10_i}
\end{flalign}
where $\hat{\mathbf{R}}_{k}(-i,-i)$ is the submatrix of ${\mathbf{R}}_{k}({\boldsymbol{ \hat{\mathcal{I}} }_{k}})$ corresponding to entries of $\hat{\boldsymbol{\mathcal{I}}}^{i-}_k$. $\mathbf{R}_{k}(i,-i)$ and $\mathbf{R}_{k}(-i,i)$ contain the measurement covariances between \textit{i}{th} and rest of the dimensions.

\begin{algorithm}[t!]
	\SetAlgoLined
	Initialize\ $\hat{\mathbf{x}}^{+}_0,\mathbf{P}^{+}_0$\;
	\For{$k=1,2...K$}{
		Initialize $\theta_k(i),{\boldsymbol{ \hat{\mathcal{I}} }_{k}},\mathbf{Q}_k,\mathbf{R}_k$\;
		\textbf{Prediction:} \\
		Evaluate $\hat{\mathbf{x}}^{-}_k,\mathbf{P}^{-}_k$ with \eqref{eqn_fl_pr1} and \eqref{eqn_fl_pr2}\;
		\textbf{Update:} \\
		\While{not converged}{
			Update ${\hat{\mathbf{x}}^{+}_k}$ and  ${\mathbf{P}^{+}_k}$ with \eqref{eqn_fl_up4b_x}-\eqref{eqn_fl_up5_x}\;
			Update ${\hat{\mathcal{I}}_k(i)}$ $\forall\ i$ with \eqref{If}\;
		}
	}
	\caption{The proposed filter: EMORF}
	\label{Algo1_EM}
\end{algorithm} 

Though we can directly evaluate $\triangle{\hat{\bm{\mathbf{R}}}}_{k}^{-1}$ in \eqref{eqdelR}, we can save computations by avoiding repetitive calculations. To this end, we first need to compute 
\begin{align}
	&\triangle{\hat{\bm{\mathfrak{R}}}}_{k}^{-1}=\begin{bmatrix}
		\bm{\Xi}({i,i}) & \bm{\Xi}({i,-i}) \\
		\bm{\Xi}({-i,i}) & \bm{\Xi}({-i,-i})  \label{del_t}
	\end{bmatrix}
\end{align}
with
\begin{align}
	\bm{\Xi}({i,i})&=\frac{1}{R_k(i,i)-\mathbf{R}_k(i,-i)(\hat{\mathbf{R}}_{k}(-i,-i))^{-1} \mathbf{R}_k(-i,i)}-\frac{\epsilon}{R_k(i,i)}\label{xi1}\\
	\bm{\Xi}({i,-i})&=-\frac{\mathbf{R}_k(i,-i) (\hat{\mathbf{R}}_{k}(-i,-i))^{-1}}{R_k(i,i)-\mathbf{R}_k(i,-i)(\hat{\mathbf{R}}_{k}(-i,-i))^{-1} \mathbf{R}_k(-i,i)}\label{xi2}\\
	\bm{\Xi}({-i,i})&=-\frac{ (\hat{\mathbf{R}}_{k}(-i,-i))^{-1}\mathbf{R}_k(-i,i)}{R_k(i,i)-\mathbf{R}_k(i,-i)(\hat{\mathbf{R}}_{k}(-i,-i))^{-1} \mathbf{R}_k(-i,i)}\label{xi3}\\
	\bm{\Xi}({-i,-i})&=\frac{ (\hat{\mathbf{R}}_{k}(-i,-i))^{-1}\mathbf{R}_k(-i,i) \mathbf{R}_k(i,-i) (\hat{\mathbf{R}}_{k}(-i,-i))^{-1}}{R_k(i,i)-\mathbf{R}_k(i,-i)(\hat{\mathbf{R}}_{k}(-i,-i))^{-1} \mathbf{R}_k(-i,i)}\label{xi4}
\end{align}
where the $i$th row/column entries in $\triangle{\hat{\bm{\mathbf{R}}}}_{k}^{-1}$ have been conveniently swapped with the first row/column elements to obtain $\triangle{\hat{\bm{\mathfrak{R}}}}_{k}^{-1}$. By swapping the first row/column entries of $\triangle{\hat{\bm{\mathfrak{R}}}}_{k}^{-1}$ to the $i$th row/column positions we can reclaim $\triangle{\hat{\bm{\mathbf{R}}}}_{k}^{-1}$. Appendix \ref{ThirdAppendix} provides  further details in this regard.

The resulting EM-based outlier-robust filter (EMORF) is outlined as Algorithm \ref{Algo1_EM}. For the convergence criterion, we suggest using the ratio of the L2 norm of the difference of the state estimates from the current and previous VB iterations and the L2 norm of the estimate from the previous iteration. This criterion has been commonly chosen in similar robust filters \cite{8398426}.
\subsection{\texorpdfstring{VB factorization of $p(\mathbf{x}_k,\boldsymbol{\mathcal{I}}_k|\mathbf{y}_{1:{k}})$ and the associated computational overhead}{} }\hspace{.5cm}
Note that for better accuracy we can factorize $p(\mathbf{x}_k,\bm{\mathcal{I}}_k|\mathbf{y}_{1:{k}})\approx q^f(\mathbf{x}_k) q^f(\bm{\mathcal{I}}_k)$ instead of forcing independence between all ${\mathcal{I}}_k(i)$. In this case, expression for evaluating  $q^f(\mathbf{x}_k)$ remains same as in \eqref{eqn_vb_1_EM}. However, this choice leads to the following VB marginal distribution of $\bm{\mathcal{I}}_k$
\begin{align}
	q^f(\bm{\mathcal{I}}_k)&\propto \exp \big( \big\langle \mathrm{ln} ( p(\mathbf{x}_k,\bm{\mathcal{I}}_k|\mathbf{y}_{1:k}))\big\rangle_{q^f(\mathbf{x}_k)  }\ \big) \label{eqn_vb_I}
\end{align} 

This results in a modified M-Step in the EM algorithm given as
\subsubsection*{\textnormal{M-Step}}
\begin{equation}
	{ \hat {\bm{\mathcal{I}}} }_k= \underset{\bm{{\mathcal{I}}}_k}{\mathrm{argmax}}\big\langle\mathrm{ln}(p(\mathbf{x}_k,{\bm{\mathcal{I}}}_k|\mathbf{y}_{1:{k}}))\big\rangle_{q^f(\mathbf{x}_k)} \label{eqn_fl_M2}
\end{equation}  

Proceeding further, with the prediction and update steps during inference, we can arrive at 
\begin{flalign}
	{\hat{\bm{\mathcal{I}}}}_k=& \Big\{\ \underset{{\bm{\mathcal{I}}}_k}{\mathrm{argmax}} \Big( {-}\mfrac{1}{2} \mathrm{tr} \big( \mathbf{W}_k {\mathbf{R}}_{k}^{-1}(  {\bm{\mathcal{I}}}_k )     \big) {-}\mfrac{1}{2}	\ln|{\mathbf{R}}_{k}( {\bm{\mathcal{I}}}_k ) | \nonumber \\
	& \ln\big( (1-\theta_{k}(i))  \delta(\mathcal{I}_{k}(i)-\epsilon)+\theta_{k}(i) \delta ({\mathcal{I}}_{k}(i)-1) \big) \Big) \Big\} \label{eqn_fl_I2}
\end{flalign}

It is not hard to notice that determining ${\hat{\bm{\mathcal{I}}}}_k$ using \eqref{eqn_fl_I2} involves tedious calculations. In fact, we run into the same computational difficulty that we have been dodging. To arrive at the result, we need to evaluate the inverses and determinants, for each of the $2^m$ combinations corresponding to the entries of ${\bm{\mathcal{I}}}_k$. This entails the bothering complexity level of $\mathcal{O}(m^3 2^m)$. 

With the proposed factorization in \eqref{eqn_fl_2}, we obtain a more practical and scalable algorithm. The resulting complexity is $\mathcal{O}(m^4)$ following from the evaluation of matrix inverses and determinants for calculating each of the $\hat{\mathcal{I}}_k(i)$ $\forall\ i=1\cdots m$ in \eqref{If}.

\subsection{Connection between EMORF and SORF (Chapter \ref{chap-4})}\hspace{.5cm}
Since the construction of EMORF is motivated by selective observations rejecting filter (SORF), it is insightful to discuss their connection. We derived SORF considering a diagonal measurement covariance matrix $\mathbf{R}_k$. In SORF, we used distributional estimates for $\bm{\mathcal{I}}_k$ since it did not induce any significant computational strain. It is instructive to remark here that the if we use point estimates of $\bm{\mathcal{I}}_k$ in SORF it becomes a special case of EMORF. In particular, the point estimates for ${\mathcal{I}}_k(i)\ \forall\ i$ in (Chapter \ref{chap-4}) can be obtained with the following criterion

\begin{align}
	{\hat{\mathcal{I}}}_k(i) &= 
	\begin{cases}
		1 & \text{if } \bar{\tau}_k(i) \leq 0,\\
		\epsilon & \text{if } \bar{\tau}_k(i) >0
	\end{cases}\label{Isor}
\end{align}

To deduce ${\hat{\mathcal{I}}}_k(i)=1$, the following should hold 
\begin{align}
	\Omega_k(i)&\geq1-\Omega_k(i)\label{sorf2}\\
	\Omega_k(i)&\geq0.5 \label{sorf3}
\end{align}
where $\Omega_k(i)$ denotes the posterior probability of ${{\mathcal{I}}}_k(i)=1$. Using the expression of $\Omega_k(i)$ from \eqref{eqn_vb_12_}, we can write \eqref{sorf3} as
\begin{align} 
	&\frac {1}{1+{\sqrt {\epsilon }}\left({\frac {1}{ \theta_k(i) } -1}\right){\exp\left({\frac {W_{k}(i,i) }{2R_{k}(i,i)} (1-\epsilon)}\right)}}\geq0.5\label{sorf4}
\end{align}
leading to 
\begin{align}
	&\bar{\tau}_k(i)  \leq 0
\end{align}
where
\begin{align}
	\bar{\tau}_k(i)  =&{{\frac {W_{k}(i,i)}{R_{k}(i,i)} (1-\epsilon)}}+\ln({\epsilon }) + 2\ln \Big({\frac {1}{\theta_{k}(i)}-1}\Big) \label{sorf6}
\end{align}
which can be recognized as the particular case of \eqref{eqn_fl_up10_i} given $\mathbf{R}_k(i,-i)$ and $\mathbf{R}_k(-i,i)$ vanish as $\mathbf{R}_k$ is considered to be diagonal. The first term in $\eqref{eqn_fl_up10_i}$ reduces to ${{\frac {W_{k}(i,i)}{R_{k}(i,i)} (1-\epsilon)}}$ given $	\bm{\Xi}({i,i})=\frac{1-\epsilon}{R_k(i,i)}$ and $\bm{\Xi}({i,-i}), \bm{\Xi}({-i,i})$ and $\bm{\Xi}({-i,-i})$ all reduce to zero. Moreover, the second term in \eqref{eqn_fl_up10_i} also disappears resulting in \eqref{sorf6}. 
\subsection{Choice of the parameters $\theta_{k}(i)$ and $\epsilon$}\hspace{.5cm}
For EMORF we propose setting the parameters $\theta_{k}(i)$ and $\epsilon$ the same as in SORF. Specifically, we suggest choosing a neutral value of 0.5 or an uninformative prior for $\theta_{k}(i)$. The Bayes-Laplace and the maximum entropy approaches for obtaining uninformative prior for a finite-valued parameter lead to the choice of the uniform prior distribution \cite{9931968,turkman2019computational}. Moreover, the selection has been justified in the design of outlier-resistant filters assuming no prior information about the outliers statistics is available \cite{8869835}. For $\epsilon$ we recommend its value to be close to zero since the exact value of $0$ denies the VB/EM updates as in Chapter \ref{chap-4}.

\section{Proposed Robust Smoothing}\label{smooth}\hspace{.5cm}
In smoothing, our interest lies in determining the posterior distribution of all the states $\mathbf{x}_{1:K}$ conditioned on the batch of all the observations $\mathbf{y}_{1:{K}}$. With that goal, we take a similar approach to filtering and approximate the joint posterior distribution as a product of marginals
\begin{equation}
	p(\mathbf{x}_{1:K},\bm{\mathcal{I}}_{1:K}|\mathbf{y}_{1:{K}})\approx q^s(\mathbf{x}_{1:K})\prod_{k}\prod_{i} q^s({\mathcal{I}}_k(i))
	\label{eqn_sm_1}
\end{equation}
where the true marginals are approximated as $p(\mathbf{x}_{1:K}|\mathbf{y}_{1:{K}})\approx{q^s(\mathbf{x}_{1:K})}$ and $	p(\bm{\mathcal{I}}_{1:K}|\mathbf{y}_{1:{K}})\approx{\prod_{k}\prod_{i} q^s({\mathcal{I}}_k(i))}$. Let us assume $ q^s({\mathcal{I}}_k(i))= \delta({{\mathcal{I}}}_k(i)-{\breve{{\mathcal{I}}}}_k(i))$ where ${\breve{{\mathcal{I}}}}_k(i)$ denotes the point approximation of ${{{\mathcal{I}}}}_k(i)$. Consequently, the EM steps are given as 
\subsubsection*{\textnormal{E-Step}}
\begin{align}
	{q^s(\boldsymbol{\mathbf{x}_}{1:K})}&=p(\mathbf{x}_{1:K}|\mathbf{y}_{1:{K}},\breve{\bm{\mathcal{I}}}_{1:K})	\propto	p(\mathbf{x}_{1:K},\breve{\bm{\mathcal{I}}}_{1:K}|\mathbf{y}_{1:{K}})	\label{eqn_sm_E}
\end{align}
\subsubsection*{\textnormal{M-Step}}
\begin{equation}
	{\breve{\mathcal{I}}}_k(i)= \underset{{{\mathcal{I}}}_k(i)}{\mathrm{argmax}}\big\langle\mathrm{ln}(p(\mathbf{x}_{1:K},{\mathcal{I}}_k(i),\breve{\bm{\mathcal{I}}}_k(i-),\breve{\bm{\mathcal{I}}}_{k{-}} |\mathbf{y}_{1:{K}})\big\rangle_{q^s(\mathbf{x}_{1:K})} \label{eqn_sm_M}
\end{equation}
where all ${\hat{\mathcal{I}}}_k(i)$ in the M-Step are sequentially updated using the latest estimates. 
\subsubsection{Derivation of $q^s\left(\mathbf{x}_{1:K}\right)$}\hspace{.5cm}
With the E-Step in \eqref{eqn_sm_E} we can write 
\begin{align}
	{q^s(\boldsymbol{\mathbf{x}_}{1:K})} &\propto p(\mathbf{y}_{{1:K}}|\mathbf{x}_{{1:K}},\breve{\bm{\mathcal{I}}}_{1:K})p(\mathbf{x}_{{1:K}})\label{eqn_sm_x1}\\	
	{q^s(\boldsymbol{\mathbf{x}_}{1:K})}&\propto \prod_{k} p(\mathbf{y}_{{k}}|\mathbf{x}_{{k}},\breve{\bm{\mathcal{I}}}_{k})p(\mathbf{x}_{{k}}|\mathbf{x}_{{k-1}})\label{eqn_sm_x2}
\end{align}

We can identify that ${q^s(\boldsymbol{\mathbf{x}_}{1:K})}$ can be approximated as a Gaussian distribution from the results of general Gaussian RTS smoothing \cite{sarkka2023bayesian}. Using the forward and backward passes, we can determine the parameters of ${q^s(\boldsymbol{\mathbf{x}_}{k})}\sim\mathcal{N}(\hat{\bm{\mathit{x}}}^{s}_{k},\bm{\mathcal{P}}^{s}_{k})$, which denotes the marginalized densities of ${q^s(\boldsymbol{\mathbf{x}_}{1:K})}$. 
\subsubsection*{Forward pass}\hspace{.5cm}
The forward pass essentially involves the filtering equations given as 
\begin{flalign}
	 \hat{\bm{\mathit{x}}} ^{-}_{k}=&\int \mathbf{f}(\mathbf{x}_{k-1})\  {\mathcal{N}}\left(\mathbf{x}_{k-1}|\hat{\bm{\mathit{x}}}^{+}_{k-1},\bm{\mathcal{P}}^{+}_{k-1}\right)	d\mathbf{x}_{k-1}&\label{eqn_sm_x3a}\\
	\bm{\mathcal{P}}^{-}_{k}=&\int( \mathbf{f}(\mathbf{x}_{k-1})-\hat{\bm{\mathit{x}}}^{-}_{k})(\mathbf{f}(\mathbf{x}_{k-1})-\hat{\bm{\mathit{x}}}^{-}_{k})^{\top} {\mathcal{N}}({\mathbf{x}_{k-1}}|\hat{\bm{\mathit{x}}}^{+}_{k-1},\bm{\mathcal{P}}^{+}_{k-1}) d\mathbf{x}_{k-1}+\mathbf{Q}_{k-1}&\label{eqn_sm_x3c}\\
	\hat{\bm{\mathit{x}}}^{+}_k=&\ \hat{\bm{\mathit{x}}}^{-}_k+\bm{\mathcal{K}}_k
	(\mathbf{y}_k-\bm{\nu}_k)	&\label{eqn_sm_x4}\\
	\bm{\mathcal{P}}^{+}_{k}=&\ \bm{\mathcal{P}}^{-}_{k}-\bm{\mathcal{C}}_k\bm{\mathcal{K}}^\top_k &\label{eqn_sm_x5}
\end{flalign}
where
\begin{align}
	\bm{\mathcal{K}}_k=&\ \bm{\mathcal{C}}_k (\bm{\mathcal{U}}_k+{\mathbf{R}}_{k}({\boldsymbol{ \breve{\mathcal{I}} }_{k}}) )^{-1} = \bm{\mathcal{C}}_k ( {\mathbf{R}}_{k}^{-1}({\boldsymbol{ \breve{\mathcal{I}} }_{k}}){-} {\mathbf{R}}_{k}^{-1}({\boldsymbol{ \breve{\mathcal{I}} }_{k}})(\mathbf{I}+\bm{\mathcal{U}}_k {\mathbf{R}}_{k}^{-1}({\boldsymbol{ \breve{\mathcal{I}} }_{k}}) )^{-1}\bm{\mathcal{U}}_k {\mathbf{R}}_{k}^{-1}({\boldsymbol{ \breve{\mathcal{I}} }_{k}}) ) \nonumber\\
	\bm{\nu}_k=&\int \mathbf{h}(\mathbf{x}_k)\  {\mathcal{N}}\left(\mathbf{x}_k|\hat{\bm{\mathit{x}}}^{-}_{k},\bm{\mathcal{P}}^{-}_{k}\right)	d\mathbf{x}_k \nonumber\\
	\bm{\mathcal{U}}_k=&\int(\mathbf{h}(\mathbf{x}_k)-\bm{\nu}_k)(\mathbf{h}(\mathbf{x}_k)-\bm{\nu}_k)^{\top}{\mathcal{N}}({\mathbf{x}_k}|\hat{\bm{\mathit{x}}}^{-}_k,\bm{\mathcal{P}}^{-}_{k})d\mathbf{x}_k \nonumber\\
	\bm{\mathcal{C}}_k=&\int(\mathbf{x}_k-\hat{\bm{\mathit{x}}}^{-}_{k})(\mathbf{h}(\mathbf{x}_k)-\bm{\nu}_k)^{\top}{\mathcal{N}}({\mathbf{x}_k}|\hat{\bm{\mathit{x}}}^{-}_{k},\bm{\mathcal{P}}^{-}_{k})d\mathbf{x}_k \nonumber
\end{align}

Note that ${\mathbf{R}}_{k}({\boldsymbol{ \breve{\mathcal{I}} }_{k}})$ and ${\mathbf{R}}_{k}^{-1}({\boldsymbol{ \breve{\mathcal{I}} }_{k}})$ can be evaluated similar to ${\mathbf{R}}_{k}({\boldsymbol{ \hat{\mathcal{I}} }_{k}})$ and ${\mathbf{R}}_{k}^{-1}({\boldsymbol{ \hat{\mathcal{I}} }_{k}})$. 
\subsubsection*{Backward pass}\hspace{.5cm}
The backward pass can be completed as
\begin{flalign}
	\bm{\mathcal{L}}_{k+1}  =&\int(\mathbf{x}_k-\hat{\bm{\mathit{x}}}^{+}_k)(\mathbf{f}\left(\mathbf{x}_k\right)-\hat{\bm{\mathit{x}}}_{k+1}^{-})^{\top}{\mathcal{N}}\left(\mathbf{x}_{k}|\hat{\bm{\mathit{x}}}^{+}_{k},\bm{\mathcal{P}}^{+}_{k}\right)d\mathbf{x}_{k}  &\label{eqn_sm_x11}\\
	\bm{\mathcal{G}}_k  = &\ \bm{\mathcal{L}}_{k+1}\left(\bm{\mathcal{P}}_{k+1}^{-}\right)^{-1} \\
	\hat{\bm{\mathit{x}}}_k^{{s}}  = &\ \hat{\bm{\mathit{x}}}^{+}_k+\bm{\mathcal{G}}_k\left(\hat{\bm{\mathit{x}}}_{k+1}^{{s}}-\hat{\bm{\mathit{x}}}_{k+1}^{-}\right) &\label{eqn_sm_m}\\
	\bm{\mathcal{P}}_k^{{s}}  =&\ \bm{\mathcal{P}}^{+}_k+\bm{\mathcal{G}}_k\left(\bm{\mathcal{P}}_{k+1}^{{s}}-\bm{\mathcal{P}}_{k+1}^{-}\right) \bm{\mathcal{G}}_k^{\top} &\label{eqn_sm_P}
\end{flalign}
\begin{algorithm}[ht!]
	\SetAlgoLined
	Initialize\ $\hat{\bm{\mathit{x}}}^{+}_0,\bm{\mathcal{P}}^{+}_0$;

	\While{not converged}{
		\For{$k=1,2...K$}{
			Initialize $\theta_k(i),{\boldsymbol{ \breve{\mathcal{I}} }_{k}},\mathbf{Q}_k,\mathbf{R}_k$\;	
			Evaluate $\hat{\bm{\mathit{x}}}^{-}_k,\bm{\mathcal{P}}^{-}_k$ with \eqref{eqn_sm_x3a}-\eqref{eqn_sm_x3c}\;
			Evaluate $\hat{\bm{\mathit{x}}}^{+}_k,\bm{\mathcal{P}}^{+}_k$ with \eqref{eqn_sm_x4}-\eqref{eqn_sm_x5}\;
		}
		$\hat{\bm{\mathit{x}}}^{s}_K=\hat{\bm{\mathit{x}}}^{+}_K$\;
		$\bm{\mathcal{P}}^{s}_K=\bm{\mathcal{P}}^{+}_K$ \;
		\For{$k=K-1,...1$}{
			Evaluate $\hat{\bm{\mathit{x}}}^{s}_k,\bm{\mathcal{P}}^{s}_k$ with \eqref{eqn_sm_m}-\eqref{eqn_sm_P}\;
		}	
		\For{$k=1,2...K$}{
			Update ${\breve{\mathcal{I}}_k(i)}$ $\forall\ i$ with \eqref{Is}\;
		}
	}
	\caption{The proposed smoother: EMORS}
	\label{Algo2_EM}
\end{algorithm} 
\subsubsection{Derivation of $\breve{\mathcal{I}}_k(i)$}\hspace{.5cm}
With the M-Step in \eqref{eqn_sm_M} we can write 
\begin{align}
	{\breve{\mathcal{I}}}_k(i)= \underset{{{\mathcal{I}}}_k(i)}{\mathrm{argmax}}\big\langle\mathrm{ln}(p(\mathbf{x}_{1:K},{\mathcal{I}}_k(i),\breve{\bm{\mathcal{I}}}_k(i-),\breve{\bm{\mathcal{I}}}_{k{-}}|\mathbf{y}_{1:{k}})\big\rangle_{q^s(\mathbf{x}_{1:K})} \label{eqn_sm_i1}
\end{align}	

Using the Bayes rule we can proceed as
\begin{flalign}
	{\breve{\mathcal{I}}}_k(i)=&\ \underset{{{\mathcal{I}}}_k(i)}{\mathrm{argmax}}\big\langle\mathrm{ln}(p(\mathbf{y}_{{k}}|\mathbf{x}_{k},{\mathcal{I}}_k(i),\breve{\bm{\mathcal{I}}}_k(i-),\stkout{\breve{\bm{\mathcal{I}}}_{k{-}},{\mathbf{y}_{k-} }} ))\nonumber\\
	&+\mathrm{ln} (p(\mathbf{x}_{1:K},{\mathcal{I}}_k(i),\breve{\bm{\mathcal{I}}}_k(i-),\breve{\bm{\mathcal{I}}}_{k{-}}|\mathbf{y}_{k-} )) \big\rangle_{q^s(\mathbf{x}_{1:K})}\nonumber \\
	&+const. & \label{eqn_sm_i2}
\end{flalign}
which leads to
\begin{flalign}
	{\breve{\mathcal{I}}}_k(i)=&\ \underset{{{\mathcal{I}}}_k(i)}{\mathrm{argmax}}\big\langle\mathrm{ln}(p(\mathbf{y}_{{k}}|\mathbf{x}_{k},{\mathcal{I}}_k(i),\breve{\bm{\mathcal{I}}}_k(i-) ))\big\rangle_{q^s(\mathbf{x}_{k})}  \nonumber\\
	&+\mathrm{ln} (p({\mathcal{I}}_k(i)|\stkout{{\mathbf{x}_{1:K}},{\breve{\bm{\mathcal{I}}}_k(i-)},{\hat{\bm{\mathcal{I}}}_{k{-}}},{\mathbf{y}_{k-}}} ))+const.  & \label{eqn_sm_i3}
\end{flalign}
which is similar to \eqref{eqn_fl_up2_i} except that the expectation is taken with respect to the marginal smoothing distribution ${q^s(\boldsymbol{\mathbf{x}_}{k})}$. Consequently, $\breve{\mathcal{I}}_k(i)$ can be determined as
\begin{align}
	{\breve{\mathcal{I}}}_k(i) &= 
	\begin{cases}
		1 & \text{if } \breve{\tau}_k(i) \leq 0,\\
		\epsilon & \text{if } \breve{\tau}_k(i) >0
	\end{cases} \label{Is}
\end{align}
where
\begin{flalign}
	\breve{\tau}_k(i) =&\  \Big\{ \mathrm{tr} \big( \bm{\mathcal{W}}_k \triangle{\breve{\mathbf{R}}}_{k}^{-1}  \big) + 
	\ln \Bigg|\textbf{I}-\frac{\mathbf{R}_k(-i,i)\mathbf{R}_k(i,-i) (\breve{\mathbf{R}}_{k}(-i,-i))^{-1}}{{R}^{i,i}_k}\Bigg| + \ln(\epsilon)\nonumber\\
	&+2\ln\Big(\frac{1}{\theta_k(i)}-1  \Big) \Big\} & \label{eqn_sm_up5_i}
\end{flalign}
with
\begin{align}
	\triangle{\breve{\mathbf{R}}}_{k}^{-1}=({\mathbf{R}}_{k}^{-1}( {\mathcal{I}_k(i)=1} , \breve{\bm{\mathcal{I}}}_k(i-) )-{\mathbf{R}}_{k}^{-1}( {\mathcal{I}_k(i)=\epsilon} , \breve{\bm{\mathcal{I}}}_k(i-) ))	
\end{align}
which can be be calculated similar to $\triangle{\hat{\mathbf{R}}}_{k}^{-1}$ in \eqref{eqdelR}. $\breve{\mathbf{R}}_{k}(-i,-i)$ denotes the submatrix of ${\mathbf{R}}_{k}({\boldsymbol{ \breve{\mathcal{I}} }_{k}})$ corresponding to entries of $\breve{\boldsymbol{\mathcal{I}}}^{i-}_k$ and
\begin{align}
	\bm{\mathcal{W}}_k=\int \left(\mathbf{y}_{k}-\mathbf{h}\left(\mathbf{x}_{k}\right)\right) \left(\mathbf{y}_{k}-\mathbf{h}\left(\mathbf{x}_{k}\right)\right)^{\top} {\mathcal{N}}({\mathbf{x}_k}|\hat{\bm{\mathit{x}}}^{s}_{k},\bm{\mathcal{P}}^{s}_{k})d\mathbf{x}_k\nonumber
\end{align}

The resulting EM-based outlier-robust smoother (EMORS) is outlined as Algorithm \ref{Algo2_EM}. We suggest using the same convergence criterion and parameters as for robust filtering.

\section{Performance Bounds}\label{Bounds}
\hspace{.5cm}
It is useful to determine the performance bounds of outlier-discarding state estimators considering correlated measurement noise. We evaluate the estimation bounds of filtering and smoothing approaches that are perfect outlier rejectors, having complete knowledge of the instances of outlier occurrences. In particular, we assume that the measurement covariance matrix is a function of perfectly known values of $\bm{\mathcal{I}}_k$ given as $\mathbf{R}_k(\bm{\mathcal{I}}_k)$. In this case, ${\mathcal{I}}_k(i)=0$ means rejection of the $i$th corrupted dimension, whereas ${\mathcal{I}}_k(i)=1$ denotes inclusion of the $i$th measurement. Resultingly, $\mathbf{R}^{-1}_k(\bm{\mathcal{I}}_k)$ has zeros at the diagonals, rows, and columns corresponding to dimensions for which ${\mathcal{I}}_k(i)=0$. Remaining submatrix of $\mathbf{R}^{-1}_k(\bm{\mathcal{I}}_k)$ can be evaluated as the inverse of submatrix of $\mathbf{R}_k$ considering the dimensions with ${\mathcal{I}}_k(i)=1$.

Note that we set ${\mathcal{I}}_k(i)=\epsilon$, not exactly as $0$, for outlier rejection in the proposed state estimators as it declines inference. However, for evaluating performance bounds this choice is appropriate resulting in perfect outlier rejection. Also note that during robust state estimation, we do not exactly know $\bm{\mathcal{I}}_k$ apriori and model it statistically for subsequent inference. The use of perfectly known $\bm{\mathcal{I}}_k$ for estimation bounds gives us an idea of how well we can estimate the state if outliers are somehow perfectly detected and rejected. 

We evaluate BCRBs for the perfect rejector for the model in \eqref{eqn_model_1_EM}-\eqref{eqn_model_2_EM} that have been corrupted with measurement outliers for both filtering and smoothing.

\subsection{Filtering}\hspace{.5cm}
For the estimation error of $\mathbf{x}_{k}$ during filtering, the BCRB matrix can be written as \cite{van2004detection}

\begin{equation*}
	\text{BCRB}^f_{k}\triangleq ({\mathbf {J}^{+}_{k}})^{-1} \tag{49}
\end{equation*}
where the corresponding filtering Fisher information matrix (FIM) denoted as $\mathbf{J}^{+}_{k}$ can be evaluated recursively as 
\begin{align}
	&\mathbf{J}^{-}_{k}=\mathbf {D}_{k-1}^{22}(1)-\mathbf {D}_{k-1}^{21}\left(\mathbf {J}^{+}_{k-1}+\mathbf {D}_{k-1}^{11}\right)^{-1} \mathbf {D}_{k-1}^{12} \\
	&\mathbf{J}^{+}_{k}=\mathbf{J}^{-}_{k}+\mathbf {D}_{k-1}^{22}(2)
\end{align}
where $\mathbf{J}^{+}_{0}=\langle -\Delta _{\mathbf {x}_{0}}^{\mathbf {x}_{0}} \ln p(\mathbf {x}_{0}) \rangle _{p(\mathbf {x}_{0})}$ and
\begin{align}
	&\Delta _{\Psi }^{\Theta }=\nabla _{\Psi } \nabla _{\Theta }^{\top } 
	\\
	&\nabla _{\Theta }=\left[\frac{\partial }{\partial \Theta _{1}},\ldots, \frac{\partial }{\partial \Theta _{r}}\right]^{\top }
\end{align}

\begin{flalign}
	\mathbf{D}_{k}^{11}&=\langle -\Delta _{\mathbf {x}_{k}}^{\mathbf {x}_{k}} \ln p\left(\mathbf {x}_{k+1} \mid \mathbf {x}_{k}\right)\rangle _{p(\mathbf {x}_{k+1},\mathbf {x}_{k})}
	&\\
	\mathbf {D}_{k}^{12}&=\langle -\Delta _{\mathbf {x}_{k}}^{\mathbf {x}_{k+1}} \ln p\left(\mathbf {x}_{k+1} \mid \mathbf {x}_{k}\right)\rangle _{p(\mathbf {x}_{k+1},\mathbf {x}_{k})}
	&\\
	\mathbf {D}_{k}^{21}&=\langle -\Delta _{\mathbf {x}_{k+1}}^{\mathbf {x}_{k}} \ln p\left(\mathbf {x}_{k+1} \mid \mathbf {x}_{k}\right)\rangle _{p(\mathbf {x}_{k+1},\mathbf {x}_{k})}=\left(\mathbf {D}_{k}^{12}\right)^{\top }
	&\\
	\mathbf {D}_{k}^{22}&= \mathbf {D}_{k}^{22}(1)+\mathbf {D}_{k}^{22}(2)
	&\\
	\mathbf {D}_{k}^{22}(1)&=\langle -\Delta _{\mathbf {x}_{k+1}}^{\mathbf {x}_{k+1}} \ln p\left(\mathbf {x}_{k+1} \mid \mathbf {x}_{k}\right)\rangle _{p(\mathbf {x}_{k+1},\mathbf {x}_{k})} &\\
	\mathbf{D}_{k}^{22}(2)&= \langle -\Delta _{\mathbf {x}_{k+1}}^{\mathbf {x}_{k+1}} \ln p\left(\mathbf {y}_{k+1} \mid \mathbf {x}_{k+1}\right)\rangle _{p(\mathbf {y}_{k+1},\mathbf {x}_{k+1})}& 
\end{flalign}

The bound is valid given the existence of the following derivatives and expectations terms for an asymptotically unbiased estimator \cite{668800}. For the perfect rejector considering the system model in \eqref{eqn_model_1_EM}-\eqref{eqn_model_2_EM} that is infested with observation outliers we can write
\begin{align}
	\mathbf {D}_{k}^{11}&=\langle \tilde{\mathbf{F}}^{\top}(\mathbf{x}_k) \mathbf {Q}_{k}^{-1} \tilde{\mathbf{F}}(\mathbf{x}_k) \rangle_{p(\mathbf{x}_k)} 
	\\
	\mathbf {D}_{k}^{12}&= -\langle \tilde{\mathbf{F}}^{\top}(\mathbf{x}_k) \rangle_{p(\mathbf{x}_k)} \mathbf {Q}_{k}^{-1}
	\\
	\mathbf {D}_{k}^{22}(1)&= \mathbf {Q}_{k}^{-1} \\
	\mathbf {D}_{k}^{22}(2)&= \langle \tilde{\mathbf {H}}^{\top}(\mathbf{x}_{k+1}) \mathbf {R}_{k+1}^{-1}(\bm{\mathcal{I}}_{k+1}) \tilde{\mathbf {H}}(\mathbf{x}_{k+1}) \rangle _{p(\mathbf {x}_{k+1})}
\end{align}
where $\tilde{\mathbf {F}}(.)$ and $\tilde{\mathbf {H}}(.)$ are the Jacobians of the transformations $\mathbf{f}(.)$ and $\mathbf{h}(.)$ respectively. Note that these bounds are general and can be applied to benchmark the performance of any rejection-based outlier robust filters e.g. presented in Chapter \ref{chap-3}~-~\ref{chap-4}.
\subsection{Smoothing}
\hspace{.5cm}
Similarly, for the estimation error of $\mathbf{x}_{k}$ during smoothing, the BCRB matrix can be written as \cite{SIMANDL20011703}
\begin{equation}
	\text{BCRB}^s_{k}\triangleq ({\mathbf {J}^{s}_{k}})^{-1} 
\end{equation}
where $\mathbf{J}^s_{K}=\mathbf{J}^{+}_{K}$. We can compute the associated smoothing FIM denoted as $\mathbf{J}^{s}_{k}$ recursively as
\begin{align}
	&\mathbf{J}^{s}_{k}=\mathbf{J}^{+}_{k}+\mathbf {D}_{k}^{11}-\mathbf {D}_{k}^{12}\left(\mathbf {D}_{k}^{22}(1)+\mathbf{J}^s_{k+1}+\mathbf {J}^{-}_{k+1}\right)^{-1} \mathbf {D}_{k}^{21}
\end{align} 

\section{Numerical Experiments}\label{exp}
To test the performance of the proposed outlier-resilient state estimators, we carry out several numerical experiments.  We use Matlab on a computer powered by an Intel i7-8550U processor. All the experiments were conducted while considering SI units.

\begin{figure}[h!]
	\centering
	\includegraphics[width=.7\linewidth]{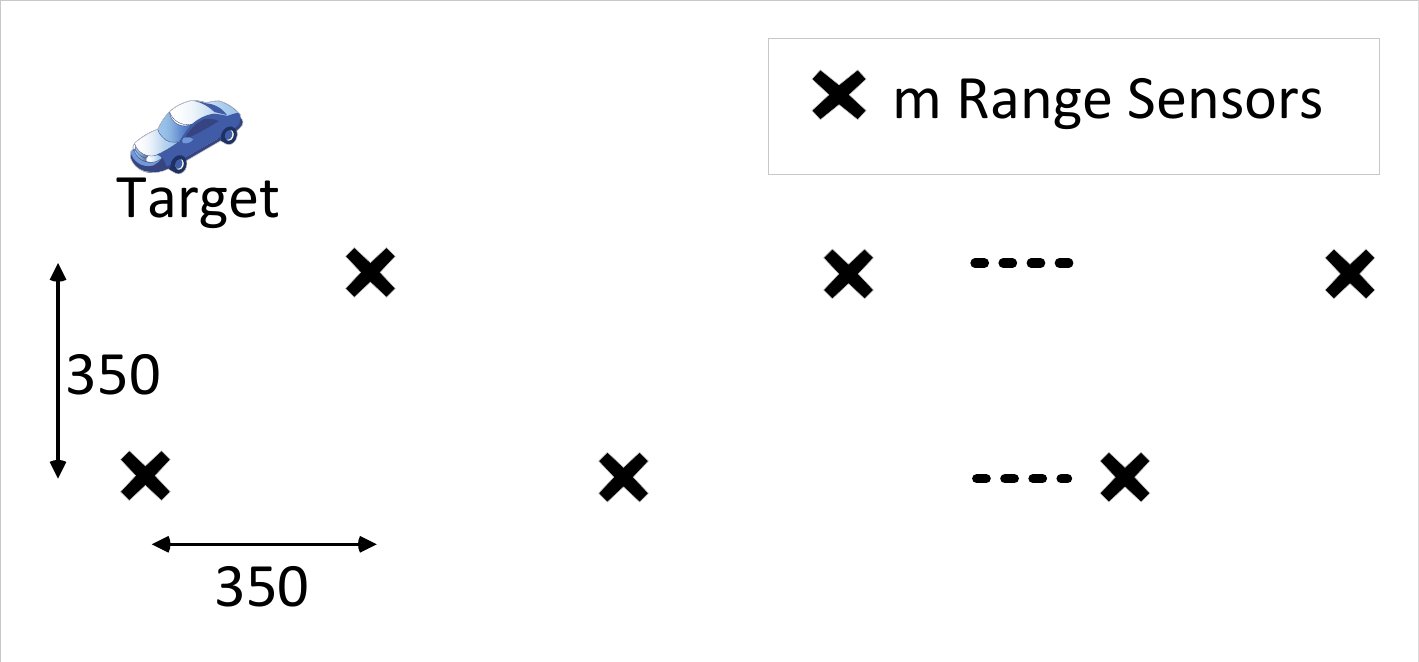}
	\caption{Target tracking test example setup } 
	\label{fig:target}
\end{figure}

For performance evaluation, we resort to a target tracking problem with TDOA-based range measurements inspired by \cite{5977569}. Fig.~\ref{fig:target} shows the setup of the considered example. Owing to the use of a common reference sensor to obtain the TDOA observations, from the difference of the time of arrival (TOA) measurements, the resulting covariance matrix becomes fully populated. 

We consider the process equation for the target assuming an unknown turning rate as \cite{8398426}

\begin{align}
	\mathbf{x}_k=\mathbf{f}(\mathbf{x}_{k-1})+\mathbf{q}_{k-1}\label{eqn_res1a}
\end{align}
with
\begin{align}
	\mathbf{f}(\mathbf{x}_{k-1}) &= \begin{bmatrix} \text{1} & \frac{\text{sin}(\omega_{k-1}{{\zeta}})}{\omega_{k-1}} & \text{0} &  \frac{\text{cos}(\omega_{k-1}{{\zeta}})-\text{1}}{\omega_{k-1}} & \text{0} \\  \text{0} & \text{cos}(\omega_{k-1}{{\zeta}}) &  \text{0} & -\text{sin}(\omega_{k-1}{{\zeta}}) & \text{0}\\  \text{0} & \frac{\text{1}-\text{cos}(\omega_{k-1}{{\zeta}})}{\omega_{k-1}} & \text{1} &\frac{\text{sin}(\omega_{k-1}{{\zeta}})}{\omega_{k-1}} & \text{0} \\  \text{0} &  \text{\text{sin}}(\omega_{k-1}{{\zeta}}) &  \text{0} &  \text{cos}(\omega_{k-1}{{\zeta}}) & \text{0}\\  \text{0} & \text{0} & \text{0} & \text{0} &\text{1} \end{bmatrix} \mathbf{x}_{k-\text{1}}\label{eqn_res1}
\end{align}
where the state vector $\mathbf{x}_k= [a_k,\dot{{a_k}},b_k,\dot{{b_k}},\omega_{k}]^{\top}$
is composed of the 2D position coordinates \(({a_k} , {b_k} )\), the corresponding velocities \((\dot{{a_k}} , \dot{{b_k}} )\), the angular velocity $\omega_{k}$ of the target at time instant $k$, \( {{\zeta}} \) denotes the sampling period, and $\mathbf{q}_{k-\text{1}} \sim N\left(0,\mathbf{Q}_{k-\text{1}}\right)$. $\mathbf{Q}_{k-\text{1}}$ is given in terms of scaling parameters $\eta_1$ and $\eta_2$ as \cite{8398426}
\begin{equation}
	\mathbf{Q}_{k-\text{1}}=\begin{bmatrix} \eta_1 \mathbf{M} & 0 & 0\\0 &\eta_1 \mathbf{M}&0\\0&0&\eta_2
	\end{bmatrix}, \mathbf{M}=\begin{bmatrix} {{\zeta}}^3/3 & {{\zeta}}^2/2\\{{\zeta}}^2/2 &{{\zeta}}
	\end{bmatrix}\nonumber
\end{equation}

Range readings are obtained using $m$ sensors installed in a zig-zag fashion as depicted in Fig.~\ref{fig:target}. The $i$th sensor is located at $\big(a^{\rho_i}=350(i-1),b^{\rho_i}=350\ ((i-1)\mod2)\big)$ for $i=1 \cdots m$. We assume the first sensor as the common sensor for reference resulting in $m-1$ TDOA-based measurements. The nominal measurement equation can be expressed as
\begin{equation}
	\mathbf{y}_k = \mathbf{h}(\mathbf{x}_{k})+\mathbf{r}_k\label{eqn_res2}
\end{equation}
with
\begin{flalign}
	{h^j(\mathbf{x}_k )} =& \Big\{ \sqrt{ (a_{k} - a^{\rho_1})^{2} + (b_{k} - b^{\rho_1})^{2} }  - \sqrt{ (a_{k} - a^{\rho_{j+1}})^{2} + (b_{k} - b^{\rho_{j+1}})^{2} } \Big\} & \label{eqn_res2b}
\end{flalign}
for $j=1 \cdots m-1$. The corresponding nominal covariance measurement matrix is fully populated given as \cite{5977569}
\begin{align}
	\mathbf{R}_k= \begin{bmatrix}
		{\sigma^2_1}+{\sigma^2_2}  & \dots &  {\sigma^2_1} \\
		\vdots &  \ddots& \vdots\\
		{\sigma^2_1}  & \cdots &{\sigma^2_1}+{\sigma^2_m}
	\end{bmatrix} \label{sim_R}
\end{align}
where $\sigma^2_i$ is the variance contribution of the $i$th sensor in the resulting covariance matrix. To consider the effect of outliers the measurement equation can be modified as
\begin{equation}
	\mathbf{y}_k = \mathbf{h}(\mathbf{x}_{k})+\mathbf{r}_k+\mathbf{o}_k\label{sim_o}
\end{equation}
where $\mathbf{o}_k$ produces the effect of outliers in the measurements and is assumed to obey the following distribution 
\begin{align}
	p(\mathbf{o}_k)&=\prod_{j=1}^{m-1} \mathcal{J}^j_k {\mathcal{N}}(o^{j}_k|0,\gamma ({\sigma^2_1}+{\sigma^2_j})) \label{eq_sim_o}
\end{align}
where $\mathcal{J}^j_k$ is a Bernoulli random variable, with values $0$ and $1$, that controls whether an outlier in the $j$th dimension occurs. Let $\lambda$ denote the probability that a sensor's TOA measurement is affected. Therefore, the probability that no outlier appears in the $j$th dimension, corresponding to $\mathcal{J}^j_k=0$, is $(1-\lambda)^2$ since the first sensor is a common reference for the TDOA-based measurements. We assume that the TOA measurements are independently affected and the corruption of the first TOA observation affects all the measurements. Similarly, the parameter $\gamma$ controls the variance of an outlier in each dimension respectively. Using the proposed model we generate the effect of outliers in the data. 

For filtering performance comparisons, we choose a hypothetical Gaussian filter that is a perfect rejector having apriori knowledge of all outlier instances. We also consider the generalized and independent VBKF estimators \cite{9286419}, referred from hereon as {Gen. VBKFs} and Ind. VBKF, for comparisons. In Gen. VBKFs, we set the design parameter as $N=1$ and {$N=10$ as originally reported}. Lastly, we use the derived BCRB-based filtering lower bounds to benchmark the performance of all the filters. Similarly, for smoothing we use the counterparts of all the considered filters i.e. a perfect outlier-rejecting general Gaussian RTS smoother and the generalized/independent VBKF-based RTS smoothers denoted as {Gen. VBKSs} and Ind. VBKS.

For simulations the following values of parameters are used: the initial state $\mathbf{x}_0= [0,1,0,-1,-0.0524] ^\text{T}$, $\zeta=1$, $\eta_1=0.1$, $\eta_2=1.75\times10^{-4}$, ${\sigma^2_j}=10$. The initialization parameters of estimators are: {${\mathbf{m}}^+_{0} \sim \mathcal{N}(\mathbf{x}_0,\mathbf{P}^+_{0}$), $\mathbf{P}^+_{0}=\mathbf{Q}_{k}$}, $\epsilon=10^{-6}$ and $\theta^i_k=0.5~\forall~i$. For {fairness we use UT in each method} for approximating the Gaussian integrals \cite{wan2001unscented}, in all the considered methods. Resultingly, the Unscented Kalman Filter (UKF) becomes the core inferential engine for all the techniques.  

In each method, UT parameters are set as $\alpha=1$, $\beta=2$, and $\kappa=0$. Moreover, we use the same threshold of $10^{-4}$ for the convergence criterion in each algorithm. Other parameters for VBKFs/VBKSs are assigned values as originally documented. All the simulations are repeated with a total time duration {{$K=100$}} and $100$ independent MC runs. Moreover, we use box and whisker plots to visualize all the results.  


\subsection{Filtering Performance}
We assess the relative filtering performance under different scenarios. 
\begin{figure}[h!]
	\centering
	\includegraphics[width=.7\linewidth]{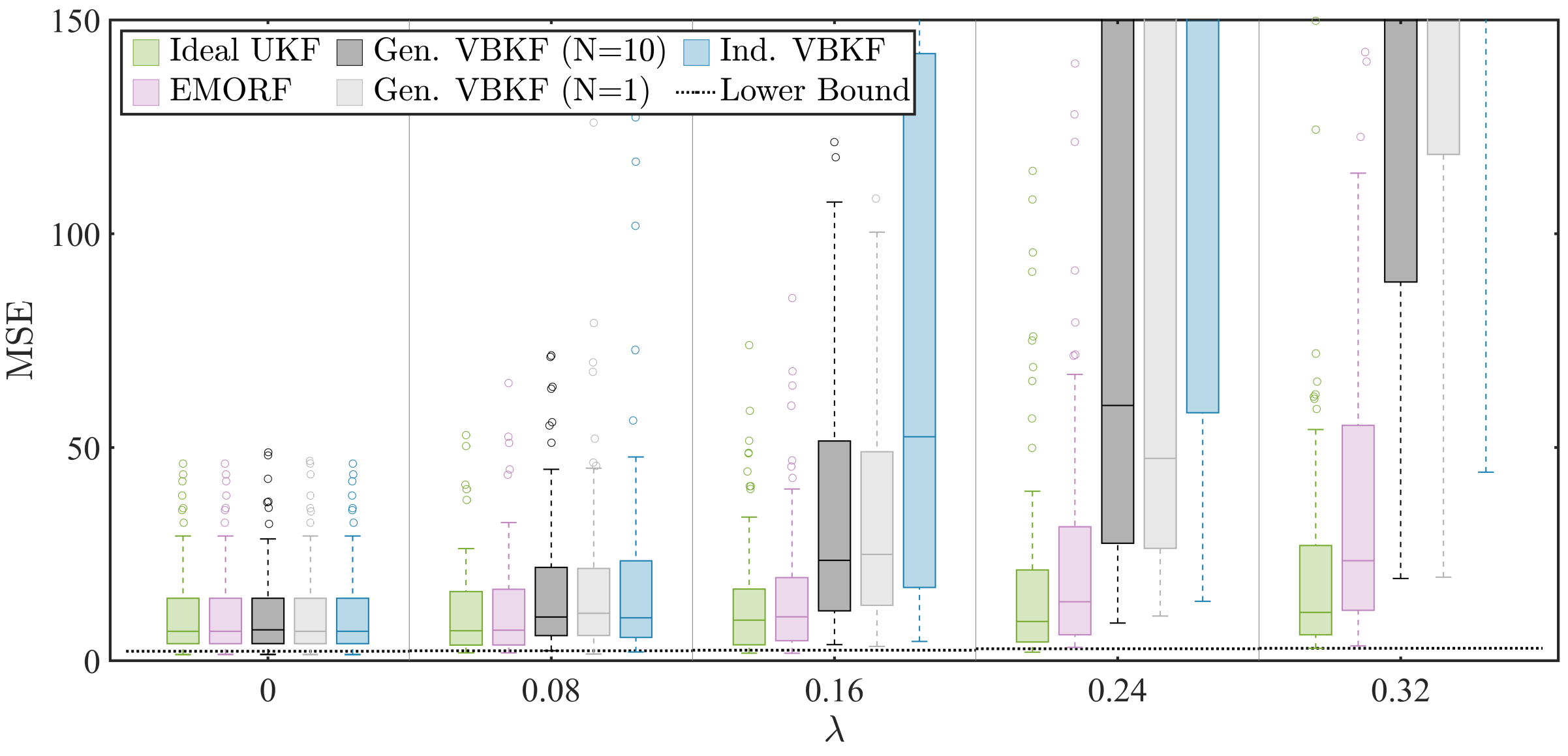}
	\caption{MSE vs $\lambda$ $(m=10,\gamma=1000)$} 
	\label{fig:fil_lambda}
\end{figure}

First, we choose $10$ number of sensors with $\gamma=1000$ and increase the TOA contamination probability $\lambda$. Fig.~\ref{fig:fil_lambda} shows the mean squared error (MSE) of the state estimate of each filter as $\lambda$ is increased. For $\lambda=0$ all the filters essentially work as the standard UKF having similar performance. As $\lambda$ increases, MSE of each method and the lower bound value are seen to increase. The hypothetical ideal UKF exhibits the best performance followed by the proposed EMORF, Gen. VBKFs, and Ind. VBKF respectively. {Gen. VBKF ($N=10$) generally tends to exhibit slightly improved performance as compared to Gen. VBKF ($N=1$) especially at higher values of $\lambda$}. The trend remains the same for each $\lambda$. Similar patterns have been observed for other combinations of $m$ and $\gamma$. Performance degradation of Ind. VBKF as compared to EMORF and Gen. VBKFs is expectable as it ignores the measurement correlations during filtering. We find EMORF to be generally more robust in comparison to Gen. VBKFs. Our results are not surprising given that we found the modified selective observation rejecting (mSOR)-UKF to be more resilient to outliers as compared to the modified outlier-detecting (mOD)-UKF \cite{chughtai2022outlier}, which are designed for independent measurements having similar structures to EMORF and VBKF respectively. 

\begin{figure}[h!]
	\centering
	\includegraphics[width=.7\linewidth]{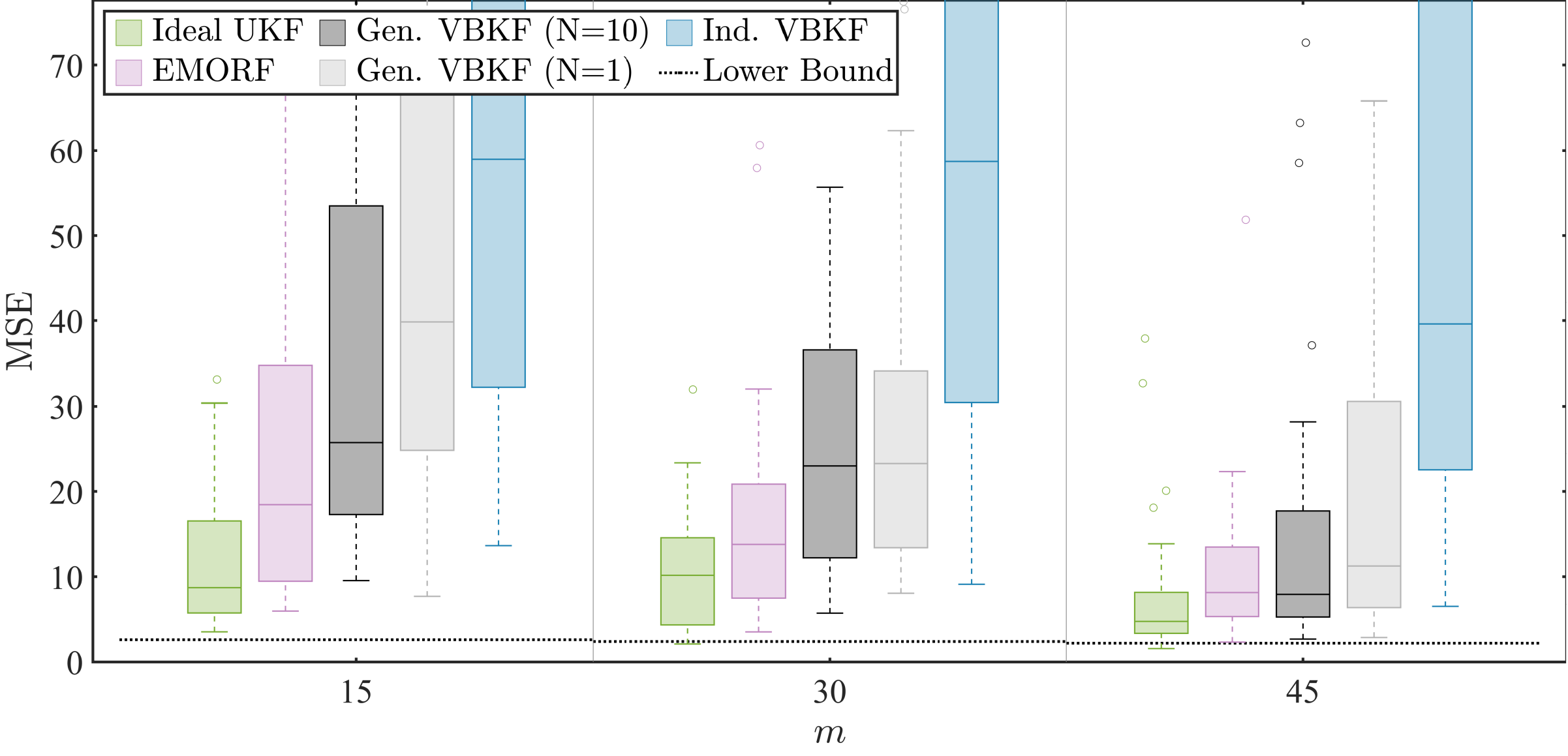}
	\caption{MSE vs $m$ ($\lambda=0.3$ and $\gamma=200$) } 
	\label{fig:fil_sensors}
\end{figure}
Next, we vary the number of sensors and assess the estimation performance of the filters. Fig.~\ref{fig:fil_sensors} shows the MSE of each method as the number of sensors is increased with {$\lambda=0.3$} and {$\gamma=200$}. As expected, the error bound and MSE of each filter decrease with increasing number of sensors since more sources of information become available. We see a pattern similar to the previous case with the best performance exhibited by the hypothetical ideal UKF followed by EMORF, Gen. VBKFs, and Ind. VBKF respectively. {In this simulation setting also, we find Gen. VBKF ($N=10$) to generally have marginally  improved performance as compared to Gen. VBKF ($N=1$).} Moreover, we have observed similar trends for other values of $\lambda$ and $\gamma$ as well.
\begin{figure}[h!]
	\centering
	\includegraphics[width=.7\linewidth]{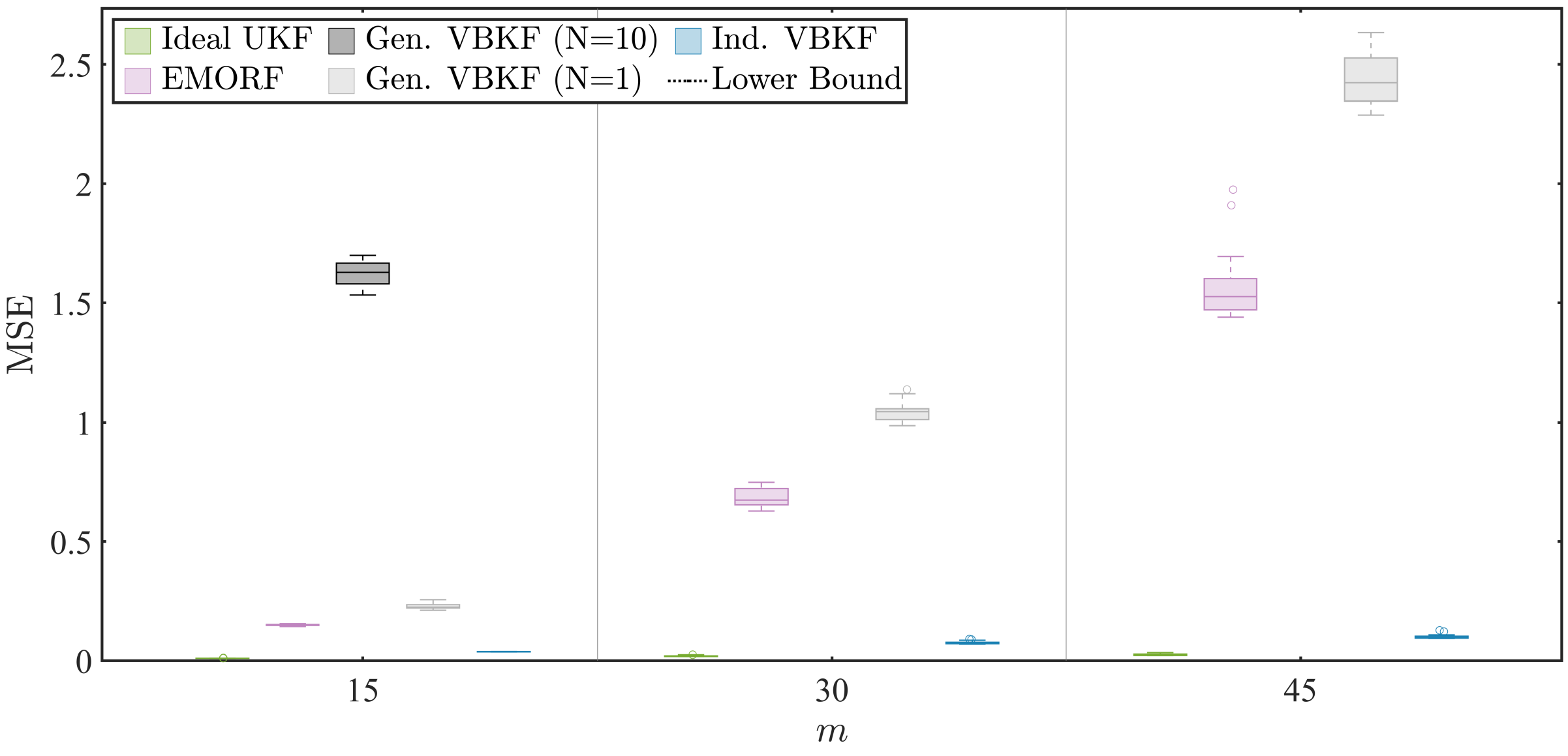}
	\caption{Computational Time vs $m$ ($\lambda=0.3$ and $\gamma=200$) } 
	\label{fig:fil_times}
\end{figure} 

Subsequently, we evaluate the processing overhead of each algorithm by varying the number of sensors. Fig.~\ref{fig:fil_times} shows the execution time taken by each algorithm as the number of sensors is increased with {$\lambda=0.3$} and {$\gamma=200$}. We observe that the ideal UKF and Ind. VBKF take lesser time for execution having a complexity of $\mathcal{O}(m^3)$. However, EMORF and Gen. VBKFs induce more computational overhead, having a complexity of $\mathcal{O}(m^4)$, due to utilization of matrix inverses and determinants for evaluating each of the $\mathcal{I}^i_k$ and $\mathbf{z}^{(i)}_t$ $\forall\ i=1\cdots m$ in EMORF and Gen. VBKFs respectively. This is the cost we pay for achieving robustness with correlated measurement noise. {Gen. VBKF ($N=10$) results in much higher computational cost as compared to Gen. VBKF ($N=1$) due evaluation of complicated expectation expressions. The gains in terms of error reduction do not appear to justify the increase in computational cost in this setting.} We find that EMORF generally takes less processing time as compared to {Gen. VBKF $(N=1)$} as shown in Fig.~\ref{fig:fil_times}. Moreover, similar performance has been observed for other combinations of $\lambda$ and $\gamma$. This can be attributed to a simpler model being employed in EMORF resulting in reduced computations. 

\subsection{Smoothing Performance}
For smoothing we perform analogous experiments and observe similar performance.  
\begin{figure}[h!]
	\centering
	\includegraphics[width=.7\linewidth]{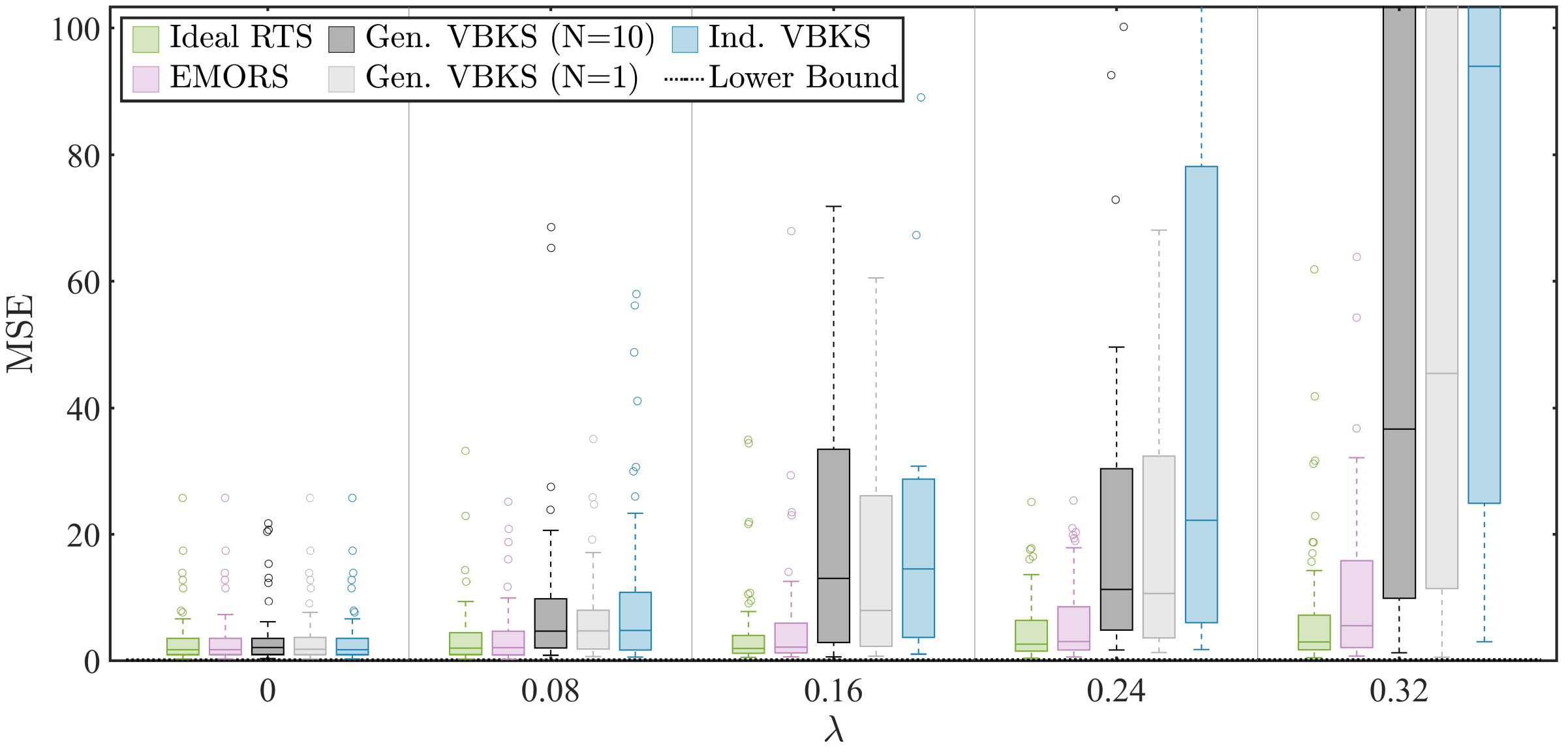}
	\caption{MSE vs $\lambda$ $(m=10,\gamma=500)$} 
	\label{fig:lam_smo}
\end{figure}

First, we choose {$10$} number of sensors and increase the TOA contamination probability $\lambda$ with {$\gamma=500$}. Fig.~\ref{fig:lam_smo} shows how MSE of the state estimate of each smoother changes as $\lambda$ is increased. Similar to filtering, we observe that MSE of each estimator grows with increasing $\lambda$ including the BCRB-based smoothing lower bound. The hypothetical RTS smoother performs the best followed by the proposed EMORS, {Gen. VBKS ($N=10$), Gen. VBKS ($N=1$)} and Ind. VBKS respectively. The trend remains the same for each $\lambda$. Similar patterns have been seen for other combinations of $m$ and $\gamma$. 

\begin{figure}[h!]
	\centering
	\includegraphics[width=.7\linewidth]{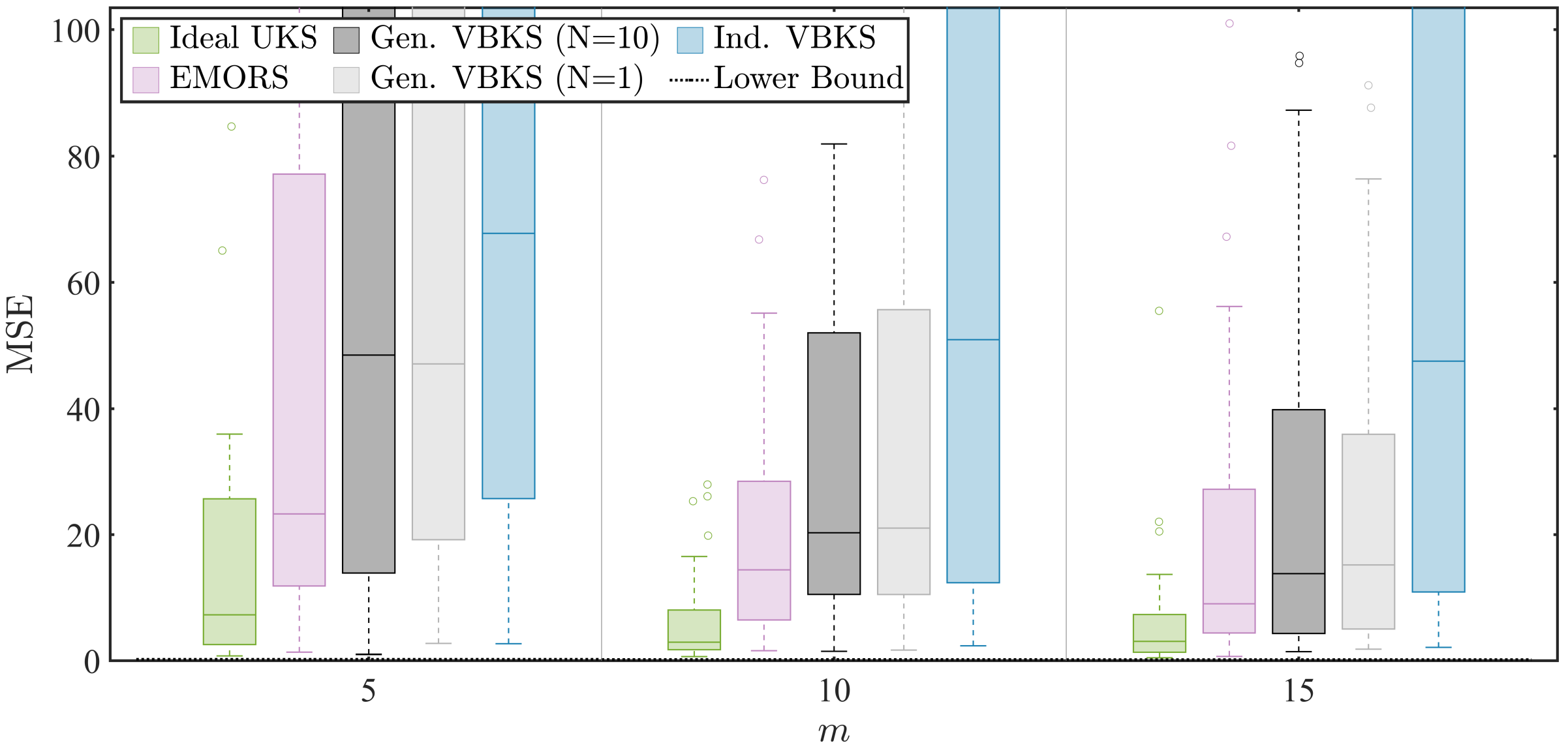}
	\caption{MSE vs $m$ ($\lambda=0.4$ and $\gamma=100$) } 
	\label{fig:smo_sensors}
\end{figure}
Thereafter, we assess the estimation performance of the filters by varying the number of sensors. Fig.~\ref{fig:smo_sensors} depicts MSE of each estimator as the number of sensors increase with {$\lambda=0.4$} and {$\gamma=100$}. MSE for each smoother decreases with growing number of sensors including the BRCB-based lower bound. The hypothetical RTS smoother is {generally} the best performing followed by EMORS, {Gen. VBKS ($N=10$), Gen. VBKS ($N=1$)}, and Ind. VBKS respectively. We have observed similar trends for other values of $\lambda$ and $\gamma$ as well.

\begin{figure}[h!]
	\centering
	\includegraphics[width=.7\linewidth]{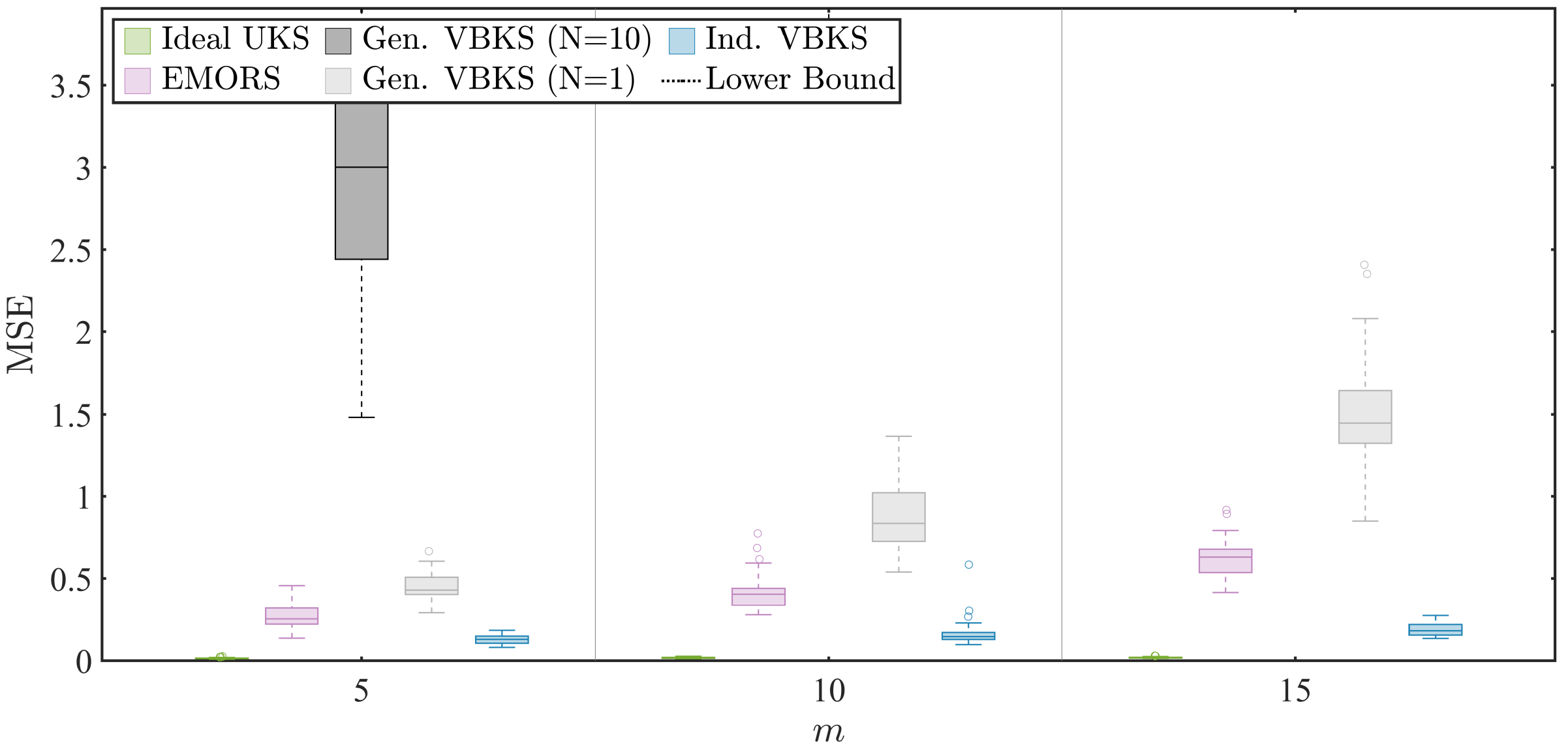}
	\caption{Computational Time vs $m$ ($\lambda=0.4$ and $\gamma=100$) } 
	\label{fig:smo_times}
\end{figure}
Lastly, we evaluate the computational overhead of each algorithm by varying the number of sensors. Fig.~\ref{fig:smo_times} shows the time each method takes as the number of sensors is increased with {$\lambda=0.4$} and {$\gamma=100$}. We observe similar patterns as for filtering. The ideal RTS smoother and Ind. VBKS having a complexity of $\mathcal{O}(m^3)$ take lesser execution time. EMORS and Gen. VBKSs having a complexity $\mathcal{O}(m^4)$ are more time consuming {with Gen. VBKS $(N=10)$ taking much higher computational time than Gen. VBKS $(N=1)$}.  EMORS generally induces lesser computing overhead as compared to {Gen. VBKS ($N=1$)} as depicted in Fig.~\ref{fig:smo_times}. We have observed similar patterns for different combinations of $\lambda$ and $\gamma$.

{\subsection*{Other simulations} 
	We repeated the numerical experiments for all the considered filtering and smoothing methods under other scenarios as well. In particular, we replaced the Gaussian distribution in \eqref{eq_sim_o} to Uniform and Laplace distributions and experimented with different distributional parameters. We generally find a similar trend of comparative performance as depicted in Fig.~\ref{fig:fil_lambda}-Fig.~\ref{fig:smo_times}. We also tested the performance of EMORF/S with other choice of the prior parameter $\theta_k^i$. Following the lines of SORF, we repeated the simulations with the prior parameter $\theta_k^i$ drawn from a Uniform density defined over a range $0.05$ to $0.95$ and got similar results as above. These results indicate that the performance of EMORF/S is robust to the changes in the prior parameter. However, the default value of $\theta_k^i=0.5$ is preferable unless some specific prior information regarding the rate of outliers is available.}

\section{Conclusion}\label{conc}
We consider the problem of outlier-robust state estimation assuming the existence of measurement noise correlation. Given their advantages, resorting to tuning-free learning-based approaches is an attractive option in this regard. Identifying the shortcomings of such existing VB-based tractable methods, we propose EMORF and EMORS. Since the standard VB approach entails significant processing complexity, we adopt EM in our algorithmic constructions. We can conclude that the presented methods are simpler and hence more practicable as compared to the state-of-the-art Gen. {VBKFs/VBKSs}, devised for the same conditions. This is possible due to the reduction of inference parameters resulting from the proposal of an uncomplicated model. Also, the need of the specialized digamma function during implementation is obviated. In addition, numerical experiments in an illustrative TDOA-based target tracking example suggest further merits of the proposed methods. We find that EMORF/EMORS generally exhibit lesser errors as compared to Gen. and Ind. {VBKFs/VBKSs} in different scenarios of the example. Moreover, though the complexity order of EMORF/EMORS and Gen. {VBKFs/VBKSs} is the same, the proposed estimators are found to be computationally more competitive in general for different test conditions. These merits make the proposed state estimators worthy candidates for implementation in relevant scenarios.   

\chapter{Robust Filtering with Biased Observations} \label{chap-6b}
\hspace{.5cm}
In this chapter, we turn our attention to the occurrence of measurement biases and propose a robust filter for their detection and mitigation during state estimation of nonlinear dynamical systems. We model the presence of bias in each dimension within the generative structure of the state-space models. Subsequently, employing the VB theory and general Gaussian filtering, we devise a recursive filter which we call the Bias Detecting and Mitigating (BDM) filter. As the error detection mechanism is embedded within the filter structure its dependence on any external detector is obviated. Simulations verify the performance gains of the proposed BDM filter compared to similar Kalman filtering-based approaches in terms of robustness to temporary and persistent bias presence.

\section{{Bias} Modeling}\label{modelling_sec}\hspace{.5cm}
\textcolor{black}{As the standard SSM does not consider the possibility of \textcolor{black}{measurement biases} in its \textit{generative} structure \cite{6266757} it needs to be modified}. At the same time, the model should remain amenable for VB inference. To this end, we propose the SSM model as 
\begin{align}
	\mathbf{x}_k= & \mathbf{f}(\mathbf{x}_{k-1})+\mathbf{q}_{k-1}\label{eqn1_VB}\\
	\mathbf{y}_k =& \mathbf{h}(\mathbf{x}_{k})+\mathbf{r}_k + \boldsymbol{{\mathcal{I}}}_{k}{\mathbf{\Theta}}_k \label{eqn2_VB}
\end{align}
where $k$ denotes the time-index,  $\mathbf{x}_k\in\mathbb{R}^n$ and $\mathbf{y}_k\in\mathbb{R}^m$ are the state and measurement vectors respectively, $\mathbf{q}_{k-1}\in\mathbb{R}^n$ and $\mathbf{r}_k\in\mathbb{R}^m$ are white process and  measurement noise vectors, $\mathbf{f}(.)$ and $\mathbf{h}(.)$ represent nonlinear process and measurement dynamics respectively, $\mathbf{\Theta}_k\in\mathbb{R}^m$ models the effect of biases in the measurements and $\boldsymbol{\mathcal{I}}_{k} \in \mathbb{R}^{m\times m}$ is a diagonal matrix with Bernoulli elements ${\mathcal{I}}_{k}(i,i)$ used to indicate the occurrence of bias in different dimensions. $\boldsymbol{ \mathcal{I}}_{k} {\mathbf{\Theta}}_k$ models the combined effect of the occurrence and magnitude of bias appearing the data. We assume the following noise distributions: $\mathbf{q}_{k-1} \sim \mathcal{N}(\mathbf{0},\mathbf{Q}_{k-1})$ and $\mathbf{r}_k  \sim \mathcal{N}(\mathbf{0}, \mathbf{R}_k)$. We assume that measurements are obtained from independent sensors making $\mathbf{R}_k$ diagonal. The bias evolution is expressed as follows 
\begin{equation}
	{\mathbf{\Theta}}_k = (\mathbf{I}-\boldsymbol{ {\mathcal{I}}}_{k-1}){\widetilde{\mathbf{\Theta}}}_{k} +\boldsymbol{{\mathcal{I}}}_{k-1}({\mathbf{\Theta}}_{k-1} +  {\Delta_k})\label{eqn3}
\end{equation}
where in \eqref{eqn3}, each entry of ${\Delta_k}$ allows for any drifts/changes in the bias value over time, in the corresponding dimension, given bias was present at the previous time step. On the contrary, if no bias occurred in any given dimension, at the preceding instant, it can possibly occur with a very large variance ${\sigma^2_{\widetilde{{\Theta}}}}$ (assuming an uninformative prior) described by the respective entries of a zero mean random vector ${\widetilde{\mathbf{\Theta}}}_{k}$. The distributions of ${\Delta_k}$ and ${\widetilde{\mathbf{\Theta}}}_{k}$ are supposed to be White and normally distributed given as
\begin{align}
	{\Delta_k} &\sim \mathcal{N}(\mathbf{0},\breve{\mathbf{\Sigma}}_{k}) \text{ with } \breve{\mathbf{\Sigma}}_{k}= \mathrm{diag}\left(\sigma^2_{\vartriangle{1}}, \cdots, \sigma^2_{\vartriangle{m}} \right) \label{PPFeqn30}&\\
	{\widetilde{\mathbf{\Theta}}}_{k} &\sim \mathcal{N}(\mathbf{0},\widetilde{\mathbf{\Sigma}}_{k}) \text{ with } \widetilde{\mathbf{\Sigma}}_{k} = \mathrm{diag}\left({\sigma^2_{\widetilde{{\Theta}}}}, \cdots, {\sigma^2_{\widetilde{{\Theta}}}} \right) \label{PPFeqn31}
\end{align}

\section{Recursive Bayesian Inference}\label{inference}\hspace{.5cm}
Considering the inference model in \eqref{eqn1_VB}-\eqref{eqn3}, the Bayes rule can be employed recursively to express the joint posterior distribution of $\mathbf{x}_k$, $\bm{\mathcal{I}}_k$ (considering only the random entries ${\mathcal{I}}_k(i)$) and ${\mathbf{\Theta}}_k$ conditioned on the set of all the observations $\mathbf{y}_{1:{k}}$ analytically as 
\begin{equation}
	p(\mathbf{x}_k,\bm{\mathcal{I}}_k,{\mathbf{\Theta}}_k|\mathbf{y}_{1:{k}})=\frac{p(\mathbf{y}_k|\mathbf{x}_{k},\bm{\mathcal{I}}_k,{\mathbf{\Theta}}_k)	p(\mathbf{x}_k,\bm{\mathcal{I}}_k,{\mathbf{\Theta}}_k|\mathbf{y}_{1:{k-1}})}{p(\mathbf{y}_k|\mathbf{y}_{1:{k-1}})}
	\label{eqn_vb_1}
\end{equation} 

Theoretically, the joint posterior can be marginalized to obtain the expression for $p(\mathbf{x}_k|\mathbf{y}_{1:{k}})$. With this approach, the exact sequential Bayesian processing becomes computationally infeasible. Therefore, we adopt the VB method \cite{vsmidl2006variational} for inference where the product of VB marginals is conveniently used to approximate the joint posterior as
\begin{equation}
	p(\mathbf{x}_k,\bm{\mathcal{I}}_k,{\mathbf{\Theta}}_k|\mathbf{y}_{1:{k}})\approx q(\mathbf{x}_k)q(\bm{\mathcal{I}}_k)q({\mathbf{\Theta}}_k)
	\label{eqn_vb_2}
\end{equation} 

With an objective to minimize the KL divergence between the marginal product and the true posterior, the VB method leads to the following marginals 
\begin{align}
	q(\mathbf{x}_k)&\propto \exp( \big\langle\mathrm{ln}(p(\mathbf{x}_k,\bm{\mathcal{I}}_k,{\mathbf{\Theta}}_k|\mathbf{y}_{1:{k}})\big\rangle_{ q({{\bm{\mathcal{I}}}_k}) {q(\mathbf{\Theta}}_k)})\label{eqn_vb_3}\\
	q(\bm{\mathcal{I}}_k)&\propto \exp ( \big\langle\mathrm{ln}(p(\mathbf{x}_k,\bm{\mathcal{I}}_k,{\mathbf{\Theta}}_k|\mathbf{y}_{1:{k}})\big\rangle_{{q(\mathbf{x}_k){q(\mathbf{\Theta}}_k)}})\label{eqn_vb_4}\\
	q({\mathbf{\Theta}}_k)&\propto \exp ( \big\langle\mathrm{ln}(p(\mathbf{x}_k,\bm{\mathcal{I}}_k,{\mathbf{\Theta}}_k|\mathbf{y}_{1:{k}})\big\rangle_{q(\mathbf{x}_k)q(\bm{\mathcal{I}}_k)})\label{eqn_vb_5_}
\end{align} 
where $ \langle.\rangle_{q(\bm{\psi}_k)}$ denotes the expectation of the argument with respect to a distribution $q(\bm{\psi}_k)$. The VB marginals can be updated iteratively until convergence, using \eqref{eqn_vb_3}-\eqref{eqn_vb_5_} in turn. The procedure provides a convenient way to approximate the true marginals of the joint posterior by approximating these as $ p(\mathbf{x}_k|\mathbf{y}_{1:{k}})\approx {q^c(\mathbf{x}_k)}$, $	p(\bm{\mathcal{I}}_k|\mathbf{y}_{1:{k}})\approx{q^c(\bm{\mathcal{I}}_k)}$ and $	p({\mathbf{\Theta}}_k|\mathbf{y}_{1:{k}})\approx	{q^c({\mathbf{\Theta}}_k)}$ where ${q^c(.)}$ denotes the VB marginals obtained after convergence.
\vfill
\subsection{Prediction}\hspace{.5cm}
Assuming that at each time step the posterior is approximated with a product of marginals, the predictive density can be \textcolor{black}{approximated} as
\begin{align}
	&p(\mathbf{x}_k,\bm{\mathcal{I}}_k,{\mathbf{\Theta}}_k|\mathbf{y}_{1:{k-1}})=\int\int\int p(\mathbf{x}_k,\bm{\mathcal{I}}_k,{\mathbf{\Theta}}_k|\mathbf{x}_{k-1},\bm{\mathcal{I}}_{k-1},{\mathbf{\Theta}}_{k-1})\nonumber \\
	&p(\mathbf{x}_{k-1},\bm{\mathcal{I}}_{k-1},{\mathbf{\Theta}}_{k-1}|\mathbf{y}_{1:{k-1}})d\mathbf{x}_{k-1}d\bm{\mathcal{I}}_{k-1}d{\mathbf{\Theta}}_{k-1}\label{eqn_vb_6}\\
	&p(\mathbf{x}_k,\bm{\mathcal{I}}_k,{\mathbf{\Theta}}_k|\mathbf{y}_{1:{k-1}})\approx\int\int\int p(\mathbf{x}_k|\mathbf{x}_{k-1})p({\mathbf{\Theta}}_k|,\bm{\mathcal{I}}_{k-1},{\mathbf{\Theta}}_{k-1})\nonumber \\
	&p(\bm{\mathcal{I}}_k) p(\mathbf{x}_{k-1}|\mathbf{y}_{1:{k-1}})p(\bm{\mathcal{I}}_{k-1}|\mathbf{y}_{1:{k-1}})p({\mathbf{\Theta}}_{k-1}|\mathbf{y}_{1:{k-1}}) d\mathbf{x}_{k-1}d\bm{\mathcal{I}}_{k-1}d{\mathbf{\Theta}}_{k-1}\label{eqn_vb_7}\\
	&p(\mathbf{x}_k,\bm{\mathcal{I}}_k,{\mathbf{\Theta}}_k|\mathbf{y}_{1:{k-1}})\approx p(\bm{\mathcal{I}}_k)p(\mathbf{x}_k|\mathbf{y}_{{1:k-1}}) p({\mathbf{\Theta}}_k|\mathbf{y}_{1:k-1}) \label{eqn_vb_8}
\end{align} 
with 
\begin{align}
	p(\mathbf{x}_k|\mathbf{y}_{1:{k-1}})&= \int p(\mathbf{x}_k|\mathbf{x}_{k-1}) p(\mathbf{x}_{k-1}|\mathbf{y}_{1:{k-1}}) d\mathbf{x}_{k-1} \label{eqn_vb_9_}\\
	p({\mathbf{\Theta}}_k|\mathbf{y}_{1:k-1}) &\approx \int\int p({\mathbf{\Theta}}_k|\bm{\mathcal{I}}_{k-1},{\mathbf{\Theta}}_{k-1}) p(\bm{\mathcal{I}}_{k-1}|\mathbf{y}_{1:{k-1}})p({\mathbf{\Theta}}_{k-1}|\mathbf{y}_{1:{k-1}}) d\bm{\mathcal{I}}_{k-1}d{\mathbf{\Theta}}_{k-1}\label{eqn_vb_10_}
\end{align}

We assume that the occurrence of bias is independent for each dimension and its historical existence. Using $\theta_k(i)$ to denote the prior probability of occurrence of bias in the $i$th observation, the distribution of $\bm{\mathcal{I}}_k$ is defined as product of independent Bernoulli distributions of each element
\begin{equation}
	p(\bm{\mathcal{I}}_k)=\prod_{i=1}^{m}p({{\mathcal{I}}}_k(i,i))=\prod_{i=1}^{m} (1-{\theta_k(i)}) \delta({{{\mathcal{I}}}_k(i,i)})+{\theta_k(i)}\delta( {{{\mathcal{I}}}_k(i,i)}-1)
	\label{eqn_model_11}
\end{equation}
where $\delta(.)$ denotes the delta function.

To evaluate \eqref{eqn_vb_9_}-\eqref{eqn_vb_10_}, suppose the following distributions for the posterior marginals at time step $k-1$ 
\begin{align}
	p(\mathbf{x}_{k-1}|\mathbf{y}_{1:{k-1}})&\approx {q^c(\mathbf{x}_{k-1})}\approx \mathcal{N}(\mathbf{x}_{k-1}|\mathbf{\hat{x}}^+_{k-1},\mathbf{{P}}^+_{k-1}) \label{eqn_vb_11}\\
	p(\bm{\mathcal{I}}_{k-1}|\mathbf{y}_{1:{k-1}})&\approx {q^c(\bm{\mathcal{I}}_{k-1})}=\prod_{i=1}^{m}p({{\mathcal{I}}}_{k-1}(i,i)|\mathbf{y}_{1:{k-1}})\nonumber \\  &= \prod_{i=1}^{m} (1-{\Omega_{k-1}(i,i)}) \delta({{{\mathcal{I}}}_{k-1}(i,i)})+{\Omega_{k-1}(i,i)}\delta( {{{\mathcal{I}}}_{k-1}}(i,i)-1) \label{eqn_vb_12}\\
	p({\mathbf{\Theta}}_{k-1}|\mathbf{y}_{1:{k-1}})&\approx {q^c({\mathbf{\Theta}}_{k-1})} \approx\mathcal{N}(\mathbf{\Theta}_{k-1}|\mathbf{\hat{\Theta}}^+_{k-1},\mathbf{{\Sigma}}^+_{k-1})\label{eqn_vb_13}
\end{align}
where ${\Omega_{k}(i,i)}$ denotes the posterior probability of bias occurrence in the $i$th dimension. The notation $\mathcal{N}(\mathbf{x}|\mathbf{m},\mathbf{{\Sigma}})$ represents a multivariate normal distribution with mean $\mathbf{m}$ and covariance $\mathbf{{\Sigma}}$, evaluated at $\mathbf{x}$. The verification of the functional forms of the distributions and the expressions of their parameters are provided in the subsequent update step of the Bayesian filter.

Since $\mathbf{f}(.)$ is assumed to \textcolor{black}{be} nonlinear, $p(\mathbf{x}_k|\mathbf{y}_{1:k-1})$ can be approximated, using general Gaussian filtering results \cite{sarkka2023bayesian}, as $\mathcal{N}(\mathbf{x}_{k}|\mathbf{\hat{x}}^-_{k},\mathbf{{P}}^-_{k})$ with the parameters predicted as follows 
\begin{align}
	\mathbf{\hat{x}}^{-}_{k}&=\big\langle \mathbf{f}(\mathbf{x}_{k-1})\big\rangle_{p(\mathbf{x}_{k-1}|\mathbf{y}_{1:k-1})}\label{eqn_vb_14}\\
	\mathbf{P}^{-}_{k}&=\big\langle(\mathbf{f}(\mathbf{x}_{k-1})-\mathbf{\hat{x}}^{-}_{k})(\mathbf{f}(\mathbf{x}_{k-1})-\mathbf{\hat{x}}^{-}_{k})^{\top}\big\rangle_{p(\mathbf{x}_{k-1}|\mathbf{y}_{1:{k-1}})}+\mathbf{Q}_{k-1}\label{eqn_vb_15}
\end{align}

The remaining term required to approximate the predictive density recursively in \eqref{eqn_vb_8} is $p({\mathbf{\Theta}}_k|\mathbf{y}_{k-1})$. Observing \eqref{eqn3} and \eqref{eqn_vb_10_}, it is evident that $p({\mathbf{\Theta}}_k|\mathbf{y}_{k-1})$ is a sum of $2^m$ Gaussian densities scaled by the probabilities of combinations of bias occurrence at previous time instance. Obviously this makes recursive inference intractable, so we propose to approximate this distribution with a single Gaussian density $\mathcal{N}(\mathbf{\Theta}_{k}|\mathbf{\hat{\Theta}}^-_{k},\mathbf{{\Sigma}}^-_{k})$ using moment matching \cite{sarkka2023bayesian}. The parameters of the distribution are updated as
\begin{align}
	\mathbf{\hat{\Theta}}^{-}_{k}&=\mathbf{\Omega}_{k-1}\mathbf{\hat{\Theta}}^{+}_{k-1}\label{eqn_vb_16}\\
	\mathbf{\Sigma}^{-}_{k}&=(\mathbf{I}-\mathbf{\Omega}_{k-1}) \widetilde{\mathbf{\Sigma}}_{k}+\mathbf{\Omega}_{k-1}\breve{\mathbf{\Sigma}}_{k}+\mathbf{{\Sigma}}^+_{k-1}\odot(\mathrm{diag}(\mathbf{\Omega}_{k-1}){\mathrm{diag}(\mathbf{\Omega}_{k-1})}^{\top}+\mathbf{\Omega}_{k-1}(\mathbf{I}-\mathbf{\Omega}_{k-1}))&\nonumber \\
	&\ \ +\mathbf{\Omega}_{k-1}(\mathbf{I}-\mathbf{\Omega}_{k-1})(\mathrm{diag}(\mathbf{\hat{\Theta}}^{+}_{k-1}))^2 &	
	\label{eqn_vb_17}
\end{align}
where $\mathbf{\Omega}_{k-1}$ is a diagonal matrix with entries ${\Omega}_{k-1}(i,i)$ denoting the posterior probability of bias occurrence at time step $k-1$. The operator $\odot$ is the Hadamard product and $\mathrm{diag}$ is used for vector to diagonal matrix conversion and vice versa. The reader is referred to the Appendix \ref{bias_pred} for detailed derivations of \eqref{eqn_vb_16}-\eqref{eqn_vb_17}.
\subsubsection*{Remarks}\hspace{.5cm}
We note the following in \eqref{eqn_vb_16}-\eqref{eqn_vb_17}
\begin{itemize}
	\item $\mathbf{\Omega}_{k-1}$ dictates how the parameters $\mathbf{\hat{\Theta}}^-_{k}$ and $\mathbf{{\Sigma}}^-_{k}$ are predicted.
	\item For the case when $\mathbf{\Omega}_{k-1}=\mathbf{I}$, i.e. bias is inferred in each dimension at time step $k-1$ with probability 1, $	\mathbf{\hat{\Theta}}^{-}_{k}=\mathbf{\hat{\Theta}}^{+}_{k-1}$ and  $\mathbf{\Sigma}^{-}_{k}=\mathbf{{\Sigma}}^+_{k-1}+\breve{\mathbf{\Sigma}}_{k}$. In other words, the mean of the bias prediction (for each dimension) is retained and its covariance is predicted as sum of previous covariance and the covariance considered for the amount of drift/change in the bias.
	\item For the case when $\mathbf{\Omega}_{k-1}=\mathbf{0}$, i.e. no bias is inferred in each dimension at time step $k-1$ with probability 1, $	\mathbf{\hat{\Theta}}^{-}_{k}=\mathbf{0}$ and  $\mathbf{\Sigma}^{-}_{k}=\widetilde{\mathbf{\Sigma}}_{k}$. In other words, the mean of the bias prediction (for each dimension) is $\mathbf{0}$ and its covariance is predicted with very large entries.
	\item Similarly, the prediction mechanism can be understood if only some dimensions are inferred to be disturbed with probability 1. The bias in the particular dimensions are predicted with the mean retained and covariance updated as addition of previous covariance plus the covariance allowed for the drift/change.
	\item Lastly, if there is partial confidence on the occurrence of bias in any dimension at $k-1$, the predicted Gaussian distribution is shifted to the mean of bias estimate at $k-1$ scaled with a factor of ${\Omega}_{k-1}(i,i)$. In addition, the covariance gets inflated by addition of scaled elements of $\widetilde{\mathbf{\Sigma}}_{k}$ and squared terms of mean at $k-1$. In other words, it can be interpreted in a sense that unless there is a very high confidence of occurrence of bias at the previous time instance, the bias would be predicted with a large covariance.       
\end{itemize}

\subsection{Update}\hspace{.5cm}
For the update step, we resort to \eqref{eqn_vb_1}-\eqref{eqn_vb_5_} and use \eqref{eqn_vb_8} for approximating the predictive density. For detailed derivations, the reader is referred to Appendices \ref{bias_up1}~-~\ref{bias_up3}. 

Parameters of $q(\mathbf{x}_k)$ are updated iteratively as
\begin{flalign}
	\hat {\mathbf{x}} _{k} ^{+} &= \hat {\mathbf{x}}_{k}^{-} + \mathbf{K}_k(\mathbf{y}_{k} - {\boldsymbol{\Omega}} _{k} \mathbf{\hat{\Theta}}^+_{k} - \bm{\mu}_k) &\label{eqn_vb_18} \\
	\bm{\mu}_k&=\langle \mathbf{h}(\mathbf{x}_{k})  \rangle_{p(\mathbf{x}_k|\mathbf{y}_{k-1})}& \label{eqn_vb_19}\\
	\mathbf{P}_{k}^{+} &= \mathbf{P}_{k}^{-} - \mathbf{C}_{k}\mathbf{K}_{k}^{\top}& \label{eqn_vb_20}\\
	\mathbf{K}_{k} &= \mathbf{C}_{k}\mathbf{S}_{k}^{-1}& \label{eqn_vb_21}\\
	\mathbf{C}_{k}&=\big \langle(\mathbf{x}_{k} - \hat {\mathbf{x}}_{k}^{-} )(\mathbf{h}(\mathbf{x}_{k}) - \bm{\mu}_k) \rangle_{p(\mathbf{x}_k|\mathbf{y}_{k-1})}& \label{eqn_vb_22}\\
	\mathbf{S}_{k}&=\big \langle(\mathbf{h}(\mathbf{x}_{k}) - \bm{\mu}_k)(\mathbf{h}(\mathbf{x}_{k}) - \bm{\mu}_k)^{\top}  \rangle_{p(\mathbf{x}_k|\mathbf{y}_{k-1})} + \mathbf{R}_{k}& \label{eqn_vb_23}
\end{flalign}

Parameters of ${q(\bm{\mathcal{I}}_k)}$ are updated iteratively as
\begin{flalign}
	\Omega_{k}(i,i) &= \frac{{\Pr({\mathcal{I}}_{k}(i,i) = 1)}}{({{\Pr}({\mathcal{I}}_{k}(i,i) = 1) + \Pr({\mathcal{I}}_{k}(i,i) = 0)})}&\label{eqn_vb_29}
\end{flalign}
where denoting $k(i)$ as the proportionality constant and
\begin{flalign}
	&\Pr({\mathcal{I}}_{k}(i,i) = 0) = k(i) (1 - \theta_{k}(i)) \exp{\big({-}\frac {1}{2} \big(\frac {(y_k(i) - {\nu}_k(i) )^{2}}{R_{k}(i,i)} + \bar{h}^2_k\big)\big)}& \label{eqn_vb_24}\\
	&\Pr({\mathcal{I}}_{k}({i,i}) = 1) = k(i) \theta_{k}(i) \exp{ \big({-}\frac{1}{2} \frac{\bar{h}^2_k + \bar{\Theta}^2_k + ( {\nu}_k(i) + {\hat{\Theta}}_k^{+}(i) - y_k(i))^{2}}{R_{k}(i,i)}\big)}&\label{eqn_vb_25}\\
	&\bm{\nu}_k= \langle {\mathbf{h}}(\mathbf{x}_{k})  \rangle_{q(\mathbf{x}_k)}&\label{eqn_vb_26}\\
	&\bar{h}^2_k = \langle(h(\textbf{x}_{k})(i) - {\nu}_k(i) )^{2}\rangle_{q(\mathbf{x}_k)}& \label{eqn_vb_27}\\
	&\bar{\Theta}^2_k = \langle({\Theta}_{k}(i) - { {\hat{\Theta}}_k^{+}(i) } )^2\rangle_{q(\mathbf{\Theta}_k)}&\label{eqn_vb_28}
\end{flalign}

Parameters of $q(\mathbf{\Theta}_k)$ are updated iteratively as
\begin{flalign}
	\hat {\mathbf{\Theta}}_{k}^{*} &= \hat{\mathbf{\Theta}}_{k}^{-} + \bm{\mathcal{K}}_{k}(\mathbf{y}_{k} - ( \bm{\nu}_k + \boldsymbol{\Omega}_{k} \hat{\mathbf{\Theta}}_{k}^{-}))&\label{eqn_vb_30}\\
	\mathbf{{\Sigma}}^*_{k} &= \mathbf{{\Sigma}}^-_{k} - \bm{\mathcal{C}}_{k}\bm{\mathcal{K}}_{k}^{\top}&\label{eqn_vb_31}\\
	\bm{\mathcal{K}}_{k} &= \bm{\mathcal{C}}_{k}\bm{\mathcal{S}}_{k}^{-1}&\label{eqn_vb_32}\\
	\bm{\mathcal{C}}_{k}&=\mathbf{{\Sigma}}^-_{k}\boldsymbol{\Omega}_{k}^{\top}&\label{eqn_vb_33}\\
		\bm{\mathcal{S}}_{k} &= \boldsymbol{\Omega}_{k}\mathbf{{\Sigma}}^-_{k}\boldsymbol{\Omega}_{k}^{\top} + \mathbf{R}_{k}&\label{eqn_vb_34}
\end{flalign}
\begin{flalign}
	\hat{\mathbf{\Theta}}_{k}^{+}&=\mathbf{\Sigma}_{k}^{+}{\mathbf{{\Sigma}}^*_{k}}^{-1}\hat{\mathbf{\Theta}}_{k}^{*}&\label{eqn_vb_35}\\
	\mathbf{\Sigma}_{k}^{+} &= \big(\boldsymbol{\Omega}_{k}(\mathbf{I}-\boldsymbol{\Omega}_{k})\mathbf{R}_{k}^{-1} + {\mathbf{\Sigma}_{k}^*}^{-1}\big)^{-1}&\label{eqn_vb_36}
\end{flalign}

\begin{algorithm}[ht!]
	\SetAlgoLined
	Initialize\ $\hat {\mathbf{x}} _{0} ^{+},\mathbf{P}^{+}_0,\hat{\mathbf{\Theta}}_{0}^{+},
	\mathbf{\Sigma}_{0}^{+}
	,\widetilde{\mathbf{\Sigma}}_{k},\breve{\mathbf{\Sigma}}_{k}$;
	
	\For{$k=1,2...K$}{
		Initialize $\theta_k(i)$, $\Omega_k(i)$,$\hat {\mathbf{\Theta}} _{k} ^{+}$, $\mathbf{Q}_k$, $\mathbf{R}_k$\;
		\textbf{Prediction}\\
		Evaluate $\hat {\mathbf{x}} _{k}^{-},\mathbf{P}^{-}_k$ with \eqref{eqn_vb_14} and \eqref{eqn_vb_15}\;
		Evaluate $\mathbf{\hat{\Theta}}^{-}_{k},\mathbf{\Sigma}^{-}_{k}$ with \eqref{eqn_vb_16} and \eqref{eqn_vb_17}\;
		\textbf{Update}\\
		\While{not converged}{
			Update $\hat {\mathbf{x}} _{k} ^{+}$ and  ${\mathbf{P}^{+ }_k}$ with \eqref{eqn_vb_18}-\eqref{eqn_vb_23}\;
			Update $\Omega_k(i)\ \forall\ i$ with \eqref{eqn_vb_29}-\eqref{eqn_vb_28}\;
			Update $\hat {\mathbf{\Theta}} _{k} ^{+}$ and  ${\mathbf{\Sigma}^{+ }_k}$ with \eqref{eqn_vb_30}-\eqref{eqn_vb_36}\;
		}
	}
	\caption{The proposed BDM filter}
	\label{Algo1_BDM}
\end{algorithm} 
\subsection{BDM Filter}

\hspace{.5cm}
Using the proposed approximations, in the prediction and update steps, we have devised a recursive filter referred as the BDM filter. \textcolor{black}{Unless real-world experiments reveal any prior information regarding the occurrence of bias in each dimension, we propose using an uninformative prior for ${\mathcal{I}}_{k}(i,i)\ \forall\ i$ which is commonly adopted for such cases. The Bayes-Laplace and the maximum entropy methods for obtaining uninformative prior for a parameter with finite values lead to the uniform prior distribution \cite{martz199414,turkman2019computational}. We adopt this choice of prior for our case i.e. $\theta_k(i)=0.5\ \forall\ i$ which has been advocated in the literature for designing robust filters \cite{8869835,chughtai2022outlier}.} For the convergence criterion, we suggest using the ratio of the L2 norm of the difference of the state estimates from the current and previous VB iterations and the L2 norm of the estimate from the previous iteration. Algorithm \ref{Algo1_BDM} outlines the devised BDM filter.



\section{Numerical Experiments}\label{simulation_sec}\hspace{.5cm}
To evaluate the comparative performance of the devised algorithm, numerical experiments have been conducted on an Apple MacBook Air with a 3.2 GHz M1 Processor and 8 GBs of unified RAM using Matlab R2021a.

For comparative fairness, we consider methods based on the UKF as their basic algorithmic workhorse. The following filters have been taken into account for comparisons: the standard UKF, the selective observations rejecting UKF (SOR-UKF) \cite{chughtai2022outlier}, the Unscented Schmidt Kalman Filter (USKF) \cite{stauch2015unscented} and the constrained Unscented Kalman filter (CUKF). The CUKF is devised by modifying the CSRUKF \cite{filtermobile} by using the standard UKF instead of SRUKF as its core algorithm. 

In terms of handling data corruption, the standard UKF has no bias compensation means in its construction. By contrast, the SOR-UKF is an outlier-robust filter, with inherent data anomaly detection mechanism, which discards the observations found to be corrupted. Lastly, the USKF and CUKF both compensate for the bias partially and require an external bias detection mechanism. Bias mitigating filters with inherent detection mechanism are generally scant in the literature and most of these are based on the PFs. The USKF is a modified version of the SKF, adapted for nonlinear systems, where the bias is not exactly estimated rather its correlations with the state are updated. The CUKF is based on two major functional components. First, it resorts to the UKF for estimation. Subsequently, it draws sigma points based on these estimates which are projected onto a region, by solving an optimization problem, supposing a constraint that the measurements can only be positively biased. In our numerical evaluation, we assume perfect detection for these two algorithms. Note that in the implementation of the USKF and CUKF, we switch to the standard UKF when no bias is detected. Also note that the proposed BDM filter has no limitations in terms of whether any bias positively or negatively disturbs the measurements. However, since the CUKF assumes a positive bias we keep this restriction in our simulations. In particular, bias \textcolor{black}{in each dimension }is added as a shifted Gaussian $\mathcal{N}(\mu,\sigma^2)$ where $\mu \geq 0$ \cite{filtermobile}.

\textcolor{black}{For performance evaluation, we resort to a target tracking problem in a wireless network where the range measurements are typically biased \cite{weiss2008network}. The observations get biased owing to the transitions between LOS and NLOS conditions \cite{4027766}. Such tracking systems find applications in emergency services, fleet management, intelligent transportation, etc. \cite{4960267}.}

\textcolor{black}{We consider the process equation for the target assuming an unknown turning rate as \cite{8398426}}

\begin{flalign}
	\mathbf{x}_k=\mathbf{f}(\mathbf{x}_{k-1})+\mathbf{q}_{k-1}\label{eqn_res1a}
\end{flalign}
with
\begin{align}
	\mathbf{f}(\mathbf{x}_{k-1}) &= \begin{pmatrix} \text{1} & \frac{\text{sin}(\omega_{k}{\textcolor{black}{\zeta}})}{\omega_{k}} & \text{0} &  \frac{\text{cos}(\omega_{k}{\textcolor{black}{\zeta}})-\text{1}}{\omega_{k}} & \text{0} \\  \text{0} & \text{cos}(\omega_{k}{\textcolor{black}{\zeta}}) &  \text{0} & -\text{sin}(\omega_{k}{\textcolor{black}{\zeta}}) & \text{0}\\  \text{0} & \frac{\text{1}-\text{cos}(\omega_{k}{\textcolor{black}{\zeta}})}{\omega_{k}} & \text{1} &\frac{\text{sin}(\omega_{k}{\textcolor{black}{\zeta}})}{\omega_{k}} & \text{0} \\  \text{0} &  \text{\text{sin}}(\omega_{k}{\textcolor{black}{\zeta}}) &  \text{0} &  \text{cos}(\omega_{k}{\textcolor{black}{\zeta}}) & \text{0}\\  \text{0} & \text{0} & \text{0} & \text{0} &\text{1} \end{pmatrix} \mathbf{x}_{k-\text{1}}\label{eqn_res1}
\end{align}
where the state vector $\mathbf{x}_k= [a_k,\dot{{a_k}},b_k,\dot{{b_k}},\omega_{k}]^{\top}$
contains the 2D position coordinates \(({a_k} , {b_k} )\), the respective velocities \((\dot{{a_k}} , \dot{{b_k}} )\), the angular velocity $\omega_{k}$ of the target at time instant $k$, \( {\textcolor{black}{\zeta}} \) is the sampling period, and $\mathbf{q}_{k-\text{1}} \sim N\left(0,\mathbf{Q}_{k-\text{1}}.\right)$. $\mathbf{Q}_{k-\text{1}}$ is given in terms of scaling parameters $\eta(1)$ and $\eta(2)$ as \cite{8398426}
\begin{equation}
	\mathbf{Q}_{k-\text{1}}=\begin{pmatrix} \eta(1) \mathbf{M} & 0 & 0\\0 &\eta(1) \mathbf{M}&0\\0&0&\eta(2)
	\end{pmatrix}, \mathbf{M}=\begin{pmatrix} {\textcolor{black}{\zeta}}^3/3 & {\textcolor{black}{\zeta}}^2/2\\{\textcolor{black}{\zeta}}^2/2 &{\textcolor{black}{\zeta}}
	\end{pmatrix}\nonumber
\end{equation}

	Range readings are obtained from $m$ sensors installed around a rectangular area where the $i$th sensor is located at $\big(\mathfrak{a}(i)=350(i-1),\mathfrak{b}(i)=350\ ((i-1)\mod2)\big)$. The nominal measurement equation can therefore be expressed as
	\begin{equation}
		\mathbf{y}_k = \mathbf{h}(\mathbf{x}_{k})+\mathbf{r}_k\label{eqn_res2_VB}
	\end{equation}
	with
	\begin{equation}
		{h^i(\mathbf{x}_k )} = \sqrt{\big( (a_{k} - \mathfrak{a}(i))^{2} + (b_{k} - \mathfrak{b}(i))^{2} \big)}\label{eqn_res2_VBb}
	\end{equation}

	{For the duration of bias presence the following observation equations, based on the random bias model \cite{park2022robust}, are assumed} 
	\begin{align}
		\mathbf{y}_k &= \mathbf{h}(\mathbf{x}_{k})+\mathbf{r}_k+ \overbrace{\boldsymbol{{\mathcal{J}}}_{k} (\mathbf{o}_k+\triangle \mathbf{o}_k)}^{\mathbf{b}_k}\label{eqn_res2_VBc}\\
		\mathbf{y}_{k+1} &= \mathbf{h}(\mathbf{x}_{k+1})+\mathbf{r}_{k+1}+ \boldsymbol{{\mathcal{J}}}_{k} (\mathbf{o}_k+\triangle \mathbf{o}_{k+1})\label{eqn_res2_VBd}
	\end{align}
	{where $\mathbf{o}_k\in\mathbb{R}^m$ models the effect of statistically independent biases in the measurements and $\boldsymbol{\mathcal{J}}_{k} \in \mathbb{R}^{m\times m}$ is a diagonal matrix with independent Bernoulli elements with parameter $\lambda$.} \textcolor{black}{Our measurement model for evaluation is inspired by the LOS and NLOS models in \cite{4027766,4960267} where shifted Gaussian distributions are used to model the biased measurements. In effect, bias with magnitude ${o}_k(i)$ with some additional uncertainty $\triangle o_k(i) \sim \mathcal{N}({0},\sigma_{o}(i))$ affects the \textit{i}th dimension at time step $k$ if ${\mathcal{J}}_{k}(i,i)=1$. Since the exact magnitudes of biases are not generally known apriori and can occur randomly in a given range we assume $o_k(i)$ to be uniformly distributed i.e. ${o}_k(i)\sim\mathcal{U}(0,\xi(i))$. This is in contrast to the approaches in \cite{4027766,4960267} where the magnitudes of biases are assumed to be known perfectly. Similarly at time step $k+1$, the bias sustains with the same magnitude of the previous time step $k$ in the \textit{i}th dimension i.e. ${o}_k(i)$ with some uncertainty $\triangle o_{k+1}(i)$ if the \textit{i}th dimension at the previous time step $k$ was affected i.e. ${\mathcal{J}}_{k}(i,i)=1$. Note that the evaluation model allows us to compare different methods under extreme conditions varying from the case of no biased dimension to the case where every observation can possibly get biased.}

	
	For evaluation we assign the following values to different parameters: ${\sigma_{o}(i)}  = {0.4}, \xi(i)=90, \mathbf{x}_{0} = [0, 10, 0, -5, \frac{3 \pi}{180}]^{\top}, \mathbf{R}_k = 4\mathbf{I},  {\textcolor{black}{\zeta}}  = 1, \eta(1)=0.1, \eta(2)=1.75\times10^{-4}$. We assume $m=4$, since for higher-dimensional problems even rejection-based methods like the SOR-UKF can have acceptable performance for a larger probability of errors owing to the redundancy of useful information in other uncorrupted dimensions. However, this does not limit the applicability of the proposed method for large $m$. Our point is to emphasize how properly utilizing information from an affected dimension, a characteristic of {analytical} redundancy approaches, is more useful in contrast to completely rejecting the information, a characteristic of {hardware} redundancy approaches \cite{9239326}. In addition, we evaluate the relative performance of the proposed filter with similar {analytical} redundancy approaches. 
	
	\begin{figure*}[ht!]
		\centering
		\begin{subfigure}[h!]{0.4\linewidth}
			\centering
			\includegraphics[width=\linewidth,trim=0 0 20cm 0,clip=true]{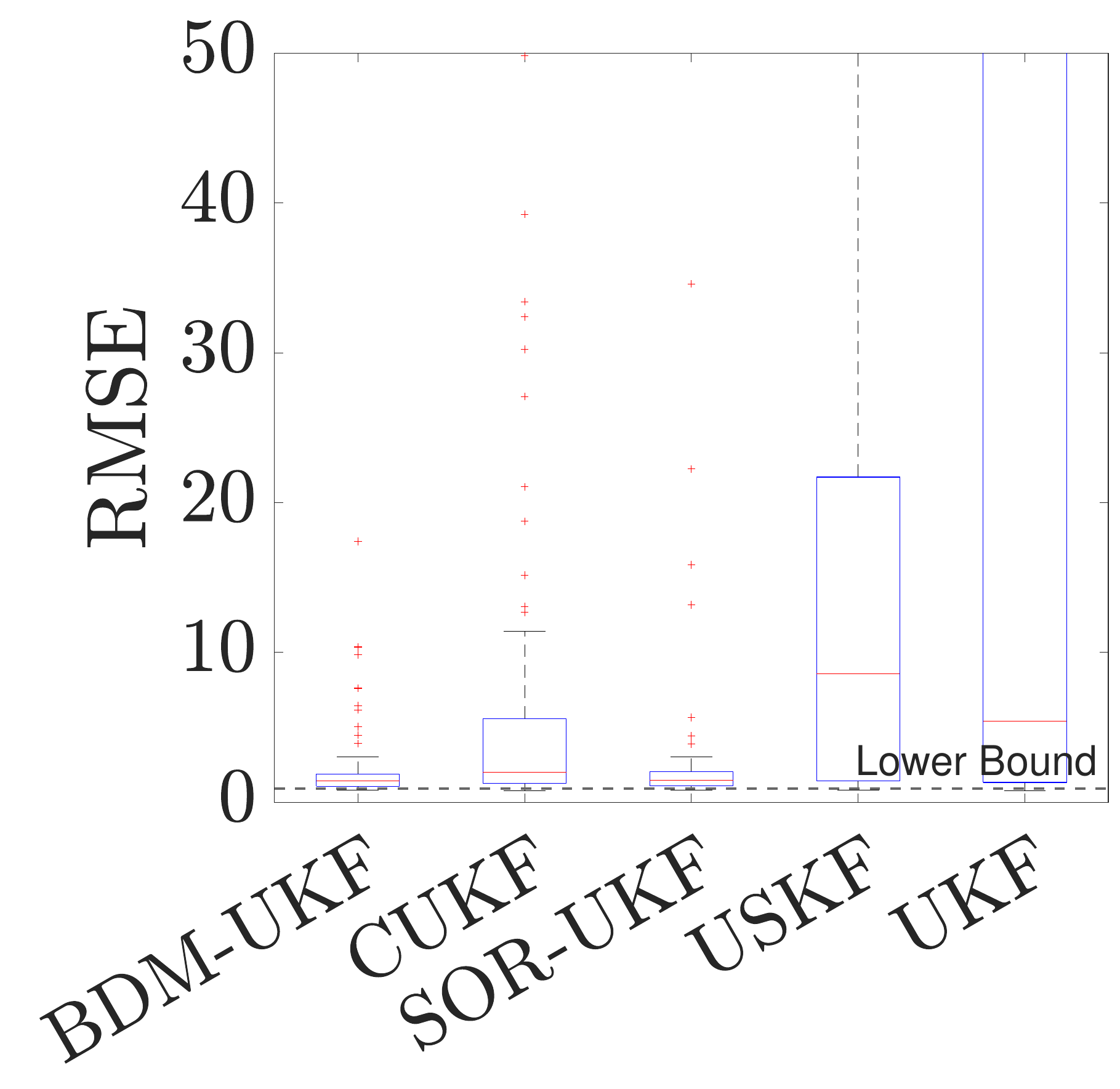}
			\caption{$\lambda = 0.2$}
			\label{Box11}
		\end{subfigure}
		\begin{subfigure}[h!]{0.4\linewidth}
			\centering
			\includegraphics[width=\linewidth,trim=0 0 20cm 0,clip=true]{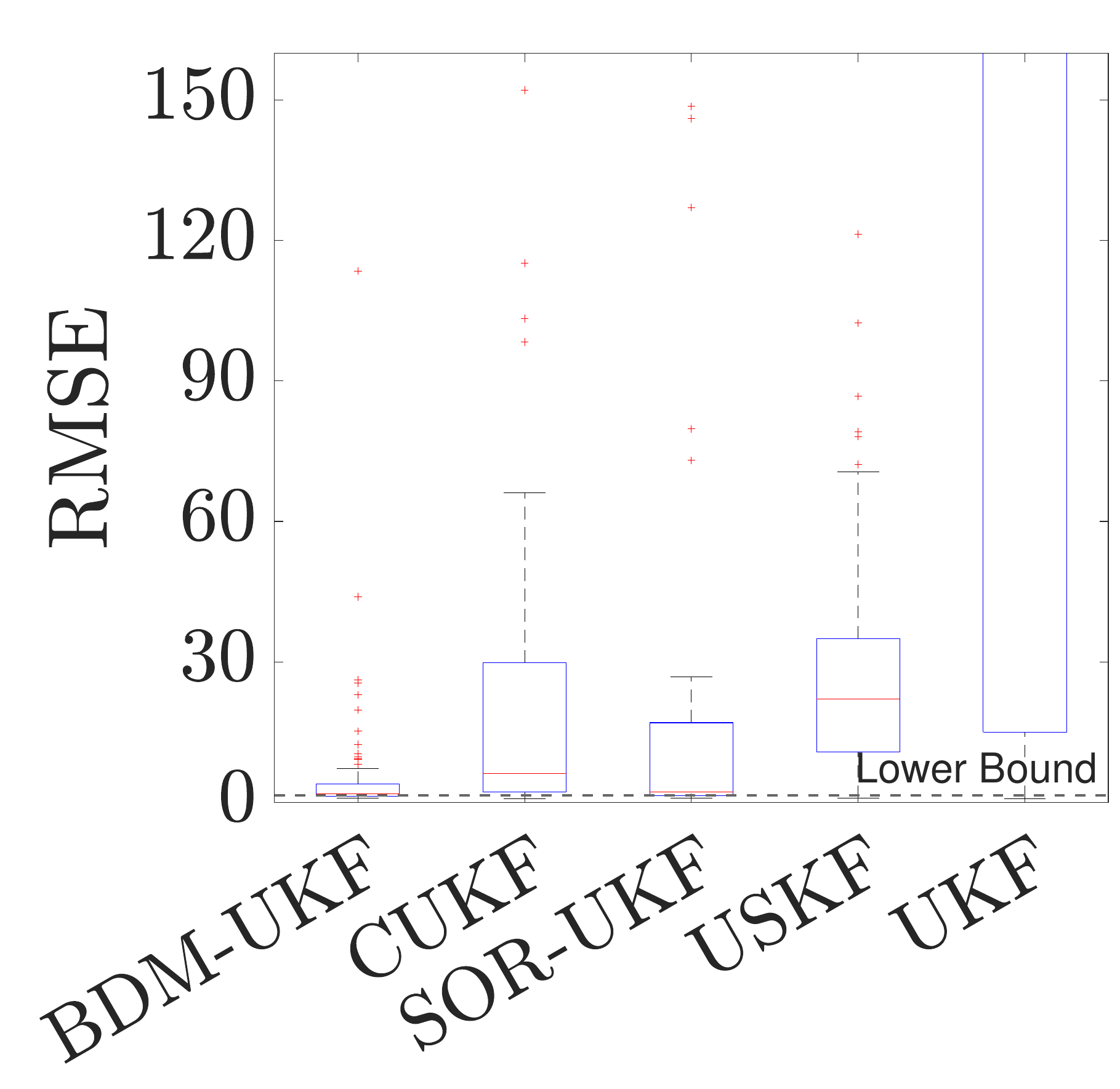}
			\caption{$\lambda = 0.4$}
			\label{Box12}
		\end{subfigure}
		\begin{subfigure}[h!]{0.4\linewidth}
			\centering
			\includegraphics[width=\linewidth,trim=0 0 20cm 0,clip=true]{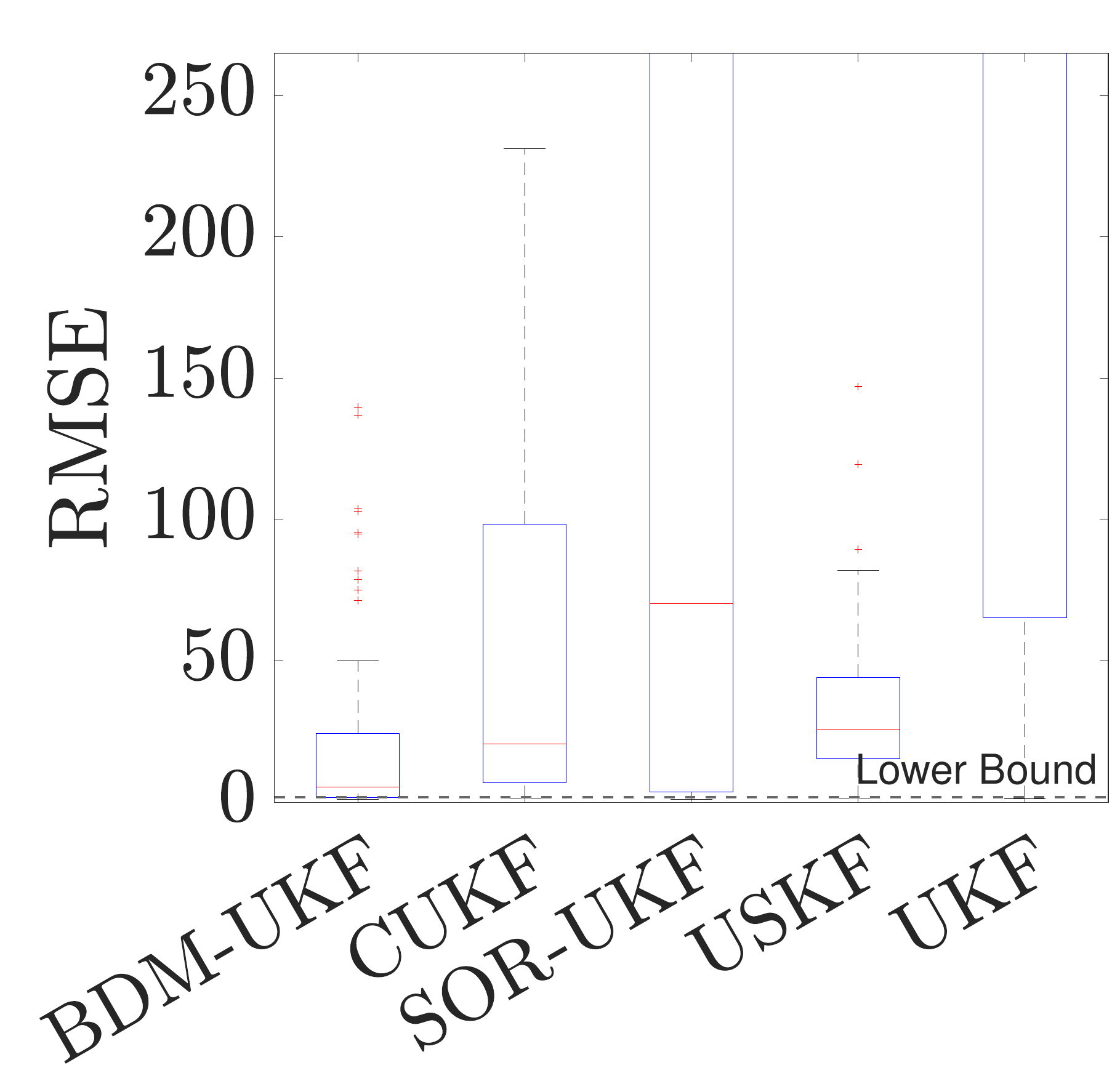}
			\caption{$\lambda = 0.6$}
			\label{Box13}
		\end{subfigure}
		\begin{subfigure}[h!]{0.4\linewidth}
			\centering
			\includegraphics[width=\linewidth,trim=0 0 20cm 0,clip=true]{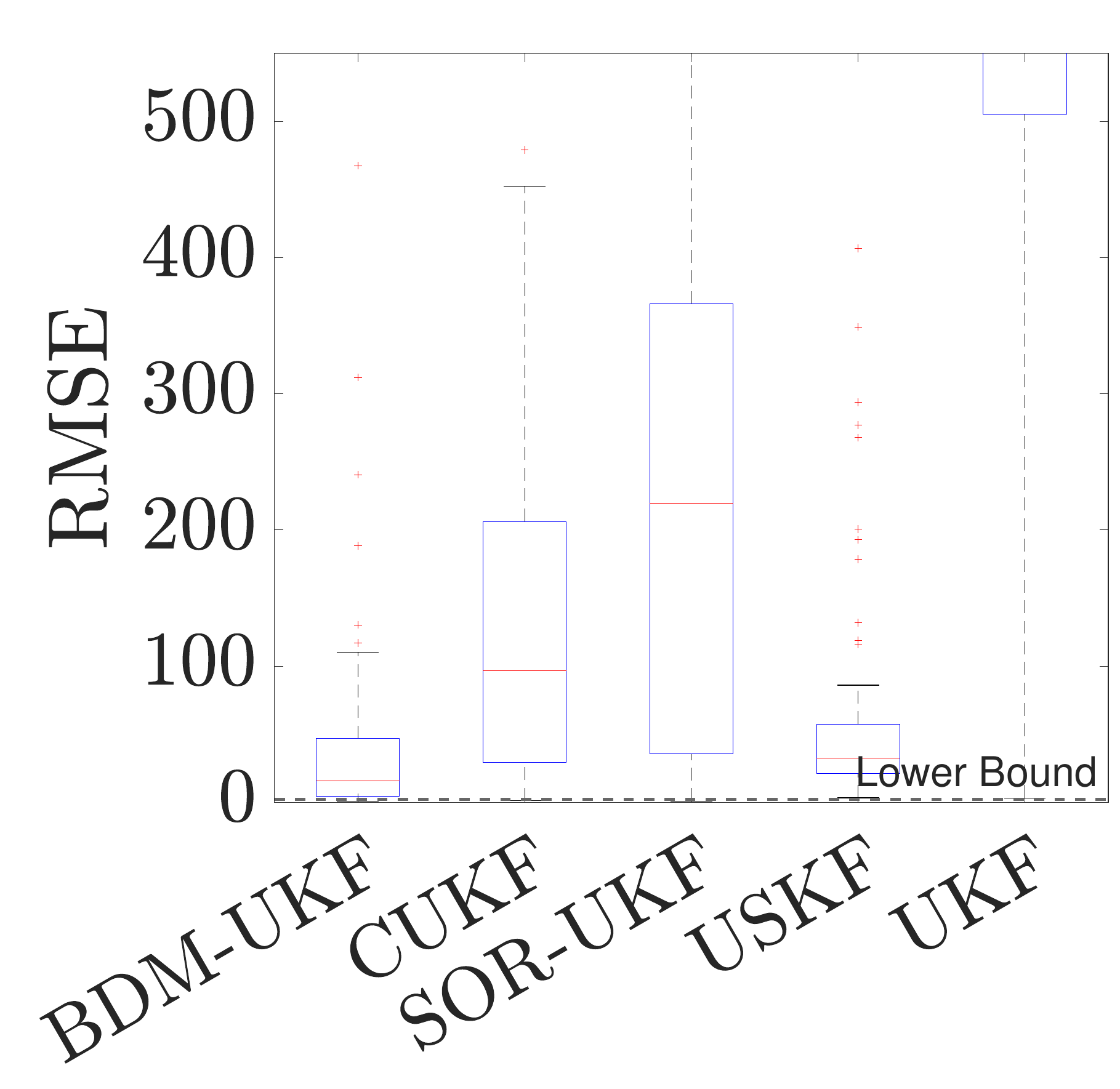}
			\caption{$\lambda = 0.8$}
			\label{Box14}
		\end{subfigure}
		\caption{Box plots of state RMSE for Case 1 with increasing $\lambda$} \label{Box1}
	\end{figure*}
	\begin{figure}[ht!]
		\centering
		\includegraphics[width=.8\linewidth]{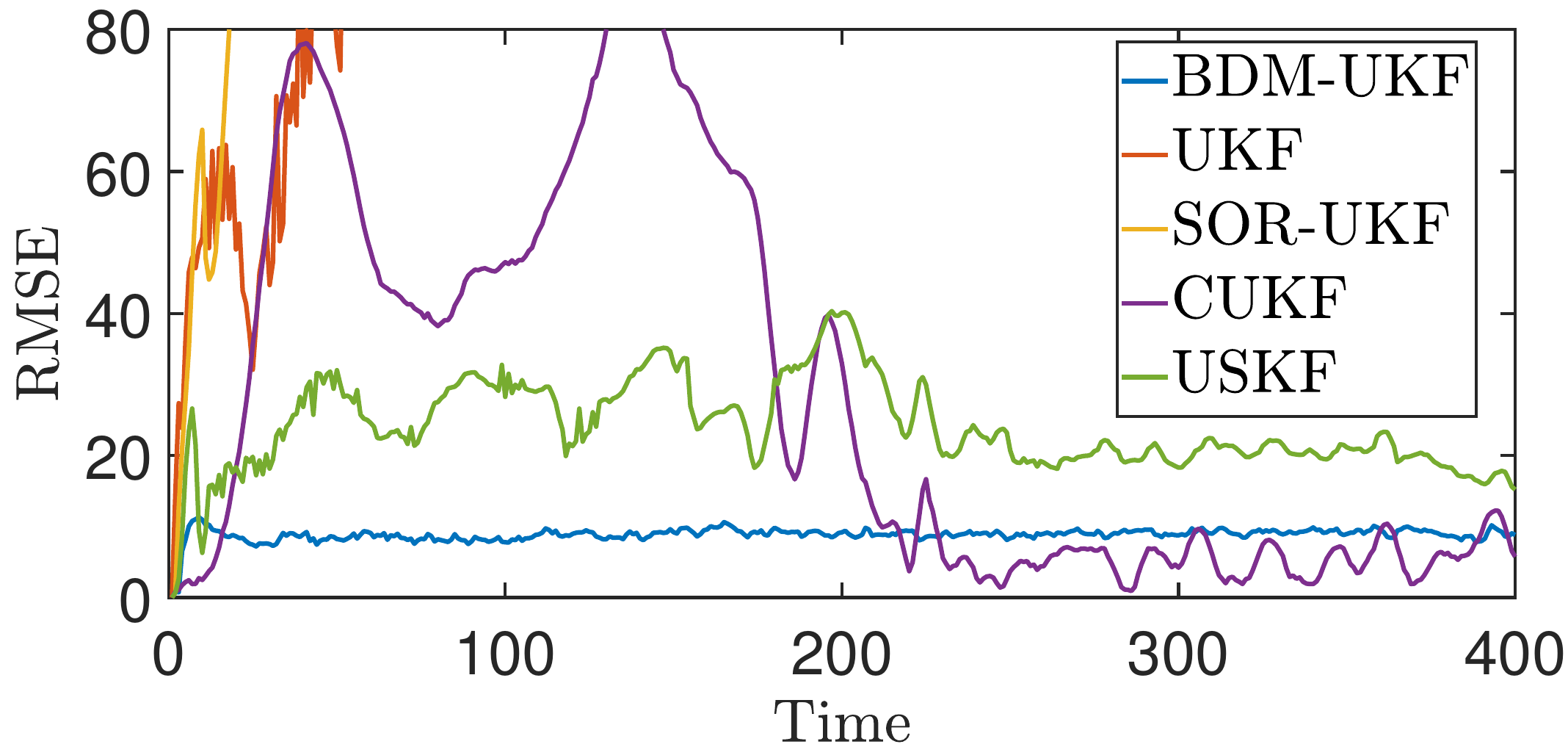}
		\caption{Tracking performance of algorithms over time for an example MC run for Case 1} 
		\label{fig:Case1_run}
	\end{figure}

	For each method, the UKF parameters are set as $\alpha=1$, $\beta=2$ and $\kappa=0$. We consider $T = 400$ time steps and carry out 100 MC simulations for each case. All the filters are initialized with a state estimate equal to $\mathbf{x}_{0}$ its error covariance as $\mathbf{Q}_{k}$ and the bias estimate is set equal to zero initially. The other parameters supposed for the BDM-UKF are: $\mathbf{\Sigma}_{0}^{+}=0.001\mathbf{I},\widetilde{\mathbf{\Sigma}}_{k}=1000\mathbf{R}_k,\breve{\mathbf{\Sigma}}_{k}=0.1\mathbf{R}_k$ and $\theta_k(i)=0.5~\forall~i$ and threshold of $10^{-4}$ for the
	convergence criterion. In CUKF the auxiliary parameters are assigned values as: $\epsilon=1,\sigma_n=2$. Lastly, the USKF is initialized with a zero mean bias with a covariance of $500\mathbf{I}$ having zero cross-covariance with the state. Other parameters for all the filters are kept the same as originally reported. 
	\subsection*{Performance Bounds}\hspace{.5cm}
	\textcolor{black}{To benchmark the relative performance of the considered algorithms, we evaluate the posterior (Bayesian) Cramer Rao bound (PCRB) for the target tracking model in \eqref{eqn_res1a}-\eqref{eqn_res2_VBc}. The bound remains valid where system dynamics can be modeled as \eqref{eqn_res1a} and where the measurement model can be considered to switch between \eqref{eqn_res2_VB} and \eqref{eqn_res2_VBc}-\eqref{eqn_res2_VBd} since we use generalized models in our derivation.} We evaluate the PCRB assuming perfect apriori knowledge of the occurrence time, duration of the bias and $\lambda$, resorting to the approach presented in \cite{668800,van2013detection} for a generalized nonlinear dynamical system, with white process and measurement noise, given as follows
	\begin{align}
		\mathbf{x}_{k+1}&=\mathbf{f}\left(\mathbf{x}_{k}, \mathbf{q}_{k}\right)\\
		\mathbf{y}_{k}&=\mathbf{h}\left(\mathbf{x}_{k}, \mathbf{r}_{k},\mathbf{y}_{k-1},\cdots,\mathbf{y}_{k-z}\right)
	\end{align} 
	where $z$ is a positive integer.		
	The PCRB matrix for the estimation error of $\mathbf{x}_k$ can be written as 
	\begin{equation}
		\text{PCRB}_k\triangleq{\mathbf{J}_k}^{-1}\label{PCRB_1}
	\end{equation}	
	where $\mathbf{J}_k$ can be expressed recursively as 
	\begin{align}
		\mathbf{J}_{k+1}&=\mathbf{D}_{k}^{22}-\mathbf{D}_{k}^{21}\left(\mathbf{J}_{k}+\mathbf{D}_{k}^{11}\right)^{-1} \mathbf{D}_{k}^{12}\label{PCRB_2}
	\end{align}
	with
	\begin{align}
		\mathbf{D}_{k}^{11}&=\langle-\Delta_{\mathbf{x}_{k}}^{\mathbf{x}_{k}} \log p\left(\mathbf{x}_{k+1} \mid \mathbf{x}_{k}\right)\rangle_{p(\mathbf{x}_{k+1},\mathbf{x}_{k})}\label{PCRB_3}\\
		\mathbf{D}_{k}^{12}&=\langle-\Delta_{\mathbf{x}_{k}}^{\mathbf{x}_{k+1}} \log p\left(\mathbf{x}_{k+1} \mid \mathbf{x}_{k}\right)\rangle_{p(\mathbf{x}_{k+1},\mathbf{x}_{k})}\label{PCRB_4}\\
		\mathbf{D}_{k}^{21}&=\langle-\Delta_{\mathbf{x}_{k+1}}^{\mathbf{x}_{k}} \log p\left(\mathbf{x}_{k+1} \mid \mathbf{x}_{k}\right)\rangle_{p(\mathbf{x}_{k+1},\mathbf{x}_{k})}=\left[\mathbf{D}_{k}^{12}\right]^{\top}\label{PCRB_5}\\
		\mathbf{D}_{k}^{22}&= \mathbf{D}_{k}^{22}(1)+\mathbf{D}_{k}^{22}(2)\\
		\mathbf{D}_{k}^{22}(1)&=\langle-\Delta_{\mathbf{x}_{k+1}}^{\mathbf{x}_{k+1}} \log p\left(\mathbf{x}_{k+1} \mid \mathbf{x}_{k}\right)\rangle_{p(\mathbf{x}_{k+1},\mathbf{x}_{k})} \nonumber\\
		\mathbf{D}_{k}^{22}(2)&=\langle-\Delta_{\mathbf{x}_{k+1}}^{\mathbf{x}_{k+1}} \log p\left(\mathbf{y}_{k+1} \mid \mathbf{x}_{k+1},\mathbf{y}_{k},\cdots,\mathbf{y}_{k-z+1}\right)\rangle_{p(\mathbf{y}_{k+1},\mathbf{x}_{k+1},\cdots)} \label{PCRB_6}
	\end{align}
	where
	\begin{align}
		\Delta_{\Psi}^{\Theta}&=\nabla_{\Psi} \nabla_{\Theta}^{\top}	\label{PCRB_8}\\
		\nabla_{\Theta}&=\left[\frac{\partial}{\partial \Theta_{1}}, \cdots, \frac{\partial}{\partial \Theta_{r}}\right]^{\top}\label{PCRB_9}
	\end{align}

	The bound is valid given the existence of derivatives and expectations terms in \eqref{PCRB_1}-\eqref{PCRB_9} for an asymptotically unbiased estimator \cite{668800}. Using results from \cite{668800,van2013detection} we obtain 
	\begin{align}
		\mathbf{D}_{k}^{11}&=\left[\nabla_{\mathbf{x}_{k}} \mathbf{f}^{\top}\left(\mathbf{x}_{k}\right)\right] \mathbf{Q}_{k}^{-1}\left[\nabla_{\mathbf{x}_{k}} \mathbf{f}^{\top}\left(\mathbf{x}_{k}\right)\right]^{\top}\label{PCRB_10}\\
		\mathbf{D}_{k}^{12}&=-\nabla_{\mathbf{x}_{k}} \mathbf{f}^{\top}\left(\mathbf{x}_{k}\right) \mathbf{Q}_{k}^{-1}\label{PCRB_11}\\
		\mathbf{D}_{k}^{22}(1)&=\mathbf{Q}_{k}^{-1}
	\end{align}
	
	For the period where the nominal equation \eqref{eqn_res2_VB} remains applicable, $\mathbf{D}_{k}^{22}(2)$ can be expressed as \cite{van2013detection}
	\begin{align}
		\mathbf{D}_{k}^{22}(2)=&-\mathbf{Q}_{k}^{-1} \langle \tilde{\mathbf{F}}_{k} \rangle_{p(\mathbf{x}_{k})}  \left[\mathbf{J}_{k}+ \langle \tilde{\mathbf{F}}_{k}^{\top} \mathbf{Q}_{k}^{-1} \tilde{\mathbf{F}}_{k} \rangle_{p(\mathbf{x}_{k})} \right]^{-1} \langle \tilde{\mathbf{F}}_{k}^{\top}\rangle_{p(\mathbf{x}_{k})}\mathbf{Q}_{k}^{-1} \nonumber \\ &+\langle \tilde{\mathbf{H}}_{k+1}^{\top} \mathbf{R}_{k+1}^{-1} \tilde{\mathbf{H}}_{k+1}\rangle_{p(\mathbf{x}_{k+1})}
	\end{align}
	where $\tilde{\mathbf{F}}_{k}$ and $\tilde{\mathbf{H}}_{k}$ are the Jacobians of $\mathbf{f}(\mathbf{x}_{k})$ and $\mathbf{h}(\mathbf{x}_{k})$ respectively. 
	
	For the duration of bias occurrence, $\mathbf{D}_{k}^{22}(2)$ can be evaluated using \eqref{PCRB_6}, equivalently written as follows using a technique similar to \cite{6916255}
	\begin{equation}
		\mathbf{D}_{k}^{22}(2)=\big\langle \frac{[\nabla_{\mathbf{x}_{k+1}} p\left(\mathbf{y}_{k+1} \mid \mathbf{x}_{k+1},\mathbf{y}_{k},\mathbf{x}_{k}\right)][.]^\top}{[p\left(\mathbf{y}_{k+1} \mid \mathbf{x}_{k+1},\mathbf{y}_{k},\mathbf{x}_{k}\right)]^2}	\big\rangle_{p(\mathbf{y}_{k+1},\mathbf{x}_{k+1},\cdots)}
	\end{equation}
	
	First consider \eqref{eqn_res2_VBc} to evaluate $\mathbf{D}_{k}^{22}(2)$ at the instance of bias occurrence for which \\ $p\left(\mathbf{y}_{k+1} \mid \mathbf{x}_{k+1},\mathbf{y}_{k},\mathbf{x}_{k}\right) $ can be approximated using Monte Carlo method as
	\begin{align}
		&p\left(\mathbf{y}_{k+1} \mid \mathbf{x}_{k+1},\mathbf{y}_{k},\mathbf{x}_{k}\right)=	p\left(\mathbf{y}_{k+1} \mid \mathbf{x}_{k+1}\right)=\int p\left(\mathbf{y}_{k+1} \mid \mathbf{x}_{k+1},\mathbf{b}_{k+1}\right)p(\mathbf{b}_{k+1})d\mathbf{b}_{k+1}\\
		&=\int \mathcal{N}\left(\mathbf{y}_{k+1}|\mathbf{h}(\mathbf{x}_{k+1})+\mathbf{b}_{k+1},\mathbf{R}_k\right) p(\mathbf{b}_{k+1})d\mathbf{b}_{k+1}\\
		&\approx \frac{1}{N_{mc1}}\sum_i\mathcal{N}\left(\mathbf{y}_{k+1}|\mathbf{h}(\mathbf{x}_{k+1})+\mathbf{b}^{(i)}_{k+1},\mathbf{R}_k\right) 
	\end{align}
	where $\mathbf{b}^{(i)}_{k+1}$, $i=1,\cdots,N_{mc1}$,  are i.i.d. samples such that $\mathbf{b}^{(i)}_{k+1}\sim p(\mathbf{b}_{k+1})$.
	
	Accordingly, $\mathbf{D}_{k}^{22}(2)$ can be approximated as
	\begin{equation}
		\mathbf{D}_{k}^{22}(2)=\big\langle \frac{[\nabla_{\mathbf{x}_{k+1}} \sum_i \exp(\phi^{(i)})][.]^\top}{[\sum_i \exp(\phi^{(i)})]^2}	\big\rangle_{p(\mathbf{y}_{k+1},\mathbf{x}_{k+1})}
	\end{equation}
	where
	\begin{equation*}
	\phi^{(i)}=-0.5(\mathbf{y}_{k+1}-(\mathbf{h}(\mathbf{x}_{k+1})+\mathbf{b}^{(i)}_{k+1}))^{\top}\mathbf{R}_{k+1}^{-1}(\mathbf{y}_{k+1}-(\mathbf{h}(\mathbf{x}_{k+1})+\mathbf{b}^{(i)}_{k+1}))
	\end{equation*}

	 We can further write
	\begin{equation}
		\mathbf{D}_{k}^{22}(2)=\big\langle \frac{\sum_i [\exp(\phi^{(i)})\nabla_{\mathbf{x}_{k+1}}\phi^{(i)}  ][.]^\top}{[\sum_i \exp(\phi^{(i)})]^2}	\big\rangle_{p(\mathbf{y}_{k+1},\mathbf{x}_{k+1})}
	\end{equation}
	where $\nabla_{\mathbf{x}_{k+1}}\phi^{(i)}=\tilde{\mathbf{H}}_{k+1}^{\top}\mathbf{R}_{k+1}^{-1}(\mathbf{y}_{k+1}-\mathbf{h}(\mathbf{x}_{k+1})-\mathbf{b}^{(i)}_{k+1})$
	
	Resultingly, $\mathbf{D}_{k}^{22}(2)$ is approximated as
	\begin{equation}
		\mathbf{D}_{k}^{22}(2)\approx\frac{1}{N_{mc2}}\sum_j \frac{\sum_i [\exp(\phi^{(i,j)})\nabla_{\mathbf{x}_{k+1}}\phi^{(i,j)} ][.]^\top}{[\sum_i \exp(\phi^{(i,j)})]^2}
	\end{equation}
	with
	\begin{equation}
		\phi^{(i,j)}=-0.5(\mathbf{y}^{(j)}_{k+1}-(\mathbf{h}(\mathbf{x}^{(j)}_{k+1})+\mathbf{b}^{(i)}_{k+1}))^{\top}\mathbf{R}_{k+1}^{-1}(\mathbf{y}^{(j)}_{k+1}-(\mathbf{h}(\mathbf{x}^{(j)}_{k+1})+\mathbf{b}^{(i)}_{k+1}))\nonumber
	\end{equation}
	 and 
	 \begin{equation*}
	 	\nabla_{\mathbf{x}_{k+1}}\phi^{(i,j)} = \tilde{\mathbf{H}}_{k+1}^{\top}\mathbf{R}_{k+1}^{-1}(\mathbf{y}^{(j)}_{k+1}-\mathbf{h}(\mathbf{x}^{(j)}_{k+1})-\mathbf{b}^{(i)}_{k+1})
	 \end{equation*}
where $\mathbf{y}^{(j)}_{k+1},\mathbf{x}^{(j)}_{k+1}$, $j=1,\cdots,N_{mc2}$, are independent and identically distributed (i.i.d.) samples such that $(\mathbf{y}^{(j)}_{k+1},\mathbf{x}^{(j)}_{k+1})\sim p(\mathbf{y}_{k+1},\mathbf{x}_{k+1})$. 

Lastly, to evaluate $\mathbf{D}_{k}^{22}(2)$ for the bias persistence period we consider \eqref{eqn_res2_VBd} and the difference of \eqref{eqn_res2_VBc}-\eqref{eqn_res2_VBd}.
	Therefore, $p\left(\mathbf{y}_{k+1} \mid \mathbf{x}_{k+1},\mathbf{y}_{k},\mathbf{x}_{k}\right) $ can be approximated for this case using Monte Carlo method as
	\begin{align}
		&p\left(\mathbf{y}_{k+1} \mid \mathbf{x}_{k+1},\mathbf{y}_{k},\mathbf{x}_{k}\right)\\ 
		&=\int p\left(\mathbf{y}_{k+1} \mid \mathbf{x}_{k+1},\mathbf{y}_{k},\mathbf{x}_{k},\boldsymbol{{\mathcal{J}}}_{k}\right)p(\boldsymbol{{\mathcal{J}}}_{k})d\boldsymbol{{\mathcal{J}}}_{k}\\
		&\approx \frac{1}{N_{mc3}}\sum_i\mathcal{N}(\mathbf{y}_{k+1}|\mathbf{h}(\mathbf{x}_{k+1})+\boldsymbol{{\mathcal{J}}}^{(i)}_{k}(\mathbf{y}_k-\mathbf{h}(\mathbf{x}_{k})) , \mathbf{R}_{k+1}+\boldsymbol{{\mathcal{J}}}^{(i)}_{k}(\mathbf{R}_{k}+2\Sigma_\mathbf{o})) 
	\end{align}
	where $\boldsymbol{{\mathcal{J}}}^{(i)}_{k}$, $i=1,\cdots,N_{mc3}$,  are i.i.d. samples such that $\boldsymbol{{\mathcal{J}}}^{(i)}_{k}\sim p(\boldsymbol{{\mathcal{J}}}_{k})$. Resultingly, $\mathbf{D}_{k}^{22}(2)$ is approximated as
	\begin{equation}
		\mathbf{D}_{k}^{22}(2)=\big\langle \frac{[\nabla_{\mathbf{x}_{k+1}} \sum_i \exp(\theta^{(i)})][.]^\top}{[\sum_i \exp(\theta^{(i)})]^2}	\big\rangle_{p(\mathbf{y}_{k+1},\mathbf{x}_{k+1},\cdots)}
	\end{equation}
	where
\begin{align*}
\theta^{(i)}=&-0.5(\mathbf{y}_{k+1}-(\mathbf{h}(\mathbf{x}_{k+1})+\boldsymbol{{\mathcal{J}}}^{(i)}_{k}(\mathbf{y}_k-\mathbf{h}(\mathbf{x}_{k})))^{\top}{(\mathbf{R}_{k+1}+\boldsymbol{{\mathcal{J}}}^{(i)}_{k}(\mathbf{R}_{k}+2\Sigma_\mathbf{o}))}^{-1}\nonumber\\ &(\mathbf{y}_{k+1}-(\mathbf{h}(\mathbf{x}_{k+1})+\boldsymbol{{\mathcal{J}}}^{(i)}_{k}(\mathbf{y}_k-\mathbf{h}(\mathbf{x}_{k})))	
\end{align*}	

Furthermore
	\begin{equation}
		\mathbf{D}_{k}^{22}(2)=\big\langle \frac{\sum_i [\exp(\theta^{(i)})\nabla_{\mathbf{x}_{k+1}}\theta^{(i)}  ][.]^\top}{[\sum_i \exp(\theta^{(i)})]^2}	\big\rangle_{p(\mathbf{y}_{k+1},\mathbf{x}_{k+1},\cdots)}
	\end{equation}
	where
	\begin{equation*}
		\nabla_{\mathbf{x}_{k+1}}\theta^{(i)}=\tilde{\mathbf{H}}_{k+1}^{\top}{(\mathbf{R}_{k+1}+\boldsymbol{{\mathcal{J}}}^{(i)}_{k}(\mathbf{R}_{k}+2\Sigma_\mathbf{o}))}^{-1}(\mathbf{y}_{k+1}-(\mathbf{h}(\mathbf{x}_{k+1})+\boldsymbol{{\mathcal{J}}}^{(i)}_{k}(\mathbf{y}_k -\mathbf{h}(\mathbf{x}_{k}) )))
	\end{equation*}

Resultingly, $\mathbf{D}_{k}^{22}(2)$ is approximated as
	\begin{equation}
		\mathbf{D}_{k}^{22}(2)\approx\frac{1}{N_{mc4}}\sum_j \frac{\sum_i [\exp(\theta^{(i,j)})\nabla_{\mathbf{x}_{k+1}}\theta^{(i,j)} ][.]^\top}{[\sum_i \exp(\theta^{(i,j)})]^2}
	\end{equation}
	with 
\begin{align*}
	\theta^{(i,j)}=&-0.5(\mathbf{y}^{(j)}_{k+1}-(\mathbf{h}(\mathbf{x}^{(j)}_{k+1})+\boldsymbol{{\mathcal{J}}}^{(i)}_{k}(\mathbf{y}^{(j)}_k-\mathbf{h}(\mathbf{x}^{(j)}_{k})))^{\top}{(\mathbf{R}_{k+1}+\boldsymbol{{\mathcal{J}}}^{(i)}_{k}(\mathbf{R}_{k}+2\Sigma_\mathbf{o}))}^{-1}\nonumber\\&(\mathbf{y}^{(j)}_{k+1}-(\mathbf{h}(\mathbf{x}^{(j)}_{k+1})+\boldsymbol{{\mathcal{J}}}^{(i)}_{k}(\mathbf{y}^{(j)}_k-\mathbf{h}(\mathbf{x}^{(j)}_{k})))
\end{align*}
%
    and
    \begin{align*}
    	\nabla_{\mathbf{x}_{k+1}}\theta^{(i,j)}=\tilde{\mathbf{H}}_{k+1}^{\top}{(\mathbf{R}_{k+1}+\boldsymbol{{\mathcal{J}}}^{(i)}_{k}(\mathbf{R}_{k}+2\Sigma_\mathbf{o}))}^{-1}(\mathbf{y}^{(j)}_{k+1}-(\mathbf{h}(\mathbf{x}^{(j)}_{k+1})+\boldsymbol{{\mathcal{J}}}^{(i)}_{k}(\mathbf{y}^{(j)}_k -\mathbf{h}(\mathbf{x}^{(j)}_{k}) )))
    \end{align*}
	$\mathbf{y}^{(j)}_{k+1},\mathbf{x}^{(j)}_{k+1},\mathbf{y}^{(j)}_{k},\mathbf{x}^{(j)}_{k}$, $j=1,\cdots,N_{mc4}$, are i.i.d. samples such that $(\mathbf{y}^{(j)}_{k+1},\mathbf{x}^{(j)}_{k+1},\mathbf{y}^{(j)}_{k},\mathbf{x}^{(j)}_{k})\sim p(\mathbf{y}_{k+1},\mathbf{x}_{k+1},\mathbf{y}_{k},\mathbf{x}_{k})$.

	\begin{figure*}[t!]
		\centering
		\begin{subfigure}[t!]{0.4\linewidth}
			\centering
			\includegraphics[width=\linewidth,trim=0 0 20cm 0,clip=true]{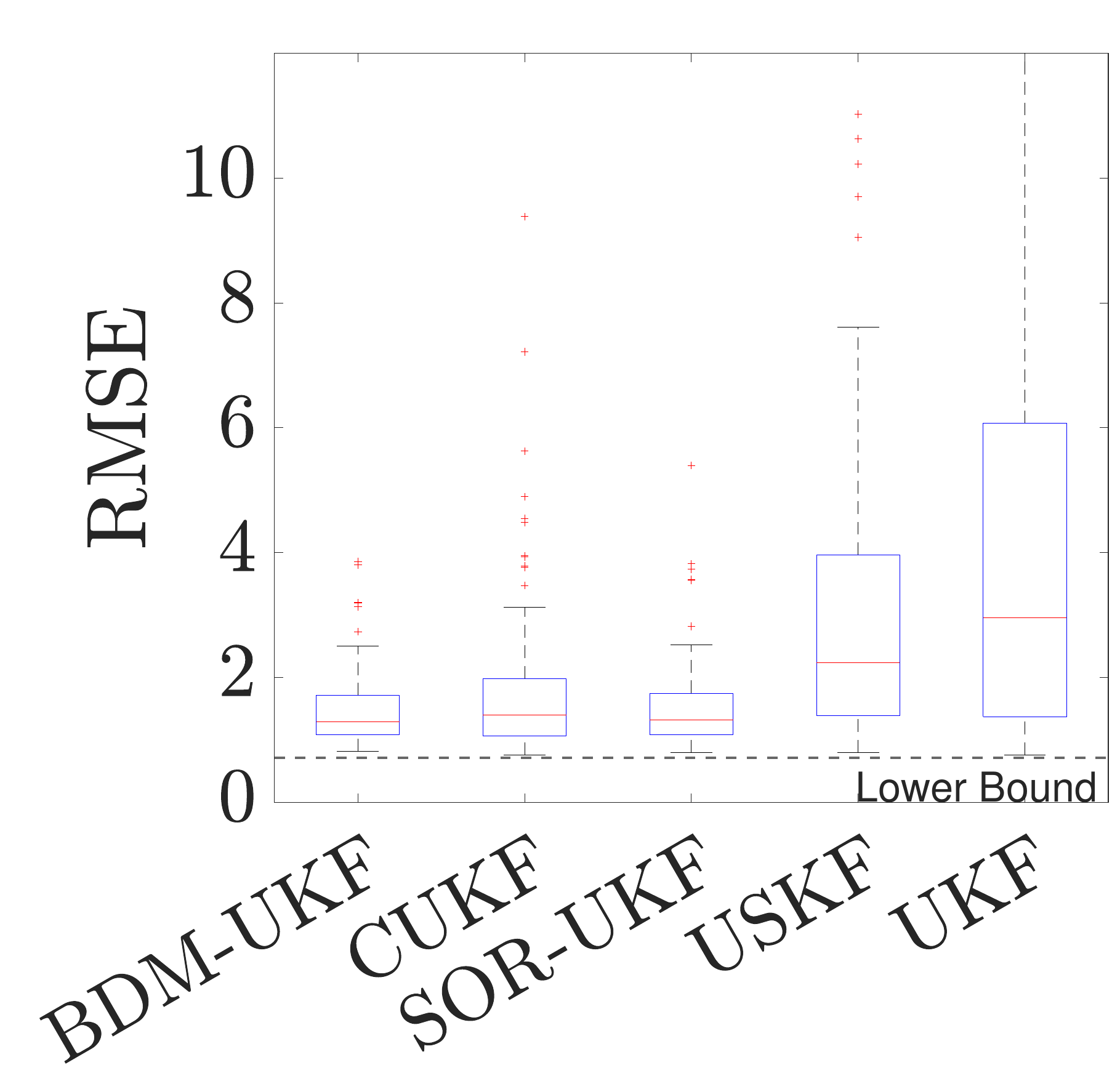}
			\caption{$\lambda = 0.2$}
			\label{Box21}
		\end{subfigure}
		\begin{subfigure}[t!]{0.4\linewidth}
			\centering
			\includegraphics[width=\linewidth,trim=0 0 20cm 0,clip=true]{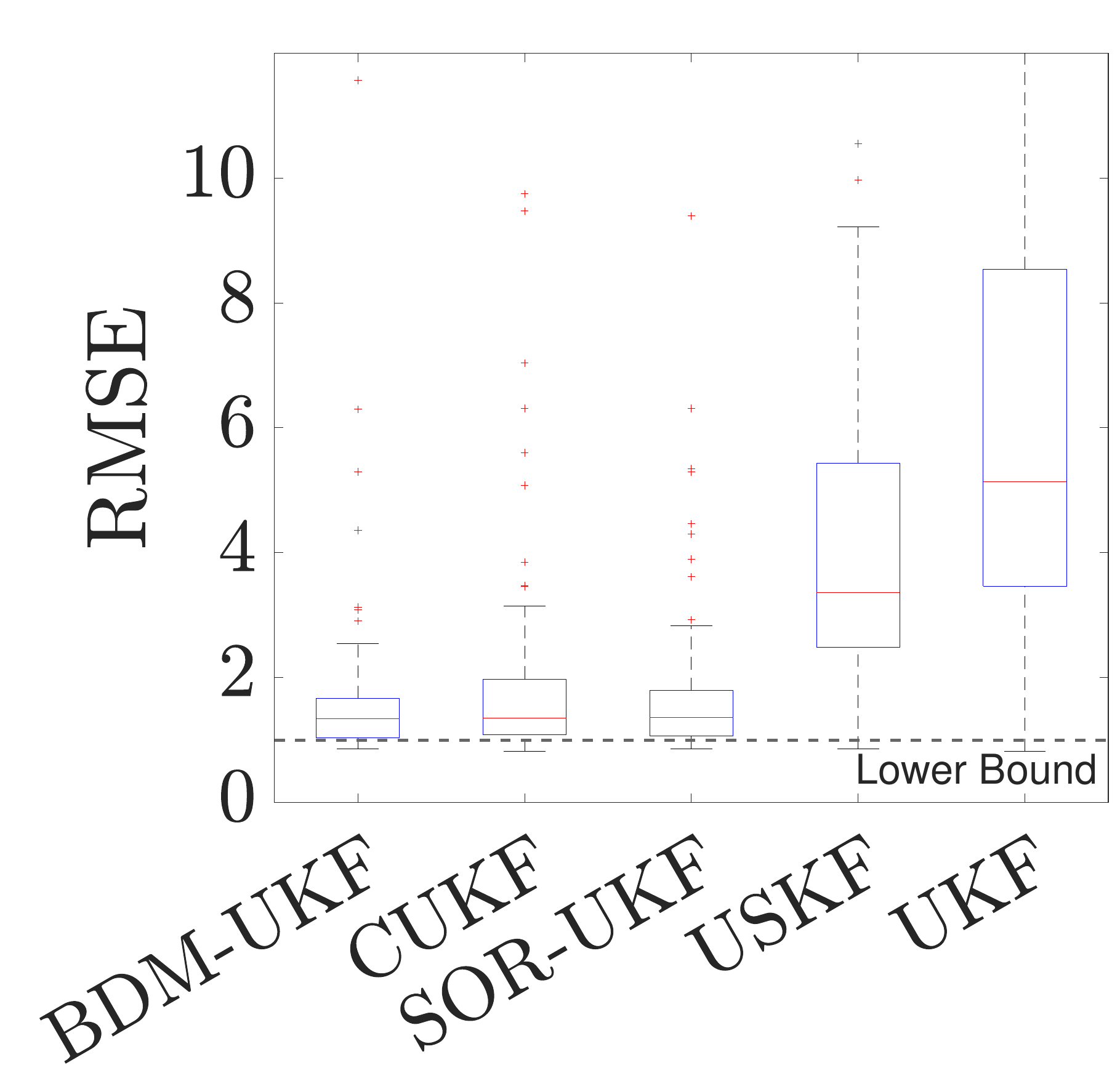}
			\caption{$\lambda = 0.4$}
			\label{Box22}
		\end{subfigure}
		\begin{subfigure}[t!]{0.4\linewidth}
			\centering
			\includegraphics[width=\linewidth,trim=0 0 20cm 0,clip=true]{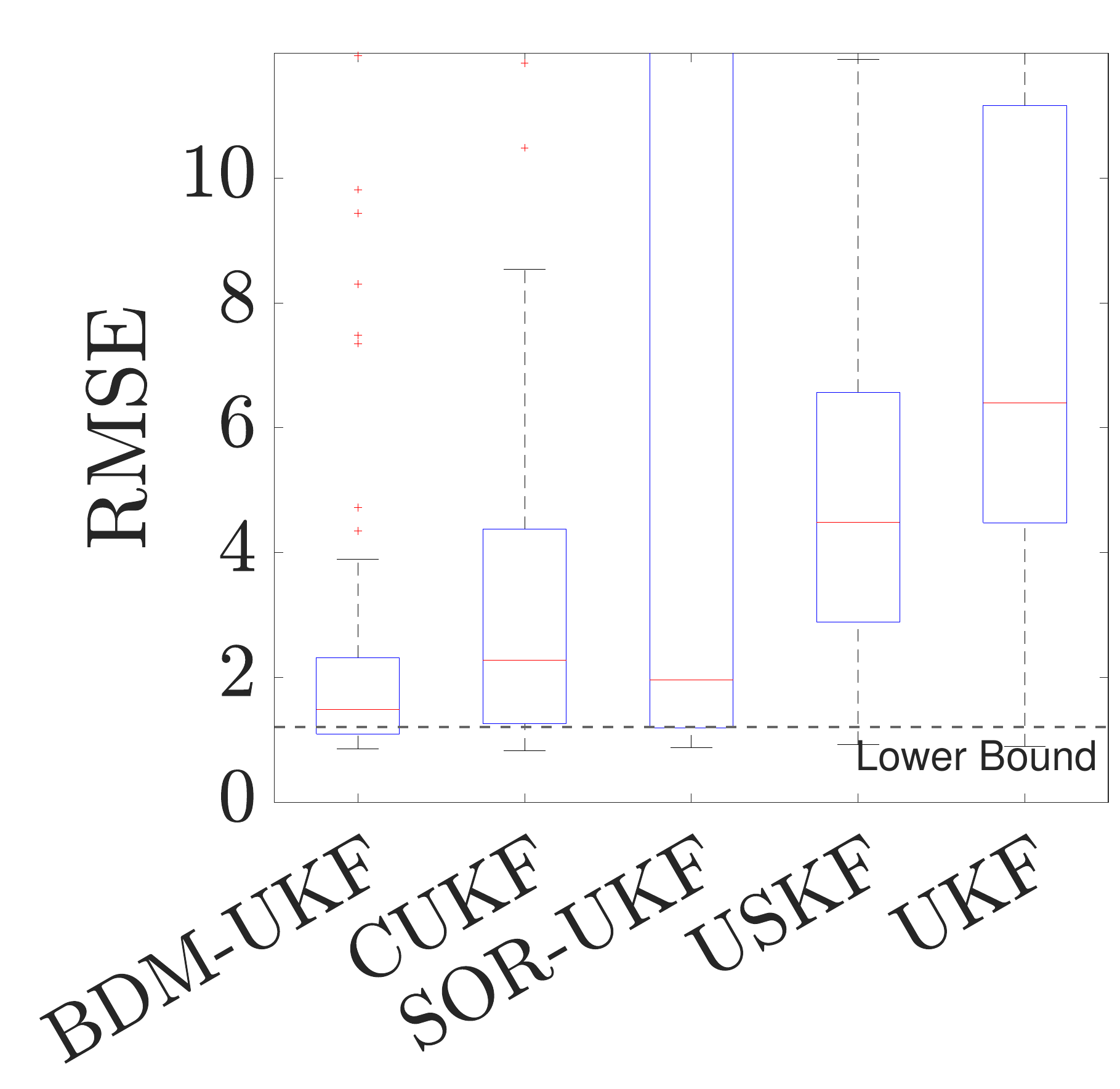}
			\caption{$\lambda = 0.6$}
			\label{Box23}
		\end{subfigure}
		\begin{subfigure}[t!]{0.4\linewidth}
			\centering
			\includegraphics[width=\linewidth,trim=0 0 20cm 0,clip=true]{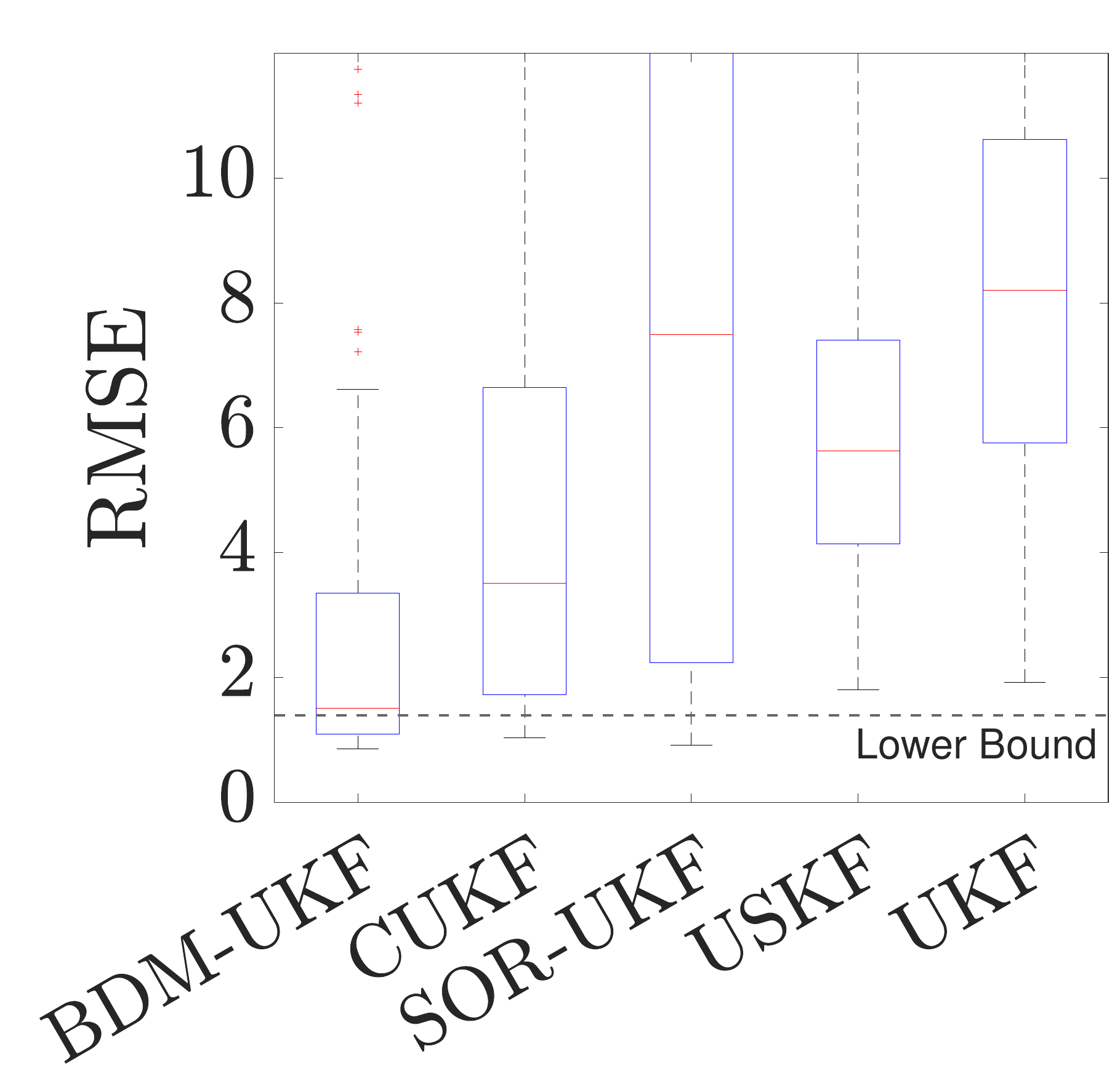}
			\caption{$\lambda = 0.8$}
			\label{Box24}
		\end{subfigure}
		\caption{Box plots of state RMSE for Case 2 with increasing $\lambda$}\label{Box2}
	\end{figure*}
	
	\begin{figure}[ht!]
		\centering
		\vspace{.07cm}
		\includegraphics[width=.8\linewidth]{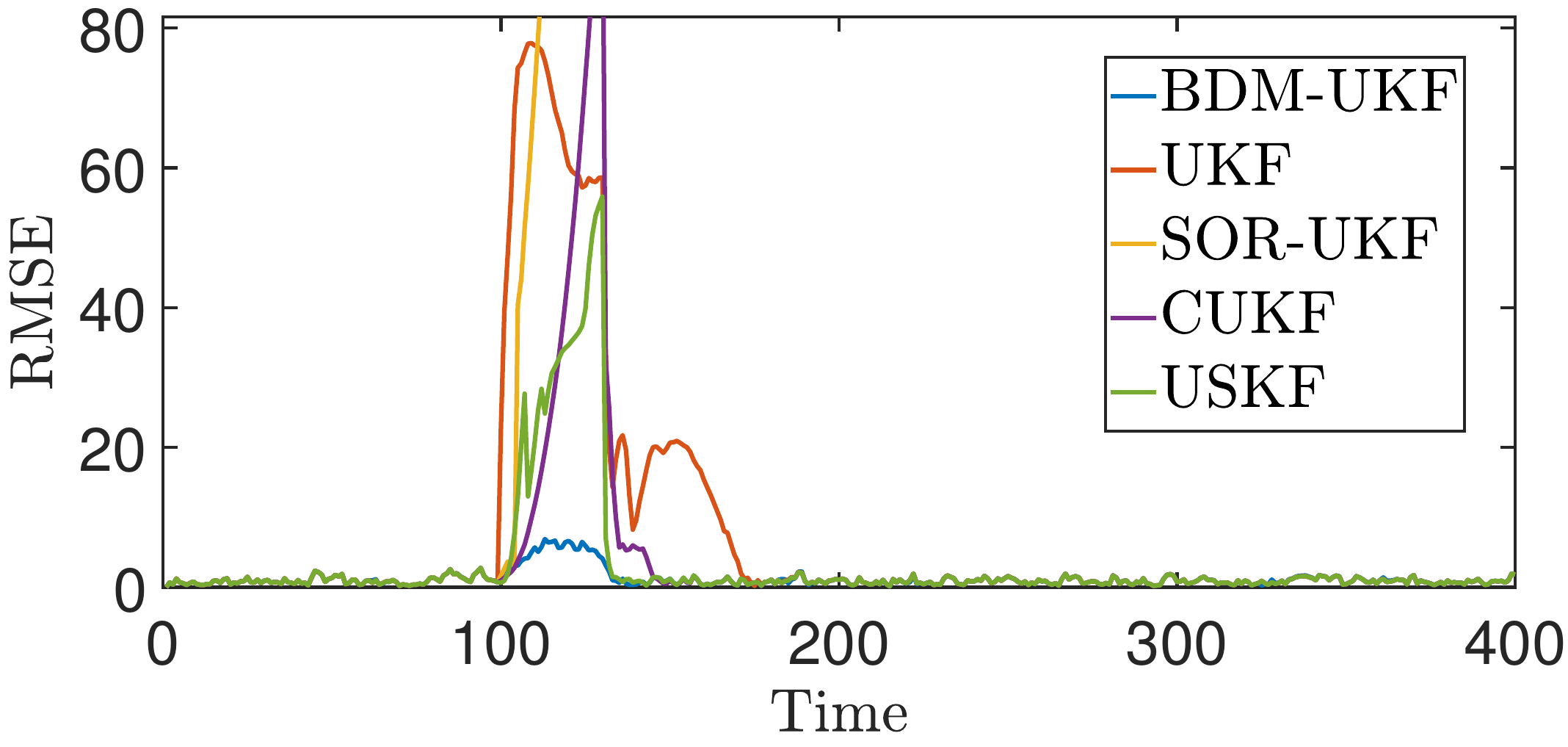}
		\caption{Tracking performance of algorithms over time for an example MC run for Case 2}
		\label{fig:Case2_run}
	\end{figure}

	\subsection{Case 1: Consistent Bias Presence}
	\begin{table}[b!]
		\centering
		\begin{tabular}{|c c c c c c|} 
			\hline
			$\boldsymbol{\lambda}$   &\textbf{UKF}&\textbf{USKF}& \textbf{BDM-UKF} & \textbf{SOR-UKF} & \textbf{CUKF}  \\ 
			\hline\hline
			0.2  &0.0469&0.0781& 0.0866 & 0.0821& 0.5762  \\ 
			0.4  &0.0470&0.0849& 0.1000 & 0.1199& 1.2499 \\
			0.6 &0.0471&0.0883& 0.1121 & 0.1416& 1.8012  \\
			0.8 &0.0470&0.0903& 0.1492 & 0.1637& 2.2305  \\ 
			\hline
		\end{tabular}
		\caption{Average Time for 100 MC Runs with Constant Bias at Different Values of $\lambda$ .}
		\label{table:1}
	\end{table}
	\hspace{.5cm}
	First, we consider the case of \textcolor{black}{biases consistently appearing in measurements} representative of real-world scenarios where the observations are systematically biased. We consider that the bias corrupts each observation according to \eqref{eqn_res2_VB} and bias occurs in each dimension from the start of the simulation with probability $\lambda$ and sustains for the complete run time. 
	
	Fig.~\ref{Box1} shows the distribution of the root mean squared error (RMSE) calculated over 100 MC runs for the algorithms under consideration at different values of $\lambda$. {The lower bounds based on the diagonal entries of $\text{PCRB}_k$ depict the benchmark performance for different methods theoretically achievable considering different values of $\lambda$. For evaluation of $\text{PCRB}_k$ we assume $N_{mc1}=N_{mc2}=N_{mc3}=N_{mc4}=100$.} We can observe an intuitive trend in the relative performance of different methods. The standard UKF exhibits the worst estimation quality since it has no means for bias detection and compensation. For the data rejection-based method i.e. the SOR-UKF, we see that for lower probabilities of existence of bias discounting the observations performs satisfactorily. However, for higher values of $\lambda$ the rejection scheme does not work well since consistently rejecting large number of the measurements leads to loss of essential information. The CUKF and the USKF generally perform better than the SOR-UKF. \textcolor{black}{As compared to the other robust filters} the USKF is found to have more error at lower values of $\lambda$ since it does not treat each dimension selectively.  It rather uses the entire vector of measurements for compensation even if one of the dimensions is corrupted unlike the CUKF which offers a selective treatment. Importantly, it can be observed that BDM-UKF results in the least error among all the methods. 
	
	Fig.~\ref{fig:Case1_run} shows the state RMSE of the algorithms over time for one MC run depictive of the general trend in Fig.~\ref{Box1} for large values of $\lambda$ for this case. The BDM-UKF deals with biased observations more effectively followed by the partially compensating methods namely the CUKF and the USKF. The SOR-UKF, on the other hand, loses track of the ground truth due to permanent rejection of essential information. Similarly, the standard UKF with its inability to deal with \textcolor{black}{biases} also exhibits large errors.

	Lastly, we evaluate the computational overhead of each algorithm for Case 1. The mean processing time, considering 100 MC runs, for each method has been summarized in Table \ref{table:1}. We find the standard UKF to be the most economical and does not exhibit any significant change with different values of $\lambda$. The USKF takes more time since it involves updating the state and bias correlation terms. We observe a rise in the processing time of the USKF with increasing $\lambda$ since the USKF is invoked more frequently than the standard UKF we utilize when no bias is detected. The BDM-UKF and the SOR-UKF, both having an inbuilt detection mechanism, have a similar processing burden. Lastly, we find the CUKF to be the most computationally expensive algorithm. This can be owed to the presence of a convex quadratically constrained quadratic program (QCQP) which we solve using the MATLAB optimization toolbox.

	\subsection{Case 2: Momentary Bias Presence}\hspace{.5cm}
	We also consider the case where bias randomly appears for a short duration, characterizing practical scenarios e.g. where ambient effects disturb the data signals briefly. We again use \eqref{eqn_res2_VB} for each observation contamination and suppose bias occurs in each dimension with probability $\lambda$ at instant $t=100$ and sustains till $t=130$ before disappearing.
	
	Fig.~\eqref{Box2} depicts the spread of the RMSEs for 100 MC runs for each algorithm with varying $\lambda$ {along with lower bounds based on the PCRB using 100 MC samples for all the calculations. The plotting scale in this case allows us to clearly see a rise in the lower bound with increasing values of $\lambda$ which has also been observed for the previous case. This can be attributed to more chances of occurrence of \textcolor{black}{biased observations} with increasing $\lambda$ leading to increased estimation uncertainty}. In terms of the relative performance of different methods we see a similar pattern as for the previous case, owing to the same rationale regarding the functionality of the methods. The UKF generally has the largest RMSE, followed by the USKF, the SOR-UKF and the CUKF. The momentary appearance of bias does not degrade the performance of SOR-UKF as compared to the last case, except for $\lambda=0.8$ where it mostly diverges. SUKF is found to be relatively less effective at lower values of $\lambda$ due to its non-selective nature. Lastly, the BDM-UKF results in the lowest RMSE. Fig.~\ref{fig:Case2_run} shows the state RMSE of the algorithms over time for one MC run representative of the general trend in Fig.~\ref{Box2} for large values of $\lambda$ for this case. We can observe similar results as in the preceding case.

	\begin{table}[h!]
		\centering
		\begin{tabular}{|c c c c c c|} 
			\hline
			$\boldsymbol{\lambda}$ &\textbf{UKF}&\textbf{USKF}& \textbf{BDM-UKF} & \textbf{SOR-UKF} & \textbf{CUKF}  \\ 
			\hline\hline
			0.2 &0.0458&0.0515& 0.0682 & 0.0669& 0.09114  \\ 
			0.4  &0.0489&0.0521& 0.0687 & 0.0672& 0.1110 \\
			0.6 &0.0482&0.0520& 0.0687 & 0.0828& 0.1470  \\
			0.8 &0.0478&0.0520& 0.0690 & 0.0902& 0.2059  \\ 
			\hline
		\end{tabular}
		\caption{Average Time for 100 MC Runs with Momentary Bias at Different Values of $\lambda$ .}
		\label{table:2}
	\end{table}
	
	Lastly, the mean processing overhead of each algorithm for this case is presented in Table \ref{table:2}. We can see that the UKF takes approximately similar times as for Case 1. The overhead of remaining methods is reduced since the bias duration has now decreased. The order in which different algorithms appear in terms of the relative computational expense remains the same as observed in the previous case following from the same reasoning.

	\section{Conclusion}\label{Conc}\hspace{.5cm}
	In this chapter, we have considered the problem of robust filtering with biased observations. The performances of standard filtering approaches degrade when the measurements are disturbed by noise with unknown statistics. Focusing on the presence of measurement \textcolor{black}{biases}, we devise the BDM filter with inherent error detection and mitigation functionality. Performance evaluation reveals the efficacy of the BDM in dealing with both persistently and temporarily present biases. We find the BDM filter more accurate compared to rejection-based KF methods i.e. SOR-UKF. Moreover, owing to better utilization of the measurements, the BDM filter has lower estimation errors as compared to the methods with similar KF based approaches, the USKF and the CUKF, aiming to exploit information from the corrupted dimensions. The BDM filter is easier to employ as it avoids the use of external detectors and any optimization solver. The gains come at the expense of increased computational overhead which is comparatively higher compared to the UKF and USKF. However, it is comparable to SOR-UKF and lower than the CUKF which requires an additional optimization solver.
	
	%
	
\chapter{Robust Filtering Considering both Biases and Outliers} \label{chap-7}

\hspace{.5cm}
In this chapter, we propose an online scheme for state estimation of a generic class of nonlinear dynamical systems in the presence of abnormal measurement data from sensors including both outliers and biases. We illustrate why the performance of standard recursive Bayesian inference degrades in the presence of each type of measurement distortion. After demonstrating how different abnormalities can be accommodated explicitly inside a generic state–space model,
we propose a robust mechanism to perform recursive Bayesian inference on the presented model to not only detect but also mitigate the effect of corrupted measurements in the final state estimates. We leverage the sampling-based recursive inference engine namely the particle filter since the model and underlying distributions get complicated. Using simulations and experimental evaluation, we demonstrate the success of the proposed framework in reducing the impact of different types of distortions in measurements. The ability to tackle different kinds of measurement abnormalities during online inference sets the proposed method apart from the existing techniques. 

\section{Problem Formulation and Preliminaries}
\label{sec2}
\subsection{Standard SSM}\hspace{.5cm}
The standard non-linear discrete-time SSM, having $n$ states and $m$ measurements, with additive process and measurement noises can be expressed as
\begin{align}
	\mathbf{x}_k&=\mathbf{f}(\mathbf{x}_{k-1})+\mathbf{q}_{k-1}\label{eqn1_PF}\\
	\mathbf{y}_k&=\mathbf{h}(\mathbf{x}_{k})+\mathbf{r}_k\label{eqn2_PF}
\end{align} 
where $k$ denotes the time-index, $\mathbf{x}_k\in\mathbb{R}^{n}$ and $\mathbf{y}_k\in\mathbb{R}^{m}$ are the state and measurement vectors respectively, $\mathbf{f(.)}$ is the vector valued dynamic model function, and $\mathbf{h(.)}$ is the vector valued measurement model function. The vectors $\mathbf{q}_{k-1}\in\mathbb{R}^{n}$ and $\mathbf{r}_k\in\mathbb{R}^{m}$ are white process noise and the measurement noise vectors distributed respectively as: $\mathbf{q}_{k-1} \sim \mathcal{N}(\mathbf{0},\mathbf{Q}_{k-1})$ and $\mathbf{r}_k \sim \mathcal{N}(\mathbf{0},\mathbf{R}_k)$. 

\label{secPF2}
\subsection{Recursive Bayesian Inference During Measurement Abnormalities}\label{sectPPF3c}
\hspace{.5cm}
If recursive Bayesian estimation is employed on the standard SSM model \eqref{eqn1_PF}-\eqref{eqn2_PF}, in the presence of measurement abnormalities, the resulting state estimates are erroneous due to the use of corrupted measurements without any rejection/compensation strategy. In the following, we systematically demonstrate how the state estimates become erroneous due to measurement abnormalities. This analysis  motivates the need to introduce modifications to the standard SSM for explicitly modeling the occurrence of different types of abnormalities. Although, we consider a simple linear SSM with single state and single measurement for illustration purposes, we note that  the presented results can be generalized to higher-dimensional nonlinear SSM models as well. Consider a linear SSM  as
\begin{align}
	{x}_k&=a\ {x}_{k-1}+{q}_{k-1}\label{PPFeqn1}\\
	{y}_k&=b\ {x}_{k}+{r}_k\label{PPFeqn2}
\end{align} 
where $a$ and $b$ are positive pre-known scalar values and ${q}_{k-1} \sim \mathcal{N}({0},{\sigma_q}^2)$ and ${r}_k \sim \mathcal{N}({0},{\sigma_r}^2)$. Since the SSM model \eqref{PPFeqn1}-\eqref{PPFeqn2} is linear, the expression for the posterior distribution $p$(${x}_{k}|{y}_{1:k})$ can be found analytically  using recursive Bayesian estimation procedure. For this purpose, assume that $p({x}_{k-1}|{y}_{1:k-1})=\mathcal{N}({x}_{k-1}|\hat{x}^+_{k-1},P^+_{k-1})$.
\subsection*{{Prediction}}
\hspace{.5cm} Using \eqref{BF_eq2},\eqref{NP_eq2} the predictive distribution, at time index $k$, can be written as
\begin{align}
	p({x}_{k}|{y}_{1:{k-1}})&=\mathcal{N}({x}_{k}|\overbrace{a\ \hat{x}^+_{k-1}}^{\hat{x}^-_{k}},\overbrace{a^2\ P^+_{k-1}+{\sigma_q}^2}^{P^-_{k}})=k_1\exp[-\frac{1}{2}(x_k-\hat{x}^-_{k})^2/{P^-_{k}}]\label{PPFeqn4}
\end{align}
where $\hat{x}^-_{k}$ and ${P^-_{k}}$ denote the mean and variance of the predictive distribution respectively and $k_1$ is the normalizing constant for $p({x}_{k}|{y}_{1:{k-1}})$.
\subsection*{{Update}} 
\hspace{.5cm}
Subsequently, using \eqref{BE_eq4}, the posterior distribution,  at time step $k$, can be found as  
\begin{align}\hspace{0.9cm}
	p({x}_{k}|{y}_{1:{k}})&=\overbrace{\mathcal{N}({y}_{k}|b\ x_k,{\sigma_r}^2)}^{p({y}_{k}|{x}_{k})}\overbrace{\mathcal{N}({x}_{k}|\hat{x}^-_{k},P^-_{k}}^{p({x}_{k}|{y}_{k-1})})/\overbrace{{Z}_k}^{p({y}_{k}|{y}_{k-1})}\label{PPFeqn5}
\end{align}
where $Z_k$ is the normalizing factor given in \eqref{BF_eq3}. The functional form of ${p({y}_{k}|{x}_{k})}$ can be expressed as 
\begin{align}\hspace{1.4cm}
	{p({y}_{k}|{x}_{k})}&=\mathcal{N}({y}_{k}|b\ x_k,{\sigma_r}^2)=k_2\exp[-\frac{1}{2}(x_k-y_{k}/b)^2/({\sigma_r}/b)^2]\label{PPFeqn6}
\end{align}
where $k_2$ is a normalizing constant for $p({y}_{k}|{x}_{k})$.

\begin{figure}[h!]
	\centering
	\includegraphics[width=.5\linewidth,,frame]{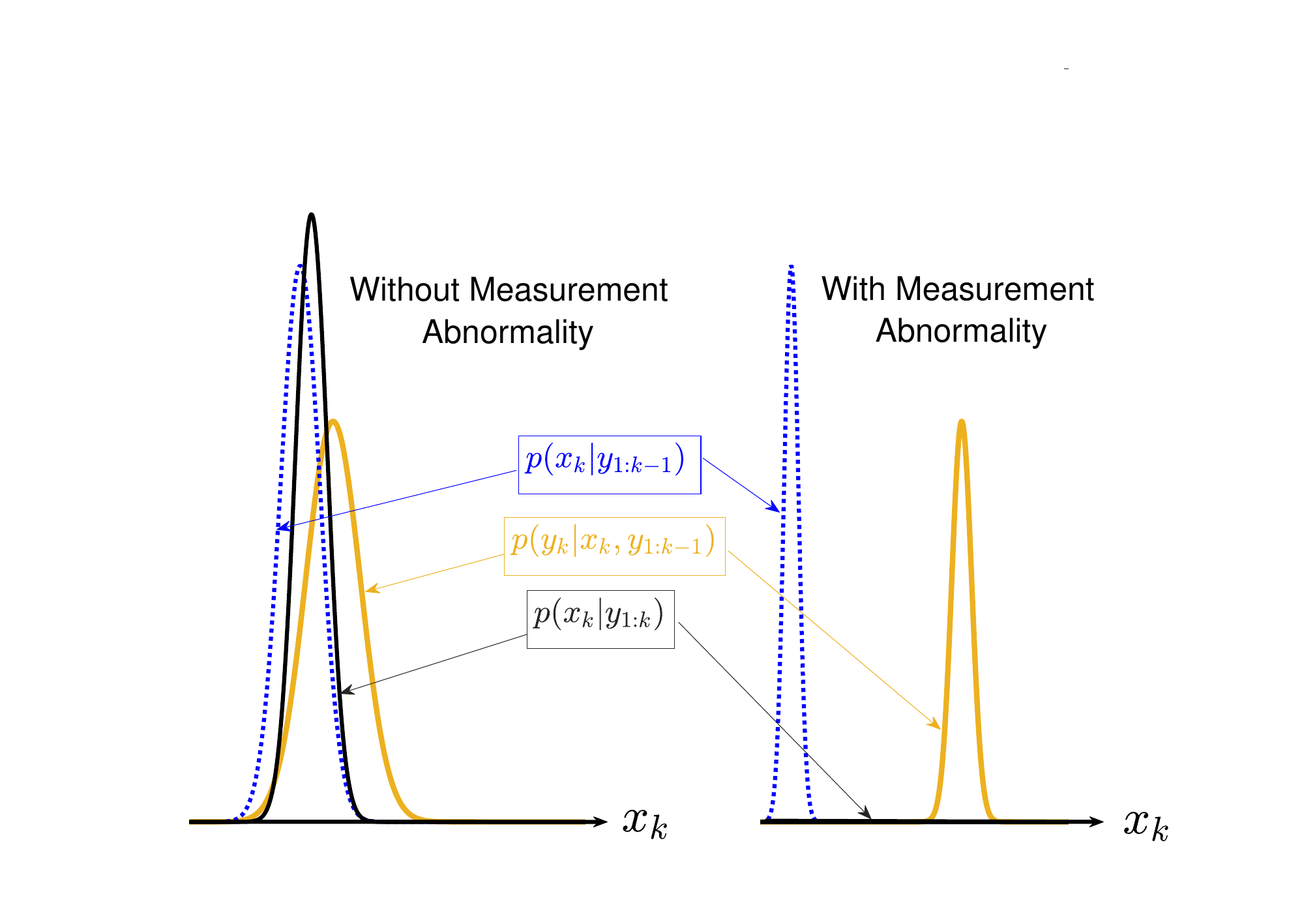}
	\caption{Visual illustration of recursive Bayesian inference using the standard model in the absence and presence of a  measurement abnormality at time $k$}
	\label{fig_PPF_1a}
\end{figure}

 Using \eqref{PPFeqn4} and \eqref{PPFeqn6}, the evolution of $p({x}_{k}|{y}_{1:{k}})$, in the normal working conditions (without any abnormality) and in the presence of measurement abnormality,  is depicted in Figure \ref{figI}. Two cases are reported;   the normal working conditions when measurement does not have any kind of abnormality at time step $k$, and  the case a measurement abnormality occurs at time step $k$. From Figure \ref{figI}, it can be observed that  the prior \textit{regularizes} the likelihood to result in the posterior distribution with increased confidence of the estimate as compared to both distributions individually, hence resulting in correct state estimates.  When an anomaly occurs,  the envelopes of the prior and likelihood distributions no longer coincide, therefore $p({x}_{k}|{y}_{1:{k}})$ is no longer well-defined. As a result, the estimation process fails with no single estimate being the clear confident choice. Essentially, the measurement abnormality does not follow the assumed statistics of $r_k$ in \eqref{PPFeqn2}. With a similar line of reasoning, it can be understood why recursive Bayesian estimation fails when the standard multidimensional nonlinear SSM is used for inference  in the presence of measurement abnormalities. 

If there is a mechanism to
\begin{enumerate}
	\item detect the abnormality in the measurement, and 
	\item compensate/mitigate the impact of this abnormality after detection
\end{enumerate}
the performance of the recursive Bayesian estimation can be improved. In the following section, we modify the standard SSM by proposing the usage of {Bernoulli random variables/vectors} for \textit{detecting} abnormal measurements, and a suitable statistical model for each type of abnormality to \textit{mitigate} its impact on the final estimate. We also provide a detailed analysis that demonstrates how these proposed changes in standard SSM improve the performance of recursive Bayesian estimation in the presence of measurement abnormalities. 

\section{Proposed Inference Model}\label{sec_model} 
\hspace{.5cm}
In this section, we present the development of a multidimensional inference model  which allows the recursive Bayesian inference framework to tackle the presence of both types of measurement abnormalities successfully. We first motivate our proposed model for both types of abnormalities for a single state and single measurement case before generalizing to a multidimensional model. 

\subsection{ Case I: Outliers with $n=1$, $m=1$  }
\textbf{Inference Model: }
We modify the standard SSM, with single state and single measurement, to include a statistical model for outliers and {a Bernoulli random variable} for their detection. More precisely, we have  
\begin{align}
	&{x}_k=f( {x}_{k-1})+{q}_{k-1}\label{PPFeqn7}&\\
	&{y}_k=h( {x}_{k})+{r}_k+{{\mathcal{I}}}^1_{k}\  {\mu}_k\label{PPFeqn8} &\\
	&{\mathcal{I}}^1_{k}=\delta(\mathcal{J}_{k}-1)&\label{PPFeqn9}
\end{align}
where an outlier is represented using ${\mu}_k$ in \eqref{PPFeqn8}, and is assumed to follow a zero mean Gaussian distribution as $\mathcal{N}({\mu}_k|0,{\sigma_{\mu}}^2)$. A very large value for ${\sigma_{\mu}}^2$  is chosen which essentially results in an uninformative prior for ${\mu}_k$. In \eqref{PPFeqn9}, {the Bernoulli random variable ${\mathcal{I}}^1_{k}$ is} defined in terms of a discrete random variable $\mathcal{J}_{k}\in\{0,1\}$ and  the delta function $\delta$(.). The  random variable $\mathcal{J}_{k}$ is used for modeling convenience and can be considered as a \textit{nuisance} parameter\cite{gelman2013bayesian}. In \eqref{PPFeqn9}, $J_k$ determines the possibility of an outlier at time instant $k$. In particular,  $\mathcal{J}_{k}=1$ (which implies $\mathcal{I}^1_{k}=1$), denotes the presence of an outlier whereas $\mathcal{J}_{k}=0\ ({\mathcal{I}}^1_{k}=0)$ indicates its absence. We further consider that $\mathcal{J}_k$ evolves in a Markovian manner with the transition probability $p$($\mathcal{J}_{k}=\beta({\mathcal{J}})|\mathcal{J}_{k-1}=\alpha({\mathcal{J}})$). At each time step, we assume  that there is no prior information on the occurrence of outlier. Hence, we set $p$($\mathcal{J}_{k}=\beta({\mathcal{J}})|\mathcal{J}_{k-1}=\alpha({\mathcal{J}})$)\  equal to 0.5  $\forall\ \alpha({\mathcal{J}})$, $\beta({\mathcal{J}})$. If any prior knowledge, of the frequency and statistics of outliers, is available, it can   be incorporated in the model for inference. Moreover, $\mathcal{J}_{k}$ and ${\mu}_k$ are assumed to be statistically independent.

\textbf{Form of Posterior: }
Recursive Bayesian inference can be performed on the proposed model in \eqref{PPFeqn7}-\eqref{PPFeqn9} to determine $p(x_k|y_{1:k})$. For illustration, we again resort to a linear SSM represented as $f(x_{k-1})=a\ {x}_{k-1}$, and  $h(x_{k})=b\ {x}_{k}$ as used in \eqref{PPFeqn1}-\eqref{PPFeqn2}. For analysis, we consider that an outlier occurs at an arbitrary time index $k$. Assuming $p(x_{k-1}|{y}_{1:{k-1}})$ $=$ $ \mathcal{N}({x}_{k-1}|\hat{x}^+_{k-1},P^+_{k-1})$ yields $p(x_{k}|{y}_{1:{k-1}})= \mathcal{N}({x}_{k}|\hat{x}^-_{k},P^-_{k})$ where $\hat{x}^-_{k}$ = $a\ \hat{x}^+_{k-1}$ and $P^-_{k}=a^2\ P^+_{k-1}+{\sigma_q}^2$. We can write the posterior $p(x_{k}|{y}_{1:{k}})$ in an analytical form given in \eqref{PPFeqn10} where $k_2$ and $k_3$ are constants. For its detailed derivation, the reader is referred  to the appendix. 

\begin{figure*}[t]
	\begin{align}
		&p({x_k}|y_{1:k})\propto \overbrace{(k_2\exp[-\frac{(x_k-y_{k}/b)^2}{2({\sigma_r}/b)^2}]\ 
			+k_3\exp[-\frac{(x_k-y_{k}/b)^2}{2(({\sigma_r}+{\sigma_{\mu}})/b)^2}])}^{p(y_k|x_k,y_{1:k-1})} 
		\overbrace{\exp[-\frac{(x_k-\hat{x}^-_{k})^2}{2{P^-_{k}}}]}^{p(x_k|y_{1:{k-1}})}\label{PPFeqn10} 
	\end{align}
\end{figure*}

\begin{figure}
	\centering
	\includegraphics[width=.5\linewidth,,frame]{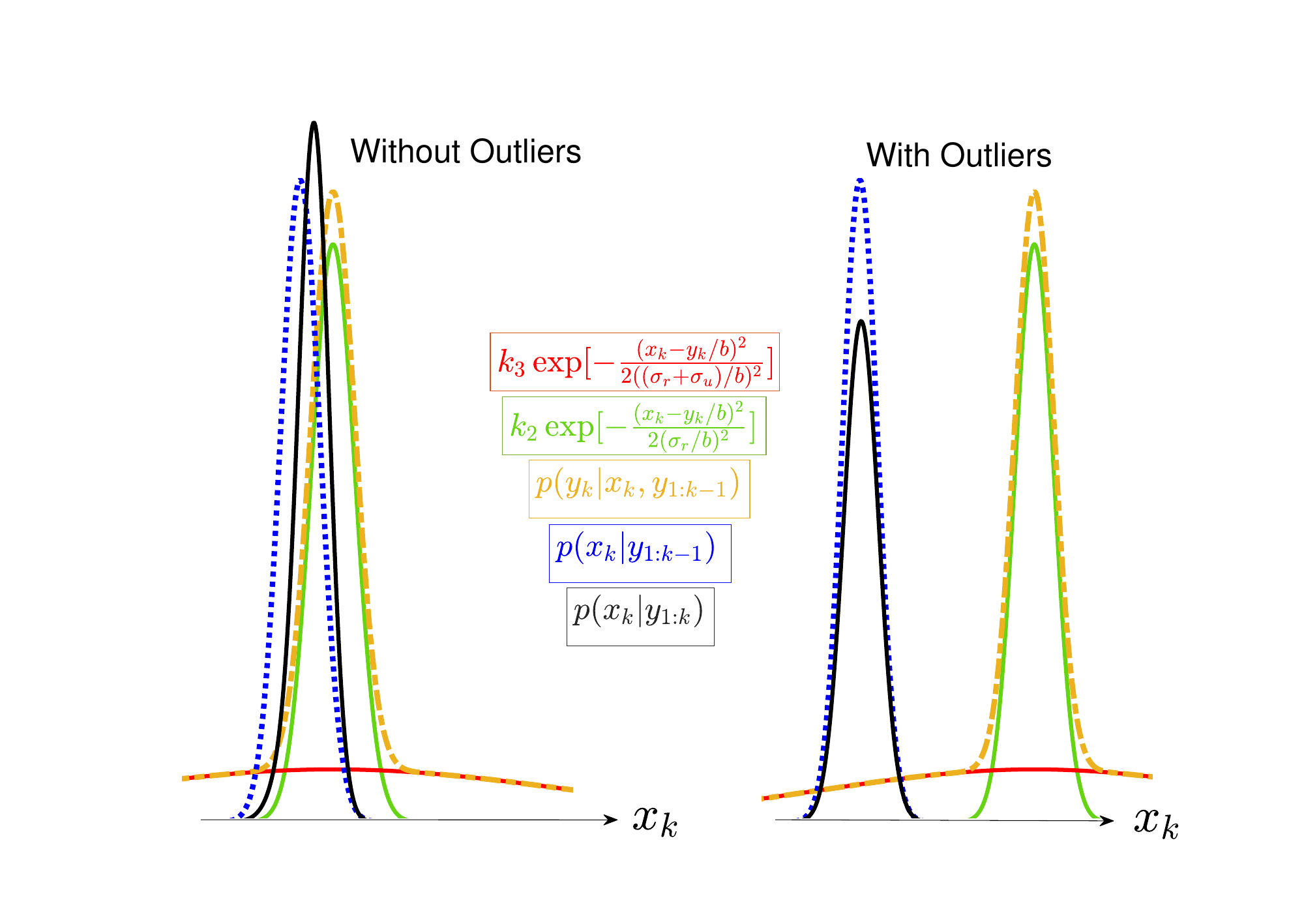}
	\caption{Visual illustration of recursive Bayesian inference using the proposed model in the absence and presence of a measurement outlier at time $k$}
	\label{fig:IIa}
\end{figure}

\textbf{Analysis of \eqref{PPFeqn10}: }
In order to analyze the expression in  \eqref{PPFeqn10} and highlight its efficacy in dealing with outliers, we plot its evolution in Figure \ref{fig:IIa} for two different scenarios: no outlier at time $k$, and with an outlier at time $k$. In both cases, the likelihood $p({y}_{k}|x_k,{y}_{1:{k-1}})$ is sum of two modes which have same means but different variances. For the case when there is no outlier, there is a significant overlap between the prior and likelihood distributions, which results in a posterior almost as in the normal working conditions. On the other hand, when an outlier occurs, there still exists an overlap between the resultant likelihood and prior. It leads to the form of posterior, which is essentially (approximately) the same as $p({x}_{k}|{y}_{1:{k-1}})$ and hence, no update takes place. Consequently, the model caters for the outliers in essentially a detect-and-reject fashion, i.e., whenever an outlier occurs, it is ignored for the state estimation.

\subsection{Case II: Biases with $n=1$, $m=1$}
\hspace{.5cm}
Now, we turn our attention to modeling biases in the standard SSM by considering single state and single measurement model. \\

\textbf{Inference Model: } Consider the case where bias appears at time step $k$ and sustains to the next time step $k+1$. We consider that at time step $k$ a bias of magnitude $\Theta_k$ occurs with some uncertainty. This effect is modeled using a random bias model \cite{park2020robust} as  $\nu_k\sim\mathcal{N}(\Theta_k,{\sigma_{\nu}}^2)$ where ${\sigma_{\nu}}^2$ models the
uncertainty of the bias leading to 
\begin{align}
	\hspace{1cm}
	{x}_{k} &=f( {x}_{k-1})+{q}_{k-1}\label{PPFeqn11}&\\
	{y}_{k} &=h( {x}_{k})+{r}_{k}+{\nu}_{k}  \label{PPFeqn12}
\end{align}
Similarly, at time $k+1$, the bias effect is captured using $\nu_{k+1} \sim \mathcal{N} (\Theta_{k+1}, {\sigma_{\nu}}^2)$, where $\Theta_{k+1}$ is assumed to evolve as a first order Markovian process as described in \eqref{PPFeqn15}. Hence, we obtain
\begin{align}
	{x}_{k+1} &=f( {x}_{k})+{q}_{k}\label{PPFeqn13}\\
	{y}_{k+1} &=h( {x}_{k+1})+{r}_{k+1}+{\nu}_{k+1}\label{PPFeqn14} \\
	\Theta_{k+1} &=\Theta_{k}+\triangle{\Theta}_{k}  \text{ with } \triangle \Theta_k\sim\mathcal{N}(0, { {\sigma_{\triangle}}}^2 ) \label{PPFeqn15}
\end{align}
where $\triangle \Theta_k$ incorporates the effect of the possible drift error and ${{\sigma_{\triangle}}}^2$ represents the variance of the drift. We assume that $\nu_k$ and $\triangle \Theta_k$ are independent and White. The model described in \eqref{PPFeqn11}-\eqref{PPFeqn15} is appropriate for inference only when an \textit{apriori} information is available about the time indices where bias occurs and then sustains. However, in practice, the actual instance of occurrence of bias is not known beforehand. Therefore, we introduce a {Bernoulli random variable} ${\mathcal{I}}^{\xi}_{k}$ defined as ${\mathcal{I}}^{\xi}_{k}=\delta(\mathcal{J}_{k}-{\xi})$. Here, the discrete random variable $\mathcal{J}_{k}\in\{0,2\}$ determines if a bias has occurred. For example, $\mathcal{J}_{k}=2\ ({\mathcal{I}}^2_{k}=1)$ denotes the presence of bias. Further, we assume $J_k$  evolves in a Markovian manner with the transition probability $p$($\mathcal{J}_{k}=\beta({\mathcal{J}})| \mathcal{J}_{k-1}=\alpha({\mathcal{J}})$). We assume $p$($\mathcal{J}_{k}=\beta({\mathcal{J}})|\mathcal{J}_{k-1}=\alpha({\mathcal{J}})$)\  equal to 0.5  $\forall\ \alpha({\mathcal{J}})$, $\beta({\mathcal{J}})$. Hence, the model in \eqref{PPFeqn11}-\eqref{PPFeqn15} is modified as
\begin{align}
	\hspace{1cm}
	{x}_k &=f( {x}_{k-1})+{q}_{k-1}\label{PPFeqn16}\\
	{y}_k &=h( {x}_{k})+{r}_k+{\mathcal{I}}^2_{k}\  {\nu}_k \text{ with } \nu_k \sim \mathcal{N}(\Theta_k,{\sigma_{\nu}}^2) \label{PPFeqn17}\\
	{\Theta}_k &={\mathcal{I}}^0_{k-1} \overset{{\sim}}{\Theta}_k+{\mathcal{I}}^2_{k-1}(\Theta_{k-1}+\triangle \Theta_{k-1})  \label{PPFeqn18} \\
	{\mathcal{I}}^{\xi}_{k} &=\delta(\mathcal{J}_{k}-\xi),\  \xi\in(0,2)\label{PPFeqn19}
\end{align}
where in \eqref{PPFeqn18}, $\Theta_k$ follows a Markovian model if at the previous time step any bias occurs in the measurement, i.e. if ${\mathcal{I}}^{2}_{k-1}=1$ then ${\Theta}_k={\Theta}_{k-1}+\triangle{{\Theta}_{k-1}}$. The  Markovian model is used to incorporate the memory effect of the sustained bias in the measurement, including  the possibility of  drift in the bias. Otherwise, if no outlier occurs at the previous time step, i.e. ${\mathcal{I}}^{0}_{k-1}=1$, then ${\Theta}_k=\overset{{\sim}}{\Theta}_k$, where $\overset{{\sim}}{\Theta}_k$ is assumed to be uniformly distributed in a predefined range  i.e. $\overset{{\sim}}{\Theta}_k\sim\mathcal {U}(c,d)$. This models the possibility of occurrence of any bias in the range $c$ to $d$. In addition, $\mathcal{J}_{k}$ and $\nu_k$ are assumed to be statistically independent. 

\scalebox{.85}{\parbox{1.07\linewidth}{%
	\begin{flalign}
		&{p({x_k}|{y}_{1:{k}})}\propto \overbrace{(3k_4\exp[-\frac{(x_k-y_{k}/b)^2}{2({\sigma_r}/b)^2}]\ 
			+k_5(\int_{c}^{d}\exp[-\frac{(x_k-(y_k-{\Theta}_k)/b)^2}{2({\sigma_r}^2+{{\sigma_{\nu}}^2})}]d{\Theta}_k))}^{p(y_k|x_k,y_{1:{k-1}})}\ 
		\overbrace{\exp[-\frac{(x_k-\hat{x}^-_{k})^2}{2{P^-_{k} }}]}^{p(x_k|y_{1:{k-1}})}\label{PPFeqn20} &\\
		&{p({x_{k+1}}|{y}_{1:{k+1}})}\propto \overbrace{(k_6\exp[-\frac{(x_{k+1}-(y_{k+1}/b))^2}{2({\sigma_r}^2/b^2)}]\ 
			+k_7\exp[-\frac{(x_{k+1}-{{{\hat{x}}^{1}_{k+1}} })^2}{2 {{P}^{1}_{k+1}} }]}^{p(y_{k+1}|x_{k+1},y_{1:{k}})}\ 
		\overbrace{\exp[-\frac{(x_{k+1}-a \hat{x}^-_{k})^2}{2(a^2 P^-_{k} +{\sigma_q}^2)}]}^{p(x_{k+1}|y_{1:{k}})} \label{PPFeqn21}&
		\\
		&{p({x_{k+1} }|{y}_{1:{k+1}})}\propto k_6\exp[-\frac{(x_{k+1}-{{{\hat{x}}^{2}_{k+1}}})^2}{2 {{P}^{2}_{k+1}} }]+k_7\exp[-\frac{(x_{k+1}-{{{\hat{x}}^{3 }_{k+1}}})^2}{2 {{P}^{3}_{k+1}} }] \label{PPFeqn22}& 
	\end{flalign}
}}

\textbf{Form of Posterior at $k$ and $k+1$:}  For illustration, we  again resort  to a linear SSM represented as $f(x_{k-1})=a\ {x}_{k-1}$, and  $h(x_{k})=b\ {x}_{k}$ as used in \eqref{PPFeqn1}-\eqref{PPFeqn2}. Using this information inside the model given in \eqref{PPFeqn16}-\eqref{PPFeqn19}, we obtain the posterior at step $k$ and $k+1$ as given in \eqref{PPFeqn20} and \eqref{PPFeqn22} respectively, where the values of different constants, mean of the density functions, and corresponding variances are provided in the appendix along with complete derivation of the posterior form. 

\textbf{Analysis of \eqref{PPFeqn20} and \eqref{PPFeqn22}:} Figure \ref{fig:IIIa} plots the evolution of posterior ${p({x_k}|{y}_{1:{k}})}$ in cases when there is no bias error at time $k$, and also when there is a bias error at time $k$. We observe that likelihood is a sum of two different modes at the time step $k$. In case of no bias, the likelihood distribution stays closer to the prior as compared to the case when there is a bias error.  As a result, the measurement is correctly used to update the posterior when there is no bias. In case when bias occurs at time $k$, the posterior is almost the same as prior, and the effect of the biased measurement does not corrupt the state estimates. The interesting case turns out to be at time $k+1$ where two possibilities exist: either the bias sustains or it disappears. In both cases, the expression for the posterior is given in \eqref{PPFeqn22}. In both cases, the likelihood $p(y_{k+1}|x_{k+1},y_{1:{k}})$, given in \eqref{PPFeqn21}, remains bi-modal. In case the bias sustains at $k+1$, one of the modes of likelihood keeps coinciding with the prior to produce a well-defined posterior distribution. In this way, the impact of the sustained bias error is mitigated in the final state estimates. The same is true for the other mode of the likelihood in the case when the bias disappears at $k+1$, as depicted in Figure \ref{fig:IIIa}.  The results for this simple case are intuitively appealing. In particular, on detection of bias at any time instance, the measurement is ignored. Once any bias sustains, the estimates are based on the difference of measurements. Otherwise, if the bias disappears, the estimates are based on the current measurement.  

\begin{figure}[h!]
	\hspace{0cm}
	\includegraphics[width=1\linewidth,trim=50 0 0 0,frame]{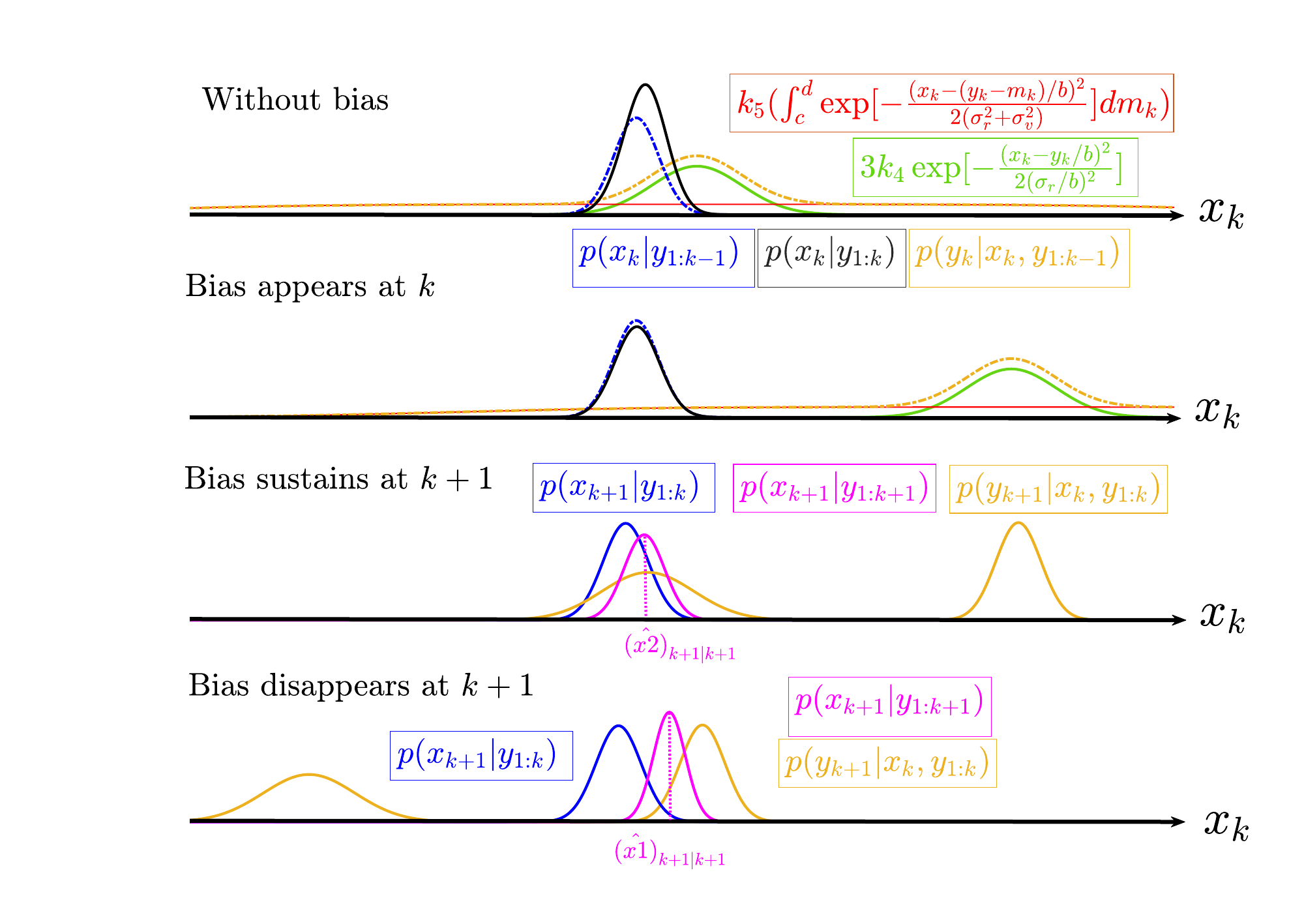}
	\caption{Visual illustration of recursive Bayesian inference using the proposed model in the absence and presence of bias at time $k$, and evolution of ${p({x_{k+1} }|{y}_{1:{k+1}})}$ when bias is sustained/not-sustained at $k+1$}
	\label{fig:IIIa}
\end{figure}

\subsection{Case III: Outliers and Biases for Generic $n$ and $m$: }\label{model_multi}\hspace{.5cm}
Now, we merge both individual models, given in \eqref{PPFeqn7}-\eqref{PPFeqn9} and \eqref{PPFeqn16}-\eqref{PPFeqn19}, into a single generalized model as
\begin{align}
	\mathbf{x}_k&=  \mathbf{f}(\mathbf{x}_{k-1})+\mathbf{q}_{k-1}\label{PPFeqn23}&\\
	\mathbf{y}_k &= \mathbf{h}(\mathbf{x}_{k})+\mathbf{r}_k + {\bm{\mathcal{I}}^1_{k}\odot\bm{\mu}_k+\bm{\mathcal{I}}^2_{k}\odot\bm{\nu}_k} \label{PPFeqn24}&\\
	\bm{{\Theta}_k} &=(\bm{\mathcal{I}}^0_{k-1}+\bm{\mathcal{I}}^1_{k-1})\odot \overset{\sim}{\bm{\Theta}}_k+\bm{\mathcal{I}}^2_{k-1}\odot(\bm{\Theta}_{k-1}+\triangle{\bm{\Theta}_{k-1}}) \label{PPFeqn25}&\\
	\mathbf{q}_{k-1} &\sim \mathcal{N}(\mathbf{0},\mathbf{Q}_{k-1}) \label{PPFeqn26}&\\
	\mathbf{r}_k  &\sim \mathcal{N}(\mathbf{0}, \mathbf{R}_k) \label{PPFeqn27}&\\
	\bm{\mu}_{k} &\sim \mathcal{N}(\mathbf{0}, \mathbf{U}_k) \text{ with } \mathbf{U}_k = \text{diag}\left( \sigma^2_{u_{1}} , \cdots, \sigma^2_{u_{m}} \right)\label{PPFeqn28}\\
	\bm{\nu}_k &\sim \mathcal{N}(\mathbf{\Theta}_k, \mathbf{\Upsilon}_k) \text{ with } \mathbf{\Upsilon}_k = \text{diag}\left( \sigma^2_{\nu_{1}}, \cdots, \sigma^2_{\nu_{m}} \right)&\label{PPFeqn29}\\
	\triangle{\bm{\Theta}_{k-1}} &\sim \mathcal{N}(\mathbf{0},\mathbf{\Delta}_{k-1}) \text{ with } \mathbf{\Delta}_{k-1} = \text{diag}\left(\sigma^2_{\triangle_{1}}, \cdots, \sigma^2_{\triangle_{m}} \right) \label{PPFeqn30}&\\
	\overset{\sim}{\bm{\Theta}}_{k} &= [\overset{\sim}{{\Theta}}_k(1)\cdots \overset{\sim}{{\Theta}}_k(m) ]^{\top}  \text{ with } \overset{\sim}{{\Theta}}_k(i)\sim\mathcal{U}(c(i),d(i))\label{PPFeqn31}&\\
	\bm{\mathcal{I}}^{\xi}_{k}&=[\delta( \mathcal{J}_{k}(1)-\xi)\  \cdots \ \delta(\mathcal{J}_{k}(m)-\xi)]^{\top} \text{ with } \xi\in\{0,1,2\} \label{PPFeqn32} & \\
	\bm{\mathcal{J}}_{k}&=[ \mathcal{J}_{k}(1)\   \cdots \ \mathcal{J}_{k}(m)]^{\top} \text{ with } \mathcal{J}_{k}(i)\in\{0,1,2\} \label{PPFeqn34}
\end{align}

where $k$ denotes the time-index, { $\mathbf{x}_k\in \mathbb{R}^{n}$ and $\mathbf{y}_k\in\mathbb{R}^{m}$ are the state and measurement vectors respectively, $\mathbf{q}_{k-1}\in \mathbb{R}^{n}$ and $\mathbf{r}_k\in \mathbb{R}^{m}$ are White process and  measurement noise vectors, $\mathbf{f}(.)$ and $\mathbf{h}(.)$ represent nonlinear process and measurement dynamics respectively, $\bm{\mu}_k\in \mathbb{R}^{m}$ model the effect of outliers, $\bm{\nu}_k\in \mathbb{R}^{m}$ and $\bm{\Theta}_k\in \mathbb{R}^{m}$ model the effect of biases in the measurement respectively} and the symbol $\odot$ denotes the Hadamard product.

The occurrence of an abnormality, in different dimensions of the measurement vector, is captured by the {multivariate Bernoulli random vectors} $\bm{\mathcal{I}}^{\xi}_{k} \in \mathbb{R}^{m}$ for $\xi=0,1,2$. For example, $\bm{\mathcal{I}}^0_{k}=[1\ 1\ 1\  ....\ 1]^{\top}$ indicates that no abnormality has occurred, $\bm{\mathcal{I}}^1_{k}=[1\ 0\ 0\  ....\ 0]^{\top}$ denotes the occurrence of an outlier in the first dimension and $\bm{\mathcal{I}}^2_{k}=[0\ 0\ 1\  ....\ 0]^{\top}$ shows the occurrence of bias in the third dimension of the measurement vector, and so on. Further, we assume $\mathcal{J}_{k}(i)$ is  statistically independent of $\mathcal{J}_{k}(j)$ for $i \neq j$ where the event $\mathcal{J}_{k}(i)$$\ \in(0,1,2)$ denotes the occurrence of any abnormality in dimension $i$ at time $k$. We again assume the evolution of $\mathcal{J}_{k}(i)$ to be Markovian $\forall \ i$, governed by the transition probability $p$($\mathcal{J}_{k}(i)=\beta({\mathcal{J}})|\mathcal{J}_{k-1}(i)=\alpha({\mathcal{J}})$). Note that we use $\bm{\mathcal{J}}_k$ and $\bm{\mathcal{I}}^n_{k}$ interchangeably for conciseness as these are alternate forms of representing the types of outliers. 

It is assumed that an abnormality occurring in each measurement is statistically independent of each other which suggests a diagonal structure for $\mathbf{U}_k$ and $\mathbf{\Upsilon}_k$. The diagonal entries of $\mathbf{U}_k$ can be predefined to the variance of the heavy-tailed noise if they are known \textit{apriori}. However, in most practical situations no exact prior knowledge is available, therefore, we assume the variances to be very large. Similarly, $\mathbf{\Upsilon}_k$  is predefined with small entries to include the effect of uncertainty in bias. 

\begin{figure} [t!]
	\centering
	\includegraphics[scale=0.35,frame,trim=-15 -15 -15 -15 ]{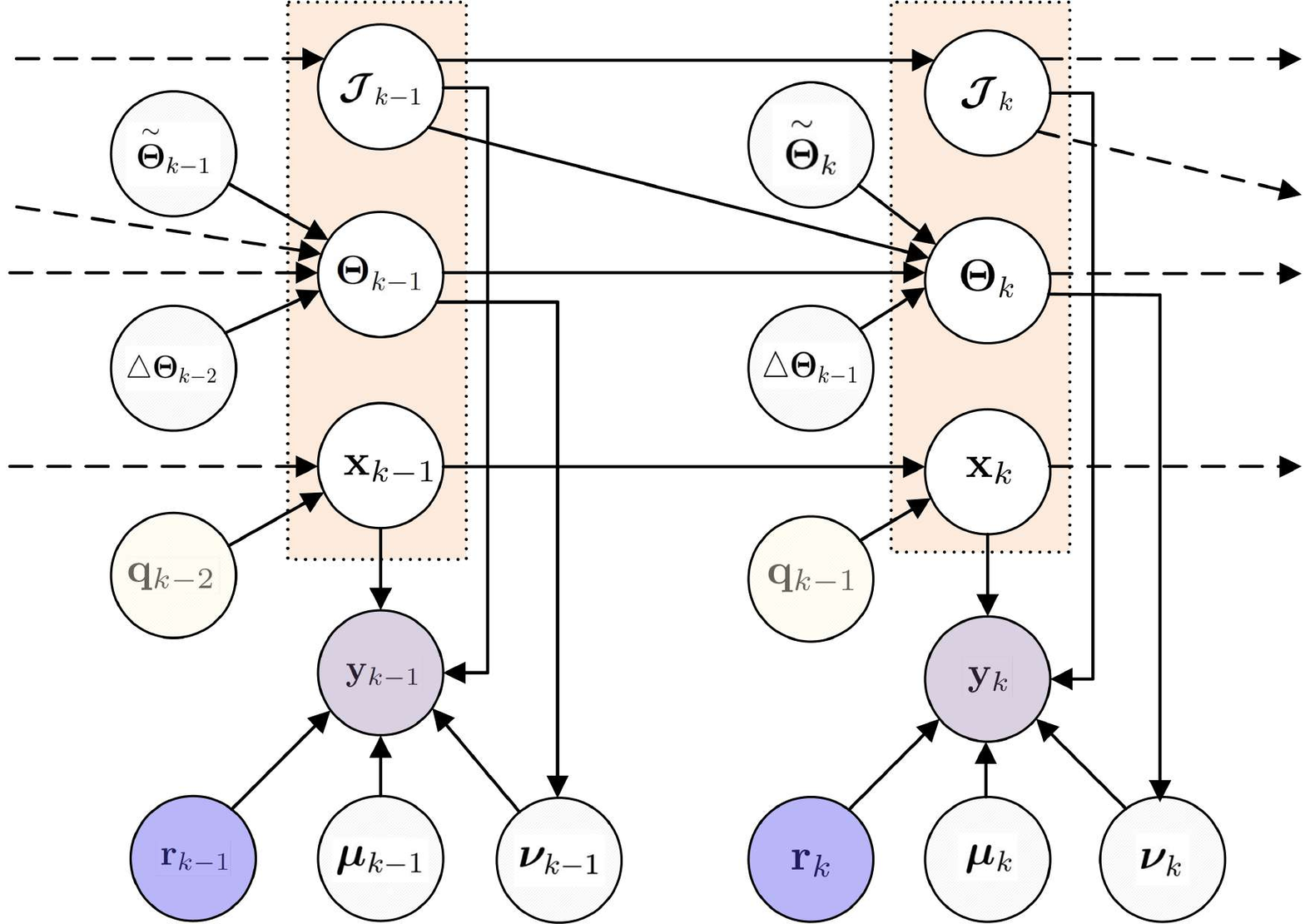}
	\caption{PGM of the proposed framework  }
	\label{fig:PGM}
\end{figure}

In \eqref{PPFeqn25}, the $i${th} element of  $	\bm{{\Theta}_k}$, denoted as ${{\Theta}_k}(i)$, follows a Markovian model (to incorporate the memory effect of the sustained bias) if a bias occurs in the $i${th} dimension of the measurement at time $k-1$, i.e. if $\bm{\mathcal{I}}^{2}_{k-1}(i)=1$ then ${\Theta}_k(i)={\Theta}_{k-1}(i)+\triangle {\Theta}_{k-1}(i)$. To account for any drift, we assume $\triangle{\bm{\Theta}_{k-1}}$ to be White and normally distributed  as $\mathcal{N}(\mathbf{0},\mathbf{\Delta}_{k-1})$. If no abnormality has occurred in the $i${th} dimension at the previous time step, i.e. $\bm{\mathcal{I}}^{0}_{k-1}(i)=1$ or an outlier has occurred in the $i${th} dimension at the previous time step, i.e. $\bm{\mathcal{I}}^{1}_{k-1}(i)=1$, then ${\Theta}_k(i)=\overset{\sim}{{\Theta}}_{k}(i)$ where $\overset{\sim}{{\Theta}}_{k}(i)$ is assumed to be uniformly distributed in a predefined range. This models the possibility of occurrence of bias in any given dimension $i$ in the range $c(i)$ to $d(i)$. Since biases can lie in a positive or negative range, $c(i)$ can be set equal to $-d(i)$ conveniently unless any prior knowledge about the outlier is available. We also assume that $\overset{\sim}{{\Theta}}_{k}(i)$ is White and statistically independent of $\overset{\sim}{{\Theta}}_{k}(j)$ for $i \neq j$.

The PGM of the proposed state-space framework is shown in Figure \ref{fig:PGM}.  

\subsubsection*{Remarks}\hspace{.5cm}
Note that the proposed model is generic and does not assume any prior information of the statistics of outliers; the transition probabilities that can be easily incorporated if available. Moreover, the presented model assumes the extreme case where each measurement can be corrupted with any type of measurement abnormalities. However, we can easily modify the proposed inference model depending on any particular characteristics of the specific measurement dimension. For example, the model can easily be modified considering that particular sensors are known to produce outliers only, biases only, or neither of these by ignoring all the relevant parameters (of outliers or biases) in different dimensions. This can reduce the computational overhead of the inference and is more suitable for larger dimensional systems.

\vfill
\newpage

\section{State Estimation Using SMC Method on the Proposed Model}\label{sect_PF}

\begin{figure} [h!]
	\centering
	\includegraphics[scale=1.2,frame,trim=-15 -15 -15 -15]{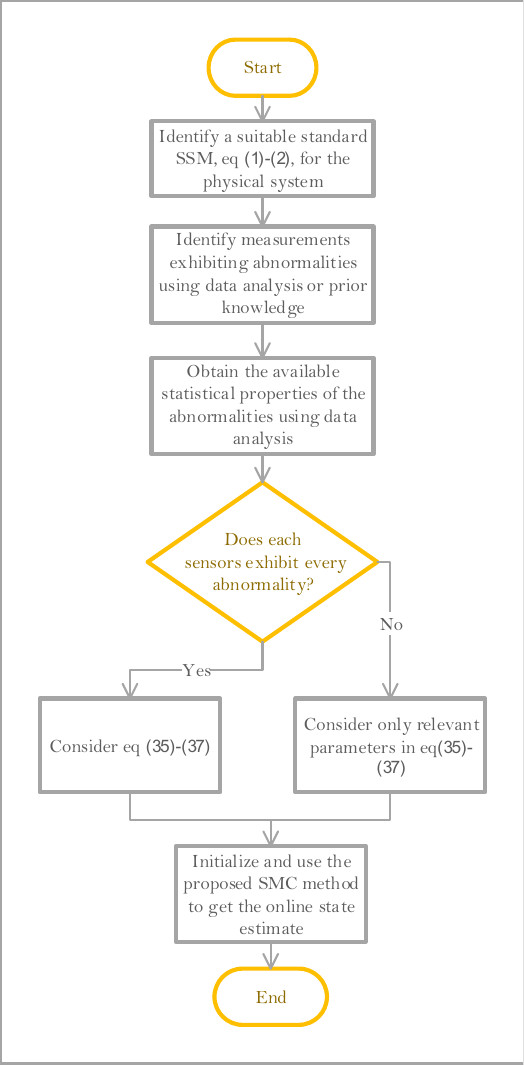}
	\caption{Flowchart of the proposed method}
	\label{fig:flow}
\end{figure}
\hspace{.5cm}
We are interested in the state estimate  at each time step using all the previous measurements. To this end, we perform joint inference of the state and the \textit{nuisance} parameters, considering the model in \eqref{PPFeqn23}-\eqref{PPFeqn24}. In particular, if $\mathbf{s}_k=[\mathbf{x}_k \ \bm{\Theta}_k \ \bm{\mathcal{J}}_{k}]^{\top}$ is the state and nuisance parameters augmented vector and $\mathbf{y}_{1:k}$ is the set of measurements up to time $k$, the inference is carried out by determining the joint posterior distribution of  $\mathbf{s}_{k}$ conditioned on the set of measurements i.e. $p$($\mathbf{s}_{k}|\mathbf{y}_{1:k})$.

Since $p$($\mathbf{s}_{k}|\mathbf{y}_{1:k})$ becomes analytically intractable for nonlinear systems,  we use an SMC based approach to approximate it by an empirical distribution using MC samples. However, drawing samples from an arbitrary distribution is not trivial, hence an importance distribution is sought for sampling. Conveniently, the importance distribution can be chosen as $p$($\mathbf{s}_{k}|\mathbf{y}_{1:{k-1}})$ which is not necessary in general \cite{tulsyan2016-eps-converted-to.pdf}. Consequently, we can empirically represent the required distribution as       
\begin{align}
	&p(\mathbf{s}_{k}|\mathbf{y}_{1:k}) \approx \sum_{l=1}^N w_{k}^{-(l)} \delta(\mathbf{s}_{k}-\hat{\mathbf{s}}_{k}^{-(l)})& \label{PPFeqn35}
\end{align}
where total $N$ samples $\hat{\mathbf{s}}_{k}^{-(l)}$ are drawn from the importance distribution and the importance weights $w_{k}^{-(l)}$ cater for the mismatch between the target and sampling distribution. We  determine the importance weights as 
\begin{align}
	\hspace{1cm}
	w_{k}^{-(l)} \propto& \quad  p(\hat{\mathbf{s}}_{k}^{-(l)}|\mathbf{y}_{1:k})/p(\hat{\mathbf{s}}_{k}^{-(l)}|\mathbf{y}_{1:{k-1}})=p(\mathbf{y}_{k}|\hat{\mathbf{s}}_{k}^{-(l)})\label{PPFeqn36}
\end{align}
which are subsequently normalized to sum to unity. For recursion, we have
\begin{flalign}
	\hspace{3.2cm}
	&p(\mathbf{s}_{k}|\mathbf{y}_{1:{k-1}})=\frac{1}{N}\sum_{l=1}^{N}p(\mathbf{s}_{k}|\hat{\mathbf{s}}^{+(l)}_{{k-1}})&\label{PPFeqn37}
\end{flalign}
where $\hat{\mathbf{s}}_{{k-1}}^{+(l)}$ denote samples drawn from $p(\mathbf{s}_{k-1}|\mathbf{y}_{1:{k-1}})$. Therefore, we  use the system model described by \eqref{PPFeqn23}-\eqref{PPFeqn25} to propagate $\hat{\mathbf{s}}_{{k-1}}^{+(l)}$ to obtain $\hat{\mathbf{s}}_{{k}}^{-(l)}$. Under the assumptions considered in Section \ref{model_multi}, we obtain the expressions \eqref{PPFeqn38}-\eqref{PPFeqn42} to be used inside SMC algorithm for inference whose details are provided in Algorithm \ref{algo1}.  Note that we have selected the transition probabilities   $p(\mathcal{J}_{k}(i)=\beta({\mathcal{J}})| \mathcal{J}_{k-1}(i)=\alpha({\mathcal{J}})) = 1/3$ for all $\alpha({\mathcal{J}})$ and $\beta({\mathcal{J}})$. We consider there are total $K$ time steps. A resampling step is required to address the degeneracy of  SMC methods \cite{sarkka2023bayesian}. 

{Figure \ref{fig:flow} summarizes the general sequence for using the proposed method of online filtering in different applications.}

\vfill
\newpage
\scalebox{.9}{\parbox{1.07\linewidth}{%
	\begin{align}
		&p\left(\mathbf{s}_k|\hat{\mathbf{s}}^{+(l)}_{k-1}\right) =  p\left(\mathbf{x}_k|\mathbf{\hat{x}}^{+(l)}_{k-1} \right) p\left(\mathbf{\Theta}_k|\mathbf{\Theta}^{+(l)}_{k-1},\bm{\mathcal{J}}^{+(l)}_{k-1}\right)  p\left(\bm{\mathcal{J}}_k|\bm{\mathcal{J}}^{+(l)}_{k-1} \right)\label{PPFeqn38} \\
		&p\left(\mathbf{x}_k|\hat{\mathbf{x}}^{+(l)}_{k-1}\right) = \mathcal{N}\left(\mathbf{x}_k|\mathbf{f} \left(\hat{\mathbf{x}}^{+(l)}_{k-1} \right),\mathbf{Q}_{k} \right)\label{PPFeqn39} \\
		&p(\mathbf{\Theta}_k|\mathbf{\Theta}^{+(l)}_{k-1},{\bm{\mathcal{J}}}^{+(l)}_{k-1})
		=\prod_{i=1}^{m}\left[\delta(\mathcal{J}^{+(l)}_{k-1}(i)) 
		+  \delta(\mathcal{J}^{+(l)}_{k-1}(i)-1) \right]\ \mathcal{U}({\Theta}_k(i)|c(i),d(i)) \quad 
		+ \nonumber \\ & \hspace{6cm} \delta(\mathcal{J}^{+(l)}_{k-1}(i)-2) \mathcal{N}({\Theta}_k(i)|{\Theta}^{+(l)}_{k-1}(i),{\sigma^{2}_{\triangle i}})\label{PPFeqn40} \\
		&p(\bm{\mathcal{J}}_k|\bm{\mathcal{J}}^{+(l)}_{k-1}) =\prod_{i=1}^{m} p({\mathcal{J}}_k(i)|\mathbf{\mathcal{J}}^{+(l)}_{k-1}) =\prod_{i=1}^{m}\sum_{\xi}p(\mathbf{\mathcal{J}}_k(i)=\xi|\mathbf{\mathcal{J}}^{+(l)}_{k-1})\delta(\mathbf{\mathcal{J}}_k(i)-\xi)  \nonumber \\&\hspace{2.22cm}=\prod_{i=1}^{m} \sum_{\xi}\frac{1}{3}\delta(\mathbf{\mathcal{J}}_k(i)-\xi)
		\label{PPFeqn41} \\
		&p(\mathbf{y_k}|\hat{\mathbf{s}}^{-(l)}_{k})=\mathcal{N} \left(\mathbf{y}_k| \mathbf{h}(\mathbf{x}^{-(l)}_{k})+\bm{\mathcal{I}}^{-2(l)}_{k}\odot\mathbf{\Theta}^{-(l)}_{k},
		\mathbf{R_k}+\text{diag}(\bm{\mathcal{I}}^{-1(l)}_{k})\odot\mathbf{U}_k+\text{diag}(\bm{\mathcal{I}}^{-2(l)}_{k})\odot\mathbf{\Upsilon}_k \right) \label{PPFeqn42}
	\end{align}
}}


\begin{algorithm}[h!]
	\SetAlgoLined
	Generate \{$\hat{\mathbf{s}}^{+(l)}_{0}$\}$^N_{l=1}$ distributed according to the initial state density $p(\mathbf{s_0})$;
	
	\For{$k=1\  to\  K$}{
		Predict \{$\hat{\mathbf{s}}^{-(l)}_{k}$\}$^N_{l=1}$ according to $\hat{\mathbf{s}}^{-(l)}_{k}$\ $\sim p(\mathbf{s}_k$$|\hat{\mathbf{s}}^{+(l)}_{k-1})$, for $l=1,...,N$ using \eqref{PPFeqn38}-\eqref{PPFeqn41}\;
		Compute importance weights according to $w^{-(l)}_{k}=\frac{p(\mathbf{y_k}|\hat{\mathbf{s}}^{-(l)}_{k})}{\sum_{l=1}^{N} p(\mathbf{y_k}|\hat{\mathbf{s}}^{-(l)}_{k})}$, for $l=1,...,N$ using \eqref{PPFeqn42}\;
		Compute the estimates as $\hat{\mathbf{s}}^+_{k}=\frac{1}{N}\sum_{l=1}^{N}w^{-(l)}_{k} \hat{\mathbf{s}}^{-(l)}_{k}$\;
		Resample -eps-converted-to.pdfs as \{$\hat{\mathbf{s}}^{+(l)}_{k}$\}$^N_{l=1}$\; 
	}
	\caption{Proposed SMC based state estimation algorithm}
	\label{algo1}
\end{algorithm}
\section{Simulation Results}\label{sec_results}\hspace{.5cm}
To evaluate the performance of the proposed method for robust state estimation, we carry out a number of  simulations in two  settings. First, we consider a uni-dimensional example to demonstrate the effectiveness of the proposed method in successfully dealing with both types of measurement abnormalities: outliers and biases.  Subsequently, we consider a more practical example of a dynamic target tracking where the occurrence of different types of measurement abnormalities is not so uncommon.  In both cases, the performance of the proposed method is compared with two different filters: the standard bootstrap PF with no changes in the standard SSM and a hypothetical/ideal PF that has perfect knowledge about the exact instances and statistics of the outliers and biases.
\vfill
\subsection{Case 1}\hspace{.5cm}
For the first case, we consider the nonlinear model  given as \cite{8869858}
\begin{align}
	x_k&=1+\sin(0.04\pi k)+0.5x_{k-1}+q_{k-1}\label{PPFeqn43}\\
	y_k&=0.2x_k^2+r_k+o_k \label{PPFeqn44}
\end{align}
where the nominal process noise follows a Gamma distribution as $q_{k-1}\sim\Gamma(3,2)$, the nominal measurement noise obeys a Gaussian distribution as $r_k\sim\mathcal{N}(0,5)$ and the term $o_k$ dictates the type, duration and amount of the contamination in the measurements. 

For this case, we sequentially add drift, bias and outliers in the measurements by making $o_k$ a piece-wise function given as 
\begin{align*}
	\hspace{3cm}
	o_k &= \begin{cases}
		50+0.25(k-100) \hspace{.3cm}&\forall\hspace{.5cm}  100\leq k \leq 500 \\
		150 \hspace{2.62cm}&\forall\hspace{.5cm} 600\leq k \leq 800 \\
		{\Xi}_k\ z_k \hspace{2.25cm}&\forall\hspace{.5cm} 900\leq k \leq 1300 \\
		0 \hspace{4cm} &\text{otherwise} 
	\end{cases} \\
	\text{ with } {\Xi}_k &\sim \text{Bernoulli}\left(0.4 \right)  \\
	z_k &\sim \mathcal{N}\left(0, 50^2\times5 \right)
\end{align*}

We carry out the simulations for $L=100$ independent MC runs and evaluate the TRMSE of the state estimate for the three methods, where
\begin{align}
	\hspace{-2cm}
	\text{TRMSE}=\frac{1}{K}\sum_{k=1}^{K}\sqrt{\frac{1}{L}\sum_{i=1}^{L}({x_k}^i-{\hat{x}_k}^i)^2}
	\label{PPFeqn45}
\end{align} 

We use systemic resampling in the implementation of Algorithm \ref{algo1}. During simulations, we set $K = 1500$ and $N=20000$ along with other different parameters as
\begin{flalign*}
	\hspace{5.2cm}
	\mathbf{\Upsilon}_k &= 0.01 &\\
	\mathbf{\Delta}_k &= 0.25 &\\
	c(1) &=-d(1) = 1000 &\\
	\mathbf{U}_k &= 500^2\times5 &\\
	p({x}_0) &= \mathcal{N}(0.1,\sqrt{2}) &\\
	p({\Theta}_0) &= \mathcal{N}(0,0.001) &\\
	p(\mathcal{J}_{0}) &=\sum_{\xi}\frac{1}{3}\delta(\mathbf{\mathcal{J}}^1_0-\xi)& 
\end{flalign*}

\begin{figure} [h!]
	\centering
	\includegraphics[scale=0.3]{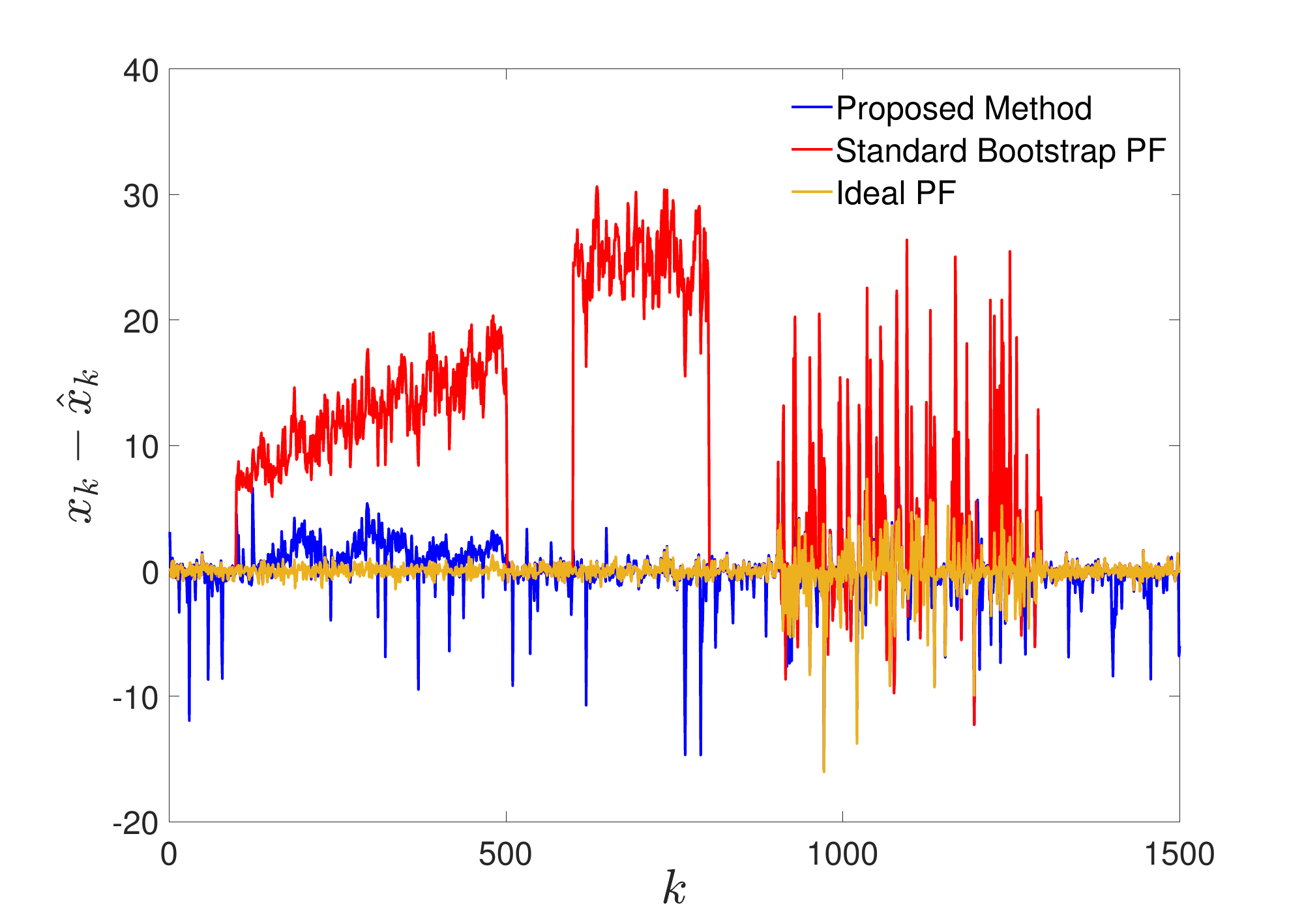}
	\caption{Instantaneous estimation error in $x_k$ in the presence of  all types of measurement abnormalities}
	\label{fig:sim_1}
\end{figure}

We first present the results of a single MC run in Figure \ref{fig:sim_1} where we plot  the instantaneous errors for each method.  The inability of the standard PF for tracking the state in presence of measurement abnormalities is evident. On the other hand, the proposed estimator is able to track the state in the presence of all types of measurement abnormalities. The ideal PF, of-course,  exhibits the least amount of error in the entire simulation run. 

The TRMSE results for all the methods are summarized in Table~\ref{tab:4}. In these simulations, we keep the same settings as described above and evaluate TRMSE for all MC runs according to \eqref{PPFeqn45}. It is clear from the  results  that the proposed method has lesser TRMSE as compared to the standard PF owing to its ability to detect and mitigate the effect of different measurement abnormalities. The ideal PF has the least TRMSE due to its knowledge of exact instances and statistics of abnormalities.

\begin {table}[h!]
\centering
\caption{Case 1: TRMSE for each method, $K = 1500$ and $N=20000$} \label{tab:4} 
\begin{tabular}{|p{3cm}|p{3cm}|p{3cm}|}
	\hline 
	Standard PF & Proposed PF & Ideal PF \\ 
	\hline 
	8.4246 & 2.21 & 0.93 \\ 
	\hline  
\end{tabular} 
\end{table}

\subsection{Case 2}\hspace{.5cm}
Now, we evaluate the performance of the proposed method when it is deployed to track a moving target  whose dynamics are simulated as \cite{7742936}
\begin{align}
	\underbrace{ \begin{bmatrix}
			{a_k} \\ {b_k} \\ \dot{a}_{k} \\ \dot{b}_{k}
	\end{bmatrix}}_{\mathbf{x}_{k}} &= \begin{bmatrix}
		1 & 0 & \triangle t & 0 \\
		1 & 1 & 0 & \triangle t \\
		0 & 0 & 1 & 0 \\
		0 & 0 & 0 & 1
	\end{bmatrix} \underbrace{ \begin{bmatrix}
			{a_{k-1}} \\ {b_{k-1}} \\ \dot{a}_{{k-1}} \\ \dot{b}_{{k-1}}
	\end{bmatrix}}_{\mathbf{x}_{k-1}} + \begin{bmatrix}
		\frac{\triangle t^2}{2} & 0 \\
		0 & \frac{\triangle t^2}{2} \\
		{\triangle t} & 0 \\
		0 & {\triangle t}
	\end{bmatrix} \mathbf{q}_{k-1} \label{PPFeqn46}  \\
	\mathbf{y}_{k} &= \begin{bmatrix}
		\sqrt{{a_k}^2+{b_k}^2} \\
		\text{atan2}\left({b_k},{a_k}\right)
	\end{bmatrix} +\mathbf{r}_{k}+\mathbf{o}_{k}\label{PPFeqn47}
\end{align} 

where the target state vector $\mathbf{x}_k$ contains  position coordinates $\left({a_k}, {b_k} \right)$ in 2-dimensions, and the corresponding velocities $\left(\dot{a}_{k}, \dot{b}_{k}\right)$, and $\triangle t$ is the sample time. The measurement vector contains the range and bearing information at each time step $k$. The vector $\mathbf{o}_{k}$ controls the amount and type of contamination. In these simulations, we use $K = 800$, $L = 100$, $N=30000$ along with other parameters  as 
\begin{align*}
	\mathbf{Q}_{k-1} &= \mathbf{I} \\
	\mathbf{R}_{k} &= \text{diag}(8, 0.002) \\
	\mathbf{\Upsilon}_k &= \text{diag}(4,0.001) \\
	\mathbf{\Delta}_k &= \text{diag}(0.01, 10^{-8}) \\
	\mathbf{U}_k &= 500 \mathbf{R}_{k} \\
	c(1) &=-d(1) = 1000\\
	c(2) &=-d(2) = \pi \\
	p(\mathbf{x}_0) &= \mathcal{N}\left([80,5,0,5]^{\top},\text{diag}(25,25,1,1) \right) \\ 
	p(\mathbf{\Theta}_0) &= \mathcal{N}\left([0,0]^{\top},\text{diag}(0.001,0.001) \right) \\
	p(\bm{{\mathcal{J}}}_{0}) &= \prod_{i=1}^{2}\sum_{\xi}\frac{1}{3}\delta(\mathbf{\mathcal{J}}_0(i)-\xi) \\
	\triangle t &= 1s 
\end{align*}

Using these settings, we carry out simulations in different scenarios  to evaluate the performance of the proposed method.

\subsection{Scenario I - Outliers} \hspace{.5cm} First, we consider the scenario where only outliers are present in sensor measurements.  To this end, the measurements are contaminated using samples drawn from $\mathbf{o}_{k}$ where 
\begin{align*}
	\mathbf{o}_{k} &={\Xi}_k\ \mathbf{z}_k &\\
	\text{ with } {\Xi}_k &\sim \text{Bernoulli}\left(\lambda  \right)  &\\
	\mathbf{z}_k &\sim \mathcal{N}\left(\mathbf{0}, \gamma \mathbf{R}_k \right)&
\end{align*}

The variables $\lambda$ and $\gamma$ control the frequency and the covariance of the outliers respectively. We carry out these simulations for $L=100$ independent MC runs and again evaluate the TRMSE  of the position estimate for each of the three methods. The TRMSE of the position estimate is defined as
\begin{align}\hspace{.8cm}
	\text{TRMSE}_\text{p}=\frac{1}{K}\sum_{k=1}^{K}\sqrt{\frac{1}{L}\sum_{i=1}^{L}({a_k}^i-\hat{a}_{k}^i)^2+({b_k}^i-\hat{b}_{k}^i)^2}
	\label{PPFeqn48}
\end{align} 
\begin{figure} [ht]
	\centering
	\includegraphics[scale=0.3]{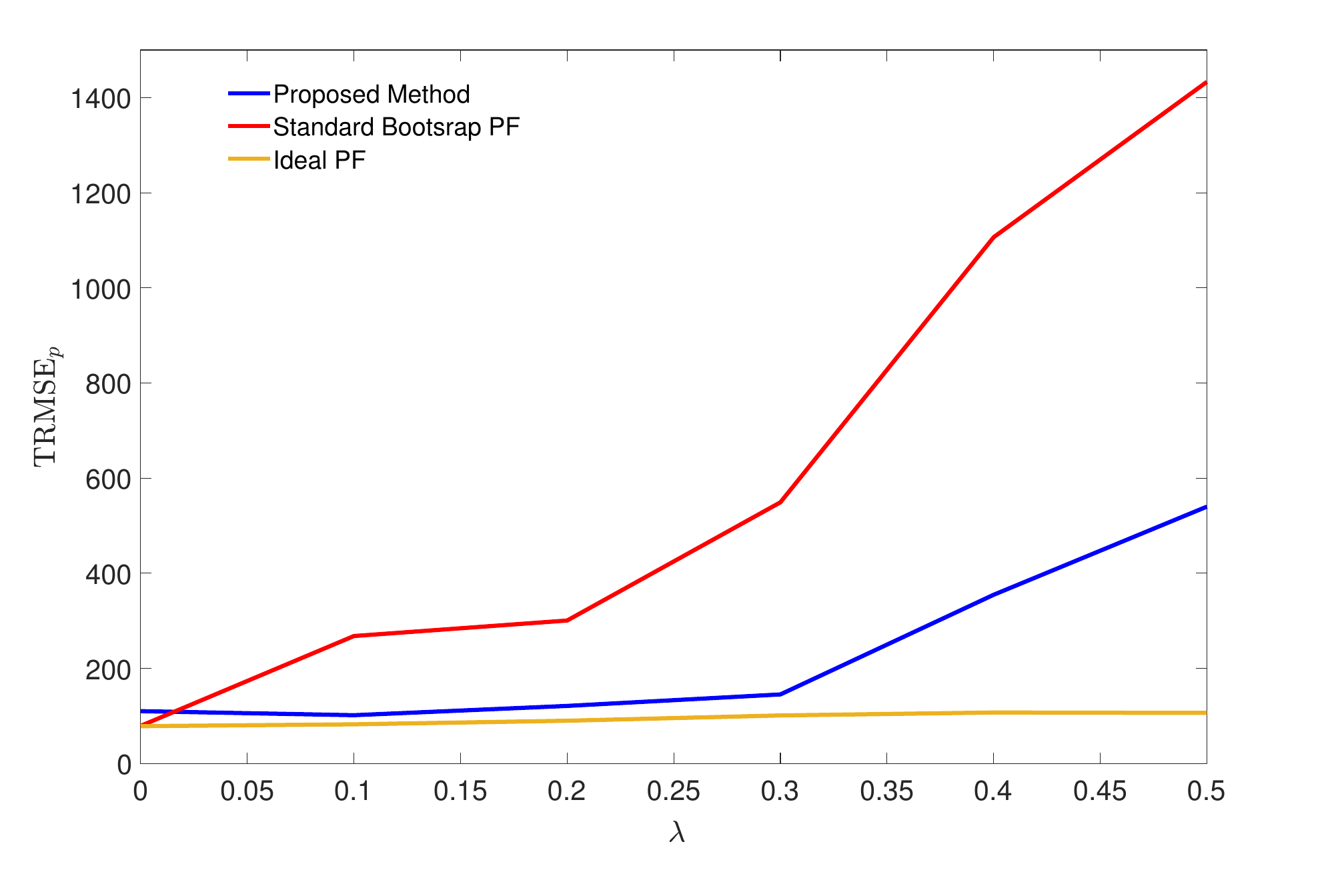}
	\caption{TRMSE of position with an increase in contamination level of measurement outliers}
	\label{fig:spo}
\end{figure}

Figure~\ref{fig:spo} shows the TRMSE results of three estimators in the presence of outliers  with varying degrees of contamination frequency specified by $\lambda$ with $\gamma=100$. It is observed that the standard bootstrap PF diverges in the presence of outliers. This can be attributed to the fact that it uses the measurements for inference even if these are corrupted. The proposed  scheme, on the other hand, performs much better even though it has no exact information about the actual statistics of heavy-tailed noise. Assuming very high variances for the outliers in the estimator model has the effect of discarding the corrupted measurements. Moreover, the use of {multivariate Bernoulli random vectors} within the model enables detection of corrupted measurements. The ideal PF exhibits the least amount of TRMSE understandably as it has perfect knowledge of the outlier statistics. 

\subsection{Scenario II - Biases} 
\hspace{.5cm}Now, we consider the scenario where the measurements are corrupted by sensor biases. For plotting the results, we  perform $L=100$ independent MC runs. For each run, sensor  biases are introduced in both measurements with values 200 and $\pi/3$ respectively, at $k=200$. The duration for which bias appears  is gradually increased from 0  to $0.5K$ i.e. $\mathbf{o}_{k}= [200,\pi/3]^{\top}\ \forall\  200\leq k \leq 200 + t_b K$ where $t_b$ is increased from 0 to 0.5. 
\begin{figure} [ht!]
	\centering
	\includegraphics[scale=0.27]{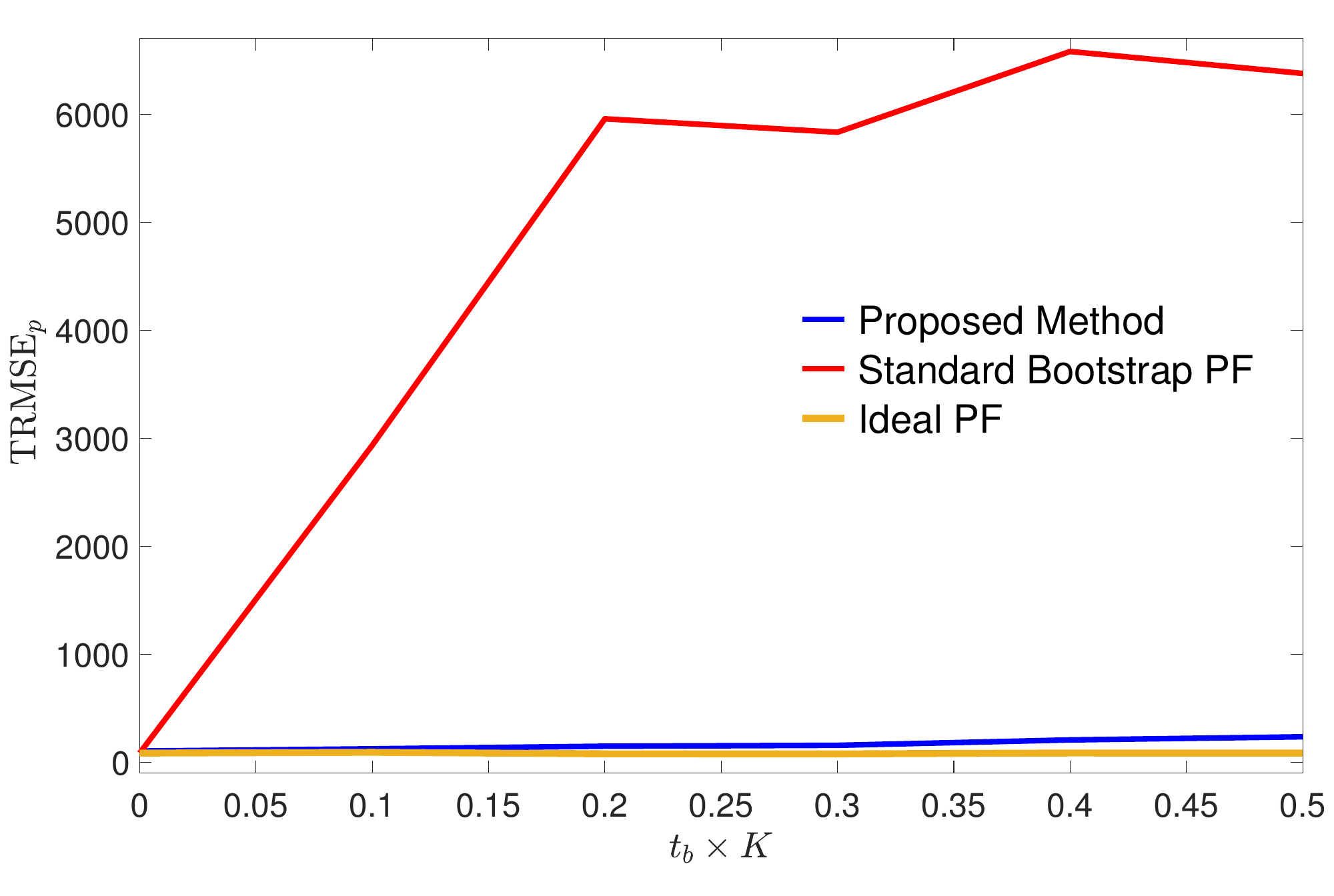}
	\caption{TRMSE of position with increase in bias duration}
	\label{fig:bias}
\end{figure}
Figure \ref{fig:bias} depicts the variation of TRMSE of position  as the duration of the applied bias, for all three estimators. It can be observed that the standard bootstrap PF again diverges in the presence of sustained biases in the measurements since it has no mechanism for bias detection and compensation. On the other hand, the proposed PF is able to detect as well as compensate for the effect of biases and its performance is comparable to the ideal PF. 

\begin{figure} [ht!]
	\centering
	\includegraphics[scale=0.5,frame]{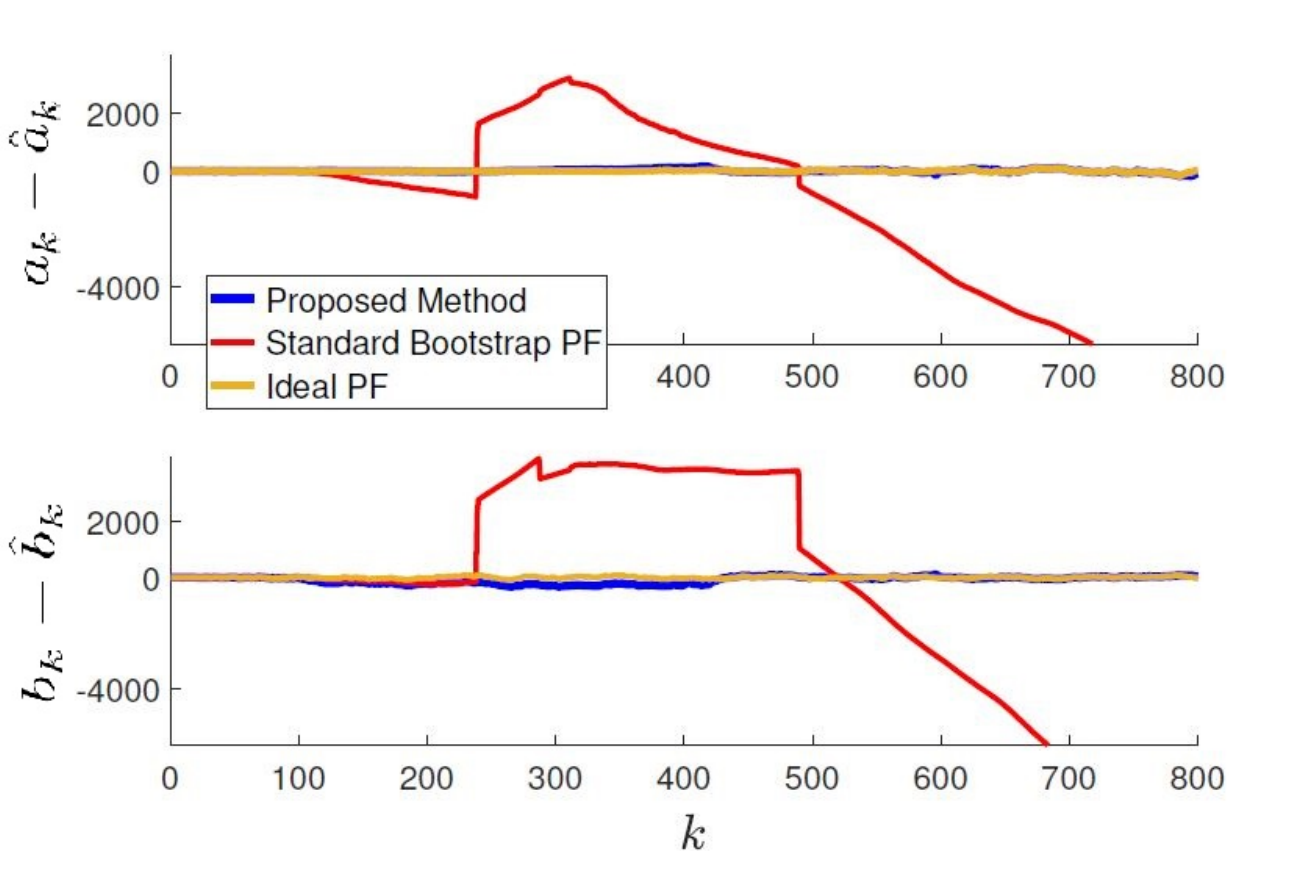}
	\caption{Instantaneous estimation error in $a_k$ and $b_k$ in the presence of both measurement outliers and biases}
	\label{fig:bands1}
\end{figure}
\subsection{Scenario III - Both Outliers and Biases} 
\hspace{.5cm} Lastly, we test the proposed algorithm in the presence of both outliers and biases that are sequentially introduced in the measurements. In this scenario, we introduce the bias for $k=100$ to $400$, in both dimensions of the measurement vector with values $400$ and $\pi/2$ respectively. Then, subsequently, we also introduce outliers in the measurements in the manner already explained, with $\gamma=150$ and $\lambda=0.4$ over the range  $k=500$ to $600$. For this scenario, we plot the instantaneous error in each position coordinate over time in Figure \ref{fig:bands1}. The estimation error in the position coordinates is largest in the bootstrap PF used on the standard model, followed by the proposed PF and the hypothetical PF. {This extreme scenario, with abnormal measurements in each dimension, shows how the proposed method provides robustness for estimation. In fact, as Figure \ref{fig:bands1}, the standard PF loses tracking of the object once the bias disappears, however, the proposed method keeps track of the target object.} Similar patterns of errors are also observed for the case where outliers and biases are introduced in a single dimension of the measurement vector.
\vfill

\section{Computational Complexity}\hspace{.5cm} Since the proposed method is based on the joint inference of the state and \textit{nuisance} parameter, the trade-off for the robustness is the computational complexity. As an example of Case 2 in the last scenario the standard and ideal PFs take around 5.2s to complete whereas the proposed SMC method requires 8.86s to execute the same simulation, using MATLAB on an Intel i7 based Windows platform. The exact computational cost generally depends on several factors such as the particular application, the system model, the dimensionality of the problem, the particular importance distribution and the number of -eps-converted-to.pdfs drawn for SMC methods, the number of parameter chosen to model the abnormalities etc. We have provided some general recommendations/remarks at the end of Section \ref{sec_model} to reduce the number of \textit{nuisance} parameters and consequently the computational cost associated with the proposed method.

\section{Experimental Results}\label{exp_results}\hspace{.5cm}
To further validate the performance of the proposed technique, we use experimental data of tracking a mobile object indoors, using UWB system \cite{web_loc,8320781}. In particular, UWB range measurements are used for the purpose that can provide high tracking accuracy but are known to suffer from different kinds of abnormalities especially those considered in this work i.e. measurement outliers and bias which is a typical characteristic of NLOS transmissions.

{In the  experimentation, the target whose 2D position is to be tracked, carries an UWB tag and traverses a single floor inside a building that is equipped with several UWB anchor nodes.  UWB nodes are based on an impulse radio UWB (IR-UWB) technology using a currently available low-cost transceiver, DecaWave DWM10001. DWM1000 is an IEEE 802.15.4-2011 IR-UWB compliant wireless transceiver module with an integrated ceramic antenna. The DWM1000 IR-UWB module is further combined with a STM32F103 microcontroller with 512 kB FLASH and 64 kB RAM memory. Each UWB node is further connected the RaspberryPi computer and are wirelessly connected to the measurement controller (e.g. laptop) using Wi-Fi and MQTT communication technologies.   The measurements are collected as measured ranges of the target from the anchor nodes. These measurements suffer from both types of abnormalities i.e. outliers as well as measurement biases \cite{8320781}.} \color{black}

For online filtering, we use the process model as defined in \eqref{PPFeqn46}. Moreover, we use single-channel range measurements from three anchor nodes installed at the following coordinates: $[4.11,4.25,1.45]^{\top}, [4.2,6.55,1.56]^{\top}, [9.4,7.31,1.7]^{\top}$. The measurement model equations are modified accordingly. Online filtering is repeated for $L=100$ independent MC runs with $K=84$ and $N=30000$. Other parameters used for state estimation are given as
{
	\begin{align*}
		\mathbf{Q}_{k-1} &= 0.005 \mathbf{I}, \mathbf{R}_{k} = \text{diag}(0.04, 0.04, 0.04) &\\
		\mathbf{\Upsilon}_k &= \text{diag}(0.01, 0.01, 0.01) &\\
		\mathbf{\Delta}_k &= \text{diag}(0.005, 0.005, 0.005), \mathbf{U}_k = 100 \mathbf{R}_{k} &\\
		c(1) &=-d(1) = c(2) =-d(2) =c(3) =-d(3) = 20 &\end{align*}
	\begin{align*}
		p(\mathbf{x}_0) &= \mathcal{N}([1.583,10.97,-0.016,-0.3]^{\top},\text{diag}(0.00125, 0.00125, 0.005, 0.005) ) &\\
		p(\mathbf{\Theta}_0) &= \mathcal{N}\left([0,0,0]^{\top},\text{diag}(0.01,0.01,0.01) \right) &\\
		p(\bm{\mathcal{J}}_{0}) &= \prod_{i=1}^{3}\sum_{\xi}\frac{1}{3}\delta(\mathbf{\mathcal{J}}_0(i)-\xi) &\\
		\triangle t &= 1s &
\end{align*}}

{Note that, 8 nodes were installed as localization anchor nodes with fixed locations in the indoor industrial environment and 1 node was used as a mobile localization tag.   All localization tag positions were generated beforehand to resemble the human walking path as closely as possible. All walking path points are equally spaced to represent the equidistant samples of a walking path in a time-domain. We consider only anchor 3 nodes for localization in our filter implementation.} \color{black}  Using the experimental data, we run the proposed filter and compare its performance with the standard Bootstrap filter in terms of TRMSE of target position. Note that since the ideal PF  requires the exact instances and statistics of measurement abnormalities, it is not possible to use that filter in experimental settings. 

\begin{figure} [ht!]
	\centering
	\includegraphics[scale=.5]{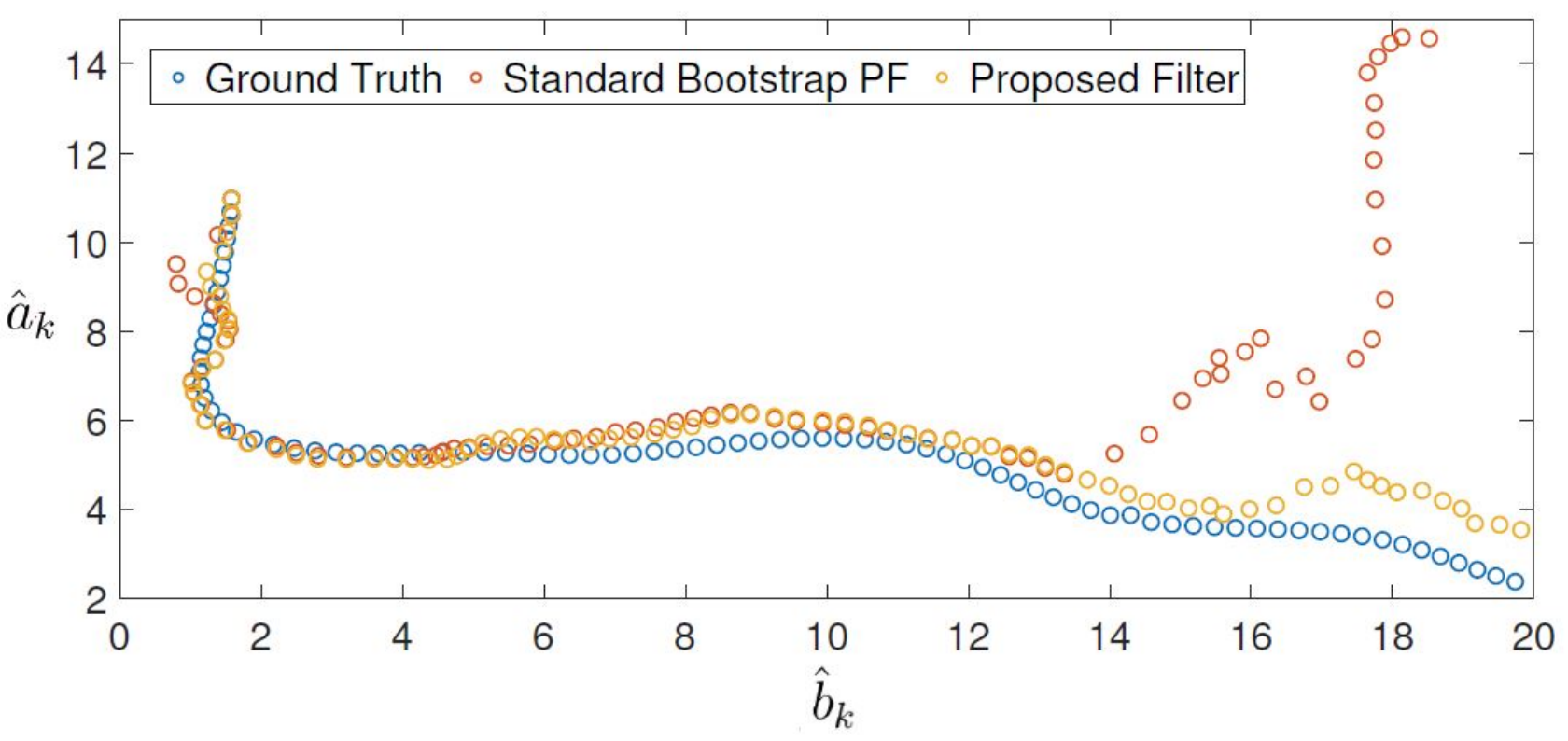}
	\caption{{Tracking performance in the experimental setting}}
	\label{fig:exp}
\end{figure}

{Figure \ref{fig:exp} shows the tracking performance of the proposed method and the standard PF along with the ground truth as the target navigates. The standard PF loses tracking of target's position when the NLOS measurements appear whereas the proposed method successfully mitigates the effect of measurement abnormalities during the filtering process. This demonstrates the usefulness of the proposed method in practical scenarios as well. Using the standard PF, we obtain the TRMSE of  position estimate equal to 2.11m. On the other hand, employing the proposed method we observe a much better tracking performance and we get the TRMSE of position estimate as 1.59m.}

\section{Conclusion}\label{sec_conc}
\hspace{.5cm}
In this chapter, we have presented a robust Bayesian method to provide accurate and online state estimates  in the presence of sensor measurement abnormalities. For this purpose, we explicitly model the possibility of these abnormalities in the standard discrete-time nonlinear state-space model, modifying the problem to a joint state and parameter estimation problem. An SMC based method has been proposed for joint inference of system state and parameters which is essentially a detect-and-reject approach for outliers and a learn-and-compensate based approach for biases. Simulations and evaluation on experimental datasets show that the proposed method can efficiently and seamlessly deal with both outliers as well as biases in an online fashion. Specifically, the proposed scheme significantly outperforms the standard bootstrap PF and can be useful for applications that are vulnerable to measurement data abnormalities especially for low-dimensionality models where the computational overhead of implementing an SMC method is not excessive.

\chapter{Conclusion, Applications and Future Directions}\hspace{.5cm}
In this dissertation, we have studied the topic of robust filtering where the measurements can be disturbed with different kinds of abnormalities. In particular, we have focused on the occurrence of gross errors namely outliers and biases in the data. In this last chapter of the dissertation we provide conclusive remarks for this work. We also highlight the applications where the algorithms we have devised can be useful. Lastly, we provide pointers for further extensions of this work.

 \label{concl} 
\section{Conclusion}
\hspace{.5cm}

In Chapters \ref{chap-3}~-~\ref{chap-6a}, we have dealt with the occurrence of outliers in the measurements during filtering. First, we approach the problem in a simplistic manner where we consider uni-dimensional data corruption devising a MAP-based EKF in Chapter \ref{chap-3}. The algorithm exhibits better performance as compared to some of the existing learning-based filters which for tractability reject the entire measurement vector on detection of outliers. However, due to the construction of the proposed filter it has scalability issues. To circumvent the problem we have proposed an outlier-robust filter namely the SOR filter using the VB theory in Chapter \ref{chap-4}. We suppose multi-dimensional data outliers and consider independent measurement noise across different channels. The algorithm has reduced computational complexity and similar performance in terms of error as compared to the similar learning-based methods. This approach is useful where we have independent measurement channels and have computational constraints. In Chapter \ref{chap-5}, we have extended the proposed VB-based robust filter to a general nonlinear estimation problem. In particular, we consider the context of robust spatial perception which is an important problem in modern intelligent systems. The proposed estimators namely ESOR and ASOR are evaluated for different perception problems where the computational gains are verified. In Chapter \ref{chap-6a}, we consider the generic setting for robust filtering where the measurement noise can be correlated across channels. We leverage the EM/VB  theory to devise a filter namely EMORF and its smoother extension called EMORS. To benchmark the performance of different outlier-rejecting estimators, we also present theoretical estimation limitations by evaluating BCRBs of a filter and smoother that perfectly detect and reject outliers. We find the propose approaches to be advantageous are in terms of simpler implementation, enhanced estimation quality, and competitive computational performance.  The results of this investigation reveal the advantages of leveraging the VB theory in the design of outlier-robust estimators.

In Chapter \ref{chap-6b}, we deal with the problem of biased observations in the filtering context. We propose a model that effectively represent the system dynamics and the measurements at the same time remaining amenable for inference. We use the VB theory and general Gaussian filtering to devise the BDM filter. As the detection mechanism is embedded within the filtering structure the dependence on any external detector is obviated. Performance evaluation reveals the merits of the BDM filter as compared to similar methods bench-marked with the PCRB-based estimation error bound. We find the performance of BDM filter better in terms of estimation error, in comparison to other methods, in the presence of both temporary and persistent bias presence.    

Lastly, in Chapter \ref{chap-7} we consider the possibility of the occurrence of both outliers and biases in the measurements. We propose an inference model and resort to stochastic sampling based PF as the basic inference algorithm for the proposed method. Though it can be computationally intensive, owing to the use of PF, it has acceptable performance and can be useful for low-dimensional problems.

\section{Applications}\hspace{.5cm}
Robust filtering has applications in many fields given the data from the sensors in dynamical systems/processes can easily get corrupted. State estimation in such scenarios remains crucial and find applications in the following domains
\begin{itemize}
	\item Vehicular technology.
	\item Machine intelligence.
	\item Robotics/Automation.
	\item Localization/Target tracking.
	\item Power system dynamic state estimation e.g. PMUs, protective relays.
	\item Healthcare/Biomedical. 
	\item Cyber-physical systems/Cybersecurity.
	\item Data science/Analytics.
\end{itemize}

\section{Future Directions}\hspace{.5cm}
We have explored the topic of robust filtering by considering the occurrence of outliers and biases in the measurements. We have identified several gaps in the literature that need further investigation and can be explored further. The work can possibly be extended in the following ways 

\begin{itemize}
	\item We have tackled the problem of the occurrence of outliers and biases based on stochastic sampling technique which can be computationally inhibiting in different applications. More computationally friendly algorithms e.g. VB-based methods can be devised in this setting by exploiting particular constraints of problems and other inference tools.
	\item There have been recent advances in the VB theory that can potentially be leveraged for more computational effectiveness of the proposed methods. For example, the concept of parallel updates of estimates during iterations can be exploited in this regard.  
	\item The proposed robust state estimators can be extended with the concept of certifiable performance. Along with evaluating the state, guarantees for the optimality of the estimates during data corruption are also provided. Moreover, we can indicate in case an estimator falters. We can possibly resort to existing methods from the literature of certifiable performance which can certify the estimates a posteriori.
	\item The proposed estimators ESOR and ASOR  can potentially be used in different inferential pipelines. We find that the error performance of these Bayesian heuristics is similar to the GNC methods and these are generally found to be faster in various scenarios of the point cloud registration problem. Therefore, these can potentially be useful in inferential pipelines like TEASER++ (while enjoying certifiable performance) but need further evaluation. We can also aim to extend the heuristics where the knowledge of nominal noise statistics is not available. 
	\item System identification in the presence of data corruption is a related problem. While it may be possible to add a preprocessing unit to filter data contamination like outliers, a holistic system can be devised where the aberrations in the observations are tackled within the inferential process. We believe that the concepts presented in this work can be useful in this direction also. For example, the devised iterative algorithms can be unfolded to devise efficient deep-learning architectures to cater to outliers in the inferential procedure.
	\item We have considered non-adversarial data corruption in our settings. The work can be expanded in the direction where the adversarial data corruption can be considered e.g. attacks on the physical layer of the cyber-physical systems.
\end{itemize}


\addtocontents{toc}{\vspace{2em}} 
\unnumberedchapter{Bibliography} 
\bibliography{Preamble/Thesis_bibliography} 


\addtocontents{toc}{\vspace{2em}} 
\appendix

\numberedchapter 
\chapter{} \label{app_VB}
\section{Predicting parameters of $\mathcal{N}(\mathbf{\Theta}_{k}|\mathbf{\hat{\Theta}}^-_{k},\mathbf{{\Sigma}}^-_{k})$}\label{bias_pred} 
	\begin{flalign}
	\hat{\mathbf{\Theta}}_{k}^{-} &=  \langle {\mathbf{\Theta}}_{k} \rangle_{p(\bm{{\Theta}}_{k}|\mathbf{y}_{1:{k-1}})}\nonumber\\
	\hat{\mathbf{\Theta}}_{k}^{-} &=  \langle (\mathbf{I}-\boldsymbol{{\mathcal{I}}}_{k-1}){\widetilde{\mathbf{\Theta}}}_{k}+ \boldsymbol{{\mathcal{I}}}_{k-1}({\mathbf{\Theta}}_{k-1} + {\Delta_k}\rangle_{\begin{subarray} {l}  p(\bm{\mathcal{I}}_{k-1}|\mathbf{y}_{1:{k-1}}).p(\widetilde{\bm{\Theta}}_k). \nonumber\\ p(\bm{{\Theta}}_{k-1}|\mathbf{y}_{1:{k-1}}).p(\Delta_{k}) 		\end{subarray}}\nonumber\\
	\hat{\mathbf{\Theta}}_{k}^{-} &= \boldsymbol{\Omega}_{k-1}\hat{\mathbf{\Theta}}_{k-1}^{+}&\nonumber\\
	\mathbf{\Sigma}_{k}^{-} &= \langle(\mathbf{\Theta}_{k}-\hat{\mathbf{\Theta}}_{k}^{-})(\mathbf{\Theta}_{k}-\hat{\mathbf{\Theta}}_{k}^{-})^{\top} \rangle_{p(\bm{{\Theta}}_{k}|\mathbf{y}_{1:{k-1}})}&\nonumber\\
	\textcolor{black}{\mathbf{\Sigma}_{k}^{-}} &=\langle \mathbf{v}_k^1 {\mathbf{v}_k^1}^{\top} \rangle_{\begin{subarray} {l}  p(\bm{\mathcal{I}}_{k-1}|\mathbf{y}_{1:{k-1}}).p(\widetilde{\bm{\Theta}}_k).  p(\bm{{\Theta}}_{k-1}|\mathbf{y}_{1:{k-1}}). 
			p({\Delta_k})	\end{subarray}}\nonumber&
\end{flalign}

{Since $\langle \widetilde{\mathbf{\Theta}}_{k}\rangle_{p(\widetilde{\bm{\Theta}}_k)}=\langle {\Delta_k} \rangle_{p({\Delta_k})}=\langle \mathbf{\Theta}_{k-1}-\hat{\mathbf{\Theta}}_{k-1}^{+} \rangle_{p(\bm{{\Theta}}_{k-1}|\mathbf{y}_{1:{k-1}})}=\mathbf{0}$, other terms in the expression of ${\mathbf{\Sigma}_{k}^{-}}$ disappear and we can write}
\begin{flalign}
{\mathbf{\Sigma}_{k}^{-}}&{={\langle} (\mathbf{I}-\boldsymbol{\mathcal{I}}_{k-1})\widetilde{\mathbf{\Theta}}_{k}\widetilde{\mathbf{\Theta}}_{k}^{\top}(\mathbf{I}-\boldsymbol{\mathcal{I}}_{k-1})^{\top} + \boldsymbol{\mathcal{I}}_{k-1}{\Delta_k} {\Delta^{\top}_k}\boldsymbol{\mathcal{I}}_{k-1}^{\top}}  \nonumber \\ 
&{+ \boldsymbol{\mathcal{I}}_{k-1}(\mathbf{\Theta}_{k-1} - \hat {\mathbf{\Theta}}_{k-1}^{+})(\mathbf{\Theta}_{k-1} -{{}\hat{\mathbf{\Theta}}_{k-1}^{+}})^{\top}\boldsymbol{\mathcal{I}}_{k-1}^{\top}} & \nonumber \\
& {+ (\boldsymbol{\mathcal{I}}_{k-1} - \boldsymbol{\Omega }_{k-1})\hat{\mathbf{\Theta}}_{k-1}^{+} {{}\hat{\mathbf{\Theta}}_{k-1}^{+}}^{\top}(\boldsymbol{\mathcal{I}}_{k-1} - \boldsymbol{\Omega}_{k-1})^{\top} {\rangle_{\begin{subarray} {l}  p(\bm{\mathcal{I}}_{k-1}|\mathbf{y}_{1:{k-1}}).  p(\bm{{\Theta}}_{k-1}|\mathbf{y}_{1:{k-1}}).p(\widetilde{\bm{\Theta}}_k). 
				p({\Delta_k})	\end{subarray}} }} \nonumber& 
\end{flalign}

{We can further write}
\begin{flalign}
{\mathbf{\Sigma}_{k}^{-}}&{=\langle \mathbf{v}_k^2 {\mathbf{v}_k^2}^{\top}+\mathbf{v}_k^3 {\mathbf{v}_k^3}^{\top}+\mathbf{v}_k^4 {\mathbf{v}_k^4}^{\top}+\mathbf{v}_k^5 {\mathbf{v}_k^5}^{\top} \rangle_{\begin{subarray} {l}  p(\bm{\mathcal{I}}_{k-1}|\mathbf{y}_{1:{k-1}}). p(\bm{{\Theta}}_{k-1}|\mathbf{y}_{1:{k-1}}).p(\widetilde{\bm{\Theta}}_k). 
			p({\Delta_k})	\end{subarray}}}  \nonumber&
\end{flalign}
with \\
$\mathbf{v}_k^2=\begin{pmatrix}
	(1-{ \mathcal{I}}_{k-1}(1,1) )\widetilde{\Theta} _{k}(1) \\
	\vdots\\
	(1-{\mathcal{I}}_{k-1}(m,m))\widetilde{\Theta} _{k}(m)
\end{pmatrix}\\
\mathbf{v}_k^3=\begin{pmatrix}
	{\mathcal{I}}_{k-1}(1,1) {\Delta_k(1)} \\
	\vdots\\
	{\mathcal{I}}_{k-1}(m,m){\Delta_k(m)} 
\end{pmatrix}$\\
$\mathbf{v}_k^4=\begin{pmatrix}
	{\mathcal{I}}_{k-1}(1,1)(\Theta_{k-1}(1) - {{}\hat{\Theta}_{k-1}^{+}}(1)) \\
	\vdots\\
	\mathcal{I}_{k-1}(m,m)({\Theta_{k-1}}(m) - {{}\hat{\Theta}_{k-1}^{+}}(m))
\end{pmatrix}\\
\mathbf{v}_k^5=\begin{pmatrix}
	(\mathcal{I}_{k-1}(1,1) - \Omega_{k-1}(1,1) ) {{}\hat{\Theta}_{k-1}^{+}}(1) \\
	\vdots\\
	(\mathcal{I}_{k-1}(m,m) - \Omega_{k-1}(m,m)) {{}\hat{\Theta}_{k-1}^{+}}(m)
\end{pmatrix}$

Resultingly $\mathbf{\Sigma}_{k}^{-}$ can be further expressed as
\begin{align}
	&\mathbf{\Sigma}_{k}^{-}\text{=} \langle 
	\begin{pmatrix}
		{(1-\mathcal{I}_{k-1}(1,1))^{2}{{} \widetilde{\Theta} _{k}}^{2}(1) } & \cdots & \prod_{i} (1-\mathcal{I}_{k-1}(i,i))\widetilde{\Theta} _{k}(i)\\
		\vdots & \ddots & \vdots \\
		. & \cdots & (1-\mathcal{I}_{k-1}(m,m))^{2}{{}\widetilde{\Theta} _{k}}^{2}(m)
	\end{pmatrix}
	\rangle_{{\begin{subarray}{l}		p(\bm{\mathcal{I}}_{k-1}|\mathbf{y}_{1:{k-1}}). p(\bm{\widetilde{\Theta}}_{k})  
			\end{subarray}
	}} & \nonumber \\
	& \text{+} \langle 
	\begin{pmatrix}
		{\mathcal{I}_{k-1}}^{2}(1,1) {\Delta_k^2(1)} & \cdots & \prod_{i} {\mathcal{I}_{k-1}}(i,i) {\Delta_k}(i) \\
		\vdots & \ddots & \vdots \\
		. & \cdots & {\mathcal{I}_{k-1}(m,m)}^2 {\Delta_k^2(m)}
	\end{pmatrix}
	\rangle_{{\begin{subarray}{l}		p(\bm{\mathcal{I}}_{k-1}|\mathbf{y}_{1:{k-1}}). p(\Delta_k)  
			\end{subarray}
	}} & \nonumber \\
	&\text{+} \langle 
	\begin{pmatrix}
		{{\mathcal{I}_{k-1}}^2(1,1)}(\Theta_{k-1}(1) - {{} \hat{\Theta}_{k-1}^{\text{+}}(1) } )^{2} & \cdots & \prod_{i} {\mathcal{I}_{k-1}(i,i)}(\Theta_{k-1}({i}) - {{}\hat{\Theta}_{k-1}^{\text{+}}(i)}) \nonumber \\
		\vdots & \ddots & \vdots \\
		. & \cdots & {\mathcal{I}_{k-1}}^2(m,m)(\Theta_{k-1}(m) - {{}\hat{\Theta}_{k-1}^{\text{+}}}(m) )^{2}
	\end{pmatrix}
	\rangle_{{\begin{subarray}{l}		p(\bm{\mathcal{I}}_{k-1}|\mathbf{y}_{1:{k-1}}).\nonumber \\p(\bm{{\Theta}}_{k-1}|\mathbf{y}_{1:{k-1}})  
			\end{subarray}
	}} &\nonumber\\
	&\text{+}\langle 
	\begin{pmatrix}
		{(\mathcal{I}_{k-1}(1,1) - \Omega _{k-1}(1,1) )}^{2}  {{{}\hat {\Theta}_{k-1}^{\text{+}}}}^{2}(1) & {\cdots} & \prod_{i} {(\mathcal{I}_{k-1}^{i} - \Omega _{k-1}(i,i) )}  {{{}\hat {\Theta}_{k-1}^{\text{+}}(i) }} \\
		\vdots & {\ddots} & \vdots \\
		. & {\cdots} & {(\mathcal{I}_{k-1}(m,m) - \Omega _{k-1}(m,m))}^{2}  {{{}\hat {\Theta}_{k-1}^{\text{+}}}}^{2}(m)
	\end{pmatrix}
	\rangle_{{\begin{subarray}{l}		p(\bm{\mathcal{I}}_{k-1}|\mathbf{y}_{1:{k-1}})\end{subarray}
	}}\nonumber&
\end{align}
where ${i\in\{1,m\}}$ 
\begin{align*}
	{=}&\mathrm{diag}	\begin{pmatrix}\langle (1-\mathcal{I}_{k-1}(1,1))^{2}\rangle_{p(\bm{\mathcal{I}}_{k-1}|\mathbf{y}_{1:{k-1}})}\\
		\vdots\\\langle(1-\mathcal{I}_{k-1}(m,m))^{2}\rangle_{p(\bm{\mathcal{I}}_{k-1}|\mathbf{y}_{1:{k-1}})}
	\end{pmatrix} \mathrm{diag}	\begin{pmatrix}\langle { {}\widetilde{\Theta} _{k}}^{2}(1) \rangle_{p(\bm{\widetilde{\Theta}}_{k})} \nonumber\\
		\vdots\\\langle{ {}\widetilde{\Theta} _{k}}^{2}(m) \rangle_{p(\bm{\widetilde{\Theta}}_{k})} \end{pmatrix} & \nonumber\\
	&{+}\mathrm{diag}	\begin{pmatrix}\langle {\mathcal{I}_{k-1}}^{2}(1,1)\rangle_{p(\bm{\mathcal{I}}_{k-1}|\mathbf{y}_{1:{k-1}})}\\
		\vdots\\\langle{{\mathcal{I}_{k-1}}^2(m,m)}\rangle_{p(\bm{\mathcal{I}}_{k-1}|\mathbf{y}_{1:{k-1}})}
	\end{pmatrix} \mathrm{diag}	\begin{pmatrix}\langle {\Delta_k^2(1)}  \rangle_{p(\Delta_{k})} \\
		\vdots\\ \langle {\Delta_k^2(m)} \rangle_{p(\Delta_{k})}
	\end{pmatrix} \nonumber\\
	&{+}\mathrm{diag}	\begin{pmatrix}\langle {(\mathcal{I}_{k-1}(1,1) - \Omega _{k-1}(1,1))^2}\rangle_{p(\bm{\mathcal{I}}_{k-1}|\mathbf{y}_{1:{k-1}})}\\
		\vdots\\\langle {(\mathcal{I}_{k-1}(m,m) - \Omega _{k-1}(m,m) )^2 }\rangle_{p(\bm{\mathcal{I}}_{k-1}|\mathbf{y}_{1:{k-1}})}
	\end{pmatrix} \mathrm{diag}	\begin{pmatrix} {{{}\hat {\Theta}_{k-1}^{+}}^{2} }(1)\nonumber \\
		\vdots\\ {{{}\hat {\Theta}_{k-1}^{+}}^2(m) } 
	\end{pmatrix} \\&+\mathbf{A}_{k-1}&	
\end{align*}
where
\begin{align}
	\langle (1-\mathcal{I}_{k-1}(i,i))^{2}\rangle_{p(\bm{\mathcal{I}}_{k-1}|\mathbf{y}_{1:{k-1}})}&=(1-\Omega^{+}_{k-1}(i,i))&\nonumber\\
	\langle { {}{\widetilde{\Theta} _{k}}}^{2}(i,i) \rangle_{p(\bm{\widetilde{\Theta}}_{k})} &=\widetilde{{\Sigma}}_{k}(i,i)&\nonumber\\
	\langle { \mathcal{I}_{k-1}}^{2}(i,i) \rangle_{p(\bm{\mathcal{I}}_{k-1}|\mathbf{y}_{1:{k-1}})}&=\Omega^{+}_{k-1}(i,i)&\nonumber\\
	\langle {\Delta_k}^2(i) \rangle_{p(\Delta_{k})}&=\breve{{\Sigma}}_{k}(i,i)&\nonumber
\end{align}
\begin{align}
	{A}_{k-1}(i,j)&=\langle 
	{\mathcal{I}_{k-1}(i,i)}{\mathcal{I}_{k-1}}(j,j)(\Theta_{k-1}(i) {-} {{}\hat{\Theta}_{k-1}^{\text{+}}(i)})(\Theta_{k-1}(j) {-} {{}\hat{\Theta}_{k-1}^{\text{+}}}(j))\rangle_{{\begin{subarray}{l}		p(\bm{\mathcal{I}}_{k-1}|\mathbf{y}_{1:{k-1}}).  \\ p(\bm{{\Theta}}_{k-1}|\mathbf{y}_{1:{k-1}})  
			\end{subarray}
	}}\nonumber&
\end{align}
for $i=j$
\begin{flalign}
	{A}_{k-1}(i,j)&=\langle 
	{ {\mathcal{I}_{k-1}}^2 (i,i)} \rangle_{{\begin{subarray}{l}		p(\bm{\mathcal{I}}_{k-1}|\mathbf{y}_{1:{k-1}})
			\end{subarray}
	}} \langle (\Theta_{k-1}(i) {-} {{}\hat{\Theta}_{k-1}^{\text{+}}(i)})^2\rangle_{{\begin{subarray}{l}		p(\bm{{\Theta}}_{k-1}|\mathbf{y}_{1:{k-1}})  
			\end{subarray}
	}} \nonumber \\ &=\Omega^{+}_{k-1}(i,i) \Sigma^{+}_{k-1}(i,i)\nonumber&
\end{flalign}
for $i \neq j$
\begin{flalign}
	{A}_{k-1}(i,j)&=\big\{ \langle 
	{\mathcal{I}_{k-1}(i,i)}\rangle_{{\begin{subarray}{l}		p(\bm{\mathcal{I}}_{k-1}|\mathbf{y}_{1:{k-1}})
			\end{subarray}
	}} \langle {\mathcal{I}_{k-1}}(j,j)\rangle_{{\begin{subarray}{l}	p(\bm{\mathcal{I}}_{k-1}|\mathbf{y}_{1:{k-1}})
			\end{subarray}
	}} \nonumber \\ 
	&\quad \ \langle (\Theta_{k-1}^{i} {-} {{}\hat{\Theta}_{k-1}^{\text{+}}(i)})(\Theta_{k-1}(j) {-} {{}\hat{\Theta}_{k-1}^{\text{+}}}(j) )\rangle_{{\begin{subarray}{l}		p(\bm{{\Theta}}_{k-1}|\mathbf{y}_{1:{k-1}})  
			\end{subarray}
	}}	\big\} &\nonumber \\
	&=\Omega^{+}_{k-1}(i,i) \Omega^{+}_{k-1}(j,j) \Sigma^{+}_{k-1}(i,j)\nonumber&
\end{flalign}
$\therefore \mathbf{A}_{k-1}$ can be written as
\begin{flalign}
	\mathbf{A}_{k-1}&=\mathbf{{\Sigma}}^+_{k-1}\odot(\mathrm{diag}(\mathbf{\Omega}_{k-1}){\mathrm{diag}(\mathbf{\Omega}_{k-1})}^{\top}+\mathbf{\Omega}_{k-1}(\mathbf{I}-\mathbf{\Omega}_{k-1}))&	\nonumber\\
	\mathbf{\Sigma}^{-}_{k}&=(\mathbf{I}-\mathbf{\Omega}_{k-1})\widetilde{\mathbf{\Sigma}}_{k}+\mathbf{\Omega}_{k-1}\breve{\mathbf{\Sigma}}_{k}+\mathbf{{\Sigma}}^+_{k-1}\odot(\mathrm{diag}(\mathbf{\Omega}_{k-1}){\mathrm{diag}(\mathbf{\Omega}_{k-1})}^{\top}+\mathbf{\Omega}_{k-1}(\mathbf{I}-\mathbf{\Omega}_{k-1}))&\nonumber \\
	&\ \ \ \ +\mathbf{\Omega}_{k-1}(\mathbf{I}-\mathbf{\Omega}_{k-1})(\mathrm{diag}(\mathbf{\hat{\Theta}}^{+}_{k-1}))^2 \nonumber &	
\end{flalign}

\section{Derivation of $q(\mathbf{x}_{k})$}\label{bias_up1}
\hspace{.5cm}
Using \eqref{eqn_vb_1}, \eqref{eqn_vb_3} and \eqref{eqn_vb_8} we can write $q(\mathbf{x}_k)$ as
\begin{flalign}
	q(\mathbf{x}_k)&\propto \exp\big( \big\langle\mathrm{ln}(p(\mathbf{y}_k|\mathbf{x}_{k},\bm{\mathcal{I}}_k,{\mathbf{\Theta}}_k) p(\bm{\mathcal{I}}_k)p(\mathbf{x}_k|\mathbf{y}_{1:k-1}) p({\mathbf{\Theta}}_k|\mathbf{y}_{1:k-1})) \rangle_{ q({{\bm{\mathcal{I}}}_k}) {q(\mathbf{\Theta}}_k)}\big)& \nonumber\\
	&\propto \exp\big( \big\langle\mathrm{ln}(p(\mathbf{y}_k|\mathbf{x}_{k},\bm{\mathcal{I}}_k,{\mathbf{\Theta}}_k)\rangle_{ q({{\bm{\mathcal{I}}}_k}) {q(\mathbf{\Theta}}_k)}\big) p(\mathbf{x}_k|\mathbf{y}_{1:k-1}) & \nonumber\\
	&\propto \exp \big( \langle \mathrm{ln}(\mathcal{N}(\mathbf{y}_{k}|\mathbf{h}(\mathbf{x}_{k}) + \boldsymbol{\mathcal{I}}_{k}\mathbf{\Theta}_{k},\mathbf{R}_{k})\rangle_{ q({{\bm{\mathcal{I}}}_k}) {q(\mathbf{\Theta}}_k)}\big)\mathcal{N}(\mathbf{x}_{k}|\hat {\mathbf{x}}_{k}^{-},\mathbf{P}_{k}^{-}) \nonumber&
	\\
	&\propto \exp ( \langle (-\frac{1}{2}{\mathbf{v}_k^6}^{\top} 
	\mathbf{R}_{k}^{-1} \mathbf{v}_k^6 ) -\frac{1}{2} \mathrm{ln}(2 \pi) ^{m} | \mathbf{R}_{k}|)\rangle_{ q({{\bm{\mathcal{I}}}_k}) {q(\mathbf{\Theta}}_k)}\big)  \mathcal{N}(\mathbf{x}_{k}|\hat {\mathbf{x}}_{k}^{-},\mathbf{P}_{k}^{-})& \nonumber\\
	&\propto \exp \big(-\frac{1}{2} \langle \mathrm{tr} (\mathbf{v}_k^6{\mathbf{v}_k^6}^{\top} 
	\mathbf{R}_{k}^{-1} )\rangle_{ q({{\bm{\mathcal{I}}}_k}) {q(\mathbf{\Theta}}_k)}\big) \mathcal{N}(\mathbf{x}_{k}|\hat {\mathbf{x}}_{k}^{-},\mathbf{P}_{k}^{-})& \nonumber
\end{flalign}
where $\mathbf{v}_k^6=\mathbf{y}_{k}-\mathbf{h}(\mathbf{x}_{k})  - \boldsymbol{\mathcal{I}}_{k}\mathbf{\Theta}_{k}$. 

Furthermore, we can write
\begin{flalign}
	& q(\mathbf{x}_k)\propto
	\exp\big({-}\frac{1}{2} \mathrm{tr}\big((\mathbf{v}_k^7{\mathbf{v}_k^7}^{\top}- \mathbf{v}_k^7  {\mathbf{v}_k^8}^{\top}  - \mathbf{v}_k^8 {\mathbf{v}_k^7}^{\top} ) \mathbf{R}_{k}^{-1}\big)\big)  \mathcal{N}(\mathbf{x}_{k}|\hat {\mathbf{x}}_{k}^{-},\mathbf{P}_{k}^{-})\nonumber&
\end{flalign}
where $\mathbf{v}_k^7=\mathbf{y}_{k}-\mathbf{h}(\mathbf{x}_{k})$ and $\mathbf{v}_k^8=\mathbf{\Omega}_{k} {{}\hat{\mathbf{\Theta}}_{k}^{+}}$. Adding a constant term $\mathbf{v}_k^8 {\mathbf{v}_k^8}^{\top}$ to complete the square in the exponential expression yields
\begin{flalign}
	&q(\mathbf{x}_k)\propto \exp \big(-\frac{1}{2}   \mathrm{tr}(\mathbf{v}_k^9{\mathbf{v}_k^9}^{\top} 
	\mathbf{R}_{k}^{-1} )\big) \mathcal{N}(\mathbf{x}_{k}|\hat {\mathbf{x}}_{k}^{-},\mathbf{P}_{k}^{-})&\nonumber\\
	&q(\mathbf{x}_k)\propto \exp \big(-\frac{1}{2}   ({\mathbf{v}_k^9}^{\top} 
	\mathbf{R}_{k}^{-1} \mathbf{v}_k^9)\big) \mathcal{N}(\mathbf{x}_{k}|\hat {\mathbf{x}}_{k}^{-},\mathbf{P}_{k}^{-})&\nonumber
\end{flalign}
where $\mathbf{v}_k^9=\mathbf{y}_{k}-\mathbf{h}(\mathbf{x}_{k})  - \mathbf{\Omega}_{k} {{}\hat{\mathbf{\Theta}}_{k}^{+}}$. 

Using general Gaussian filtering results \cite{sarkka2023bayesian}, we can further write $q(\mathbf{x}_k)$ as
\begin{flalign}
	&q(\mathbf{x}_k)\propto\mathcal{N}({\mathbf{x}}_{k}|\hat{\mathbf{x}}_{k}^{+},\mathbf{P}_{k}^{+})\nonumber&
\end{flalign}
where the parameters $\hat{\mathbf{x}}_{k}^{+}$ and $\mathbf{P}_{k}^{+}$ can be updated as
\begin{flalign*}
	\hat {\mathbf{x}} _{k} ^{+} &= \hat {\mathbf{x}}_{k}^{-} + \mathbf{K}_k(\mathbf{y}_{k} - {\boldsymbol{\Omega}} _{k} \mathbf{\hat{\Theta}}^+_{k} - \bm{\mu}_k) & \\
	\bm{\mu}_k&=\langle \mathbf{h}(\mathbf{x}_{k})  \rangle_{p(\mathbf{x}_k|\mathbf{y}_{k-1})}& \\
	\mathbf{P}_{k}^{+} &= \mathbf{P}_{k}^{-} - \mathbf{C}_{k}\mathbf{K}_{k}^{\top}& \\
	\mathbf{K}_{k} &= \mathbf{C}_{k}\mathbf{S}_{k}^{-1}& \\
	\mathbf{C}_{k}&=\big \langle(\mathbf{x}_{k} - \hat {\mathbf{x}}_{k}^{-} )(\mathbf{h}(\mathbf{x}_{k}) - \bm{\mu}_k) \rangle_{p(\mathbf{x}_k|\mathbf{y}_{k-1})}& \\
	\mathbf{S}_{k}&=\big \langle(\mathbf{h}(\mathbf{x}_{k}) - \bm{\mu}_k)(\mathbf{h}(\mathbf{x}_{k}) - \bm{\mu}_k)^{\top}  \rangle_{p(\mathbf{x}_k|\mathbf{y}_{k-1})} + \mathbf{R}_{k}& 
\end{flalign*}

\section{\texorpdfstring{Derivation of $q(\boldsymbol{\mathcal{I}}_k)$}{}}\label{bias_up2}
\hspace{.5cm}
Using \eqref{eqn_vb_1}, \eqref{eqn_vb_4} and \eqref{eqn_vb_8} we can write $q(\boldsymbol{\mathcal{I}}_{k})$ as
\begin{flalign}
	q(\boldsymbol{\mathcal{I}}_{k})\propto &\exp\big( \big\langle\mathrm{ln}(p(\mathbf{y}_k|\mathbf{x}_{k},\bm{\mathcal{I}}_k,{\mathbf{\Theta}}_k) p(\bm{\mathcal{I}}_k)p(\mathbf{x}_k|\mathbf{y}_{1:k-1}) p({\mathbf{\Theta}}_k|\mathbf{y}_{1:k-1})) \rangle_{ q(\mathbf{x}_k) {q(\mathbf{\Theta}}_k)}\big)& \nonumber\\
	\propto &\exp\big( \big\langle\mathrm{ln}(p(\mathbf{y}_k|\mathbf{x}_{k},\bm{\mathcal{I}}_k,{\mathbf{\Theta}}_k) \rangle_{ q(\mathbf{x}_k) {q(\mathbf{\Theta}}_k)}\big) p(\bm{\mathcal{I}}_k)&  \nonumber
\end{flalign}

As we consider $\textbf{R}_k$ to be diagonal we can write
\begin{flalign}
	q(\boldsymbol{\mathcal{I}}_{k})\propto 
	&\exp\big( \sum_i -\frac{1}{2 R_{k}(i,i)} a_k(i) \big) \begin{pmatrix}
		\prod_{i} (1-\theta_{k}(i))\delta(\mathcal{I}_{k}(i,i)) + \theta_{k}(i)(\mathcal{I}_{k}(i,i) - 1)\end{pmatrix} \nonumber &
\end{flalign}
where 
\begin{flalign}
	a_k(i)&=\langle {({y}_{k}({i}) - (h(\mathbf{x}_{k})(i) +   \mathcal{I}_{k}(i,i) \Theta_{k}(i)) )^{2}} \rangle_{q(\mathbf{x}_{k})q(\mathbf{\Theta}_{k})}&\nonumber\\
	& =\langle  ({h(\mathbf{x}_{k})(i) - {\nu}_k(i) + {\mathcal{I}}_{k}(i,i) ({\Theta}_{k}(i) - {{{}\hat {\Theta}_{k}^{\text{+}}}}(i) ) + {\nu}_k(i) + {\mathcal{I}}_{k}(i,i) {{{}\hat {\Theta}_{k}^{\text{+}}}}(i) - y_{k}(i) )^{2}}  \rangle_{{\begin{subarray}{l}		q(\mathbf{x}_{k})\nonumber\\.q(\mathbf{\Theta}_{k})  
			\end{subarray}
	}}  &\nonumber\\
	&=\bar{h}^2_k+{{\mathcal{I}}_{k}}^2(i,i) \bar{\Theta}^2_k+ ({\nu}_k(i) + {\mathcal{I}}_{k}(i,i) {{{}\hat {\Theta}_{k}^{\text{+}}}}(i) - y_{k}(i))^{2}\nonumber & 	
\end{flalign}

Consequently $q(\boldsymbol{\mathcal{I}}_{k})$ can be expressed as follows 
\begin{flalign}
	&q(\boldsymbol{\mathcal{I}}_{k})	=\prod_{i=1}^{m} (1-{\Omega_{k-1}}(i,i)) \delta( {{{\mathcal{I}}}_{k-1}(i,i)} )+{\Omega_{k-1}}(i,i)\delta( {{{\mathcal{I}}}_{k-1}(i,i)}-1)\nonumber&
\end{flalign}
where the parameters of ${q(\bm{\mathcal{I}}_k)}$ are updated iteratively as
\begin{flalign*}
	\Omega_{k}(i,i) &= \frac{{\Pr({\mathcal{I}}_{k}(i,i) = 1)}}{({{\Pr}({\mathcal{I}}_{k}(i,i) = 1) + \Pr({\mathcal{I}}_{k}(i,i) = 0)})}&\label{eqn_vb_29}
\end{flalign*}
where $k(i)$ denotes the proportionality constant and
\begin{flalign}
	&\Pr({\mathcal{I}}_{k}(i,i) = 0) = k(i) (1 - \theta_{k}(i)) \exp{\big({-}\frac {1}{2} \big(\frac {(y_k(i) - {\nu}_k(i) )^{2}}{R_{k}(i,i)} + \bar{h}^2_k\big)\big)}&  \nonumber \\
	&\Pr({\mathcal{I}}_{k}({i,i}) = 1) = k(i) \theta_{k}(i) \exp{ \big({-}\frac{1}{2} \frac{\bar{h}^2_k + \bar{\Theta}^2_k + ( {\nu}_k(i) + {\hat{\Theta}}_k^{+}(i) - y_k(i))^{2}}{R_{k}(i,i)}\big)}& \nonumber \\
	&\bm{\nu}_k= \langle {\mathbf{h}}(\mathbf{x}_{k})  \rangle_{q(\mathbf{x}_k)}& \nonumber\\
	&\bar{h}^2_k = \langle(h(\textbf{x}_{k})(i) - {\nu}_k(i) )^{2}\rangle_{q(\mathbf{x}_k)}&\nonumber\\
	&\bar{\Theta}^2_k = \langle({\Theta}_{k}(i) - { {\hat{\Theta}}_k^{+}(i) } )^2\rangle_{q(\mathbf{\Theta}_k)}&\nonumber
\end{flalign}
\section{Derivation of  $q(\mathbf{\Theta}_{k})$}\label{bias_up3}
\hspace{.5cm}
Using \eqref{eqn_vb_1}, \eqref{eqn_vb_5_} and \eqref{eqn_vb_8} we can write $q(\mathbf{\Theta}_k)$ as
\begin{flalign}
	q(\mathbf{\Theta}_{k})&\propto \exp\big( \big\langle\mathrm{ln}(p(\mathbf{y}_k|\mathbf{x}_{k},\bm{\mathcal{I}}_k,{\mathbf{\Theta}}_k) p(\bm{\mathcal{I}}_k)p(\mathbf{x}_k|\mathbf{y}_{1:k-1}) p({\mathbf{\Theta}}_k|\mathbf{y}_{1:k-1})) \rangle_{ q(\mathbf{\mathbf{x}_{k}}) q({{\bm{\mathcal{I}}}_k}) }\big)&\nonumber \\
	&\propto \exp\big( \big\langle\mathrm{ln}(p(\mathbf{y}_k|\mathbf{x}_{k},\bm{\mathcal{I}}_k,{\mathbf{\Theta}}_k)\rangle_{ q(\mathbf{\mathbf{x}_{k}}) q({{\bm{\mathcal{I}}}_k}) } \big) p(\mathbf{\Theta}_k|\mathbf{y}_{1:k-1}) &\nonumber\\
	&\propto \exp \big( \langle \mathrm{ln}(\mathcal{N}(\mathbf{y}_{k}|\mathbf{h}(\mathbf{x}_{k}) + \boldsymbol{\mathcal{I}}_{k}\mathbf{\Theta}_{k},\mathbf{R}_{k})\rangle_{ q(\mathbf{\mathbf{x}_{k}}) q({{\bm{\mathcal{I}}}_k}) }\big)\mathcal{N}(\mathbf{\Theta}_{k}|\mathbf{\hat{\Theta}}^-_{k},\mathbf{{\Sigma}}^-_{k}) &\nonumber
	\\
	&\propto \exp ( \langle (-\frac{1}{2}{\mathbf{v}_k^6}^{\top} 
	\mathbf{R}_{k}^{-1} \mathbf{v}_k^6 ) -\frac{1}{2} \mathrm{ln}(2 \pi) ^{m} | \mathbf{R}_{k}|)\rangle_{ q(\mathbf{\mathbf{x}_{k}}) q({{\bm{\mathcal{I}}}_k}) }\big) \mathcal{N}(\mathbf{\Theta}_{k}|\mathbf{\hat{\Theta}}^-_{k},\mathbf{{\Sigma}}^-_{k})&\nonumber\\
	&\propto \exp \big(-\frac{1}{2} \langle \mathrm{tr} (\mathbf{v}_k^6{\mathbf{v}_k^6}^{\top} 
	\mathbf{R}_{k}^{-1} )\rangle_{ q({{\bm{\mathcal{I}}}_k}) {q(\mathbf{\Theta}}_k)}\big) \mathcal{N}(\mathbf{\Theta}_{k}|\mathbf{\hat{\Theta}}^-_{k},\mathbf{{\Sigma}}^-_{k}) &\nonumber
\end{flalign}
where we write $\mathbf{v}_k^6$ in a useful form $\mathbf{v}_k^6=(\boldsymbol{\mathcal{I}}_{k}-\boldsymbol{\mathbf{\Omega}}_{k})\mathbf{\Theta}_{k}+\boldsymbol{\mathbf{\Omega}}_{k}\mathbf{\Theta}_{k}+\mathbf{h}(\mathbf{x}_{k})  -\mathbf{y}_{k}$ for the subsequent derivation of $q(\mathbf{\Theta}_{k})$ as
\begin{flalign}
	& q(\mathbf{\Theta}_k)\propto
	\exp\big({-}\frac{1}{2} \mathrm{tr}\big((\mathbf{B}_{k}+\mathbf{v}_k^{10}{\mathbf{v}_k^{10}}^{\top}+\mathbf{v}_k^{10} {\mathbf{v}_k^{11}}^{\top}+\mathbf{v}_k^{11} {\mathbf{v}_k^{10}}^{\top})\mathbf{R}_{k}^{-1}\big)\big)\mathcal{N}(\mathbf{\Theta}_{k}|\mathbf{\hat{\Theta}}^-_{k},\mathbf{{\Sigma}}^-_{k}) \nonumber &
\end{flalign}
where $\mathbf{B}_k=\mathrm{diag}(\mathbf{\Theta}_{k})\mathbf{\Omega}_k (\mathbf{I}-\mathbf{\Omega}_k)\mathrm{diag}(\mathbf{\Theta}_{k})$, $\mathbf{v}_k^{10}=\boldsymbol{\mathbf{\Omega}}_{k}\mathbf{\Theta}_{k}$ and $\mathbf{v}_k^{11}=\bm{\nu}_k-\mathbf{y}_{k}$ where $\bm{\nu}_k= \langle {\mathbf{h}}(\mathbf{x}_{k})  \rangle_{q(\mathbf{x}_k)}$. 

Adding a constant term $\mathbf{v}_k^{11}{\mathbf{v}_k^{11}}^{\top}$ to complete the square in the exponential expression and considering $\mathbf{R}_k$ to be diagonal yields
\begin{flalign}
	q(\mathbf{{\Theta}}_k)\propto\ &\overset{\mathcal{N}(\mathbf{\Theta}_{k}|\mathbf{0},(\boldsymbol{\Omega}_{k}(1-\boldsymbol{\Omega}_{k})\mathbf{R}_{k}^{-1})^{-1})}{\overbrace{\exp\big(-\frac{1}{2} \mathbf{\Theta}_{k}^{\top} (\boldsymbol{\Omega}_{k} (1 - \boldsymbol{\Omega}_{k})\mathbf{R}_{k}^{-1}\mathbf{\Theta}_{k}\big)}} \times \overset{\mathcal{N}(\mathbf{\Theta}_{k}|\hat{\mathbf{\Theta}}_{k}^{*},\mathbf{\Sigma}_{k}^{*})}{\overbrace{\exp \big(-\frac{1}{2}   ({\mathbf{v}_k^{12}}^{\top} 
			\mathbf{R}_{k}^{-1} \mathbf{v}_k^{12})\big) \mathcal{N}(\mathbf{\Theta}_{k}| \hat {\mathbf{\Theta}}_{k}^{-},\mathbf{\Sigma}_{k}^{-})}}&\nonumber \\
	\propto\ &\mathcal{N}\big( \mathbf{\Theta}_{k}| \hat{\mathbf{\Theta}}_{k}^{+}, \mathbf{\Sigma}_{k}^{+}\big)&\nonumber
\end{flalign}
where $\mathbf{v}_k^{12}=\mathbf{y}_{k}-\bm{\nu}_k  - \mathbf{\Omega}_{k} {{\mathbf{\Theta}}_{k}}$. 

Using general Gaussian filtering results \cite{sarkka2023bayesian}, we can update $\hat {\mathbf{\Theta}}_{k}^{*}$ and $\mathbf{\Sigma}_{k}^{*} $ using \eqref{eqn_vb_30}-\eqref{eqn_vb_34}. The following appears as a result of the product of two multivariate Gaussian distributions \cite{petersen2008matrix}
\begin{flalign}
	\mathcal{N}(\mathbf{x}|\mathbf{m}_{1},\boldsymbol{\Sigma} _{1})&\mathcal{N}(\mathbf{x}|\mathbf{m}_{2},\boldsymbol{\Sigma}_{2})\propto \mathcal{N}(\mathbf{x}|\mathbf{m}_{c},\boldsymbol{\Sigma}_{c})&\nonumber
	\\
	\boldsymbol{\Sigma}_{c} &= (\boldsymbol{\Sigma}_{1}^{-1} + \boldsymbol{\Sigma}_{2}^{-1})^{-1}  & \nonumber \\
	\mathbf{m}_{c} &= \boldsymbol{\Sigma}_{c} (\boldsymbol{\Sigma}_{1}^{-1} \mathbf{m}_{1} + \boldsymbol{\Sigma}_{2}^{-1}\mathbf{m}_{2}) & \nonumber
\end{flalign}

Using the above result, we can update the parameters $ \hat{\mathbf{\Theta}}_{k}^{+}$ and $\mathbf{\Sigma}_{k}^{+}$ using the following 
\begin{flalign*}
	\hat {\mathbf{\Theta}}_{k}^{*} &= \hat{\mathbf{\Theta}}_{k}^{-} + \bm{\mathcal{K}}_{k}(\mathbf{y}_{k} - ( \bm{\nu}_k + \boldsymbol{\Omega}_{k} \hat{\mathbf{\Theta}}_{k}^{-}))&\\
	\mathbf{{\Sigma}}^*_{k} &= \mathbf{{\Sigma}}^-_{k} - \bm{\mathcal{C}}_{k}\bm{\mathcal{K}}_{k}^{\top}&\\
	\bm{\mathcal{K}}_{k} &= \bm{\mathcal{C}}_{k}\bm{\mathcal{S}}_{k}^{-1}&\\
	\bm{\mathcal{C}}_{k}&=\mathbf{{\Sigma}}^-_{k}\boldsymbol{\Omega}_{k}^{\top}&
	\\
	\bm{\mathcal{S}}_{k} &= \boldsymbol{\Omega}_{k}\mathbf{{\Sigma}}^-_{k}\boldsymbol{\Omega}_{k}^{\top} + \mathbf{R}_{k}&\\
	\hat{\mathbf{\Theta}}_{k}^{+}&=\mathbf{\Sigma}_{k}^{+}{\mathbf{{\Sigma}}^*_{k}}^{-1}\hat{\mathbf{\Theta}}_{k}^{*}&\\
	\mathbf{\Sigma}_{k}^{+} &= \big(\boldsymbol{\Omega}_{k}(\mathbf{I}-\boldsymbol{\Omega}_{k})\mathbf{R}_{k}^{-1} + {\mathbf{\Sigma}_{k}^*}^{-1}\big)^{-1}&
\end{flalign*}
\chapter{} \label{app_EM}

\section{\texorpdfstring{Evaluating ${\mathbf{R}}_{k}^{-1}({\boldsymbol{ \hat{\mathcal{I}} }_{k}})$}{}  }\label{FirstAppendix_EM}
For evaluating ${\mathbf{R}}_{k}^{-1}({\boldsymbol{ \hat{\mathcal{I}} }_{k}})$ we consider that ${\mathbf{R}}_{k}({\boldsymbol{ \hat{\mathcal{I}} }_{k}})$ can easily be rearranged by swapping rows/columns depending on $\hat{\boldsymbol{\mathcal{I}}}_k$ as  
\begin{align*}
	{{\bm{\mathfrak{R}}}}_{k}({\boldsymbol{ \hat{\mathcal{I}} }_{k}})= 
	\begin{bmatrix}
		\bm{\mathtt{R}}_k / {\epsilon}  & \mathbf{0}  \\
		\mathbf{0} &  \hat{\mathbf{R}}_{k}
	\end{bmatrix}  
\end{align*}

where $\bm{\mathtt{R}}_k$ is a sub-matrix with diagonal entries of  ${\mathbf{R}}_k$.  $\hat{\mathbf{R}}_{k}$ contains the rest of the fully populated submatrix of $	{\mathbf{R}}_{k}({\boldsymbol{ \hat{\mathcal{I}} }_{k}})$ corresponding to entries of $\hat{\boldsymbol{\mathcal{I}}}_k=1$. Inversion of ${\bm{\mathfrak{R}}}_{k} ({\boldsymbol{ \hat{\mathcal{I}} }_{k}})$ results in
\begin{align*}
	{\bm{\mathfrak{R}}}_{k}^{-1}({\boldsymbol{ \hat{\mathcal{I}} }_{k}})=& \begin{bmatrix}
		{\epsilon} \bm{\mathtt{R}}^{-1}_k & \mathbf{0}  \\
		\mathbf{0} &  \hat{\mathbf{R}}^{-1}_{k}
	\end{bmatrix} 
\end{align*}
Finally, ${\bm{\mathfrak{R}}}_{k}^{-1}({\boldsymbol{ \hat{\mathcal{I}} }_{k}})$ can be swapped accordingly to obtain the required matrix ${\mathbf{R}}_{k}^{-1}({\boldsymbol{ \hat{\mathcal{I}} }_{k}})$ . 

%
\section{Simplifying $\hat{\tau}_k(i)$
}\label{SecondAppendix_EM}
We can swap the \textit{i}th row/column entries of  ${\mathbf{R}}_{k}( {\mathcal{I}_k(i)} =1 , \hat{\bm{\mathcal{I}}}_k(i-) )$ and ${\mathbf{R}}_{k}( {\mathcal{I}_k(i)} =1 , \hat{\bm{\mathcal{I}}}_k(i-) )$ with the first row/column elements to obtain  
\begin{align*}
	|{\bm{\mathfrak{R}}}_{k}( {\mathcal{I}_k(i)} =1 , \hat{\bm{\mathcal{I}}}_k(i-) )|&= 
	\begin{vmatrix} 
		R_k(i,i) & \mathbf{R}_k(i,-i) \\
		\mathbf{R}_k(-i,i) & \hat{\mathbf{R}}_{k}(-i,-i)
	\end{vmatrix}  \\
	|{\mathbf{R}}_{k}( {\mathcal{I}_k(i)} =\epsilon , \hat{\bm{\mathcal{I}}}_k(i-) )|&= 
	\begin{vmatrix}
		R_k(i,i)/\epsilon & \mathbf{0} \\
		\mathbf{0} & \hat{\mathbf{R}}_{k}(-i,-i)
	\end{vmatrix} 
\end{align*}
\vfill
\newpage
Consequently, we can write
\begin{align*}
	&\ln\Big(\frac{|{\mathbf{R}}_{k}( {\mathcal{I}_k(i)=1} , \hat{\bm{\mathcal{I}}}_k(i-) ) |}{|{\mathbf{R}}_{k}( {\mathcal{I}_k(i)=\epsilon} , \hat{\bm{\mathcal{I}}}_k(i-) )|}\Big)=\ln\Big(\frac{|{\bm{\mathfrak{R}}}_{k}( {\mathcal{I}_k(i)=1} , \hat{\bm{\mathcal{I}}}_k(i-) ) |}{|{\bm{\mathfrak{R}}}_{k}( {\mathcal{I}_k(i)=\epsilon} , \hat{\bm{\mathcal{I}}}_k(i-) )|}\Big)\nonumber \\ 
	&= \ln \Bigg|\textbf{I}-\frac{\mathbf{R}_k(-i,i)\mathbf{R}_k(i,-i) (\hat{\mathbf{R}}_{k}(-i,-i))^{-1}}{{R}^{i,i}_k}\Bigg|+ \ln(\epsilon) 
\end{align*}
where we have used the following property from matrix algebra \cite{zhang2006schur}
\begin{align}
	\begin{vmatrix}
		\bm{\mathsf{A}} & \bm{\mathsf{B}} \\
		\bm{\mathsf{C}} & \bm{\mathsf{D}}
	\end{vmatrix}= |\bm{\mathsf{A}}| |\bm{\mathsf{D}}-\bm{\mathsf{C}}\bm{\mathsf{A}}^{-1}\bm{\mathsf{B}}|  \nonumber
\end{align}
Resultingly, we can simplify \eqref{eqn_fl_up5_i} to \eqref{eqn_fl_up10_i}.

\section{Evaluating $\triangle{\hat{\mathbf{R}}}_{k}^{-1}$}\label{ThirdAppendix_EM}
To avoid redundant calculations during the evaluation of $\triangle{\hat{\mathbf{R}}}_{k}^{-1}$, we can first swap the \textit{i}th row/column elements of matrices with the first row/column entries in \eqref{eqdelR} to obtain
\begin{align*}
	\triangle{\hat{\bm{\mathfrak{R}}}}_{k}^{-1}&=({\bm{\mathfrak{R}}}_{k}^{-1}( {\mathcal{I}_k(i)=1} , \hat{\bm{\mathcal{I}}}_k(i-) )-{\bm{\mathfrak{R}}}_{k}^{-1}( {\mathcal{I}_k(i)=\epsilon} , \hat{\bm{\mathcal{I}}}_k(i-) ))\nonumber\\
	&=\begin{bmatrix} 
		R_k(i,i) & \mathbf{R}_k(i,-i) \\
		\mathbf{R}_k(-i,i) & \hat{\mathbf{R}}_{k}(-i,-i)
	\end{bmatrix}^{-1}-\begin{bmatrix} 
		R_k(i,i)/\epsilon & \mathbf{0}\\
		\mathbf{0} & \hat{\mathbf{R}}_{k}(-i,-i)
	\end{bmatrix}^{-1}
\end{align*}
which an be further simplified using the following property from matrix algebra \cite{zhang2006schur} 
\begin{align*}
	\begin{bmatrix}
		\bm{\mathsf{A}} & \bm{\mathsf{B}} \\
		\bm{\mathsf{C}} & \bm{\mathsf{D}}
	\end{bmatrix}^{-1}= 	
	\begin{bmatrix}
		\bm{\mathsf{S}}^{-1} & -\bm{\mathsf{S}^{-1} \bm{\mathsf{B}} \bm{\mathsf{D}}^{-1} } \\
		-\bm{\mathsf{D}}^{-1} \bm{\mathsf{C}} \bm{\mathsf{S}^{-1} } & \bm{\mathsf{D}}^{-1}+\bm{\mathsf{D}}^{-1} \bm{\mathsf{C}} \bm{\mathsf{S}}^{-1} \bm{\mathsf{B}} \bm{\mathsf{D}}^{-1}  
	\end{bmatrix}  \nonumber
\end{align*}
where $\bm{\mathsf{S}}$ is the Schur's complement of $\bm{\mathsf{D}}$ given as $\bm{\mathsf{S}}=\bm{\mathsf{A}}-\bm{\mathsf{B}}\bm{\mathsf{D}}^{-1}\bm{\mathsf{C}}$. As a result, we obtain
\begin{align*}
	\triangle{\hat{\bm{\mathfrak{R}}}}_{k}^{-1}&=\begin{bmatrix}
		\bm{\Xi}({i,i}) & \bm{\Xi}({i,-i}) \\
		\bm{\Xi}({-i,i}) & \bm{\Xi}({-i,-i}) 
	\end{bmatrix}
\end{align*}
where 
\begin{align*}
	\bm{\Xi}({i,i})&=\frac{1}{R_k(i,i)-\mathbf{R}_k(i,-i)(\hat{\mathbf{R}}_{k}(-i,-i))^{-1} \mathbf{R}_k(-i,i)}-\frac{\epsilon}{R_k(i,i)}\\
	\bm{\Xi}({i,-i})&=-\frac{\mathbf{R}_k(i,-i) (\hat{\mathbf{R}}_{k}(-i,-i))^{-1}}{R_k(i,i)-\mathbf{R}_k(i,-i)(\hat{\mathbf{R}}_{k}(-i,-i))^{-1} \mathbf{R}_k(-i,i)}\\
	\bm{\Xi}({-i,i})&=-\frac{ (\hat{\mathbf{R}}_{k}(-i,-i))^{-1}\mathbf{R}_k(-i,i)}{R_k(i,i)-\mathbf{R}_k(i,-i)(\hat{\mathbf{R}}_{k}(-i,-i))^{-1} \mathbf{R}_k(-i,i)}
\end{align*}
\begin{align*}
	\bm{\Xi}({-i,-i})&=\frac{ (\hat{\mathbf{R}}_{k}(-i,-i))^{-1}\mathbf{R}_k(-i,i) \mathbf{R}_k(i,-i) (\hat{\mathbf{R}}_{k}(-i,-i))^{-1}}{R_k(i,i)-\mathbf{R}_k(i,-i)(\hat{\mathbf{R}}_{k}(-i,-i))^{-1} \mathbf{R}_k(-i,i)}
\end{align*}
where the redundant calculations can be computed once and stored for further computations e.g. $({R_k(i,i)-\mathbf{R}_k(i,-i)(\hat{\mathbf{R}}_{k}(-i,-i))^{-1} \mathbf{R}_k(-i,i)})^{-1}$ and $(\hat{\mathbf{R}}_{k}(-i,-i))^{-1}$. Lastly, the first row/column entries of $\triangle{\hat{\bm{\mathfrak{R}}}}_{k}^{-1}$ are interchanged to the actual $i$th row/column positions to obtain the required  $\triangle{\hat{\mathbf{R}}}_{k}^{-1}$. 
\chapter{} \label{app_PF}
\scriptsize
\section{Derivation of $\eqref{PPFeqn10}$}
\begin{align*}&\left.{\begin{array}{l} {x}_{k}=f({x}_{k-1})+{q}_{k-1} \\ {y}_{k}=h({x}_{k})+{r}_{k}+{\mathcal{I}}^{1}_{k}~{\mu}_{k} \\ {\mathcal{I}}^{1}_{k}=\delta ({\mathcal{J}}_{k}-1)\end{array}}\right \} \begin{array}{l} \text {Proposed SSM Model} \\ \text {State Estimation and Outlier} \\ \text {Detection\ }(n=1, m=1) \end{array}\\
&p(\underbrace{x_k ,\mathcal{J}_k}_{s_k}|y_{1:k})= p({y}_{k}|\underbrace{x_k ,\mathcal{J}_k}_{s_k}\stkout{{y}_{1:{k-1}}}) p(\underbrace{ x_k ,\mathcal{J}_k}_{s_k}|{y}_{1:{k-1}})/ p({y}_{k}|{y}_{1:{k-1}})&\\	
		&{p({x_k ,\mathcal{J}_k}|{y}_{1:{k-1}})}\propto\int \sum_{\mathcal{J}_{k-1}}p({x_k ,\mathcal{J}_k}|{x_{k-1} ,\mathcal{J}_{k-1}},\stkout{{y}_{1:{k-1}}}))p({x_{k-1} ,\mathcal{J}_{k-1}}|{y}_{{1:k-1}})\ d{x_{k-1}}&\\
		&=  \int p({x_k }|{x_{k-1} ,\stkout{{\mathcal{J}_k,\mathcal{J}_{k-1}}}}) \overbrace{  \sum_{\mathcal{J}_{k-1}}\left(p({\mathcal{J}_k }|{\stkout{x_{k-1}},\mathcal{J}_{k-1}})p({ \mathcal{J}_{k-1}}|x_{k-1},{y}_{{1:k-1}})\right)}^{p({\mathcal{J}_{k}}|x_{k-1},{y}_{{1:k-1}})}p({ x_{k-1}}|{y}_{{1:k-1}})\ d{x_{k-1}}&\\
		&=  \int p(x_k|x_{k-1}) \overbrace{ \left((\frac{1}{2}\delta(\mathcal{J}_k)+\frac{1}{2}\delta(\mathcal{J}_{k}-1))
			\overbrace{ \sum_{\mathcal{J}_{k-1}}(\delta(\mathcal{J}_{k-1})+\delta(\mathcal{J}_{k-1}-1))p( \mathcal{J}_{k-1}|x_{k-1},{y}_{{1:k-1}})}^1\right)}^{p({\mathcal{J}_{k}}|x_{k-1},{y}_{{1:k-1}})}
		p({ x_{k-1}}|{y}_{{1:k-1}})d{x_{k-1}}& \\
		&  p(\underbrace{  x_k ,\mathcal{J}_k}_{s_k}|y_{1:k})\propto \delta(\mathcal{J}_k)p({y}_{k}|{x_k ,\mathcal{J}_k}) p(x_k|{y}_{1:{k-1}})+\delta(\mathcal{J}_{k}-1)p({y}_{k}|{x_k ,\mathcal{J}_{k}}) p(x_k|{y}_{1:{k-1}})\\
		&  p({x_k }|y_{1:k})\propto {(p({y}_{k}|{x_k ,\mathcal{J}_k=0}) +p({y}_{k}|{x_k ,\mathcal{J}_{k}=1}))} p(x_k|{y}_{1:{k-1}})&\\
		&  p({x_k }|y_{1:k})\propto\mathcal{N}(y_k|b\  x_k,\sigma^{2}_{r})\ 
		\mathcal{N}({x}_{k}|\hat{x}^-_k,P^-_k)+\mathcal{N}(y_k|b\  x_k,\sigma^{2}_{r}+\sigma^{2}_{\mu})\ \mathcal{N}({x}_{k}|\hat{x}^-_k,P^-_k)&\\
		&  p({x_k}|y_{1:k})\propto \overbrace{  (k_2\exp[-\frac{(x_k-y_{k}/b)^2}{2({\sigma_{r}}/b)^2}]\ 
			+k_3\exp[-\frac{(x_k-y_{k}/b)^2}{2(({\sigma_{r}}+{\sigma_{\mu}})/b)^2}])}^{p(y_k|x_k,y_{1:k-1})} 
		\overbrace{  \exp[-\frac{(x_k-\hat{x}^-_k)^2}{2{P^-_k}}]}^{p(x_k|y_{1:{k-1}})}\\
		&&\\
		&&\\
		&&\\
		&&\\
		&&\\
		&&\\
		&&\\
		&&\\
\end{align*}
\vfill
\section{Derivation of $\eqref{PPFeqn20}$}
		\begin{align*}&\left.{\begin{array}{l} {x}_{k}=f({x}_{k-1})+{q}_{k-1} \\ {y}_{k}=h({x}_{k})+{r}_{k}+{\mathcal{I}}^{2}_{k}~{\nu}_{k} \\ {\Theta}_{k}={\mathcal{I}}^{0}_{k-1}\overset{{\sim}}{\Theta}_k+{\mathcal{I}}^{2}_{k-1}(\Theta_{k-1}+\triangle \Theta_{k-1}) \\ {\mathcal{I}}^{\xi}_{k}=\delta ({\mathcal{J}}_{k}-n),~\xi\in (0,2) \end{array}}\right \}\begin{array}{l} \text {Proposed SSM Model} \\ \text {for Bias Detection and} \\ \text {State Estimation} \\ (n=1, m=1) \end{array}\\
		&p(\underbrace{x_k ,\mathcal{J}_k,\Theta_k}_{s_k}|y_{1:k})= p({y}_{k}|\underbrace{x_k ,\mathcal{J}_k,\Theta_k}_{s_k}\stkout{{y}_{1:{k-1}}}) p(\underbrace{x_k ,\mathcal{J}_k,\Theta_k}_{s_k}|{y}_{1:{k-1}})/ p({y}_{k}|{y}_{1:{k-1}})\\
		&{p({x_k ,\mathcal{J}_k,\Theta_k}|{y}_{1:{k-1}})}=\int\int \sum_{\mathcal{J}_{k-1}}p({x_k ,\mathcal{J}_k,\Theta_k}|{x_{k-1} ,\mathcal{J}_{k-1},\Theta_{k-1}},\stkout{{y}_{1:{k-1}}}))p({x_{k-1} ,\mathcal{J}_{k-1},\Theta_{k-1}}|{y}_{{1:k-1}})\ d{\Theta_{k-1}}d{x_{k-1}}\\
		&{p({x_k ,\mathcal{J}_k,\Theta_k}|{y}_{1:{k-1}})}=\int\int \sum_{\mathcal{J}_{k-1}}p({x_k |\stkout{\mathcal{J}_k,\Theta_k},x_{k-1} ,\stkout{\mathcal{J}_{k-1},\Theta_{k-1}}})p({\Theta_k|\stkout{\mathcal{J}_k,x_{k-1}} ,\mathcal{J}_{k-1},\Theta_{k-1}}))p({\mathcal{J}_k}|\mathcal{J}_{k-1},\stkout{x_{k-1},\Theta_{k-1}})\nonumber\\
		&\hspace{10em}p({\mathcal{J}_{k-1}}|x_{k-1},\Theta_{k-1},{y}_{{1:k-1}})p({\Theta_{k-1} }|\mathcal{J}_{k-1},{y}_{{1:k-1}})p({x_{k-1}}|{y}_{{1:k-1}})\ d{\Theta_{k-1}}d{x_{k-1}}\\
		&{p({x_k ,\mathcal{J}_k,\Theta_k}|{y}_{1:{k-1}})}=\int p({x_k |x_{k-1}})p({x_{k-1}}|{y}_{{1:k-1}})\int (\sum_{\mathcal{J}_{k-1}}p({\mathcal{J}_k}|\mathcal{J}_{k-1}) p({\Theta_k|\mathcal{J}_{k-1},\Theta_{k-1}})\nonumber\\
		&\hspace{10em}p({\mathcal{J}_{k-1}}|x_{k-1},\Theta_{k-1},{y}_{{1:k-1}})) p({\Theta_{k-1} }|x_{k-1},{y}_{{1:k-1}})\ d{\Theta_{k-1}}d{x_{k-1}}\\
		&{p({x_k ,\mathcal{J}_k,\Theta_k}|{y}_{1:{k-1}})}\propto\mathcal{N}(x_k|\hat{x}^-_k,P^-_k)[\delta(\mathcal{J}_k)\mathcal{U}(\Theta_k|c,d)+\delta(\mathcal{J}_k-2)\mathcal{U}(\Theta_k|c,d)\nonumber\\
		&\hspace{9em}+\delta(\mathcal{J}_k)\mathcal{N}(\Theta_k|0,{\sigma}^2_{\triangle})+\delta(\mathcal{J}_k-2)\mathcal{N}(\Theta_k|0,{\sigma}^2_{\triangle})]\\
		&{p({x_k ,\mathcal{J}_k,\Theta_k}|{y}_{1:{k}})}\propto\mathcal{N}(y_k|b x_k+\mathcal{I}^2_k \Theta_k,\sigma_{r}^2+\mathcal{I}^2_k\sigma_{\nu}^2)\mathcal{N}(x_k|\hat{x}^-_k,P^-_k)[\delta(\mathcal{J}_k)\mathcal{U}(\Theta_k|c,d)+\delta(\mathcal{J}_k-2)\mathcal{U}(\Theta_k|c,d)\nonumber\\
		&\hspace{9em}+\delta(\mathcal{J}_k)\mathcal{N}(\Theta_k|0,{\sigma}^2_{\triangle})+\delta(\mathcal{J}_k-2)\mathcal{N}(\Theta_k|0,{\sigma}^2_{\triangle})]\\
		&{p({x_k,\Theta_k}|{y}_{1:{k}})}\propto\mathcal{N}(x_k|\hat{x}^-_k,P^-_k)[\mathcal{N}(y_k|b x_k,\sigma_{r}^2)\mathcal{U}(\Theta_k|c,d)+\mathcal{N}(y_k|b x_k+ \Theta_k,\sigma_{r}^2+\sigma_{\nu}^2)\mathcal{U}(\Theta_k|c,d)\nonumber\\
		&\hspace{9em}+\mathcal{N}(y_k|b x_k,\sigma_{r}^2)\mathcal{N}(\Theta_k|0,{\sigma}^2_{\triangle})+\mathcal{N}(y_k|b x_k+\Theta_k,\sigma_{r}^2+\sigma_{\nu}^2)\mathcal{N}(\Theta_k|0,{\sigma}^2_{\triangle})] \\
	\end{align*}
Assuming $\sigma_{\nu}$ and $\sigma_{\triangle}$ are small
\begin{align*}
		&{p({x_k}|{y}_{1:{k}})}\propto \overbrace{(3k_4\exp[-\frac{(x_k-y_{k}/b)^2}{2({\sigma_{r}}/b)^2}]\ 
					+k_5(\int_{c}^{d}\exp[-\frac{(x_k-(y_k-\Theta_k)/b)^2}{2({\sigma^2_r}+{\sigma^2_{\nu}})}]d\Theta_k))}^{p(y_k|x_k,y_{1:{k-1}})}\ 
				\overbrace{\exp[-\frac{(x_k-\hat{x}^-_k)^2}{2P^-_k}]}^{p(x_k|y_{1:{k-1}})}		
\end{align*}
\vfill
\newpage


\section{Derivation of $\eqref{PPFeqn22}$}
	\begin{align*}
		&p(\underbrace{x_{k+1} ,\mathcal{J}_{k+1},\Theta_{k+1}}_{s_{k+1}}|y_{1:{k+1}})= p({y}_{k+1}|\underbrace{x_{k+1} ,\mathcal{J}_{k+1},\Theta_{k+1}}_{s_{k+1}}\stkout{{y}_{1:{k}}}) p(\underbrace{x_{k+1} ,\mathcal{J}_{k+1},\Theta_{k+1}}_{s_{k+
				1}}|{y}_{1:{k}})/ p({y}_{k+1}|{y}_{1:{k}})\\
		&{p({x_{k+1} ,\mathcal{J}_{k+1},\Theta_{k+1}}|{y}_{1:{k}})}=\int\int \sum_{\mathcal{J}_{k}}p({x_{k+1} ,\mathcal{J}_{k+1},\Theta_{k+1}}|{x_{k} ,\mathcal{J}_{k},\Theta_{k}},\stkout{{y}_{1:{k}}}))p({x_{k} ,\mathcal{J}_{k},\Theta_{k}}|{y}_{{1:k}})\ d{\Theta_{k}}d{x_{k}}\\
		&{p({x_{k+1} ,\mathcal{J}_{k+1},\Theta_{k+1}}|{y}_{1:{k}})}=\int\int \sum_{\mathcal{J}_{k}}p({x_{k+1} |\stkout{\mathcal{J}_{k+1},\Theta_{k+1}},x_{k} ,\stkout{\mathcal{J}_{k},\Theta_{k}}})p({\Theta_{k+1}|\stkout{\mathcal{J}_{k+1},x_{k}} ,\mathcal{J}_{k},\Theta_{k}}))p({\mathcal{J}_{k+1}}|\mathcal{J}_{k},\stkout{x_{k},\Theta_{k}})\nonumber\\
		&\hspace{13em}p({x_{k} ,\mathcal{J}_{k},\Theta_{k}}|{y}_{{1:k}})d{\Theta_{k}}d{x_{k}}\\
		\\
		& \text{Since bias occurs at $k$, we approximate } {p({x_k ,\mathcal{J}_k,\Theta_k}|{y}_{1:{k}})} \text{ by the most significant term i.e.}\\
		\\
		&p({x_k ,\mathcal{J}_k,\Theta_k}|{y}_{1:{k}})\approx\mathcal{N}(y_k|b x_k+\mathcal{I}^2_k \Theta_k,\sigma_{r}^2+\mathcal{I}^2_k\sigma_{\nu}^2)\mathcal{N}(x_k|\hat{x}^-_k,P^-_{k})\delta(\mathcal{J}_k-2)\mathcal{U}(\Theta_k|c,d) \\ 
		&{p({x_{k+1} ,\mathcal{J}_{k+1},\Theta_{k+1}}|{y}_{1:{k}})}\propto\int\int \sum_{\mathcal{J}_{k}}p({x_{k+1} |x_{k} })p({\Theta_{k+1}|\mathcal{J}_{k},\Theta_{k}})p({\mathcal{J}_{k+1}}|\mathcal{J}_{k})\nonumber\\
		&\hspace{13em}\mathcal{N}(y_k|b x_k+\mathcal{I}^2_k \Theta_k,\sigma_{r}^2+\mathcal{I}^2_k\sigma_{\nu}^2)\mathcal{N}(x_k|\hat{x}^-_{k},P^-_{k})\delta(\mathcal{J}_k-2)\mathcal{U}(\Theta_k|c,d)d{\Theta_{k}}d{x_{k}}\\
		&{p({x_{k+1} ,\mathcal{J}_{k+1},\Theta_{k+1}}|{y}_{1:{k}})}\propto\int\int p({x_{k+1} |x_{k} })p({\Theta_{k+1}|\mathcal{J}_{k}=2,\Theta_{k}})p({\mathcal{J}_{k+1}}|\mathcal{J}_{k}=2)\nonumber\\
		&\hspace{13em}\mathcal{N}(y_k|b x_k+ \Theta_k,\sigma_{r}^2+\sigma_{\nu}^2)\mathcal{N}(x_k|\hat{x}^-_{k},P^-_{k})\mathcal{U}(\Theta_k|c,d)d{\Theta_{k}}d{x_{k}} \\
		&{p({x_{k+1} ,\mathcal{J}_{k+1},\Theta_{k+1}}|{y}_{1:{k}})}\propto\int\int \mathcal{N}(x_{k+1}|a x_k,\sigma^2_q)\mathcal{N}(\Theta_{k+1}|\Theta_k,{\sigma}^2_{\triangle})(\delta(\mathcal{J}_{k+1})+\delta(\mathcal{J}_{k+1}-2))\nonumber\\
		&\hspace{13em}\mathcal{N}(y_k|b x_k+ \Theta_k,\sigma_{r}^2+\sigma_{\nu}^2)\mathcal{N}(x_k|\hat{x}^-_{k},P^-_{k})\mathcal{U}(\Theta_k|c,d)d{\Theta_{k}}d{x_{k}}\\
		&{p({x_{k+1} ,\mathcal{J}_{k+1},\Theta_{k+1}}|{y}_{1:{k}})}\propto(\delta(\mathcal{J}_{k+1})+\delta(\mathcal{J}_{k+1}-2))\int{\exp[-\frac{(x_{k+1}-a{x}_{k})^2}{2{\sigma^2_{q}}}]}\int_{c}^{d}{\exp[-\frac{(\Theta_{k}-{\Theta}_{k+1})^2}{2{\sigma}^2_{\triangle}}]}\nonumber\\	&\hspace{13em}\exp[-\frac{(\Theta_k-(y_k-bx_k))^2}{2({\sigma_{r}^2+\sigma_{\nu}^2})}]d{\Theta_{k}}{\exp[-\frac{(x_k-\hat{x}^-_{k})^2}{2 P^-_{k}}]}d{x_{k}}\\
		\\
		&\text{Considering ${\sigma}_{\triangle}$ is small and $\Theta_{k+1}$ lies between $c$ and $d$ we obtain}\\
		\\
		&{p({x_{k+1} ,\mathcal{J}_{k+1},\Theta_{k+1}}|{y}_{1:{k}})}\propto(\delta(\mathcal{J}_{k+1})+\delta(\mathcal{J}_{k+1}-2))\int{\exp[-\frac{(x_{k+1}-a{x}_{k})^2}{2{\sigma^2_{q}}}]}\exp[-\frac{({\Theta_{k+1}}-(y_k-b x_k))^2}{2 ({\sigma}^2_{\triangle}+\sigma _r^2+\sigma _{\nu}^2)}]\nonumber\\
		&\hspace{13em}{\exp[-\frac{(x_k-\hat{x}^-_{k})^2}{2 P^-_{k}}]}d{x_{k}}\\
		&{p({x_{k+1} ,\mathcal{J}_{k+1},\Theta_{k+1}}|{y}_{1:{k}})}\propto(\delta(\mathcal{J}_{k+1})+\delta(\mathcal{J}_{k+1}-2)){\Lambda}(x_{k+1},\Theta_{k+1})\\
		\\
		& \text{with}\\
		&{\Lambda}(x_{k+1},\Theta_{k+1})=\exp[-\frac{{{P^-_k}} (a {\Theta_{k+1}}-a y_k+b {x_{k+1}})^2+({x_{k+1}}-a \hat{x}^-_k)^2(\sigma _{{{\triangle}}}^2 +\sigma _r^2+\sigma _{\nu}^2)+\sigma _q^2 (b \hat{x}^-_k+\Theta_{k+1}-y_k)^2}{2 \left(\sigma _{{{\triangle}}}^2 \left(a^2 {P^-_k}+\sigma _q^2\right)+a^2 {P^-_k} \left(\sigma _r^2+\sigma _{\nu}^2\right)+\sigma _q^2 \left(b^2 {P^-_k}+\sigma _r^2+\sigma _{\nu}^2\right)\right)}]\\
		&{p({x_{k+1} ,\mathcal{J}_{k+1},\Theta_{k+1}}|{y}_{1:{k+1}})}\propto\exp[-\frac{(y_{k+1}-(bx_{k+1}+\mathcal{I}^2_{k+1}\Theta_{k+1}))^2}{-2(\sigma^2_{r}+\mathcal{I}^2_{k+1}\sigma^2_{{\nu}})}](\delta(\mathcal{J}_{k+1})+\delta(\mathcal{J}_{k+1}-2)){\Lambda}(x_{k+1},\Theta_{k+1})\\
			&{p({x_{k+1} }|{y}_{1:{k+1}})}\propto k_6\exp[-\frac{(x_{k+1}-(y_{k+1}/b))^2}{2(\sigma^2_{r}/b^2)}]\int {\Lambda}(x_{k+1},\Theta_{k+1})d\Theta_{k+1}\nonumber\\
		&\hspace{7em}+k_7\int\exp[-\frac{(x_{k+1}-(y_{k+1}-\Theta_{k+1})/b)^2}{2(\sigma^2_{r}+\sigma^2_{{\nu}})/b^2}]{\Lambda}(x_{k+1},\Theta_{k+1})d\Theta_{k+1}\\
		&{p({x_{k+1}}|{y}_{1:{k+1}})}\propto \overbrace{(k_6\exp[-\frac{(x_{k+1}-(y_{k+1}/b))^2}{2(\sigma^2_{r}/b^2)}]\ 
			+k_7\exp[-\frac{(x_{k+1}-{{\hat{x}}^{1}_{k+1}} )^2}{2{{P}^{1}_{k+1}} }]}^{p(y_{k+1}|x_{k+1},y_{1:{k}})}\ 
		\overbrace{\exp[-\frac{(x_{k+1}-a \hat{x}^-_{k})^2}{2(a^2 {P^-_{k}}+\sigma^2_q)}]}^{p(x_{k+1}|y_{1:{k}})}\\
	\end{align*}
 with
\begin{align*}
		&{{\hat{x}}^{1}_{k+1}} =\frac{a^2 P^-_{k} ({y_{k+1}}-y_k)+\sigma _q^2 (b \hat{x}^-_{k}+y_{k+1}-y_k)}{b \left((a-1) a P^-_{k}+\sigma _q^2\right)}\\
		&{{P}^{1}_{k+1}} =\frac{\left(a^2 P^-_{k}+\sigma _q^2\right) \left(a^2 P^-_{k} ({\sigma}^2_{{\triangle}}+2 \sigma _r^2+2 \sigma _{\nu}^2)+\sigma _q^2 \left(b^2 P^-_{k}+{\sigma}^2_{\triangle}+2 \sigma _r^2+2 \sigma _{\nu}^2\right)\right)}{b^2 \left((a-1) a P^-_{k}+\sigma _q^2\right){}^2}\\
		&{p({x_{k+1} }|{y}_{1:{k+1}})}\propto k_6\exp[-\frac{(x_{k+1}-{{\hat{x}}^{2}_{k+1}} )^2}{2 {{P}^{2}_{k+1}}  }]+k_7\exp[-\frac{(x_{k+1}-{{\hat{x}}^{3}_{k+1}} )^2}{2 {{P}^{3}_{k+1}} }]\\
\end{align*}
with
\begin{align*}				
		&{{\hat{x}}^{2}_{k+1}} =\frac{b {y_{k+1}} (a^2 {P^-_k}+\sigma _q^2)+a \hat{x}^-_k \sigma _r^2}{b^2 (a^2 {P^-_k}+\sigma _q^2)+{\sigma_r}^2}\\
		&{{P}^{2}_{k+1}} =\frac{1}{\frac{1}{a^2 {P^-_k}+\sigma^2_q}+\frac{b^2}{\sigma^2_r}} \\
\end{align*}
and
\begin{align*}
		&{{\hat{x}}^{3}_{k+1}} =\frac{a \hat{x}^-_{k}( {\sigma}^2_{\triangle}+2\sigma _r^2+2\sigma _{\nu}^2)+(a-1) a b P^-_{k} (y_{k+1}-y_k)+b \sigma _q^2 (b \hat{x}^-_{k}+y_{k+1}-y_k)}{(a-1)^2 b^2P^-_{k}+b^2 \sigma _q^2+{\sigma}^2_{\triangle}+2 \sigma _r^2+2 \sigma _{\nu}^2}\\
		&{{P}^{3}_{k+1}} =\frac{a^2 P^-_{k} ({\sigma}^2_{\triangle}+2 \sigma _r^2+2 \sigma _{\nu}^2)+\sigma _q^2 \left(b^2 P^-_{k}+{\sigma}^2_{\triangle}+2 \sigma _r^2+2 \sigma _{\nu}^2\right)}{(a-1)^2 b^2 P^-_{k}+b^2 \sigma _q^2+{\sigma}^2_{\triangle}+2 \sigma _r^2+2 \sigma _{\nu}^2}
	\end{align*}



\end{document}